\documentclass[11pt]{elegantbook}

\title{Lectures on AI for Mathematics}

\author{Xiaoyang Chen \& Xiang Jiang}
\institute{School of Mathematical Sciences, Tongji University}
\date{Apr. 8, 2026}
\version{1.0}

\usepackage{array}
\newcolumntype{L}[1]{>{\raggedright\arraybackslash}p{#1}}
\usepackage{microtype}
\usepackage{ragged2e}

\usetikzlibrary{decorations.pathreplacing}
\logo{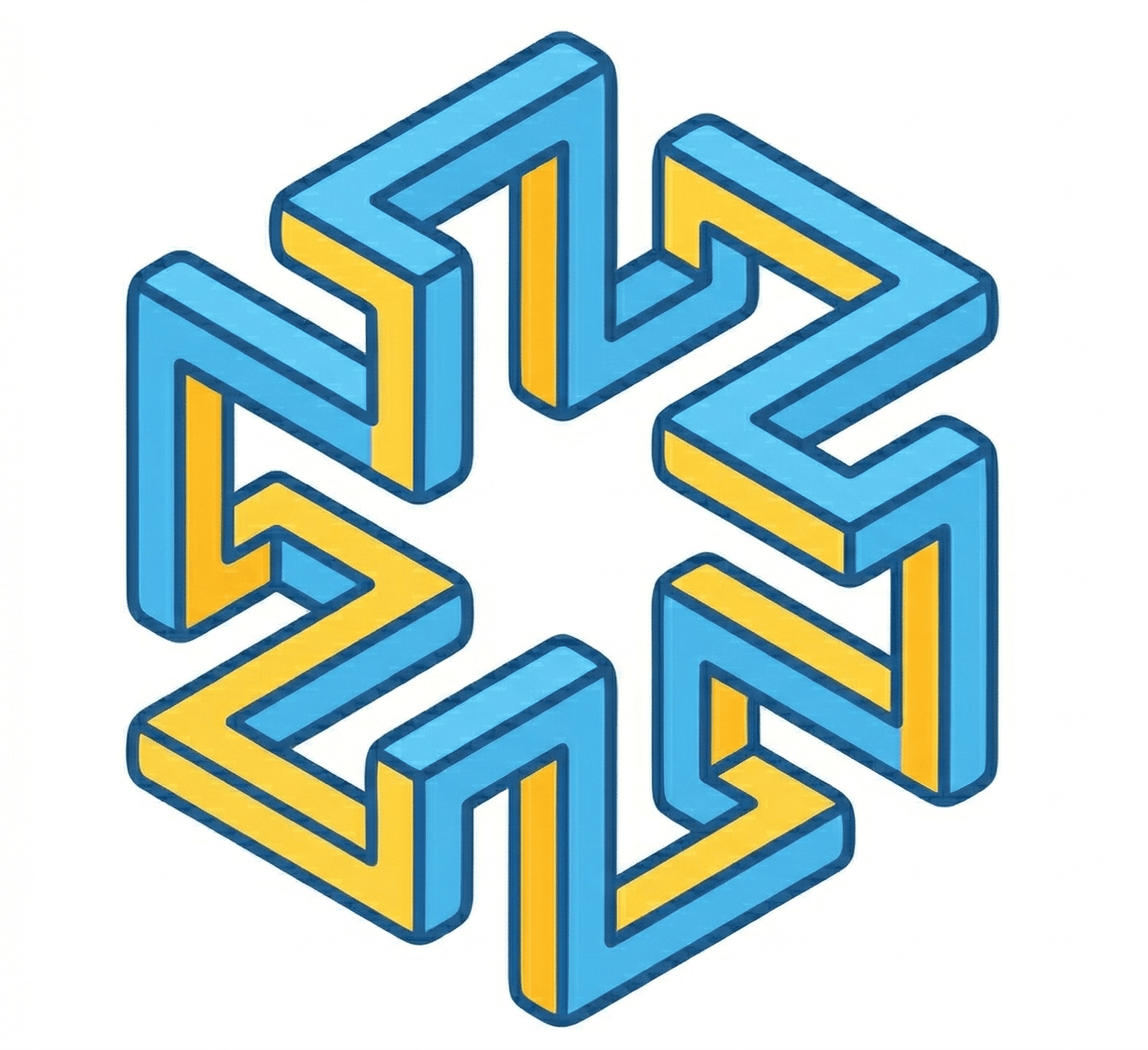}
\cover{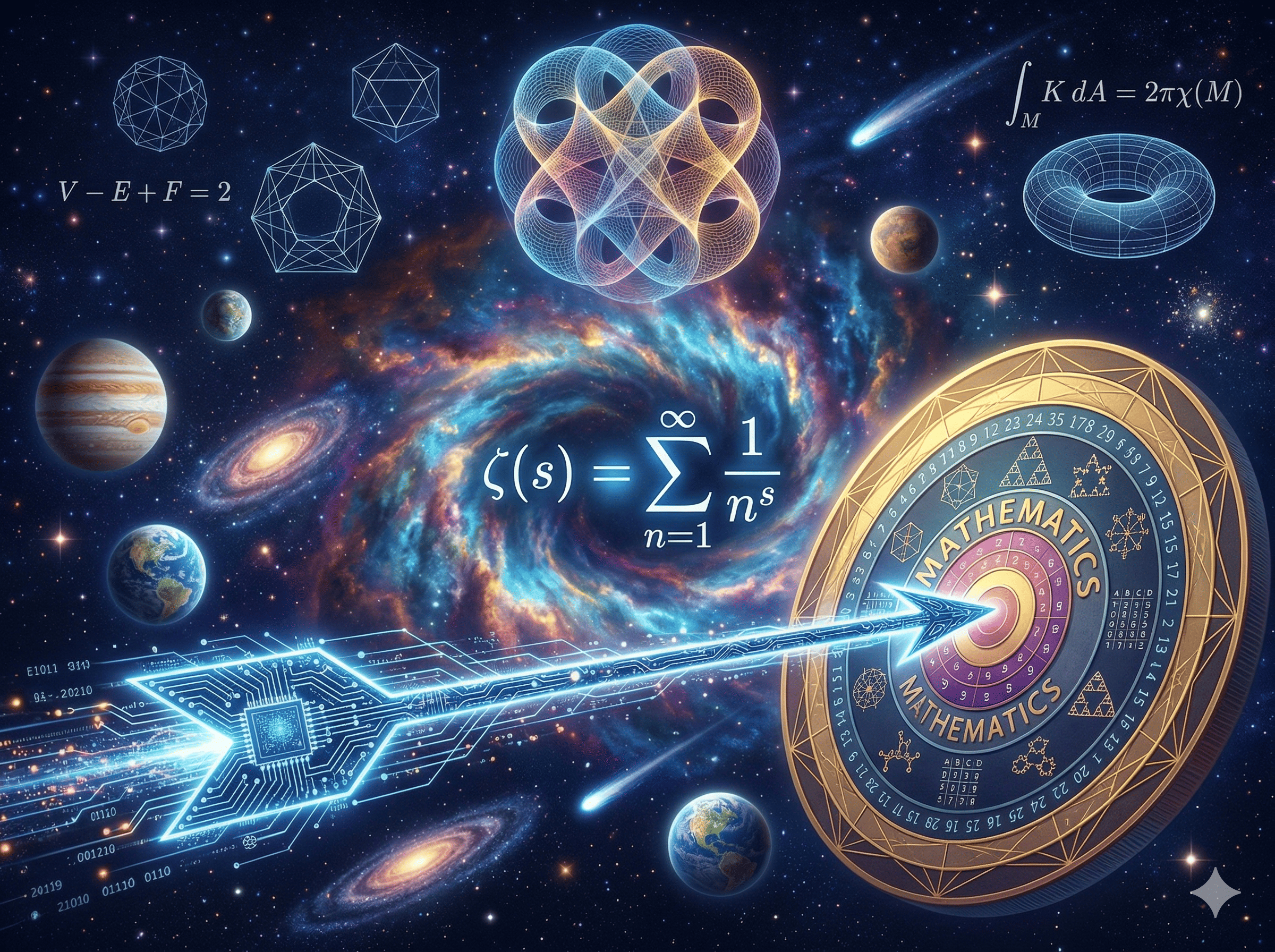}
\usetikzlibrary{arrows.meta}
\usepackage{array}

\definecolor{customcolor}{RGB}{32,178,170}
\colorlet{coverlinecolor}{customcolor}
\usepackage{cprotect}
\usepackage[ruled,vlined,linesnumbered]{algorithm2e}
\DontPrintSemicolon
\usepackage{amsmath, amssymb}
\counterwithin{algocf}{chapter}

\begin{document}
	
	\maketitle
	\frontmatter
	
	\tableofcontents
	
	\mainmatter
	\begin{refsection}[ref1.bib]
		\chapter{Overview}
		For a long time, mathematics has been regarded as the pinnacle of pure human intellect. Mathematicians rely on insight, intuition, and rigorous logic to prove conjectures and create theories in an abstract world. Currently, artificial intelligence, especially data-driven machine learning technology, has made breakthrough progress. A natural question arises: How will the combination of artificial intelligence and mathematics reshape our way of exploring the mathematical world?
		
		This book aims to systematically organize and answer this question. We will take mathematical research as the main thread to explain the significant value of artificial intelligence's involvement. This is not a simple additive combination of ``how A helps B'' in a tool-like manner, but rather an exploration of a deeper structural question: When the complexity and scale of mathematical problems reach a certain level, what inherent limitations do the traditional paradigms of discovery and proof face? What new conceptual tools and possible pathways do artificial intelligence methods provide for addressing these limitations? By constructing a clear cognitive map of the ``AI for Math'' field, we hope readers will understand that this is not merely a technological application, but a methodological transformation concerning ``how mathematics is researched.''
		
		\section{Core Elements of Mathematical Research}
		
		To understand how artificial intelligence can play a role, we first need to understand the nature of mathematical research work. Although the specific work of each mathematician varies greatly, its core activities can roughly be summarized into three interrelated yet distinct stages: discovery, proof, and refutation. These three form a cyclical, mutually reinforcing exploratory loop.
		\begin{enumerate}
			\item Discovering mathematical patterns: This is the source of mathematical creativity. Starting from concrete examples, observed patterns, or phenomena in the physical world, mathematicians propose new concepts, conjectures, or theories through intuition, analogy, and insight. For example, conjecturing a general formula after observing a pattern in a sequence of numbers, or proposing a proposition about invariants while studying geometric objects. The essence of this process is searching and filtering within a vast, potentially infinite space of possibilities, aiming to locate structures that are ``mathematically meaningful'' or ``likely true.'' Human intuition plays a central heuristic search role here, allowing mathematicians to quickly focus on promising ``regions,'' but it may also miss certain non-intuitive yet important patterns due to cognitive biases or the complexity of the problem.
			
			\item Proving theorems: Once a conjecture is proposed, the core value of mathematics—rigor—demands that we establish its truth. Proof is the process of transforming intuition and conjecture into irrefutable logical truth. It ensures the necessity of the conclusion given the axioms and definitions. The activity of proving itself can also be seen as a complex search: within the logical space constituted by axioms, theorems, and rules of inference, finding a path from known facts to the proposition to be proven. As mathematical theories become highly abstract and branches become more specialized, proofs become increasingly long, complex, and reliant on more lemmas and cross-disciplinary knowledge.
			
			\item Constructing counterexamples: Constructing counterexamples is often extremely challenging, as it requires us to precisely find a specific instance within a vast set of candidate objects. Human constructions often rely on ingenious induction or recursion techniques, but when constraints are complex or the search space is huge, constructing counterexamples typically becomes extremely difficult.
		\end{enumerate}
		
		\section{The Role of AI in Mathematical Research}
		
		\noindent\textcolor{structure3}{\textbf{1. From Automated Theorem Proving to Deep Learning}}
		\\
		1.1 The Development History of Automated Theorem Proving
		
		The history of automated theorem proving can be traced back to Leibniz's 17th-century conception of a ``calculus of reasoning,'' but significant breakthroughs were only realized in the 20th century with the development of computer technology. In the 1950s, AI pioneers like Allen Newell and others first attempted to use computer programs to prove mathematical theorems. In 1976, mathematicians, with the aid of a computer, completed the proof of the Four Color Theorem, a milestone event that garnered widespread attention in the mathematical community. The exhaustive case analysis required for the proof of the Four Color Theorem far exceeded the capacity of manual human processing. This successful case fully demonstrated the unique value of computers in solving extremely large-scale mathematical problems.
		
		The development of automated theorem proving has gone through several important stages: from early exhaustive proof methods, to Wu Wenjun's algebraic method, and later to automated reasoning and interactive proof systems. With the exponential increase in computer performance, the capabilities of automated theorem proving have also continuously strengthened. Formal verification methods represented by systems like Coq, Isabelle, and Lean can ensure absolute rigor in proofs, avoiding potential oversights present in traditional mathematical proofs.
		
		The value of automated theorem proving is reflected not only in improving the efficiency of mathematical research but also, more profoundly, in changing the way of thinking in mathematical research. It breaks through the limitations of human cognition, enabling the handling of extremely complex calculations and reasoning processes; simultaneously, it promotes the development of mathematical rigor, making mathematical proofs more reliable and verifiable. In the future, with the advancement of artificial intelligence technology, automated theorem proving will undoubtedly play an even more important role in mathematical research.
		\\
		1.2 The Deep Learning Revolution
		
		Deep learning, as an important branch of machine learning, has a development history traceable to the proposal of the neural network concept in the 1940s, but its explosive growth truly began after 2010. The core of this technological revolution lies in constructing neural network architectures with multiple layers of nonlinear transformations, endowing computers with unprecedented capabilities for feature learning and pattern recognition. The breakthrough performance of AlexNet in the 2012 ImageNet competition marked the official arrival of the deep learning era. Subsequently, with increased computing power, optimized algorithms, and the accumulation of big data, deep learning achieved a series of remarkable accomplishments in fields such as computer vision and natural language processing.
		
		In the field of mathematical research, the value of deep learning is mainly manifested in three key dimensions. First, in discovering new mathematical patterns, deep learning has shown astonishing potential. In a 2021 study published in \textit{Nature} by the DeepMind team, by analyzing invariants in knot theory, they discovered mathematical connections that human mathematicians had long failed to notice.
		
		In the verification of mathematical proofs, the combination of deep learning and formal proof systems has opened up new possibilities. By integrating neural networks with formal verification, AI systems can assist in completing complex proof processes. This ``neural-symbolic'' system retains strict logical rigor while possessing powerful mathematical reasoning capabilities.
		
		In constructing counterexamples, deep learning also performs exceptionally well. Through reinforcement learning algorithms, AI systems can efficiently search for potential counterexample structures within vast combinatorial spaces. In the field of graph theory, this method has successfully constructed counterexamples for several famous conjectures.
		
		The unique value of deep learning lies in its ability to break through the inherent limitations of human thinking. It can handle high-dimensional data spaces that are difficult for humans to manage, discover non-intuitive mathematical patterns, and provide new ideas in proof and counterexample construction. With the development of new technologies like neural-symbolic systems, deep learning is pushing mathematical research into a new era of human-machine collaboration. This technology not only improves research efficiency but, more importantly, expands human understanding of the essence of mathematics, injecting new vitality into this oldest discipline.
		\vspace{3mm}
		
		\noindent\textcolor{structure3}{\textbf{2. How Does AI Assist Mathematical Research?}}
		
		Mathematical research, as a systematic cognitive activity, mainly consists of three core stages: discovering new mathematical patterns, verifying the correctness of these patterns, and constructing counterexamples. In this subsection, we briefly summarize the capabilities currently demonstrated by AI in these three key areas:
		\\
		2.1 AI Discovers Mathematical Patterns
		
		In 2021, the DeepMind team published groundbreaking results in the journal \textit{Nature}. Their developed AI system could help mathematicians discover new connections between knot invariants. This research proved that AI can guide mathematicians' intuition to discover mathematical connections that humans might overlook.
		
		A knot is a closed curve in three-dimensional space, and mathematicians typically use invariants such as the Jones polynomial and hyperbolic volume to describe its properties. DeepMind encoded the algebraic and geometric invariants of knots as high-dimensional vectors and used a deep neural network (DNN) to learn the latent relationships between them, i.e., learning how to predict one invariant (e.g., hyperbolic volume) from another (e.g., the Jones polynomial). Based on the AI's predictions, mathematicians further constructed rigorous mathematical proofs, ultimately discovering new connections between invariants in knot theory.
		
		In another paper published in \textit{Nature}, DeepMind used deep reinforcement learning (DRL) to discover more efficient matrix multiplication algorithms, surpassing the best records held by human mathematicians in this field. This achievement demonstrates how AI can discover mathematical optimization strategies unthought of by humans through autonomous exploration. Matrix multiplication is a fundamental operation in computer science and mathematics, and traditional algorithms (like Strassen's algorithm) have been optimized for decades. For multiplying two \( n \times n \) matrices, the standard method requires \( O(n^3) \) scalar multiplications. DeepMind's goal was to find combinations requiring fewer multiplications, thereby reducing computational complexity. DeepMind employed AlphaTensor (an improved model based on AlphaZero), with core components including:
		
		\begin{itemize}
			\item State Representation
			\begin{itemize}
				\item Encoding the matrix multiplication problem as a three-dimensional tensor, where each element represents a possible multiplication combination.
				\item For example, \( 2 \times 2 \) matrix multiplication can be represented as a \( 4 \times 4 \times 4 \) tensor, where each dimension corresponds to elements of the input/output matrices.
			\end{itemize}
			\item Action Space:
			\begin{itemize}
				\item Each action corresponds to a basic multiplication operation (e.g., scalar multiply-add).
				\item The AI's goal is to decompose the tensor through a series of actions to find the minimal number of multiplication steps.
			\end{itemize}
			\item Reward Function:
			\begin{itemize}
				\item Primary optimization objective: Reduce the number of multiplications (i.e., the rank of the tensor decomposition).
				\item Additional reward: Discover structured patterns (e.g., symmetry) to facilitate generalization to larger matrices.
			\end{itemize}
		\end{itemize}
		By combining Monte Carlo Tree Search (MCTS) with deep learning, DeepMind ultimately discovered more efficient matrix multiplication algorithms.
		\\
		2.2 AI Automatically Proves Theorems
		
		Modern artificial intelligence systems have made significant progress in the field of mathematical proof, with their technical implementation primarily manifested in two important directions: solving Olympiad-level geometry problems and formal mathematical proof. These breakthrough developments demonstrate the powerful capabilities of AI in mathematical reasoning.
		
		In geometric proof, DeepMind's AlphaGeometry system represents the current state-of-the-art technology. The system adopts a neuro-symbolic hybrid architecture, combining the logical reasoning capabilities of a language model and a symbolic engine. Specifically, AlphaGeometry first analyzes the geometric diagram to generate potential auxiliary construction points (such as midpoints, foots of perpendiculars, etc.). Then, the symbolic reasoning engine incorporates these construction points into a deduction database, applying geometric axioms and theorems for rigorous logical derivation. The system was trained using over 100 million synthetic geometry problems generated by a random theorem generation algorithm, ensuring the diversity and complexity of the training data. Notably, in 2023 testing, AlphaGeometry solved 25 out of 30 IMO geometry problems.
		
		In the field of formal verification, interactive theorem provers like Lean play the core role of the verification kernel. Such systems, based on dependent type theory, can perform ultimate machine checking of mathematical reasoning—once a user or external tool provides complete proof steps, the Lean kernel verifies their logical correctness in an indisputable manner.
		
		To improve the efficiency of this verification process, AI technology has been introduced in recent years as a proof assistant to aid humans or automate the construction of proofs. The collaboration model is as follows:
		\begin{itemize}
			\item Formal Encoding and Semantic Understanding: First, informal mathematical propositions are transformed into strict, machine-readable formal statements (e.g., Lean code). AI (such as pre-trained large language models) understands this semantics to provide intelligent suggestions for the next proof strategies.
			\item Intelligent Search and Strategy Optimization: Faced with complex proof spaces, AI is used for heuristic search to predict which proof paths are more likely to succeed, thereby avoiding inefficient exhaustive attempts.
			\item Automated Filling and Final Verification: AI tools are responsible for filling in specific proof steps. Once these steps are fully submitted to the Lean kernel, the kernel performs strict formal verification, ensuring the entire proof process is flawless.
		\end{itemize}
		In this model, AI is responsible for ``exploration'' and ``conjecture,'' while Lean is responsible for ``verification'' and ``confirmation.'' The combination of the two retains the absolute reliability of formal verification while significantly improving the efficiency of constructing proofs.
		
		The core technical breakthroughs of these AI systems lie in: 1) combining the intuitive capabilities of neural networks with the strict reasoning of symbolic systems; 2) constructing large-scale mathematical problem datasets to train and validate system performance. Current research focus is expanding towards more complex mathematical fields, indicating that AI will become an indispensable intelligent assistant in mathematical research.
		\\
		2.3 AI Constructs Counterexamples
		
		In recent years, reinforcement learning algorithms have made breakthrough progress in the field of graph theory, particularly demonstrating unique advantages in constructing counterexamples to combinatorial mathematical conjectures. This method, through a carefully designed algorithmic framework, can efficiently search for potential counterexample structures within enormous combinatorial possibility spaces. Its core technologies mainly include:
		\begin{itemize}
		\item Problem Modeling and State Representation: First, the graph theory problem is transformed into a form suitable for machine learning. For graph theory conjectures, the state space is typically represented as the adjacency matrix or feature vector of a graph. For example, when constructing a counterexample, the system initializes a graph with $n$ vertices, where each possible edge is treated as an independent decision variable.
		
		\item Reinforcement Learning Framework Design: A reinforcement learning algorithm is employed, where:
		\begin{itemize}
			\item Action Space: Includes graph modification operations such as adding/deleting edges, changing vertex attributes.
			\item Reward Function: Carefully designed as a multi-objective optimization form, containing:
			\item Primary reward: The degree of violation of the target conjecture (e.g., the magnitude of violating a certain inequality).
			\item Auxiliary reward: Maintaining other properties of the graph (e.g., connectivity, regularity).
			\item Penalty term: Controlling the complexity of the graph (e.g., number of edges, vertices).
		\end{itemize}
		Through reinforcement learning, AI has successfully constructed counterexamples for several famous conjectures, demonstrating AI's unique advantages in exploring mathematical problems.
		\end{itemize}
		\vspace{3mm}
		
		\noindent\textcolor{structure3}{\textbf{3. The Mathematical Revolution in the Era of Large Models}}
		
		Mathematics, this ancient discipline exploring the truths of the universe, is undergoing a profound transformation brought about by artificial intelligence large models. From ChatGPT to AlphaGeometry, these AI systems are not only changing the way mathematical research is conducted but are also redefining the very nature of mathematical discovery. The core of this transformation lies in the fact that artificial intelligence is breaking through the limitations of human cognition, integrating mathematical knowledge, discovering hidden patterns, and creatively solving open problems in unprecedented ways. This shift not only improves the efficiency of mathematical research but, more importantly, expands the boundaries of mathematical exploration, injecting new vitality into this most rigorous science.
		
		In terms of knowledge integration, artificial intelligence large models are becoming ``super assistants'' that break down disciplinary barriers. A major challenge in traditional mathematical research is the high degree of specialization within the field. As mathematics develops, various subfields become increasingly profound. An expert in one area may know little about progress in adjacent fields, and this fragmentation of knowledge often hinders major breakthroughs. The famous mathematician Hilbert once noted: ``Mathematics is an organic whole, and its vitality stems precisely from unexpected connections between its various parts.'' However, in practice, discovering these connections often requires researchers to possess rare breadth of knowledge and extraordinary insight.
		
		Large language models, with their vast knowledge reserves and powerful associative capabilities, are changing this landscape. Large models represented by GPT can instantly draw upon knowledge from tens of thousands of mathematical papers, build cross-domain knowledge graphs, and discover deep connections between different branches. For example, when a researcher studies a difficult problem in algebraic geometry, the AI might suggest: ``This structure bears a striking resemblance to homology theory in topology''; or when studying a number theory problem, point out: ``This conjecture corresponds to the phase estimation algorithm in quantum computing.'' This cross-disciplinary associative ability, which in the past required mathematicians decades of extensive reading and deep thought to acquire, can now be achieved in a short time with AI assistance.
		
		DeepMind's FunSearch system vividly demonstrates the power of such knowledge integration. By cleverly combining the creative thinking of a large language model with the precise verification of evaluation code, the system discovered new cap set constructions in combinatorics, solving the open cap set problem that had remained unsolved for years. The key to this breakthrough lies in the AI's ability to freely establish connections between seemingly unrelated fields like discrete mathematics, algorithm design, and information theory, discovering associative patterns that human researchers might overlook. Here, AI plays the role not of a calculator, but more like a collaborator with an interdisciplinary perspective, capable of examining problems from completely different angles. The value of this knowledge integration is reflected not only in solving specific problems but, more importantly, in its potential to change the paradigm of mathematical research. First, AI assistance can significantly shorten mathematicians' ``learning curves,'' enabling researchers to quickly grasp the fundamentals of related fields; second, it can reveal ``hidden bridges'' between different mathematical branches, providing clues for new research directions; finally, this cross-domain knowledge fusion often catalyzes entirely new mathematical tools and methods, driving innovative development of the discipline.
		
		\vspace{3mm}
		
		\noindent\textcolor{structure3}{\textbf{4. Can AI Create New Mathematics?}}
		
		Creative thinking is the most dazzling jewel in the crown of human intelligence. From Euclid's axiomatic system to Gauss's differential geometry, from Riemann's complex functions to Grothendieck's scheme theory, every major breakthrough in the history of mathematics shines with the brilliance of creative thinking. What are the key elements of this thinking? And why is current artificial intelligence finding it so difficult to match?
		
		Genuine mathematical creative thinking encompasses at least three interrelated levels:
		\begin{enumerate}
			\item Conceptual abstraction ability is the cornerstone of mathematical creation. Excellent mathematicians can extract essential features from concrete problems to form new mathematical concepts. For instance, when Euler saw the Königsberg bridge problem, he did not dwell on the specific bridges and rivers but abstracted the basic concepts of topology. This leap from the concrete to the abstract requires profound intuitive insight.
			
			\item Analogical transfer ability allows mathematical ideas to flow between different fields. In the 19th century, Riemann transferred ideas from Gauss's theory of surfaces to the study of complex functions, pioneering Riemann surface theory. This cross-domain associative ability relies on a deep understanding of the essence of mathematics, not a simple correspondence of surface features.
			
			\item Constructing entirely new mathematical theories is the highest level of mathematical creation. Grothendieck's scheme theory in algebraic geometry not only solved specific problems but also built an entirely new theoretical framework. This systematic creation requires a combination of a macroscopic mathematical vision and rigorous logical thinking.
		\end{enumerate}
		The essence of mathematical creative thinking lies in perfectly combining intuitive leaps with logical rigor. Current AI cannot yet replicate the ``flash of inspiration'' creative process of human mathematicians. Studying mathematical creative thinking is not only relevant to the development of mathematics itself but also an important avenue for advancing artificial intelligence to higher levels.
		
		In summary, from Wu Wenjun's automated theorem proving to today's AlphaGeometry, the integration of AI and mathematics is creating new research paradigms. Although AI currently cannot fully replace the creativity and insight of mathematicians, it has become an indispensable research partner. In the future, with technological progress, we may witness AI proposing entirely new mathematical theories, opening a new era in mathematical research. In this new era of human-machine collaboration, the development of mathematics will no longer be limited by individual intelligence but will be jointly propelled by humans and AI. This transformation will not only change the way mathematical research is conducted but may also help us uncover deeper mathematical mysteries of the universe.
		
		\section{Core Idea: Mathematical Problems as the Main Thread}
		
		Throughout this book, we will consistently adhere to a core narrative principle: starting from the mathematical problem itself, we examine how AI technology serves as a new methodology to solve these problems. This means the logical chain of our thinking is:
		\begin{center}
			\textcolor{structure3}{\textbf{Mathematical Problem $\to$ Suitable AI Technology $\to$ Specific Implementation Process}}
		\end{center}
		For example, when facing the deep mathematical problem of ``understanding the relationship between geometric invariants and algebraic invariants of knots'', the challenge lies in learning the connections between invariants from unstructured data. This guides us to use neural networks for learning and to distill mathematical insights.
		
		This problem-oriented perspective helps us avoid falling into blind worship of AI technology or the mere piling up of tools. Instead, we consistently focus on: what new, essential help does this technology bring to our understanding of a particular mathematical structure?
		
		\section{Controversies and Reflections: New Issues in Mathematical Research in the AI Era}
		
		In the discussions of the preceding three parts, we systematically reviewed the core elements of mathematical research, the multiple roles AI plays in mathematical research, and the core idea of taking mathematical problems as the main thread. So far, we have seen the enormous potential brought by the deep integration of AI and mathematics—from assisting in conjecture generation to accelerating theorem proving, from discovering conceptual metaphors to exploring through computational experiments.
		
		However, just as every technological revolution brings profound social and academic changes, the entry of AI into the field of mathematics has also triggered a series of controversies worthy of deep thought. These issues are not simple technical challenges; they concern the most fundamental definition of mathematics as a discipline—who is the creator? How is trust established? What capabilities should future mathematicians possess?
		
		Following this line of thought, the first issue that emerges is the blurring boundary from ``assistive tool'' to ``collaborator.'' When AI systems are not merely performing calculations but can independently propose core conjectures and discover key proof steps, we are compelled to re-examine the attribution and authorship of mathematical discoveries. If the core insight of a paper is generated by AI, how should the human authors attribute it? Should it be considered a tool and not credited, or should the AI's status as a ``co-author'' be acknowledged? Furthermore, if AI generates new results based on a vast corpus of existing human knowledge, how should the intellectual property rights be assigned? Although these questions have not yet reached a general consensus in academia, as AI's role in mathematical research deepens, they are gradually moving from theoretical discussion to practical concern.
		
		Closely related to the issue of attribution is the need to reconstruct the system of verification and trust. Mathematics, as the most rigorous discipline, builds its edifice of knowledge on the cornerstone that ``proofs must be correct.'' Traditionally, the correctness of a paper relies on peer review—a few experts reading and endorsing its logical derivation. But when the proof process involves complex AI models, massive computational experiments, or even internal representations difficult for humans to directly understand, the traditional review mechanism reveals its limitations. The mathematical community needs to consider: what level of AI involvement is acceptable? To ensure the reproducibility of results, what materials do authors need to disclose—only the final results, or also including training code, data, model weights, and even hyperparameter configurations? How does one distinguish between AI's ``effective inspiration'' and ``potential errors''? These questions are sparking initial discussions in academia, and their future direction will profoundly affect the foundation of trust in mathematical research.
		
		Finally, the most far-reaching impact is reflected in the field of mathematical education. When AI can solve more and more standard mathematical problems, and even demonstrate capabilities surpassing humans in certain areas, we cannot help but ask: where should the focus of mathematics education shift? Should we continue to train students in techniques that machines can already execute perfectly, or should we shift towards cultivating abilities that AI still lacks? Critical thinking, the ability to formulate problems stemming from an understanding of the essence of mathematics, cross-domain metaphorical association, and the judgment to verify the reasonableness of AI outputs—the importance of these abilities will become increasingly prominent. Future mathematics education may no longer be about ``imparting known solutions,'' but about ``cultivating the ability to solve problems in collaboration with AI''—this requires us to rethink curriculum design, assessment methods, and even the definition of mathematical literacy.
		
		Faced with these controversies, we need to avoid two extreme attitudes: neither blindly embracing them while ignoring potential risks, nor rejecting change due to fear of difficulties. Just as the invention of the telescope did not replace astronomers but greatly expanded human vision, the goal of AI should also be to enhance mathematicians' capabilities, not to replace mathematicians themselves. The key lies in establishing a framework that can fully leverage the potential of AI while maintaining academic rigor and humanistic values.
		
		We stand at an exciting crossroads. Mathematics, the oldest and most rigorous discipline continuing from ancient Greek times, is undergoing an unprecedented deep integration with artificial intelligence. This integration will not only change the tools and methods of mathematical research but will also reshape the epistemological foundation of mathematical discovery—what is proof? What is understanding? What is creation? In the following chapters, we will explore these questions together, witnessing and participating in this transformation that profoundly affects the future face of mathematics.
		
		\nocite{*}
		
		\printbibliography[heading=subbibliography,title=References]
		
	\end{refsection}
\begin{refsection}[ref2.bib]
\chapter{Introduction to Machine Learning}
\section{Overview of Machine Learning}

In the previous chapter, we outlined the broad landscape of the emerging interdisciplinary field ``AI for Math.'' Its core foundation stems from a set of methodologies known as machine learning. Through mathematical tools, machine learning establishes a computational framework for automatically learning patterns from data and making predictions or decisions about the future based on these patterns. Understanding the mathematical essence of this framework is a prerequisite for mastering all subsequent content in this book.

The core idea of machine learning is to automatically learn an ``optimal'' mapping from an input space to an output space from data. This artificial intelligence approach of ``automatically optimizing a mapping based on data'' opens the door to using computational power to handle complex patterns. This is precisely the starting point for artificial intelligence to intervene in mathematical research: finding a good mapping that can learn the intrinsic laws of mathematics from existing mathematical objects and data.

Any machine learning task can be decomposed into three fundamental components: data, model, and algorithm. Together, they define a specific and computable mathematical optimization problem.
\begin{enumerate}
	\item Data
	
	Data is the raw material for learning, typically given as a set of samples: $\mathcal{D} = \{ { (x^{1}, y^{1}),  (x^{2}, y^{2}),  \dots, (x^{N}, y^{N}) }\}$. Here, $ N $ is the number of samples.
	Here $x^{i}$ is the input data, and $y^{i}$ is the output data.
	In practical problems, data samples are usually represented by vectors, responsible for transforming real-world problems (such as images, text, mathematical formulas) into a numerical form that algorithms can process.
	
	\item Model
	
	The model $ f(x; \theta)$ is the mapping we use to approximate the true data distribution, where $\theta$ are the parameters to be optimized. Common mathematical models include:
	\begin{itemize}
		\item Linear models: The simplest example is $ f(x; w, b) = w^T x + b $, where $w, b$ are unknown parameters. This type of model is simple in form, easy to interpret, and suitable for situations where the data has a strong linear relationship. For example:
		
		\begin{itemize}
			\item House price prediction: Predicting price ($y$) based on floor area ($x$), assuming a fixed price per square meter, then the total price is approximately linearly related to the area.
			\item Academic performance analysis: Studying the relationship between study time ($x$) and exam score ($y$). Within a reasonable range, longer study time may lead to a linear improvement in scores.
			\item Advertising effectiveness evaluation: Analyzing whether there is a stable linear growth trend between advertising expenditure ($x$) and product sales ($y$).
		\end{itemize}

		\item Nonlinear models: When the data relationship is more complex and a simple straight line cannot describe its variation pattern, nonlinear models need to be introduced. For example, polynomial models: $$f(x; w) = w_0 + w_1 x + w_2 x^2 + ... + w_M x^M ,$$ where \(w_0, w_1, \dots, w_M\) are the coefficients to be determined. It can approximate very complex functions and is a powerful tool for handling high-dimensional, unstructured mathematical objects. For example: Free-falling object: The relationship between the distance fallen ($y$) and time ($x$) is \(y \propto x^2\), requiring a quadratic polynomial for description.
		
	\end{itemize}
	
	\item Algorithm
	
	The algorithm is responsible for automatically finding the optimal parameters $ \theta$ within the predefined mathematical model $ f(x; \theta)$ based on the data $ \mathcal{D}$. This leads to the next core question: How do we define ``optimal''?
	
	We want the model to perform well on all data it might encounter. In a probabilistic framework, assuming the data is generated by an unknown true distribution $ p(x, y)$, optimality is defined as minimizing the Expected Risk or expected loss:
	$$
	\mathcal{R}(\theta) = \mathbb{E}_{(x, y) \sim p(x, y)} \mathcal{L}(y, f(x; \theta))
	$$
	where $\mathcal{L}(y, \hat{y} )$ is called the loss function, measuring the discrepancy between the predicted output $ \hat{y} = f(x; \theta ) $ and the true output $ y $. Therefore, the ideal objective is
	$$\theta^* = \arg \min_{\theta} \mathcal{R}(\theta).$$
	
	However, the true distribution $ p(x, y) $ is usually unknown. The only information we have is a sample from that distribution: the training set $\mathcal{D}$. A natural and practical approximation is to replace the expected risk with the Empirical Risk:
	$$
	\mathcal{R}^{emp}_{\mathcal{D}}(\theta) = \frac{1}{N} \sum_{n=1}^{N} \mathcal{L}(y^{(n)}, f(x^{(n)}; \theta))
	$$
	This leads to the core principle of machine learning—Empirical Risk Minimization (ERM):
	$$
	\theta^* = \arg \min_{\theta} \mathcal{R}^{emp}_{\mathcal{D}}(\theta)
	$$
	ERM transforms the learning problem into a pure mathematical optimization problem. The underlying statistical belief is that by optimizing the model's performance on the training set, we expect the model to also perform well on unseen data, i.e., to possess generalization ability. Overfitting is precisely the core challenge faced by the ERM principle with limited samples: the model may overly ``cater'' to specific samples (including noise) in the training data, harming generalization performance.
\end{enumerate}

In summary, data provides the material for learning, the model defines the form of expression, and the algorithm indicates the path for optimization. These three elements together constitute the basic framework of machine learning. However, the framework itself does not dictate the specific way of learning—just as having paper, pen, and writing rules, we still need to know what type of text to write. This leads to a higher-level distinction: based on the form of data and learning objectives, machine learning can be divided into three basic paradigms, which determine how the three elements are combined to solve problems.

\begin{enumerate}
	\item Supervised Learning
	
	The most distinctive feature of supervised learning is that the data consists of labeled input-output pairs $ (x, y) $, and the goal is to learn the best mapping $ f: \mathcal{X} \rightarrow \mathcal{Y} $ from the input space to the output space. In supervised learning, the data samples we collect contain both the real input and the labeled true output. This is a distinctive feature of supervised learning. Based on the type of output space, it is mainly divided into:
	\begin{itemize}
		\item Regression problems: $\mathcal{Y} \subseteq \mathbb{R}$, meaning the output samples are continuously varying numerical values. For example, predicting tomorrow's temperature based on historical weather data, or predicting the integral value of a function based on its values at a finite number of points.
		
		\item Classification problems: $ \mathcal{Y} = \{1, 2, ..., C\}$, meaning the output samples are discrete values. For example, determining whether an email is spam, or classifying a mathematical proposition as true or false.
	\end{itemize}
	
	\item Unsupervised Learning
	
	In unsupervised learning, we only have input data $ \{ x^{(1)}, ..., x^{(N)} \}$, without labeled true output data $y$. Since the true output $y$ is not labeled in advance, we do not know the values of $y$ beforehand. This is a significant difference between unsupervised learning and supervised learning. The goal of unsupervised learning is to discover hidden structures, patterns, or distributions in the data, for example:
	\begin{itemize}
		\item Clustering: Partitioning data into several clusters such that samples within the same cluster are similar to each other, while samples from different clusters are dissimilar. This is equivalent to discovering natural ``groupings'' or ``categories'' in the data space without predefining what these categories are.
		
		\item Dimensionality reduction: Projecting high-dimensional data into a low-dimensional space while preserving as much important information (such as variance, manifold structure) as possible. This aids in visualization, denoising, and reducing computational complexity for subsequent tasks.
		
		\item Density estimation: Generating the underlying probability distribution of the observed data.
	\end{itemize}
	
	\item Reinforcement Learning (RL)
	
	Reinforcement learning deals with an agent learning in an environment, changing the environment state by executing actions, and receiving reward signals from the environment. The goal is to learn a policy model (a mapping from states to actions) to maximize long-term cumulative reward.
	
	The training data in reinforcement learning comes from the interaction between the agent and the environment and does not need to be collected in advance. This is a significant difference between reinforcement learning and other learning paradigms. In reinforcement learning, the reward function is one of the key elements used to optimize the parameters in the policy model (in supervised and unsupervised learning, a loss function is typically used to optimize the parameters in the prediction model).
\end{enumerate}

\section{Linear Models}

This section briefly describes linear models in machine learning. An important class of nonlinear models—artificial neural networks—will be discussed in the next section.

\subsection{Linear Models in Supervised Learning}
\noindent 1. Linear Regression Model

Linear regression is the most fundamental regression model. Its core assumption is that the output variable $y$ can be expressed as a linear combination of the input feature vector $x$:
$$y = w^T x + b.$$
This model uses the squared loss function: $\mathcal{L}(y, \hat{y}) = (y - \hat{y})^2$, and its empirical risk is defined as:
\[ \mathcal{R}_{\text{emp}}(w, b) = \frac{1}{2} \sum_{n=1}^N \left( y^{(n)} - (w^T x^{(n)} + b) \right)^2 \]
The optimal parameters $w, b$ that minimize this empirical risk function can be solved using the Lagrangian multiplier method. Iterative optimization algorithms such as gradient descent can also be used for solving (refer to the subsequent material in this section for gradient descent).
\\
2. Classification Model: Perceptron

The perceptron is an important classification model. From a geometric perspective, it automatically finds a linear hyperplane that can correctly separate two classes of data through iterative learning. It is mathematically defined as follows:

A perceptron is a function \( f: \mathbb{R}^d \rightarrow \{-1, +1\} \) that maps a \(d\)-dimensional real-valued input vector \( \mathbf{x} = [x_1, x_2, ..., x_d]^T \) to a binary output (typically +1 and -1, representing two classes).

Its computation consists of two steps:

(1). Weighted Sum (Net Input): Compute the linear combination of input features.
\[
z = \sum_{i=1}^{d} w_i x_i + b = \mathbf{w}^T \mathbf{x} + b
\]
Here, \( \mathbf{w} = [w_1, w_2, ..., w_d]^T \) is called the weight vector, and \( b \) is called the bias. The weight \( w_i \) measures the importance of the corresponding feature \( x_i \) for the decision, and the bias \( b \) determines the offset of the decision plane relative to the origin.

(2). Activation Function: Apply a step function (or Heaviside function) to the net input \( z \).
\[
f(\mathbf{x}) = \text{sign}(z) =
\begin{cases}
	+1, & \text{if } z \ge 0 \\
	-1, & \text{if } z < 0
\end{cases}
\]

Interpreting the perceptron from a geometric perspective, it essentially seeks a decision hyperplane in the sample space. The perceptron's decision rule \( \mathbf{w}^T \mathbf{x} + b = 0 \) defines a decision hyperplane in the feature space, where the parameters have the following meanings:

\begin{itemize}
	\item Weight vector \( \mathbf{w} \): The normal vector perpendicular to this decision hyperplane, indicating the positive direction of classification.
	\item Bias \( b \): Determines the distance from the hyperplane to the origin. When \( b=0 \), the hyperplane passes through the origin.
	\item Decision: For any data point \( \mathbf{x} \), if it lies on the side pointed to by the normal vector \( \mathbf{w} \) (i.e., \( \mathbf{w}^T \mathbf{x} + b > 0 \)), it is classified as +1; otherwise, it is classified as -1.
\end{itemize}

Therefore, the learning objective of the perceptron is to find such a hyperplane that can correctly separate all positive samples (\( y=+1 \)) and negative samples (\( y=-1 \)) on opposite sides.

Learning Algorithm: Weight Update Rule

The perceptron employs an error-driven learning algorithm. Given a labeled training dataset \( \{ (\mathbf{x}^{(1)}, y^{(1)}), (\mathbf{x}^{(2)}, y^{(2)}), ... \} \), the algorithm proceeds as follows:
\begin{enumerate}
	\item  Initialization: Initialize the weight \( \mathbf{w} \) and bias \( b \) to zero or small random numbers.
	\item  Iteration: For each sample \( (\mathbf{x}, y) \) in the training set (or in random order):
	\begin{enumerate}
		\item Compute the prediction: \( \hat{y} = \text{sign}(\mathbf{w}^T \mathbf{x} + b) \).
		\item Update weights (if and only if the prediction is incorrect):
		\[
		\begin{aligned}
			\mathbf{w} & \leftarrow \mathbf{w} + \eta \cdot (y - \hat{y}) \cdot \mathbf{x} \\
			b & \leftarrow b + \eta \cdot (y - \hat{y})
		\end{aligned}
		\]
		Here, \( \eta > 0 \) is a crucial hyperparameter called the learning rate, which controls the step size of each update.
	\end{enumerate}
\end{enumerate}

After understanding the basic form of the perceptron, let's delve into the geometric intuition behind its update rule—why can such a seemingly simple rule gradually correct classification errors?

When a sample is misclassified, the predicted value \(\hat{y}\) must be inconsistent with the true label \(y\). In this case, the value of \((y - \hat{y})\) is \(+2\) or \(-2\) (considering \(\hat{y} = \text{sign}(w^T x + b)\) takes values \(\pm 1\)). To understand the geometric meaning of parameter adjustment more clearly, let's examine the two misclassification scenarios separately:

Scenario One: True label \(y = +1\), but model predicts \(\hat{y} = -1\)

This means the current linear output from the model is \(w^T x + b < 0\), i.e., the sample point lies on the wrong side of the decision plane.

The update rule is: \(w \leftarrow w + \eta \cdot x\), \(b \leftarrow b + \eta \cdot (1)\)

Geometric interpretation: This update effectively ``pushes'' the entire decision plane away from the current sample point. To understand this, we need to observe the change in the sample point's position relative to the new plane after the update.

Before the update, we have \(w^T x + b < 0\). After the update, the new linear output is:
\[
(w + \eta x)^T x + (b + \eta) = w^T x + \eta \|x\|^2 + b + \eta = (w^T x + b) + \eta (\|x\|^2 + 1)
\]
Since \(\|x\|^2 + 1 > 0\) and \(\eta > 0\), the new output value adds a positive term to the original negative value. This means the sample point's ``score'' relative to the new plane increases—although it may still be negative, it has moved a step towards the positive direction. If the adjustment magnitude of this step is large enough, the sample point may cross the decision plane and enter the positive half-space.

From the perspective of the plane: The direction of the weight vector \(w\) determines the normal direction of the plane, and the bias \(b\) determines the plane's position along the normal. When we increase the component of \(w\) in the direction of \(x\), it is equivalent to rotating the plane around some axis, making its normal direction more inclined towards \(x\); simultaneously increasing \(b\) is equivalent to translating the plane along the normal direction. The combined effect is to move the plane away from \(x\), attempting to place \(x\) on the correct side.

Scenario Two: True label \(y = -1\), but model predicts \(\hat{y} = +1\)

In this case, \(w^T x + b > 0\), and the sample point also lies on the wrong side of the decision plane, but in the opposite direction.

The update rule is: \(w \leftarrow w - \eta \cdot x\), \(b \leftarrow b + \eta \cdot (-1) = b - \eta\)

Geometric interpretation: This update also pushes the decision plane away from the current sample point, but in the opposite direction to Scenario One.

Before the update, \(w^T x + b > 0\). After the update, the new linear output is:
\[
(w - \eta x)^T x + (b - \eta) = w^T x - \eta \|x\|^2 + b - \eta = (w^T x + b) - \eta (\|x\|^2 + 1)
\]
The new output value subtracts a positive term from the original positive value, decreasing the sample point's ``score''. This means the sample point moves towards the negative direction, gradually approaching or even crossing the decision plane into the negative half-space.

From the plane's perspective: Subtracting the component in the \(x\) direction is equivalent to making the plane's normal \(w\) deviate from the direction of \(x\), while decreasing the bias \(b\) is equivalent to translating the plane along the negative normal direction. These two adjustments work together to move the plane away from the current sample point, attempting to place it in the negative half-space.

Understanding the above two updates, we can visualize the dynamic changes of the decision plane during the perceptron's learning process:
\\
Each time a misclassified sample is encountered, the algorithm applies a local adjustment to the decision plane based on the sample's position and the direction of misclassification. This adjustment always moves the currently misclassified sample towards the correct side—specifically manifested as the sample's linear output value relative to the new plane moving closer to zero until it crosses the boundary. However, this adjustment comes at the cost of sacrificing the classification status of other samples: some originally correctly classified samples may be pushed back to the wrong side. This is precisely the characteristic of the perceptron algorithm—it does not seek a one-step global optimum but rather achieves a balanced state through continuous trial and error and repeated corrections: all samples lie on the correct side of the decision plane, provided the data is linearly separable.

From a geometric intuition, the entire learning process is like a tilted plate (the decision plane) constantly flipping and translating among data points, slightly lifting one end each time it encounters a point that is pressed incorrectly, until all points lie on the correct side of the plate or the iteration limit is reached.

Can this repeated adjustment process eventually stop? The Perceptron Convergence Theorem [Novikoff, 1963] gives an affirmative answer: if the training data is linearly separable, i.e., there exists some hyperplane that can perfectly separate the two classes of samples, then regardless of the initial plane, the above update rule guarantees convergence to a solution that correctly classifies all training samples within a finite number of iterations.

However, the convergence theorem only guarantees the ``existence of a solution'' and does not promise that this solution is ``optimal.'' In fact, the decision plane finally found by the perceptron heavily depends on the order in which samples appear and the initial parameters. It may happen to lie in the marginal zone between the two classes—extremely close to some sample points, even tightly adhering to the boundary of one class. Although such a solution performs perfectly on the training set, it often performs poorly on unseen test data due to a lack of robustness. This limitation later gave rise to the birth of the support vector machine (SVM): SVM not only seeks a hyperplane that can classify but also pursues the one with the ``maximum margin,'' thereby mathematically guaranteeing stronger generalization ability. This difference between the perceptron and SVM precisely reflects the evolution of machine learning thought from ``finding a feasible solution'' to ``finding an optimal solution.''
\\
3. Classification Model: Support Vector Machine (SVM)

The perceptron provides us with a method to find a classification hyperplane, but it only guarantees finding ``some'' plane that can separate the two classes of samples, without caring whether this plane is ``good.'' When there are infinitely many possible classification hyperplanes, we naturally ask: Which one is the best?

\begin{figure}[htbp]
	\centering
	\includegraphics[width=0.5\linewidth]{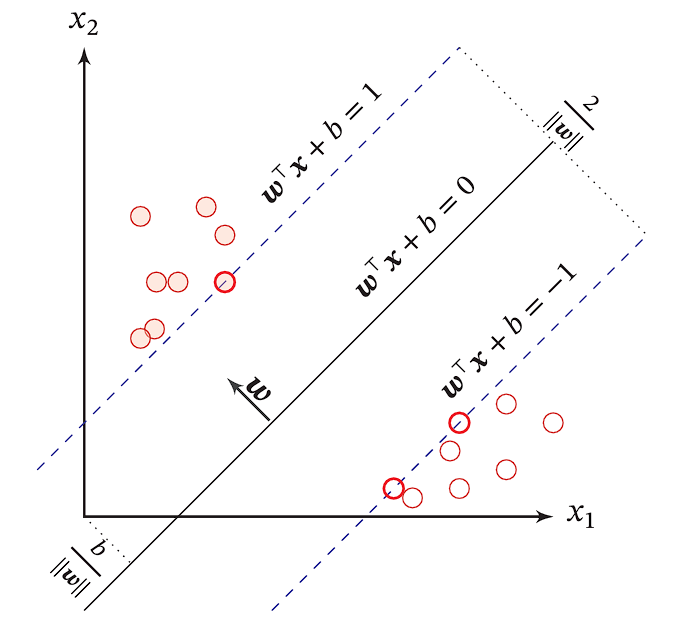}
	\caption{Support Vector Machine\label{fig:svm}}
\end{figure}

The idea of maximum margin. From a geometric perspective, the support vector machine gives a clear answer: the best classification hyperplane should be the one that is as far away as possible from both classes of samples. This ``distance'' is measured by the margin—defined as the distance from the hyperplane to the nearest sample point. The intuition behind maximizing the margin is that such a hyperplane has the strongest robustness: when new samples fluctuate slightly near the boundary, they are less likely to be misclassified, potentially leading to optimal generalization ability. It is worth noting that for linearly separable data, the hyperplane that maximizes the margin is unique, as shown in Figure\ref{fig:svm} (adapted from Qiu Xipeng's ``Neural Networks and Deep Learning'' Figure 3.6).

\noindent (1). Linearly Separable Case: Hard-Margin SVM

When the data is linearly separable, the SVM optimization problem can be formulated as:
\[ \begin{aligned} & \min_{w, b} \frac{1}{2} \| w \|^2 \\ & \text{s.t.} \quad y^{(n)}(w^T x^{(n)} + b) \geq 1, \quad \forall n = 1,\dots,N \end{aligned} \]
Here, we need to explain where the ``1'' in the constraint comes from. We know that for any hyperplane \(w^T x + b = 0\) that can correctly classify the two classes of samples, we can always scale \(w\) and \(b\) proportionally so that \(|w^T x + b| \geq 1\) holds for all samples, and equality holds at least at the points closest to the plane. This ``1'' essentially normalizes the functional margin, eliminating the uncertainty caused by parameter scaling and giving the optimization problem a definite form. At this point, the margin is exactly \(1/\|w\|\), so minimizing \(\|w\|^2\) is equivalent to maximizing the margin.

This is a convex quadratic programming problem—the objective function is quadratic, and the constraints are linear—so there exists a unique global optimal solution. The decision function obtained after solving is \(f(x) = \text{sign}(w^T x + b)\), where only those samples on the margin boundary (i.e., points satisfying \(y^{(n)}(w^T x^{(n)} + b) = 1\)) contribute to the final hyperplane; they are called support vectors.

\noindent (2). Linearly Inseparable Case: Soft-Margin SVM

Real-world data is often not perfectly linearly separable; there may be noise, overlap, or even individual outliers. If we insist on using the above ``hard-margin'' constraint, the problem will have no solution. For this reason, we need to allow the model to make mistakes on some samples—this is the starting point of soft-margin SVM.

Introduce non-negative slack variables \(\xi_n \geq 0\), each corresponding to a sample \(\xi_n\), which measures the degree to which that sample violates the constraint. Specifically, we relax the original strict constraint to:

\[ y^{(n)}(w^T x^{(n)} + b) \geq 1 - \xi_n, \quad \forall n = 1,\dots,N \]

\begin{itemize}
	\item When \(\xi_n = 0\), the sample lies on the correct side and satisfies the margin requirement;
	\item When \(0 < \xi_n < 1\), the sample is on the correct side but lies inside the margin (i.e., between the decision plane and the margin boundary);
	\item When \(\xi_n \geq 1\), the sample is misclassified (lies on the wrong side of the decision plane).
\end{itemize}
The introduction of slack variables gives the model ``tolerance space,'' but we need to control the total error. Thus, the optimization objective becomes seeking a balance between maximizing the margin and minimizing the total error:

\[ \min_{w,b,\xi} \frac{1}{2}\|w\|^2 + C \sum_{n=1}^{N} \xi_n \]

Here, \(\sum \xi_n\) is the total penalty for all samples' constraint violations, and the penalty coefficient \(C > 0\) is a hyperparameter that needs to be manually set, regulating the weight between the two objectives:
\begin{itemize}
	\item A larger \(C\) means low tolerance for misclassification; the model will try its best to classify each sample correctly, even if this means a smaller margin;
	\item A smaller \(C\) allows more misclassification in exchange for a larger margin, thereby obtaining stronger generalization ability.
\end{itemize}

The choice of \(C\) is crucial—it essentially controls the model's sensitivity to noise and is the key knob for balancing fitting ability and generalization ability.
\vspace{3mm}

\noindent\textcolor{structure3}{\textbf{Kernel Trick: From Linear to Nonlinear}}

The idea of maximum margin is elegant, but it is still limited to linear classifiers. How can we make SVM handle nonlinearly separable data? Intuition tells us that if data cannot be linearly separated in the original feature space, perhaps we can map it to a higher-dimensional space through some transformation, making them linearly separable there.

This is precisely the core idea of the kernel trick. Consider a nonlinear mapping \(\phi: \mathcal{X} \rightarrow \mathcal{F}\) that maps data points from the original space to a high-dimensional feature space \(\mathcal{F}\). In this new space, we can perform standard linear SVM. However, directly computing \(\phi(x)\) and performing inner products in the high-dimensional space is often computationally expensive and may even face the curse of dimensionality.

The key insight of the kernel trick is that in both the SVM optimization problem and the final decision function, data appears in the form of inner products—i.e., the form \(x_i^T x_j\). If we can find a function \(K(x_i, x_j)\) that directly equals the inner product in the high-dimensional space \(\langle \phi(x_i), \phi(x_j) \rangle\), then we do not need to explicitly compute \(\phi(x)\) or know the specific form of the mapping; we only need to compute this kernel function. This not only avoids the computational burden caused by high dimensionality but even allows us to implicitly work in infinite-dimensional spaces.
\\
Commonly used kernel functions include:

\begin{itemize}
	\item Linear kernel: \(K(x,z) = x^T z\), degenerating to ordinary linear SVM;
	\item Polynomial kernel: \(K(x,z) = (x^T z + c)^d\), introducing polynomial combination features;
	\item Gaussian Radial Basis Function (RBF) kernel: \(K(x,z) = \exp(-\gamma \|x-z\|^2)\), capable of approximating any nonlinear function, one of the most commonly used kernel functions, where the parameter \(\gamma\) controls the influence range of a single sample.
\end{itemize}
After introducing the kernel trick, the SVM decision function becomes:

\[ f(x) = \text{sign}\left( \sum_{i \in SV} \alpha_i y_i K(x_i, x) + b \right) \]

where \(\alpha_i\) are the Lagrange multipliers obtained by solving the dual problem, and only those samples with \(\alpha_i > 0\) (i.e., support vectors) influence the decision outcome. This reflects an elegant property of SVM: the solution has sparsity, and the final model is determined only by a few key samples.

From the perceptron to hard-margin SVM, then to soft-margin SVM and the kernel trick, we have witnessed an important leap in machine learning thought: from ``finding a feasible solution'' to ``finding an optimal solution,'' from linear to nonlinear, from parametric models to sample-based kernel methods. With its elegant mathematical form and solid theoretical foundation, the support vector machine has become a milestone in the history of machine learning development.

\subsection{Linear Models in Unsupervised Learning: Principal Component Analysis (PCA)}

Unlike supervised learning, unsupervised learning deals with data without labels—we only have inputs \(x\), but no corresponding outputs \(y\) to tell the model what to learn. So, in the absence of a ``standard answer,'' what can the model learn? The answer is: the structure, patterns, or distribution inherent in the data itself. The goal of unsupervised learning is for the model to autonomously discover this hidden information.

Unsupervised learning encompasses many different tasks, with the two most representative categories being dimensionality reduction and clustering. Dimensionality reduction attempts to compress high-dimensional data into a low-dimensional space while preserving key information; clustering attempts to automatically group similar samples together. Here, we focus on one of the most classic methods in dimensionality reduction: Principal Component Analysis.

In real-world problems, we often face high-dimensional data—an image may have thousands of pixels, a text may have tens of thousands of vocabulary words. High-dimensional data is not only difficult to visualize but also imposes a significant computational burden. More importantly, it often contains a large amount of redundant information. The goal of dimensionality reduction is to find a way to map data from a high-dimensional space to a low-dimensional space while preserving the most important information in the data as much as possible.

As an analogy: Suppose you need to describe a person's appearance to someone else. You could describe a photo pixel by pixel, but this is obviously highly inefficient. A better way is to capture key features—a few dimensions such as face shape, eye color, hairstyle—to roughly reconstruct a person's appearance. Dimensionality reduction does something similar: it finds the directions that best capture the variation in the data and discards directions with little variation or dominated by noise.
\\
The value of dimensionality reduction is reflected in several aspects:
\begin{itemize}
	\item Visualization: Reducing data to two or three dimensions allows direct observation of sample distribution and clustering;
	\item Denoising: Discarding directions with small variation often simultaneously removes noise;
	\item Computational efficiency: Low-dimensional data significantly reduces the computational cost of subsequent algorithms;
	\item Feature extraction: The new features after dimensionality reduction are sometimes more representative than the original features.
\end{itemize}

Principal Component Analysis is the most commonly used and classic linear dimensionality reduction method. Its core idea is very intuitive: If we could only choose one direction to project the data onto, which direction should we choose to preserve the variation of the original data to the greatest extent?

\begin{figure}[htbp]
	\centering
	\includegraphics[width=0.5\linewidth]{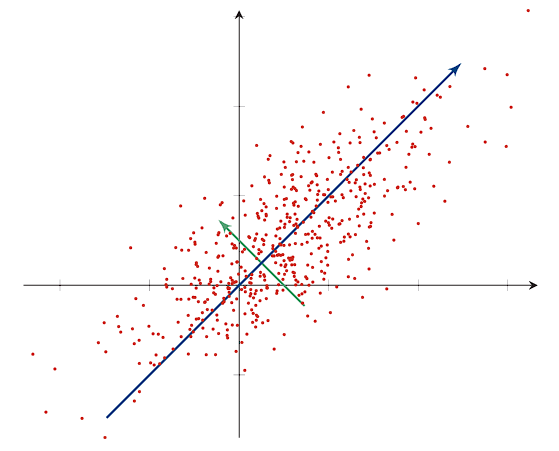}
	\caption{Schematic diagram of Principal Component Analysis \label{fig:pca}}
\end{figure}

Imagine a group of points scattered on a two-dimensional plane. If we project them onto a straight line, some projection directions will cause all points to be squeezed together (small variance), while other directions will cause the points to spread out as much as possible (large variance). Clearly, the latter preserves more information about the original data distribution—because the relative distances between points are better preserved. What Principal Component Analysis seeks are precisely such directions: the variance of the data projected onto these directions should be as large as possible, and these directions are orthogonal to each other. As shown in Figure\ref{fig:pca} (adapted from Qiu Xipeng's ``Neural Networks and Deep Learning'' Figure 9.1).

These directions are called principal components. The first principal component is the direction with the largest variance; the second principal component is the direction with the next largest variance under the constraint of being orthogonal to the first; and so on.

So, how do we find these directions mathematically? This requires the use of the covariance matrix. Given \(N\) \(d\)-dimensional samples \(x^{(1)}, x^{(2)}, \dots, x^{(N)}\), we first compute the sample mean \(\bar{x} = \frac{1}{N} \sum_{i=1}^N x^{(i)}\), then center each sample (subtract the mean). The centered data forms a matrix, and its covariance matrix is defined as:

\[
\Sigma = \frac{1}{N} \sum_{i=1}^N (x^{(i)} - \bar{x})(x^{(i)} - \bar{x})^T
\]

This matrix has dimensions \(d \times d\), and it captures the covariance relationships between the original features—the diagonal elements are the variances of each feature, and the off-diagonal elements are the covariances between features.

The key to Principal Component Analysis is: the variance of the data projected onto a certain direction is precisely equal to a certain relationship between that direction and the covariance matrix. Mathematically, it can be proven that the eigenvectors of the covariance matrix \(\Sigma\) indicate the directions of variance, and the corresponding eigenvalues measure the magnitude of variance in those directions. A larger eigenvalue means more information is preserved in that direction.
\\
Therefore, the steps for solving PCA are very clear:
\begin{enumerate}
	\item Compute the covariance matrix \(\Sigma\) of the data;
	\item Solve for all eigenvalues and eigenvectors of \(\Sigma\);
	\item Sort the eigenvalues in descending order and select the eigenvectors corresponding to the top \(k\) largest eigenvalues;
	\item Form the projection matrix \(W \in \mathbb{R}^{d \times k}\) from these eigenvectors;
	\item For any sample \(x\), the reduced-dimensional representation is \(z = W^T (x - \bar{x})\).
\end{enumerate}
Thus, we have compressed the original \(d\)-dimensional data into \(k\) dimensions (\(k \ll d\)), while preserving the main variation in the data as much as possible.

Principal Component Analysis also has an equivalent interpretation: it attempts to find a low-dimensional representation such that when reconstructing back to the original space from this representation, the resulting error is as small as possible. In other words, if we view dimensionality reduction as a form of compression, then good compression should be able to restore the original data as much as possible. PCA is optimal in this sense among all linear dimensionality reduction methods—it achieves the best in terms of minimizing the reconstruction error (i.e., the sum of squared Euclidean distances between the original data and the reconstructed data).

This perspective is equivalent to ``maximizing variance'': preserving the direction with the largest variance is also the direction that minimizes reconstruction error. Both interpret the same essence from different angles—Principal Component Analysis attempts to capture the most essential patterns of variation in the data.

Note: Another typical application of unsupervised learning is cluster analysis, which partitions data into several clusters such that samples within the same cluster are similar to each other, while samples from different clusters are more dissimilar. This is equivalent to discovering natural ``groupings'' or ``categories'' in the data space without predefining what these categories are. We briefly introduce here a special type of clustering method (although this method is not a linear model):

K-means Clustering: The goal is to partition \(N\) samples into \(K\) clusters, minimizing the sum of squared distances from each sample to its assigned cluster center:
\[ J = \sum_{n=1}^{N} \sum_{k=1}^{K} r_{nk} \| x^{(n)} - \mu_k \|^2 \]
where \(r_{nk} \in \{0,1\}\) indicates whether sample \(n\) belongs to cluster \(k\), and \(\mu_k\) is the center of cluster \(k\). The algorithm solves this by alternately executing the following steps:
\begin{enumerate}
	\item Assignment step: Fix \(\{\mu_k\}\), assign each sample to the nearest cluster center;
	\item Update step: Fix \(\{r_{nk}\}\), recalculate each cluster center as the mean of the samples in that cluster.
\end{enumerate}
K-means is simple and efficient but sensitive to initial centers, requires pre-specifying the number of clusters \(K\), and has limited effectiveness for non-spherical clusters or clusters of different densities.

\subsection{Optimization Algorithms: Gradient Descent}

In the previous discussion, we established the model and clarified the learning objective—typically minimizing a certain loss function. However, for the vast majority of models with practical value, this minimization problem cannot be solved directly through analytical methods (like solving a quadratic equation to obtain an explicit formula). The reality we face is: we need to find a method to let the model iteratively approach the optimal solution. This is precisely where optimization algorithms come into play. Gradient descent and its numerous variants form the core of machine learning optimization.

Imagine you are standing at a certain point on a mountain, with thick fog obscuring the entire landscape, but you can feel the slope of the ground beneath your feet. What would you do to reach the valley as quickly as possible? Naturally, you would take a step in the direction of the steepest descent, then reassess the slope at the new position, take another step—and repeat this process until you no longer feel a significant downward slope.

This is an intuitive depiction of gradient descent. Mathematically, the gradient \(\nabla_{\theta} \mathcal{R}(\theta)\) points in the direction of the fastest increase of the function value; its opposite direction is naturally the direction of the fastest decrease. Each component of the gradient is the partial derivative of the loss function with respect to the corresponding parameter—it tells us whether adjusting this parameter will increase or decrease the loss, and by how much.

Based on this idea, the update rule for gradient descent can be written as:

\[ \theta_{t+1} = \theta_t - \alpha \cdot \nabla_{\theta} \mathcal{R}_{\text{emp}}(\theta_t) \]
Here, \(\theta_t\) represents the parameters at the \(t\)-th iteration, \(\nabla_{\theta} \mathcal{R}_{\text{emp}}(\theta_t)\) is the gradient of the empirical risk function (i.e., the average loss on the training set) at the current parameters. The learning rate \(\alpha > 0\) controls how large a step we take—too small a step leads to slow convergence; too large a step may overshoot the valley or even diverge.

The specific implementation of gradient descent depends on how much data we use to compute the gradient each time. Based on the amount of data used, three basic variants can be distinguished.
\begin{itemize}
	\item Batch gradient descent uses the entire training dataset to compute the gradient. The advantage of this approach is that the gradient direction is accurate, with each step moving toward the global optimum; but the cost is also obvious—when the dataset is huge, each iteration requires traversing the entire dataset, making the computational cost prohibitively high.
	
	\item Stochastic gradient descent [Nemirovskietal.,2009] goes to the other extreme: each time, it randomly selects only one sample and uses its loss gradient as an estimate of the global gradient. This makes the computation extremely lightweight, allowing updates to be very frequent. More importantly, the randomness introduced by the single-sample gradient actually helps the algorithm escape local minima, offering a chance to find better solutions. However, the cost is that the update direction oscillates violently, leading to an unstable convergence process.
	
	\item Mini-batch gradient descent [Bottou,2010] is a compromise between the two: each time, it uses a small batch of data (e.g., 32, 64, or 128 samples) to compute the gradient. It both reduces variance through batch averaging, making the update direction relatively stable, and maintains high computational efficiency.
\end{itemize}
\vspace{3mm}

\noindent\textcolor{structure3}{\textbf{Advanced Optimization Methods}}

Although basic gradient descent is effective, it encounters two thorny challenges on complex problems: first, it easily oscillates back and forth in steep valleys, leading to slow convergence; second, using a uniform learning rate for all parameters cannot adapt to the update needs of different parameters. To address these, researchers have proposed a series of ``smarter'' optimization methods.

Momentum: The inspiration for momentum comes from physics: imagine a small ball rolling down a hill; it accumulates momentum, and when it encounters an opposing slope, it does not stop immediately but rushes past due to inertia. Momentum introduces a velocity variable \(v\) to simulate this effect:
\[ v_{t+1} = \beta v_t + (1-\beta) \nabla_{\theta} \mathcal{L}(\theta_t), \quad \theta_{t+1} = \theta_t - \alpha v_{t+1} \]
The current gradient is responsible for ``acceleration,'' while the velocity accumulated from historical gradients maintains the direction of motion. The hyperparameter \(\beta\) (usually set around 0.9) controls the degree of reliance on historical information. The effect of momentum is: accelerating convergence in dimensions where the gradient direction is consistent, and suppressing oscillations in dimensions where the gradient direction fluctuates—like adding inertia to the rolling ball, making it run more steadily and quickly.
\\
Adam (Adaptive Moment Estimation) combines momentum with adaptive learning rates. It maintains two state variables:

\begin{itemize}
	\item First moment estimate \(m_t\): the exponentially weighted moving average of gradients, equivalent to gradient with momentum;
	\item Second moment estimate \(v_t\): the exponentially weighted moving average of squared gradients, reflecting the variance of gradients (i.e., the historical update magnitude for each parameter).
\end{itemize}
The specific update process is as follows. First, compute the current gradient \(g_t = \nabla_{\theta} \mathcal{L}(\theta_t)\), then update the two moment estimates:

\[
m_t = \beta_1 m_{t-1} + (1-\beta_1) g_t
\]
\[
v_t = \beta_2 v_{t-1} + (1-\beta_2) g_t^2
\]
Here, \(\beta_1\) and \(\beta_2\) are decay rates. Since \(m_t\) and \(v_t\) are biased towards $0$ in the initial stages, bias correction is necessary:

\[
\hat{m}_t = \frac{m_t}{1 - \beta_1^t}, \quad \hat{v}_t = \frac{v_t}{1 - \beta_2^t}
\]
Finally, update the parameters:

\[
\theta_{t+1} = \theta_t - \alpha \cdot \frac{\hat{m}_t}{\sqrt{\hat{v}_t} + \epsilon}
\]
where \(\epsilon\) is a small constant (e.g., \(10^{-8}\)) to prevent division by zero.

The ingenuity of Adam lies in: each parameter has its own adaptive learning rate. For parameters with historically large gradients (frequent updates), \(\sqrt{\hat{v}_t}\) is large, and the effective learning rate is automatically reduced; for parameters with small gradients (slow updates), the effective learning rate is automatically increased. This adaptive mechanism makes Adam less sensitive to the choice of initial learning rate and allows it to perform well on various problems, making it one of the preferred optimizers for training neural networks (see the next section on artificial neural networks).

\subsection{Model Evaluation and Selection}

In machine learning practice, the fundamental question we face is not ``can the model memorize the training data,'' but ``can the model handle new, unseen samples''—this is generalization ability. How to accurately evaluate a model's generalization ability and, based on that, select the optimal model configuration constitutes the core issue of model evaluation and selection.
\vspace{3mm}

\noindent\textcolor{structure3}{\textbf{The Three-Way Split of Data}}

To evaluate generalization ability, we must first clarify: we cannot use the data that trained the model to test it—this is like taking an exam with a paper you've already seen, which cannot reflect true proficiency. Therefore, we need to properly split the available data.
\begin{itemize}
	\item The training set is the material for the model to learn from. The model uses it to update parameters and fit the data distribution, analogous to a student's textbooks and practice problems.
	
	\item The validation set plays the role of a ``mock exam.'' It does not participate in parameter learning but is used to guide model selection and hyperparameter tuning—when we need to decide the polynomial degree, the learning rate, etc., the performance on the validation set is the basis for decision-making. The reason a validation set is needed is that if we directly use the test set for tuning, it would be like knowing the exam answers in advance, rendering the final evaluation meaningless.
	
	\item The test set is the final ``college entrance exam.'' It must remain completely unseen throughout the entire training and tuning process, used only once after the model is fully finalized, to provide an unbiased estimate of generalization ability. The performance on the test set is the model performance we report externally.
\end{itemize}
This three-way split forms the basic framework for evaluation, ensuring clear data boundaries at every stage from model development to final assessment.

While the single split described above is simple, it harbors a risk: the result may depend on a particular, specific split. If the validation set happens to contain some easy or difficult samples, the evaluation result will be biased. To mitigate this issue, k-fold cross-validation was developed.

The procedure is as follows: randomly divide the training data into k equal parts (typically 5 or 10). Then perform k training-validation cycles—the first cycle uses the first part as the validation set and the rest as the training set; the second cycle uses the second part as the validation set, and so on. Each cycle records the model's performance metric on the validation set, and finally, the average of the k results is taken as the performance estimate.

The advantages of this method are: every sample has a chance to be in the validation set, making the performance estimate more stable and reliable; simultaneously, all data is used for both training and validation, avoiding data waste. Cross-validation is particularly valuable when the amount of data is limited.

It is important to emphasize that cross-validation is still conducted *within* the training data—it is used for model selection and hyperparameter tuning. The final test set must still be kept untouched until the very last evaluation.
\vspace{3mm}

\noindent\textcolor{structure3}{\textbf{Parameters and Hyperparameters}}

In the context of machine learning, ``parameters'' and ``hyperparameters'' are two easily confused but fundamentally different concepts. Understanding their distinction helps grasp the hierarchical structure of model learning.

Parameters are the internal variables of the model; they constitute the core of the model itself. The weight coefficients \textit{w} and bias \textit{b} in a linear model, the coefficients of various terms in a polynomial model—all belong to parameters. The characteristic of these variables is: their values are entirely driven by the training data, learned automatically through optimization algorithms (such as gradient descent). In other words, parameters are the knowledge ``summarized'' by the model from the data.

Hyperparameters are completely different. They are configuration options set manually before training begins, controlling the training process itself and the model's structure. The learning rate \textit{$\alpha$}, the batch size in gradient descent, the degree \textit{M} of a polynomial model, the regularization coefficient—all these are hyperparameters. Their values cannot be learned directly from the data; they need to be determined based on human experience or some search strategy.

One can understand the relationship between the two as follows: parameters are the kernel of the model, while hyperparameters are the external instructions that shape this kernel. Hyperparameters determine *how* and under *what constraints* the model learns the parameters. Precisely because the choice of hyperparameters directly affects learning outcomes, we need to rely on the validation set or cross-validation to ``tune'' them—trying different hyperparameter combinations, observing performance on the validation set, and finally selecting the most suitable set.

This hierarchical structure runs through the entire practice of machine learning: within the framework determined by hyperparameters, the model learns parameters from data; while the hyperparameters themselves require a higher-level validation mechanism for optimization. The two complement each other, together forming the complete chain from data to model.

\section{Artificial Neural Networks}

So far, we have introduced the fundamental paradigm of machine learning, whose core lies in learning an optimal mapping from data. Linear models serve as the cornerstone due to their simplicity and interpretability, but their ability to handle complex nonlinear relationships is limited. This chapter will focus on a powerful and revolutionary family of nonlinear models—Artificial Neural Networks (Neural Networks).

The study of artificial neural networks is inspired by a simplified simulation of how networks of neurons operate in the biological brain. Artificial neural networks play a unique and crucial role: they are an extremely flexible nonlinear function approximator and a powerful feature transformation engine. This means that when faced with complex problems—for example, finding hidden patterns in high-dimensional, unstructured data (such as recognizing geometric structures in images), or approximating an unknown function that has no explicit formula but can be described by samples (such as solving partial differential equations)—artificial neural networks provide a way to directly ``learn'' the solution from data.

Artificial neural networks, especially deep neural networks, form the foundation of most current breakthrough advances in artificial intelligence. From image recognition and natural language processing to game systems like AlphaGo, and further to mathematical discovery tools like FunSearch and AlphaTensor, artificial neural networks have played a central role. This section will start from a mathematical perspective, strip away engineering details, and delve into the three core components of neural networks: activation functions, network architecture, and learning algorithms, explaining the underlying mathematical ideas and theoretical guarantees.

Before introducing artificial neural networks, we first introduce its predecessor: logistic regression.

Despite containing ``regression'' in its name, logistic regression is actually a classic binary classification model. Its core idea is to use the model's output to fit the log-odds of a sample belonging to the positive class. The model first computes a linear score \(z = w^T x + b\), then maps it to the interval \((0,1)\) via the Sigmoid function \(\sigma(z) = 1/(1+e^{-z})\), interpreting it as the probability of the positive class:
\[ p(y=1 | x; w, b) = \sigma(w^T x + b) \]
The classification decision rule is: if \(p(y=1|x) > 0.5\), predict the positive class; otherwise, predict the negative class.

Logistic regression employs the classic cross-entropy loss function. For a single sample, with true label \(y \in \{0,1\}\) and predicted probability \(\hat{y} = p(y=1|x)\), the loss is:
\[ \mathcal{L}(y, \hat{y}) = -\left[ y \log(\hat{y}) + (1-y) \log(1-\hat{y}) \right] \]
This function measures the difference between the predicted probability distribution and the true label distribution; minimizing cross-entropy is equivalent to maximum likelihood estimation. Its gradient has a simple form: \(\partial \mathcal{L} / \partial w = (\hat{y} - y) x\).

Building upon binary classification, logistic regression can be further generalized to multi-class classification tasks, leading to Softmax regression (also known as multinomial logistic regression). When the number of classes \(C > 2\), we need to compute a score \(z_c = w_c^T x + b_c\) for each class and transform these scores into a probability distribution via the Softmax function:
\[ p(y=c | x; W, b) = \frac{\exp(z_c)}{\sum_{j=1}^{C} \exp(z_j)} = \text{softmax}(z)_c \]
Thus, Softmax regression unifies predictions for multiple classes into a normalized probabilistic framework. Correspondingly, its loss function also adopts the cross-entropy form. For the true class \(c\), the loss is:
\[ \mathcal{L}(y, \hat{y}) = -\sum_{c=1}^{C} \mathbb{I}(y=c) \log p(y=c|x) = -\log p(y=c|x) \]
where \(\mathbb{I}(\cdot)\) is the indicator function. Softmax regression is concise and effective, making it a standard configuration for the output layer in multi-class tasks in deep learning.

Artificial neural networks are a generalization of the logistic regression function, comprising the following three core components: activation functions, network architecture, and learning algorithms.
\vspace{3mm}

\noindent\textcolor{structure3}{\textbf{From Biological Neurons to Artificial Neurons}}

The basic computational unit of a neural network is the artificial neuron, whose design is inspired by biological neurons. A simplified biological neuron receives electrical signals from other neurons; when the total signal strength exceeds a certain threshold, the neuron is activated and transmits signals downstream. The artificial neuron is a mathematical abstraction and modeling of this process.

The mathematical description of a typical artificial neuron (also known as a Perceptron in early days) is as follows:

Given an input vector \(\mathbf{x} = [x_1, x_2, ..., x_d]^T \in \mathbb{R}^d\), the neuron first computes a weighted sum and adds a bias term. This intermediate result is called the neuron's net input or pre-activation value:

\[
z = \sum_{i=1}^{d} w_i x_i + b = \mathbf{w}^T \mathbf{x} + b
\]

where \(\mathbf{w} = [w_1, w_2, ..., w_d]^T \in \mathbb{R}^d\) is the weight vector, with each \(w_i\) measuring the importance or connection strength of the corresponding input \(x_i\); \(b \in \mathbb{R}\) is the bias, used to control how easily the neuron is activated. Figure\ref{fig:Neuron} shows a typical neuron structure example (adapted from Figure 4.1 in Qiu Xipeng's ``Neural Networks and Deep Learning''):
\begin{figure}[htbp]
	\centering
	\includegraphics[width=0.5\linewidth]{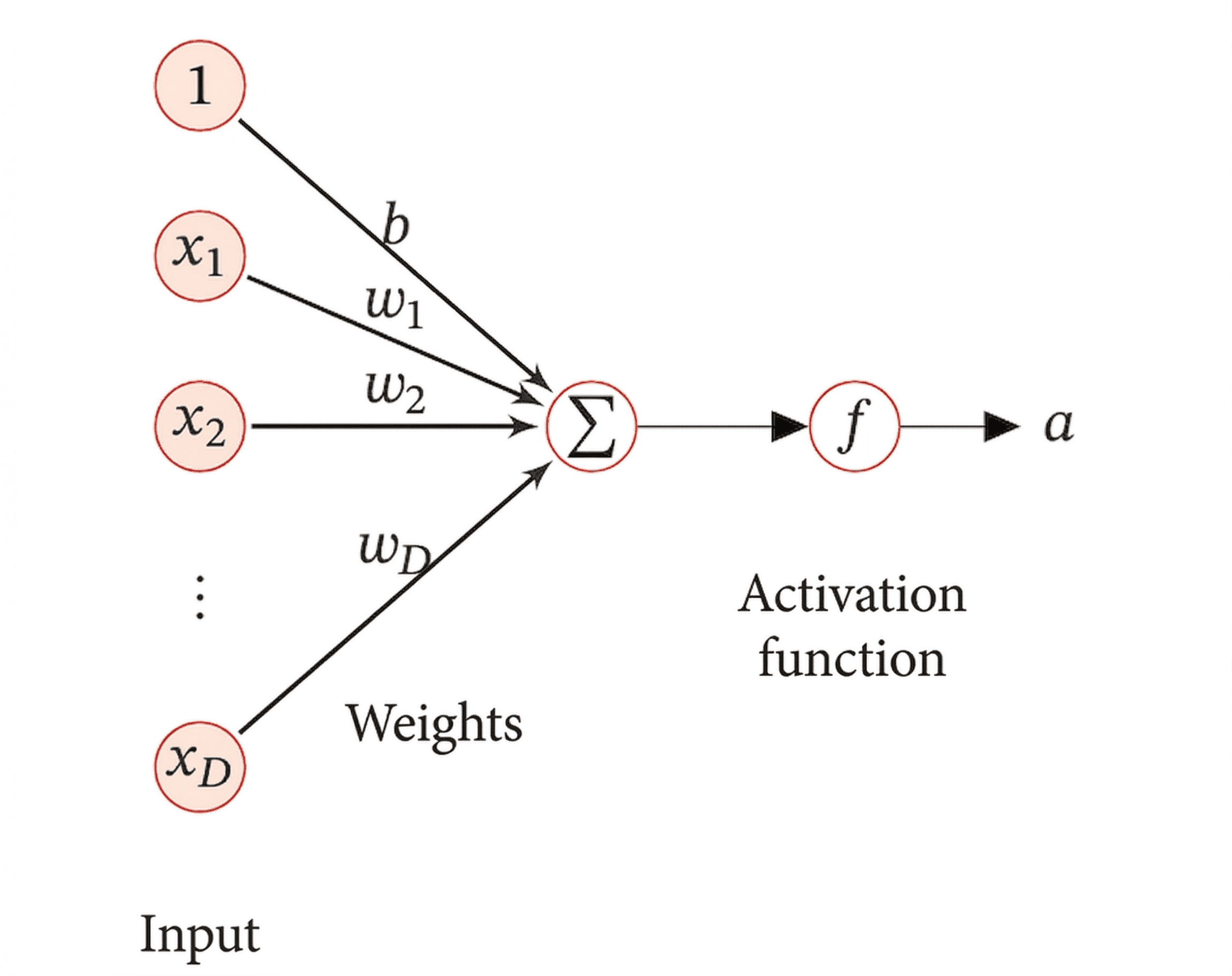}
	\caption{Neuron Structure \label{fig:Neuron}}
\end{figure}

However, if the neuron merely outputs the linear combination \(z\), then a network composed of multiple such linear units would essentially remain a combination of linear models, unable to break through the limitations of linear transformations. To endow the network with the ability to approximate arbitrarily complex functions, nonlinearity must be introduced. Therefore, the neuron passes the net input \(z\) to an activation function \(f(\cdot)\), obtaining the final output or activation value:

\[
a = f(z) = f(\mathbf{w}^T \mathbf{x} + b)
\]

It is precisely the nonlinear nature of the activation function \(f\) that gives neural networks their powerful expressive capacity, enabling them to learn complex patterns and hierarchical features in data.
\vspace{3mm}

\noindent\textcolor{structure3}{\textbf{Core Nonlinear Component: Activation Functions}}

The choice of activation function is crucial; it directly determines how the neuron responds to input signals and profoundly affects the training effectiveness and performance of the entire network. Ideal activation functions typically need to possess properties such as nonlinearity, continuous differentiability (or almost everywhere differentiability), computational simplicity, and easy-to-compute derivatives. Here are several classic and widely used activation functions.
\begin{enumerate}
	\item Sigmoid function (Logistic function)
	\[
	\sigma(z) = \frac{1}{1 + e^{-z}}
	\]
	The Sigmoid function ``squeezes'' any real number \(z\) into the interval \((0, 1)\), and its output can be interpreted as a probability. It was very popular in early neural networks. Its derivative is \(\sigma'(z) = \sigma(z)(1 - \sigma(z))\). However, when the absolute value of the input \(z\) is large, its derivative approaches 0, which can lead to the ``vanishing gradient'' problem in deep network training.
	
	\item Hyperbolic tangent function (Tanh function)
	\[
	\tanh(z) = \frac{e^{z} - e^{-z}}{e^{z} + e^{-z}}
	\]
	The Tanh function maps the input to the interval \((-1, 1)\) and is zero-centered (mean 0), which in practice can sometimes allow the network to train faster. Its derivative is \(\tanh'(z) = 1 - \tanh^2(z)\).
	
	\item Rectified Linear Unit (ReLU) [Nair et al., 2010]
	\[
	\text{ReLU}(z) = \max(0, z)
	\]
	ReLU is one of the most popular activation functions today. Its computation is extremely simple: when \(z > 0\), the output is \(z\); otherwise, the output is 0. Its derivative is 1 for \(z > 0\) and 0 for \(z < 0\). The non-saturating nature of ReLU (constant derivative of 1 in the positive region) effectively alleviates the vanishing gradient problem and speeds up training. Its drawback is that when the input is negative, the gradient is 0, which may cause some neurons to never activate (the ``dying neuron'' problem).
	
	\item Leaky ReLU and ELU
	
	To mitigate the ``dying'' problem of ReLU, variants have been proposed. Leaky ReLU gives a small slope (e.g., 0.01) in the negative region: \[f(z) = \max(0.01z, z)\]
	The Exponential Linear Unit (ELU) uses an exponentially decaying function in the negative region:
	\[f(z) = \begin{cases} z, & z \ge 0 \\ \gamma(\exp(z)-1), & z < 0 \end{cases},\]
	where \(\gamma\) is a hyperparameter.
\end{enumerate}
These nonlinear activation functions are the foundation for neural networks to construct complex, hierarchical feature representations. A single neuron acts like a ``mini-classifier'' or ``feature detector,'' and by combining countless such neurons, the network can learn and represent extremely complex features.
\vspace{3mm}

\noindent\textcolor{structure3}{\textbf{Network Architecture: From Fully Connected to Locally Connected}}

A single neuron has limited capability. Connecting a large number of neurons according to a specific topology forms an artificial neural network. The network architecture defines the paths and manner of information flow between neurons.
\vspace{3mm}

\noindent\textcolor{second}{\textbf{Fully Connected Feedforward Neural Networks}}

The fully connected feedforward neural network is one of the most basic and core network architectures. Often referred to as ``Multilayer Perceptron'' (MLP) or ``Deep Neural Network'' (DNN), it typically refers to this type of network.

Architectural characteristics: The network consists of multiple layers of neurons, typically including an input layer, one or more hidden layers, and an output layer. There are no connections between neurons within the same layer. Each neuron in a layer is connected to all neurons in the previous layer. Information starts from the input layer and propagates unidirectionally layer by layer, with no feedback, hence the term ``feedforward.'' This structure can be represented by a directed acyclic graph.

Mathematical description: Consider a network with \(L\) layers (the input layer is not counted).

\(M_l\): Number of neurons in layer \(l\) (\(l=0\) is the input layer).
\(f_l(\cdot)\): Activation function for neurons in layer \(l\).
\(\mathbf{W}^{(l)} \in \mathbb{R}^{M_l \times M_{l-1}}\): Weight matrix connecting layer \(l-1\) to layer \(l\). Its element \(W_{ij}^{(l)}\) represents the connection weight from the \(j\)-th neuron in layer \(l-1\) to the \(i\)-th neuron in layer \(l\).
\(\mathbf{b}^{(l)} \in \mathbb{R}^{M_l}\): Bias vector for layer \(l\).
\(\mathbf{z}^{(l)} \in \mathbb{R}^{M_l}\): Net input vector for neurons in layer \(l\).
\(\mathbf{a}^{(l)} \in \mathbb{R}^{M_l}\): Output (activation value) vector for neurons in layer \(l\), with \(\mathbf{a}^{(0)} = \mathbf{x}\) (input).

The forward propagation process of the network can be described recursively as:
\[
\mathbf{z}^{(l)} = \mathbf{W}^{(l)} \mathbf{a}^{(l-1)} + \mathbf{b}^{(l)}
\]
\[
\mathbf{a}^{(l)} = f_l(\mathbf{z}^{(l)})
\]
Finally, the network's output is \(\mathbf{a}^{(L)} = \phi(\mathbf{x}; \mathbf{W}, \mathbf{b})\), where \(\phi\) denotes the composite function of the entire network.

\begin{figure}[htbp]
	\centering
	\includegraphics[width=0.7\linewidth]{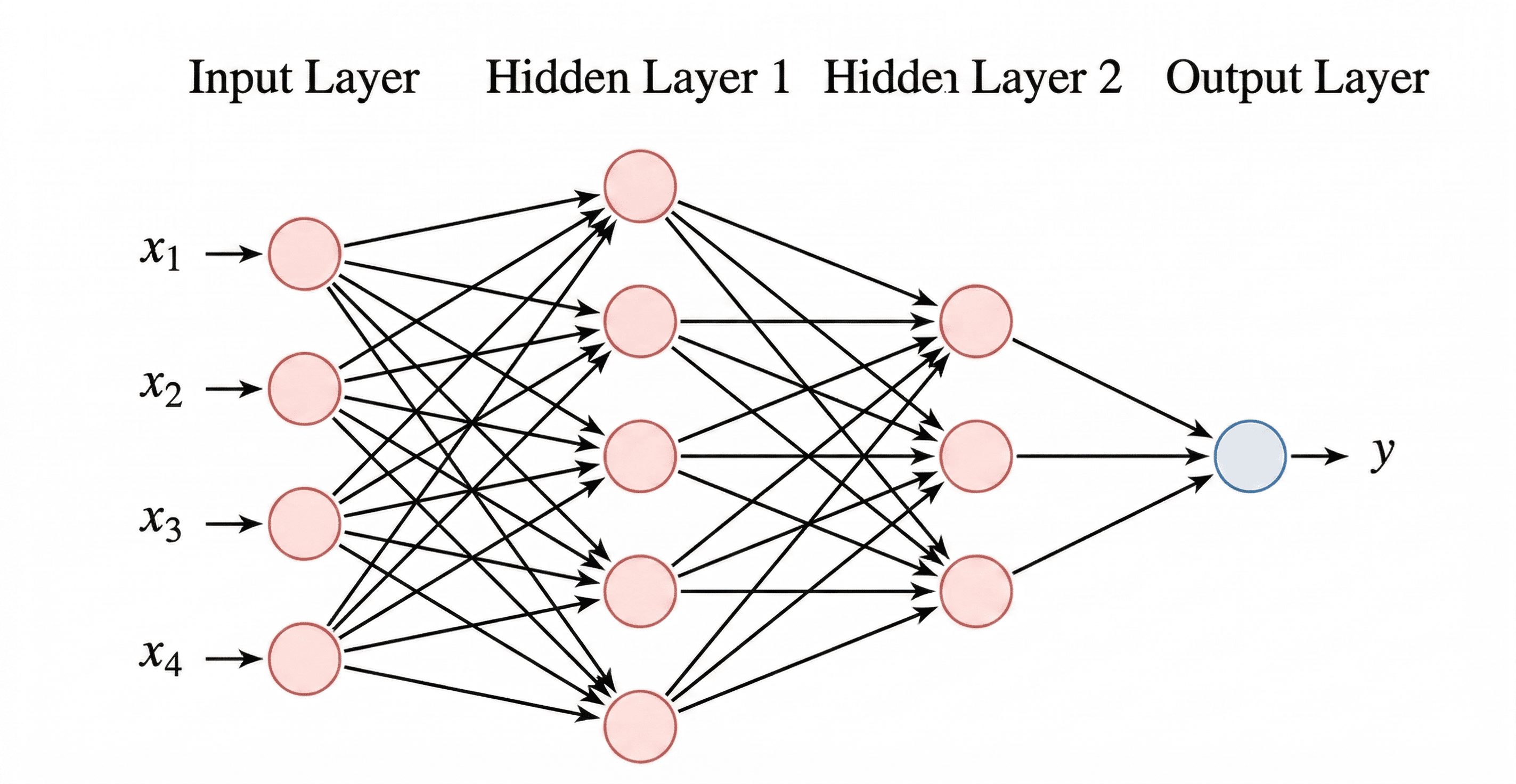}
	\caption{Feedforward Neural Network \label{fig:Feedforward}}
\end{figure}

The structure of a fully connected feedforward neural network is shown in Figure\ref{fig:Feedforward} (adapted from Figure 4.7 in Qiu Xipeng's ``Neural Networks and Deep Learning'').

The powerful capability of fully connected neural networks has solid mathematical theoretical support. The Universal Approximation Theorem states: A fully connected network with at least one hidden layer containing a sufficient number of neurons and using a nonlinear activation function (such as Sigmoid, ReLU, etc.) can approximate any continuous function defined on a compact set to arbitrary accuracy.

Theorem conditions and intuitive understanding: The classical form of this theorem usually requires the activation function \(\varphi(\cdot)\) to be non-constant, bounded, monotonic increasing, and continuous (the Sigmoid function satisfies this condition; ReLU, although unbounded, is supported by subsequent extended theories). Also, the approximation holds on a ``compact set'' (e.g., a bounded closed interval). This means that no matter how complex or irregular the function you want to approximate is, as long as the network is sufficiently wide (enough neurons in the hidden layer), there always exists a network whose input-output relationship differs from your target function by an error within any allowable range. A simple intuition is: each hidden layer neuron can be seen as a ``basis function'' (like an ``S-shaped'' peak or trough formed by the Sigmoid function). By adjusting the weights and biases of enough such basis functions, we can ``sculpt'' arbitrarily complex continuous function shapes using their linear combination. This theorem establishes the theoretical foundation for neural networks as ``universal function approximators,'' explaining why they can handle a wide variety of problems from image classification to solving mathematical equations. Of course, the theorem guarantees ``existence'' but does not tell us how to find (i.e., train) this network, which relies on the learning algorithms introduced in the next section.
\vspace{3mm}

\noindent\textcolor{second}{\textbf{Convolutional Neural Networks (CNN)}}

Fully connected networks have significant drawbacks when processing data with local correlations and translation invariance, such as images and audio: excessive parameters lead to huge computational and storage overhead, and they struggle to effectively capture local features.

Convolutional neural networks cleverly address these issues by introducing convolutional layers. Their core idea originates from the ``receptive field'' mechanism of the biological visual system. The core of a convolutional neural network is the convolution operation:

One-dimensional convolution: For an input sequence \(x\) and a filter (or convolution kernel) \(w\), the convolution output \(y_t = \sum_{k=1}^{K} w_k x_{t-k+1}\). The filter slides over the input, computing the dot product at each position, thereby extracting local features.

Two-dimensional convolution: For two-dimensional inputs like images \(\mathbf{X} \in \mathbb{R}^{H \times W}\) and a small filter \(\mathbf{W} \in \mathbb{R}^{U \times V}\) (e.g., U=3, V=3), the convolution operation \(\mathbf{Y} = \mathbf{W} * \mathbf{X}\) is defined as:
\[
y_{ij} = \sum_{u=1}^{U} \sum_{v=1}^{V} w_{uv} \cdot x_{i-u+1, j-v+1}
\]
Here is an intuitive example: Suppose the input image \(\mathbf{X}\) is a 6x6 matrix, and the filter \(\mathbf{W}\) is a 3x3 matrix. The filter starts from the top-left corner of the input image, covering a 3x3 area. The corresponding elements are multiplied and summed to obtain the first element \(y_{11}\) of the output feature map \(\mathbf{Y}\). Then the filter slides one step to the right (stride=1), computing \(y_{12}\), and so on, until the entire image is scanned, generating a new feature map.

\begin{figure}[htbp]
	\centering
	\includegraphics[width=0.6\linewidth]{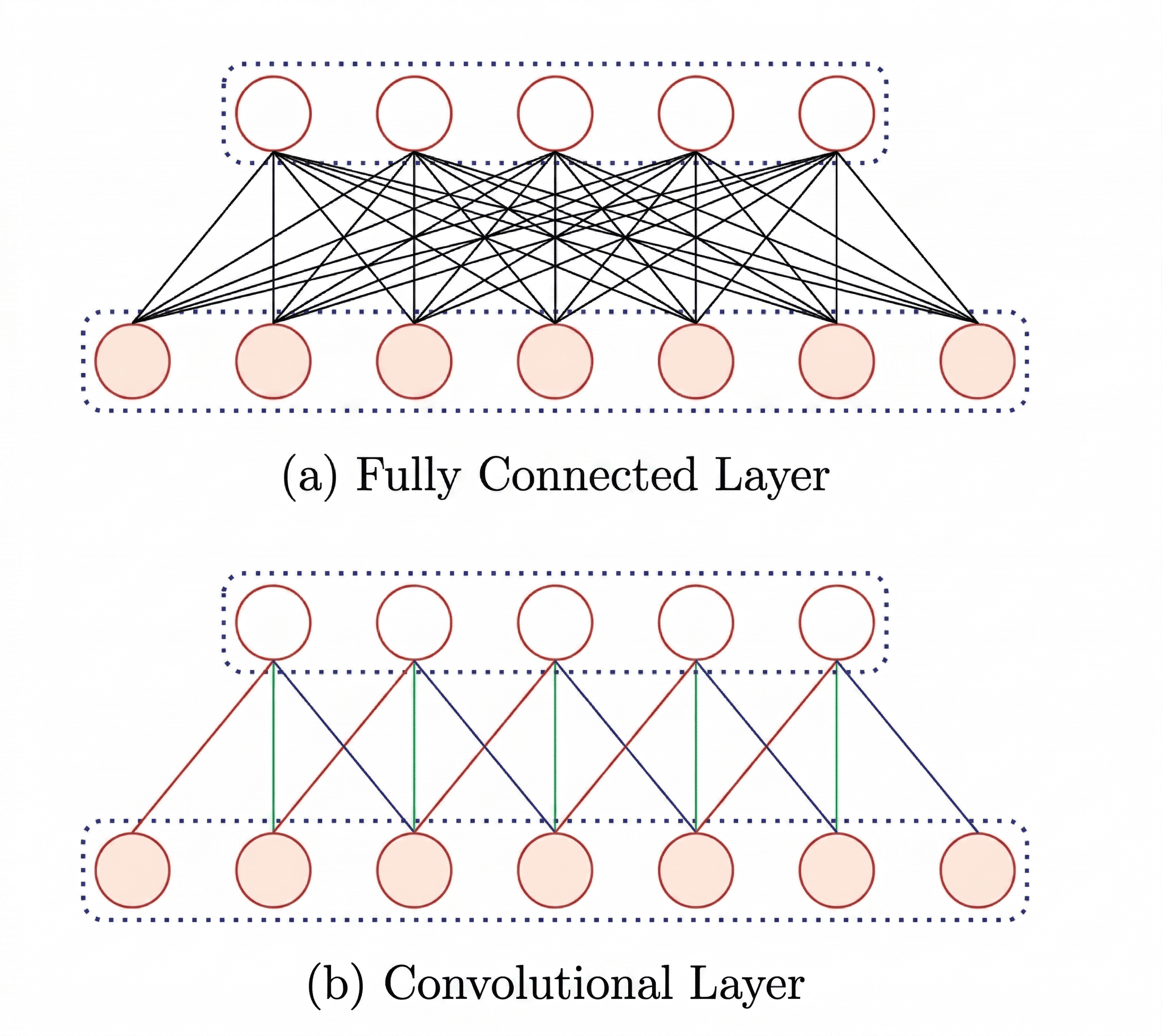}
	\caption{Convolutional Layer \label{fig:Convolutional}}
\end{figure}

A comparison between fully connected layers and convolutional layers is shown in Figure\ref{fig:Convolutional} (adapted from Figure 5.5 in Qiu Xipeng's ``Neural Networks and Deep Learning'').
\\
The advantages brought by the characteristics of the convolutional structure are:
\begin{enumerate}
	\item Local connectivity: Each neuron is only connected to a local region (receptive field) of the input, not to all of it. This significantly reduces the number of parameters.
	\item Weight sharing: The same filter shares the same set of weights \(w_{uv}\) at different positions of the input. This means that whether the target appears in the top-left or bottom-right corner of the image, it is recognized by the same ``feature detector,'' which endows the model with translation invariance. Simultaneously, this further compresses the number of parameters.
	\item Hierarchical feature extraction: A typical CNN consists of multiple convolutional layers, pooling layers (Pooling Layer, such as max pooling, used for downsampling and enhancing translation robustness), and fully connected layers (for final classification or regression) stacked alternately. Shallow convolutions learn low-level features like edges and corners; deep convolutions combine low-level features into more complex high-level features, such as object parts or wholes. This architecture has led to tremendous success for CNNs in fields like image and video analysis.
	\item Trend towards fully convolutional: Modern advanced CNN architectures (like ResNet, DenseNet) tend towards ``small convolution kernels, large depth,'' and reduce or even eliminate fully connected layers, using operations like global average pooling, making the network more flexible to input sizes.
\end{enumerate}
CNNs are essentially function approximators with strong structural priors (locality and translation invariance). They impose powerful constraints on high-dimensional input spaces (like image pixel space), making the functions learned by the network possess specific symmetries. This ``symmetry'' can be understood as: if we translate the input image, the feature representation output by the network will also shift correspondingly without changing its intrinsic semantic information.
\vspace{3mm}

\noindent\textcolor{second}{\textbf{Recurrent Neural Networks (RNN): Memory Models for Sequential Data}}

Fully connected networks and convolutional networks primarily handle independent and identically distributed samples. However, for sequential data like time series, natural language, and speech, samples have sequential dependencies. Recurrent neural networks handle this dynamic temporal information by introducing a ``memory'' mechanism.

\begin{figure}[htbp]
	\centering
	\includegraphics[width=0.7\linewidth]{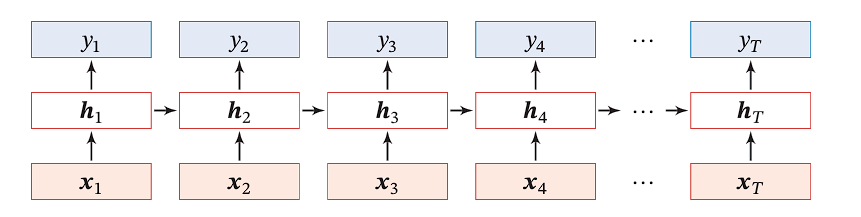}
	\caption{Recurrent Neural Network \label{fig:Recurrent}}
\end{figure}

The core idea is: RNNs have ``recurrent'' connections, allowing the network to pass information from past time steps to the current time step. At each time step \(t\), the network receives the current input \(\mathbf{x}_t\) and the hidden state from the previous time step \(\mathbf{h}_{t-1}\), computing the current hidden state \(\mathbf{h}_t\) and output \(\mathbf{y}_t\).
\[
\mathbf{h}_t = f(\mathbf{U}\mathbf{h}_{t-1} + \mathbf{W}\mathbf{x}_t + \mathbf{b})
\]
\[
\mathbf{y}_t = g(\mathbf{V}\mathbf{h}_t + \mathbf{c})
\]
where \(f, g\) are activation functions. The hidden state \(\mathbf{h}_t\) is considered the network's ``memory'' at time \(t\), encapsulating all historical input information up to the current moment.
The structure of a recurrent neural network is shown in Figure\ref{fig:Recurrent} (adapted from Figure 6.2 in Qiu Xipeng's ``Neural Networks and Deep Learning'').

Similar to feedforward networks, RNNs also have their own universal approximation theorem: A fully connected recurrent network with a sufficient number of sigmoid-type neurons can approximate any nonlinear dynamical system to arbitrary accuracy. This theoretically guarantees that RNNs have the capacity to simulate complex temporal processes.

Basic RNNs, when training on long sequences, are prone to gradient vanishing or explosion problems because error backpropagation through time involves multiple matrix multiplications, making it difficult to learn long-term dependencies. LSTM addresses this issue by introducing an ingenious ``gating mechanism.'' The core of an LSTM unit is the cell state \(\mathbf{C}_t\), which acts like a conveyor belt, maintaining information flow throughout the chain. LSTM regulates information through three ``gates'':
\begin{enumerate}
	\item Forget gate \(\mathbf{f}_t = \sigma(\mathbf{W}_f \cdot [\mathbf{h}_{t-1}, \mathbf{x}_t] + \mathbf{b}_f)\): Decides what information to discard from the cell state.
	\item Input gate \(\mathbf{i}_t = \sigma(\mathbf{W}_i \cdot [\mathbf{h}_{t-1}, \mathbf{x}_t] + \mathbf{b}_i)\): Decides what new information will be stored in the cell state.
	\item Output gate \(\mathbf{o}_t = \sigma(\mathbf{W}_o \cdot [\mathbf{h}_{t-1}, \mathbf{x}_t] + \mathbf{b}_o)\): Based on the current cell state, decides what information to output to the hidden state.
\end{enumerate}
The cell state update formula is: \(\mathbf{C}_t = \mathbf{f}_t \odot \mathbf{C}_{t-1} + \mathbf{i}_t \odot \tilde{\mathbf{C}}_t\), where \(\tilde{\mathbf{C}}_t\) is the candidate new information.

The key to LSTM lies in additive updates: the cell state is updated via element-wise multiplication with the forget gate and element-wise addition with the input gate. This additive connection allows gradients to flow through the cell state path during backpropagation without easily vanishing (gradients can be directly passed through the additive path).

GRU is a simplified variant of LSTM, merging the forget gate and input gate into an ``update gate,'' resulting in a simpler structure with comparable performance. These gating mechanisms endow RNNs with the ability to remember long-term information and selectively forget or retain it.

RNNs and their variants (LSTM, GRU) are crucial in fields like machine translation, speech recognition, and text generation. In mathematics, they can be used for generating symbolic sequences (like mathematical expressions), time series forecasting, or as components of reinforcement learning agents (e.g., AlphaGo's policy network included CNNs and RNNs to evaluate board positions).
\vspace{3mm}

\noindent\textcolor{structure3}{\textbf{Learning Algorithms: How to Find the ``Right'' Network}}

The Universal Approximation Theorem tells us that theoretically, there exists a neural network that can solve our problem. But how do we find that high-performing network from the countless possible ones? This is precisely the task of learning algorithms. The training process of neural networks can be summarized into three core steps: forward propagation to compute predictions, backpropagation to compute parameter gradients, and parameter update to optimize network performance.
\\
Step One: Forward Propagation and Loss Function—Defining ``Good'' and ``Bad''

Before training begins, we first need a quantitative criterion to measure the discrepancy between the network's output and the true target. This is the role of the loss function \(\mathcal{L}(\mathbf{y}, \hat{\mathbf{y}})\). For different tasks, we choose different loss functions: mean squared error is commonly used for regression problems, while cross-entropy loss is typical for classification.
Loss Function: Defining ``Good'' and ``Bad''

Take a \(C\)-class classification problem as an example. During forward propagation, the input sample \(\mathbf{x}\) passes through the network layers sequentially. Finally, the last layer typically uses the Softmax function to transform the net input into a class probability distribution \(\hat{\mathbf{y}} = \text{softmax}(\mathbf{z}^{(L)})\). Subsequently, the cross-entropy loss computes the difference between the predicted distribution and the true label:
\[
\mathcal{L}(\mathbf{y}, \hat{\mathbf{y}}) = -\sum_{c=1}^{C} y_c \log \hat{y}_c
\]
where \(\mathbf{y}\) is the true one-hot label vector.

Given a training set \(D = \{(\mathbf{x}^{(n)}, \mathbf{y}^{(n)})\}_{n=1}^N\) containing \(N\) samples, we typically minimize the empirical risk, which is the average loss over all samples, possibly with an added regularization term (like weight decay \(\frac{1}{2}\lambda \|\mathbf{W}\|_F^2\)) to prevent overfitting:
\[
\mathcal{R}(\mathbf{W}, \mathbf{b}) = \frac{1}{N} \sum_{n=1}^{N} \mathcal{L}(\mathbf{y}^{(n)}, \hat{\mathbf{y}}^{(n)}) + \frac{1}{2}\lambda \|\mathbf{W}\|_F^2
\]
At this point, forward propagation has completed its mission: it not only generated the prediction \(\hat{\mathbf{y}}\) but also computed the loss value \(\mathcal{R}\) that measures network performance, and cached necessary intermediate variables for subsequent steps.
\\
Step Two: Backpropagation Algorithm—The Efficient Engine for Computing Gradients

With the loss function \(\mathcal{R}\), our goal is to find a set of parameters \((\mathbf{W}, \mathbf{b})\) that minimize it. However, \(\mathcal{R}\) is a complex non-convex function with respect to all network parameters, making direct computation of gradients for each layer's parameters extremely difficult.

The backpropagation algorithm is the key to solving this challenge. Its essence is the efficient implementation of the chain rule from calculus on deep composite functions. The algorithm starts from the output layer and propagates error signals backward layer by layer:
\begin{enumerate}
	\item Compute error term: First, define the error term for layer \(l\) as \(\boldsymbol{\delta}^{(l)} = \frac{\partial \mathcal{L}}{\partial \mathbf{z}^{(l)}}\), i.e., the gradient of the loss function with respect to that layer's net input.
	\item Error backward recurrence: Using the chain rule, we obtain the recurrence formula for propagating the error term backward:
	\[
	\boldsymbol{\delta}^{(l)} = f'_l(\mathbf{z}^{(l)}) \odot \left( (\mathbf{W}^{(l+1)})^T \boldsymbol{\delta}^{(l+1)} \right)
	\]
	where \(\odot\) denotes element-wise multiplication. The intuitive meaning of this formula is: the error at the current layer depends on the derivative of the current layer's activation function (local gradient) and the weighted combination of errors from the next layer.
	\item Compute parameter gradients: Based on the error term \(\boldsymbol{\delta}^{(l)}\), the gradients for that layer's parameters can be concisely computed:
	\[
	\frac{\partial \mathcal{L}}{\partial \mathbf{W}^{(l)}} = \boldsymbol{\delta}^{(l)} (\mathbf{a}^{(l-1)})^T, \quad \frac{\partial \mathcal{L}}{\partial \mathbf{b}^{(l)}} = \boldsymbol{\delta}^{(l)}
	\]
\end{enumerate}
The brilliance of the backpropagation algorithm lies in the fact that through one forward pass and one backward pass, it efficiently computes the gradients for all parameters, with computational complexity comparable to forward propagation. This provides the crucial gradient information for subsequent parameter updates.
\\
Step Three: Parameter Update—Gradient Descent Drives Network Evolution

After obtaining the gradients, the final step is to use this gradient information to update the network parameters, thereby reducing the loss function value. The most commonly used optimization method is gradient descent and its various variants. Its core idea is simple yet profound: parameters are updated in the direction opposite to the gradient of the loss function (i.e., the direction of steepest descent).

\[
\mathbf{W}^{(l)} \leftarrow \mathbf{W}^{(l)} - \alpha \frac{\partial \mathcal{R}}{\partial \mathbf{W}^{(l)}}, \quad \mathbf{b}^{(l)} \leftarrow \mathbf{b}^{(l)} - \alpha \frac{\partial \mathcal{R}}{\partial \mathbf{b}^{(l)}}
\]

where \(\alpha > 0\) is the learning rate, controlling the step size of parameter updates—too small steps lead to slow convergence, while too large steps may overshoot the optimum or even cause divergence.

In practical training, we typically do not compute gradients on all samples at once (batch gradient descent). Instead, we use mini-batch stochastic gradient descent (Mini-batch SGD): each time, a small batch of samples is randomly selected, their average gradient is computed, and parameters are updated. This strategy achieves a good balance between computational efficiency and convergence stability.

These three steps—forward propagation to compute loss, backpropagation to compute gradients, and parameter update to optimize the network—constitute the basic training loop for neural networks. This loop repeats, the loss value gradually decreases, network performance improves step by step, and eventually, the ``right'' network capable of solving the practical problem is found.
\vspace{3mm}

\noindent\textcolor{structure3}{\textbf{Optimizers and Challenges}}

Training deep neural networks not only requires understanding the basic gradient descent framework but also faces numerous core mathematical and engineering challenges. The continuous improvement of optimization algorithms is key to addressing these challenges and driving the development of deep learning.

Standard gradient descent (also known as batch gradient descent) requires computing the gradient using the entire training set at each step. When the dataset is massive, this approach is computationally expensive and unsuitable for online learning scenarios. Therefore, stochastic gradient descent (SGD) and its variant—mini-batch stochastic gradient descent (Mini-batch SGD)—are more commonly used in practice. The latter randomly samples a small batch of data each time to compute the gradient and update parameters. Although the introduced randomness brings some noise, it greatly improves iteration efficiency, and appropriate noise can even help escape local minima.

However, standard SGD still has room for improvement. To make the training process more intelligent and efficient, researchers have proposed various adaptive optimization algorithms.
\begin{enumerate}
	\item SGD with Momentum
	
	The momentum method simulates the concept of inertia in the physical world. When updating parameters, it not only considers the current gradient but also accumulates historical gradient directions, forming a ``velocity'' effect. The update rule is:
	\[
	\mathbf{v} \leftarrow \beta \mathbf{v} - \alpha \nabla_{\theta} \mathcal{R}, \quad \theta \leftarrow \theta + \mathbf{v}
	\]
	where \(\beta\) is the momentum decay coefficient (typically around 0.9). This mechanism helps accelerate convergence, reduce oscillations, and makes it easier for the network to traverse flat regions and saddle points, especially when the loss landscape contains narrow ravines.
	
	\item Adam (Adaptive Moment Estimation)
	
	Adam is currently one of the most popular optimizers (Kingma \& Ba, 2015). It cleverly combines the advantages of momentum and adaptive learning rates. Adam maintains two state variables: an estimate of the first moment (mean) of the gradient \(m_t\) and an estimate of the second moment (uncentered variance) \(v_t\):
	\[
	m_t = \beta_1 m_{t-1} + (1-\beta_1) g_t, \quad v_t = \beta_2 v_{t-1} + (1-\beta_2) g_t^2
	\]
	where \(g_t\) is the current gradient. Then, bias correction is applied to these moment estimates, and finally, parameters are updated:
	\[
	\theta_t = \theta_{t-1} - \alpha \cdot \frac{\hat{m}_t}{\sqrt{\hat{v}_t} + \epsilon}
	\]
	Adam can automatically adjust the learning rate for each parameter, is relatively robust to hyperparameter choices, and performs excellently on numerous tasks in natural language processing, computer vision, etc., making it one of the preferred optimizers in deep learning practice.
	
	\item Learning Rate Scheduling
	
	Dynamically adjusting the learning rate \(\alpha\) is another important strategy. Common practices include: step decay (multiplying by a decay factor after a certain number of epochs), cosine annealing, or adaptive adjustment based on validation performance (e.g., reducing the learning rate when validation loss plateaus, known as ReduceLROnPlateau). Reasonable learning rate scheduling often leads to significant performance improvements.
\end{enumerate}
Through the organic combination of these techniques, modern deep learning training can achieve a better balance between convergence speed, stability, and final performance.

Deep learning also faces several other important challenges in practical training:
\begin{itemize}
	\item Non-convex optimization: The loss function of neural networks is a non-convex function in high-dimensional space, containing numerous local minima and saddle points. However, practice and some theoretical studies show that many local minima have similar loss values and may generalize well, and adaptive optimizers help escape certain saddle points.
	
	\item Vanishing and exploding gradients: In deep networks, the error signal \(\boldsymbol{\delta}^{(l)}\) needs to be backpropagated through multiple layers of multiplication. If the norm of the gradient at each layer is less than 1, successive multiplication causes exponential decay of the gradient, i.e., vanishing gradients; conversely, if greater than 1, it may cause exponential growth, i.e., exploding gradients. This problem severely hinders the training of deep networks. Techniques such as ReLU activation functions, appropriate weight initialization (e.g., Xavier, He initialization), batch normalization, and residual connections (the core idea of ResNet) have proven effective in mitigating these issues.
	
	\item Hyperparameter tuning: The choice of hyperparameters like learning rate \(\alpha\), network depth, width, regularization coefficient \(\lambda\), etc., greatly impacts model performance. Their optimization is itself a meta-optimization problem, typically addressed through methods like grid search, random search, or more advanced Bayesian optimization.
\end{itemize}
\vspace{3mm}

\noindent\textcolor{structure3}{\textbf{A Glimpse of Neural Network Applications}}

Neural networks have many important applications. Here are a few landmark results:

Image Classification and Pattern Recognition: This is the most classic application scenario for convolutional neural networks (CNNs). From a mathematical perspective, image classification essentially learns a complex function mapping from a high-dimensional pixel space to a discrete class space. This capability is widely used for recognizing mathematical symbols, parsing charts, and even automatically mining potential patterns from scientific images (like astronomical observations, biological microscopy images), greatly enhancing the processing efficiency of scientific data.

AlphaGo and Reinforcement Learning: AlphaGo, proposed by DeepMind, combined CNNs (policy network and value network for evaluating board positions) with Monte Carlo tree search. Its success demonstrated that neural networks can approximate or even surpass human expert-level complex decision functions. From a mathematical perspective, this breakthrough inspired the modeling of mathematical problems like theorem proving and algorithm search as reinforcement learning problems. In this framework, neural networks act as agents, autonomously learning optimal solving strategies through interaction with the environment (like a prover or computational environment). Representative works include AlphaProof, AlphaTensor, and FunSearch.

Symbolic Regression and Equation Discovery: Works represented by AI Feynman and Ramanujan Machine use neural networks to fit data and then attempt to extract or induce concise, human-readable mathematical expressions from the trained network structure, such as physical laws or mathematical identities. These methods build a bridge between ``black-box'' models and symbolic mathematics, aiding in the automatic discovery of new scientific laws.

Construction of Combinatorial Objects: In combinatorics, many challenging problems require constructing extremal structures with specific properties, such as large graphs or special set families. Neural networks, especially generative models, can learn to generate candidate constructions that satisfy constraints, thereby providing mathematicians with new conjectures and research directions. Related research, such as ``Constructions in combinatorics via neural networks,'' demonstrates the potential in this direction.

Scientific Computing: Physics-Informed Neural Networks (PINNs) incorporate partial differential equations (PDEs) describing physical processes as regularization terms into the loss function. The network must satisfy the physical laws defined by the PDEs while fitting observational data. This makes PINNs a flexible, mesh-free, data-driven approach suitable for scientific computing tasks like forward PDE solving and parameter inversion.

Despite numerous successes, neural networks still face a series of profound challenges:
\begin{itemize}
	\item Interpretability: The decision-making process of neural networks is often seen as a ``black box,'' making it difficult to provide human-understandable reasoning chains. How to make neural networks not only output predictions but also provide ``proofs'' or ``explanations'' that meet mathematical rigor is a core challenge in current neuro-symbolic methods research. Lack of interpretability limits their application in high-stakes fields like mathematical proof and medical diagnosis.
	
	\item Lack of Generalization Theory: Although deep neural networks exhibit strong generalization ability in practice, their theoretical foundation remains incomplete. Traditional statistical learning theory often provides generalization bounds that are too loose for over-parameterized deep models, failing to explain the observed success. Although there have been preliminary explorations, such as analyses of simplified models like linear networks or two-layer ReLU networks, and research on ``implicit regularization'' (gradient descent tends to converge to flat minima, which usually correspond to better generalization) and the ``double descent'' phenomenon, a unified, universal generalization theory for deep networks has not yet been established.
	
	\item Computational Cost: Training large neural networks requires massive computational resources and data, leading to high economic costs and raising concerns about environmental impact. High energy consumption and the concentration of hardware resources limit participation from some research groups and regions, making the accessibility of deep learning a non-negligible issue.
\end{itemize}

\section{Introduction to Reinforcement Learning}

The core idea of reinforcement learning is to allow an agent to learn how to take actions through trial and error during continuous interaction with an environment, in order to achieve long-term goals. It differs from supervised learning (which requires labeled data) and unsupervised learning (which seeks the intrinsic structure of data), as its learning signal comes from the environment's feedback on the agent's actions—rewards.

\noindent\textcolor{structure3}{\textbf{Basic Framework and Core Elements of Reinforcement Learning}}

The interaction process of reinforcement learning can be abstracted as a loop: at time \( t \), the agent observes the environment state \( s_t \) and selects an action \( a_t \) to execute accordingly. After receiving the action, the environment transitions to a new state \( s_{t+1} \) and provides an immediate reward \( r_t \). The agent's ultimate goal is to maximize the total cumulative reward throughout the entire process. Figure \ref{fig:agent-environment} (cited from Qiu Xipeng's ``Neural Networks and Deep Learning'' Figure 14.2) illustrates the interaction between the agent and the environment:
\begin{figure}[htbp]
	\centering
	\includegraphics[width=0.5\linewidth]{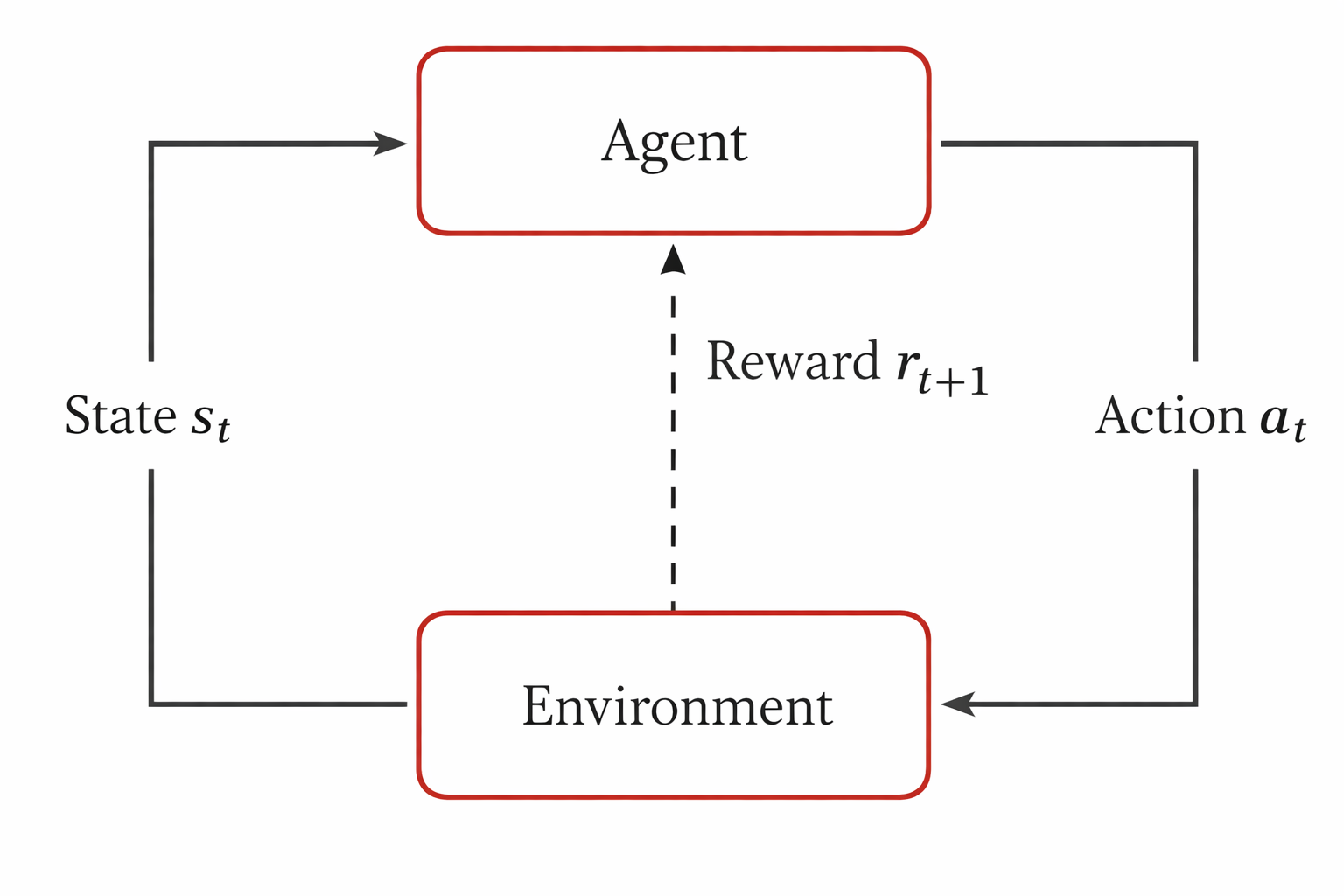}
	\caption{Schematic diagram of agent-environment interaction \label{fig:agent-environment}}
\end{figure}

\begin{figure}[htbp]
	\centering
	\includegraphics[width=0.7\linewidth]{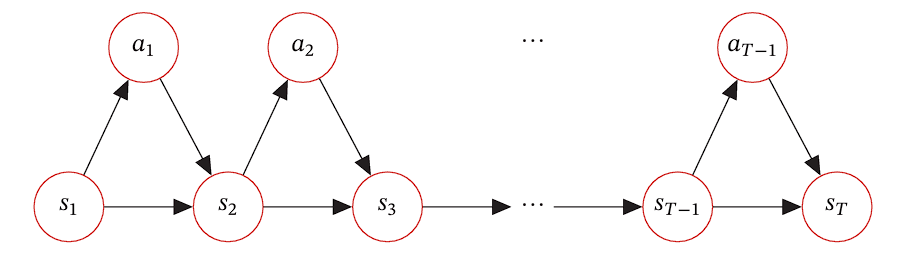}
	\caption{Markov diagram \label{fig:Markov}}
\end{figure}

This process is perfectly characterized mathematically by a Markov Decision Process (MDP). Figure \ref{fig:Markov} (cited from Qiu Xipeng's ``Neural Networks and Deep Learning'' Figure 14.3) shows the graphical model representation of an MDP. Its quintuple \(( \mathcal{S}, \mathcal{A}, P, R, \gamma )\) constitutes the fundamental elements of reinforcement learning.
\begin{enumerate}
	\item State Space (State Space, \( \mathcal{S} \)): The set of all possible environmental situations. A state \( s \) should contain all information necessary for decision-making (satisfying the Markov property).
	
	Example forms:
	
	Board games (e.g., Go, Chess): The state can be the layout of all pieces on the entire board.
	
	Video games (e.g., Atari games): The state can be raw pixel images from several consecutive frames to capture dynamic information.
	
	Robot control: The state can be the angles and angular velocities of the robot's joints, as well as external environment information collected by sensors (e.g., cameras, LiDAR, IMU).
	
	Recommendation systems: The state can include user profile features, historical click sequences, and current contextual information (e.g., time, device, location).
	
	\item Action Space (Action Space, \( \mathcal{A} \)): The set of all possible actions the agent can execute in a given state.
	
	Example forms:
	
	Discrete action space: Actions are enumerable. For example: {up, down, left, right} in a maze game; {placing a stone on one of the 361 intersections, passing} in Go.
	
	Continuous action space: Actions are real-valued vectors, typically used in scenarios requiring fine control. For example: steering wheel angle (continuous value), throttle and brake depth (continuous value) in autonomous driving; target torque or position increments for each joint motor in robotic arm control.
	
	\item State Transition Probability (Transition Probability, \( P(s' \mid s, a) \)): Defines the dynamic model of the environment. It represents the probability of the environment transitioning to state \( s' \) after executing action \( a \) in state \( s \).
	
	Example forms:
	
	In simple problems with known models (e.g., grid worlds), it can be an exact probability table. In most complex problems (e.g., real physical world, complex games), this transition model is unknown, complex, and stochastic. Reinforcement learning algorithms typically do not require prior knowledge of \( P \); instead, they learn optimal behavior indirectly through sampled experiences \((s, a, r, s')\) from interacting with the environment.
	
	\item Reward Function (Reward Function, \( R(s, a, s’) \)): The immediate feedback signal provided by the environment to the agent. It is the fundamental driving force and design core in reinforcement learning. It quantifies the immediate goodness or badness of taking action \( a \) in state \( s \) resulting in a transition to state \( s' \).
	
	The specific roles of the reward function are:
	\begin{itemize}
		\item Define the goal: The reward function essentially defines the task the agent needs to accomplish. All learning by the agent revolves around ``maximizing cumulative reward,'' so the design of the reward function directly determines what policy the agent ultimately learns.
		\item Guide learning: Through positive rewards (e.g., scoring, approaching a goal) and negative rewards (e.g., collision, losing points), it acts like a beacon, guiding the agent to explore beneficial behaviors and avoid harmful ones.
		\item Provide evaluation criteria: Value functions \( V(s) \) and \( Q(s, a) \) are both expectations of ``future cumulative reward''; reward is the foundation of these evaluation functions.
	\end{itemize}
	Below, we illustrate the reward rules of reinforcement learning with several concrete examples.
	
	For a maze navigation game: Reaching the goal gives $+10$ points; each step taken gives $-0.1$ points (encouraging finding the goal quickly); hitting a wall gives $-1$ point. The reward rule is as follows:
	\[
	R(s, a, s') = \begin{cases}
		+10, & \text{if } s’ \text{ is the goal state} \\
		-1, & \text{if } s' \text{ is a wall-collision state} \\
		-0.1, & \text{otherwise}
	\end{cases}
	\]
	For Go/Chess: Define winning as $+1$ point; losing as $-1$ point; draw or intermediate states as $0$ points. This is a typical example of sparse reward, where most moves have no immediate feedback, and the agent must learn to plan long-term for the final outcome.
	
	For autonomous driving: Safe driving within the lane gives $+0.01$ points; deviating from the lane gives $-0.1$ points; comfort (smooth acceleration/braking) gives $+0.001$ points; emergency braking or collision gives $-10$ points.
	
	For energy consumption management: Specify a fixed reward $+$ for completing a specific task; energy consumed $-$ (energy consumption amount × coefficient).
	
	\item Discount Factor (Discount Factor, \( \gamma \in [0, 1) \)): Used to calculate the present value of future rewards. A \( \gamma \) closer to 0 makes the agent more ``short-sighted,'' focusing only on immediate rewards; closer to 1 makes the agent more ``far-sighted,'' considering long-term returns. Its role is to mathematically guarantee that the cumulative reward over infinite time steps is bounded; in practice, it helps balance immediate and future gains.
\end{enumerate}
Under the definition of the above five elements, the agent's core task is to learn a policy \( \pi(a|s) \), which is the rule for selecting actions in each state (can be a deterministic function or a probability distribution). Its goal is to maximize the expected cumulative discounted reward (also called return):

\[
\mathbb{E}_{\pi}\left[ \sum_{t=0}^{\infty} \gamma^t R_t \right]
\]

To evaluate and find the optimal policy, two key value functions are introduced:
\begin{itemize}
	\item State-value function \( V^{\pi}(s) \): The expected return starting from state \( s \) under policy \( \pi \).
	\item Action-value function \( Q^{\pi}(s, a) \): The expected return after taking action \( a \) in state \( s \) under policy \( \pi \).
\end{itemize}
The \( Q^* \) function corresponding to the optimal policy \( \pi^* \) satisfies the Bellman optimality equation, which provides the theoretical foundation for many algorithms.
\vspace{3mm}

\noindent\textcolor{structure3}{\textbf{Major Algorithm Categories}}

\begin{enumerate}
	\item Value-based methods (e.g., Q-learning, DQN):
	Idea: Instead of learning the policy directly, first learn the optimal value function \( Q^*(s, a) \). The optimal policy naturally becomes selecting the action that maximizes the \( Q \) value: \( \pi^*(s) = \arg\max_a Q^*(s, a) \).
	The idea is not to learn the policy directly, but to first learn to assess ``which action is more valuable''—that is, to learn the optimal value function \(Q^*(s,a)\). The optimal policy is then: in each state, choose the action with the highest value. These methods are suitable for discrete action spaces and can efficiently reuse historical experience.
	
	\item Policy gradient methods (e.g., REINFORCE):
	The idea is to directly parameterize the policy \( \pi_\theta(a|s) \) (e.g., using a neural network), compute the gradient of the expected return with respect to the policy parameters \( \theta \), and use gradient ascent to optimize the policy. These methods are naturally suitable for continuous action spaces and can learn stochastic policies, but they often have high variance and less stable training.
	
	\item Actor-Critic methods:
	This approach combines the above two. The ``Actor'' is the policy function, responsible for selecting actions based on the state; the ``Critic'' is the value function (e.g., \( V(s) \) or \( Q(s, a) \)), responsible for evaluating the actor's performance and guiding its updates. This method uses the low-variance estimates provided by the critic to reduce the variance of the policy gradient, significantly improving learning efficiency and stability. This is the basic architecture of current mainstream reinforcement learning algorithms (e.g., A3C, PPO, SAC).
\end{enumerate}
Reinforcement learning is a goal-oriented, interactive learning paradigm. Its basic principles revolve around the MDP quintuple. The agent learns a policy that maximizes long-term return through the reward signals obtained from interacting with the environment. The design of the reward function is crucial to the success or failure of the task; it acts like the designer's ``baton'' and needs to precisely and carefully reflect the ultimate goal. From classic Q-learning to modern deep actor-critic algorithms, various methods are attempting to solve this core problem more efficiently and stably, achieving remarkable success in fields such as games, robotics, autonomous driving, and resource management.

\section{Introduction to Generative Artificial Intelligence}

According to the type of task, machine learning can be divided into two major categories: discriminative artificial intelligence and generative artificial intelligence.
The goal of discriminative artificial intelligence is to learn to ``draw boundaries'' between different data points and make judgments about unknown data. It answers the question ``What is this?''. For example, training a model to distinguish between pictures of cats and dogs, the model learns ``which combinations of pixel features are more likely to correspond to cats and which to dogs'', and establishes a decision boundary. When it sees a new picture, it judges which side of the boundary the picture falls on. Discriminative models care about the dividing line between categories, not the complete picture of the data itself.
The goal of generative artificial intelligence is even more ambitious: to understand and imitate the essential structure and distribution of data. It learns how the training data is generated—that is, the underlying probability distribution. Then, it can sample from the learned distribution to create brand new instances that conform to the data patterns. If discriminative models answer ``What is this?'', generative models answer ``How to create this kind of thing''. By learning the patterns hidden in massive data, it can generate text, images, code, music, and even scientific hypotheses that did not originally exist but conform to the patterns of that data.
\\
The achievements of generative AI have flooded into our work and lives, with its impact spanning multiple fields:
\begin{enumerate}
	\item Natural Language Processing:
	\begin{itemize}
		\item Intelligent Dialogue and Creation: Dialogue models represented by ChatGPT and Claude can engage in deep Q\&A, write articles, translate, program, etc.
		\item Code Generation: Tools like GitHub Copilot can automatically generate, complete, or explain code based on comments or context, greatly improving development efficiency.
	\end{itemize}
	\item Visual and Multimedia Creation:
	\begin{itemize}
		\item Text-to-Image/Video: Models like Midjourney, Stable Diffusion, and Sora can generate highly realistic or artistic images and dynamic videos based solely on a text description.
		\item Design Assistance: Automatically generate product prototypes, marketing materials, interior design drawings, etc.
	\end{itemize}
	\item Science and Discovery:
	\begin{itemize}
		\item Molecule and Material Design: For example, generating new drug molecules or battery material structures with specific properties, accelerating the R\&D process.
		\item Mathematics and Algorithm Discovery: Such as DeepMind's AlphaEvolve, which directly improved a core computational task that had not been optimized for 50 years by generating new algorithms.
	\end{itemize}
\end{enumerate}
The leap of generative AI is primarily built upon breakthroughs in three major technical approaches.
\vspace{3mm}

\noindent\textcolor{structure3}{\textbf{Generative Adversarial Networks: Evolving Through Gameplay}}

Generative Adversarial Networks (GANs) were proposed by Ian Goodfellow et al. in 2014. Their core idea is full of game theory wisdom: let two neural networks compete and co-evolve.

\begin{figure}[htbp]
	\centering
	\includegraphics[width=0.6\linewidth]{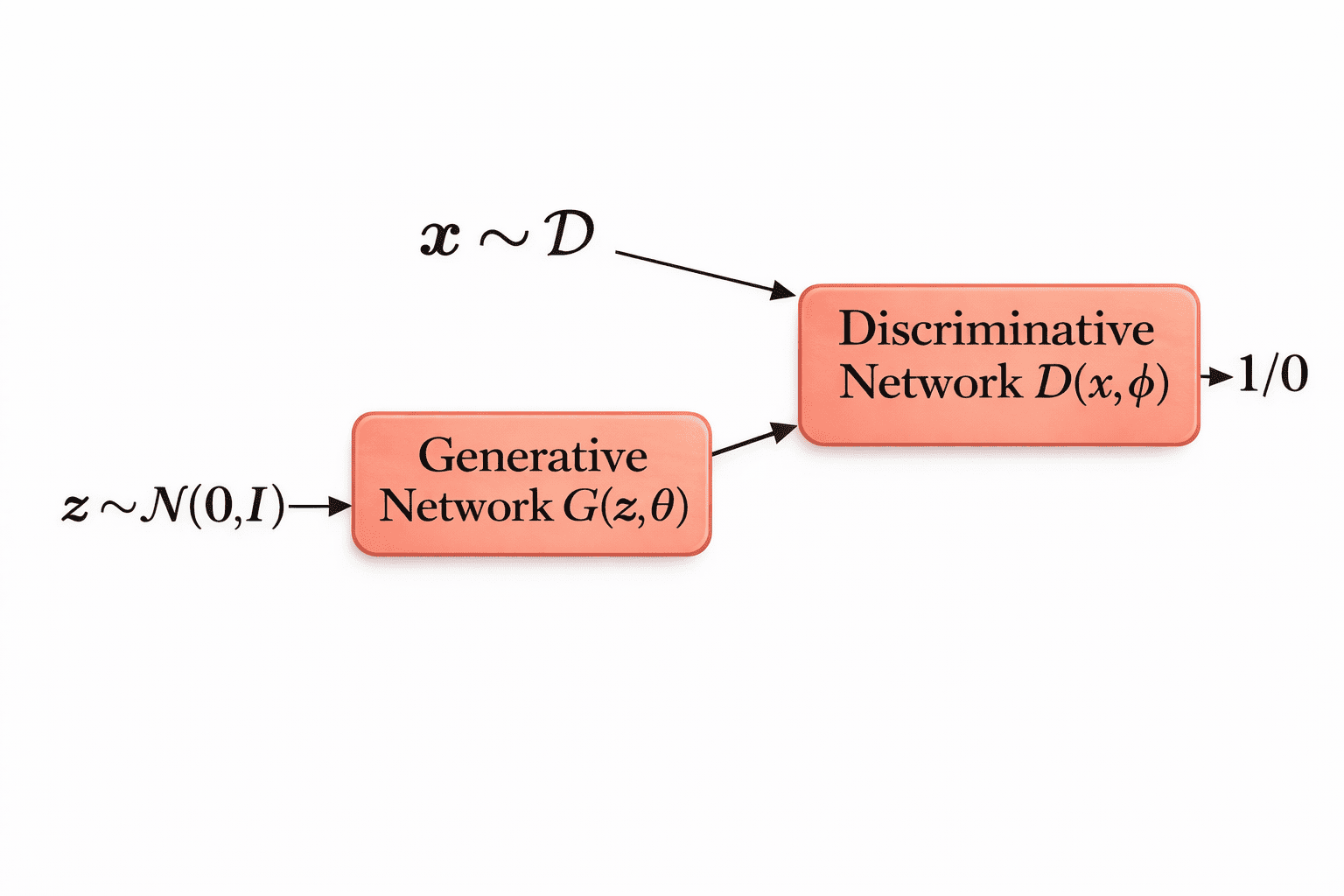}
	\caption{Generative Adversarial Network \label{fig:gan}}
\end{figure}

Basic Architecture: A GAN consists of two roles:

\begin{itemize}
	\item Generator: An artist trying to ``forge'' data. It receives a random noise vector \(z\) (usually sampled from a Gaussian or uniform distribution) and transforms it into a fake sample \(G(z)\), such as a fake image. The generator's goal is to make the generated sample realistic enough to be undetectable.
	\item Discriminator: An expert trying to ``authenticate''. It receives real samples \(x\) and fake samples \(G(z)\) produced by the generator, and judges whether each sample is real or fake. The discriminator's output is a probability value \(D(x)\), indicating the confidence that the sample is real.
\end{itemize}
The process of a generative adversarial network is shown in Figure \ref{fig:gan}.

\noindent During adversarial training, the two networks compete in a zero-sum game:

\begin{itemize}
	\item The generator continuously improves its forgery skills, trying to fool the discriminator into misclassifying fake samples as real.
	\item The discriminator continuously hones its discrimination ability, trying to expose the generator's tricks.
\end{itemize}
The entire training process can be formalized as a minimax game problem:

\[
\min_G \max_D V(D, G) = \mathbb{E}_{x \sim p_{\text{data}}}[\log D(x)] + \mathbb{E}_{z \sim p_z}[\log(1 - D(G(z)))]
\]
where \(p_{\text{data}}\) is the real data distribution and \(p_z\) is the noise distribution. Intuitive understanding: the discriminator \(D\) tries to maximize the log-likelihood of correctly classifying real and fake samples; the generator \(G\) tries to minimize the probability that the discriminator correctly identifies fake samples (i.e., maximize the chance of the discriminator making a mistake).

Dynamic Training Balance: Ideally, this game will eventually reach a Nash equilibrium—the samples produced by the generator are indistinguishable from real data, and the discriminator's judgment probability for any sample is close to 0.5 (completely uncertain about authenticity). At this point, the generator has learned the distribution of the real data.

However, the balance of this game is extremely delicate, and GAN training is therefore notorious for being unstable. The capabilities of the generator and discriminator need to grow synchronously—if the discriminator is too strong, it can easily expose all fake samples, causing the generator to stop learning due to ineffective gradients; if the generator is too strong, it may exploit certain weaknesses of the discriminator and only learn to generate a few types of samples that can fool the discriminator, a phenomenon known as mode collapse. Furthermore, when the discriminator is overconfident, its output gradients may approach zero, causing the generator to fall into the dilemma of vanishing gradients, making training unsustainable. Researchers have proposed numerous improvement schemes, such as WGAN, LSGAN, StyleGAN, etc., to alleviate these problems by improving loss functions, network architectures, and training strategies.
\\
Classic Application Examples:

\begin{itemize}
	\item Face Generation: The StyleGAN series of models can generate realistic faces with resolutions up to 1024×1024. These faces do not exist in reality, but the richness of detail is astonishing. Users can control attributes such as age, pose, hairstyle, and expression of the generated results.
	
	\item Image Translation: The Pix2Pix model can transform sketches into realistic photos (e.g., converting edge maps into pictures of cats), or convert daytime maps into nighttime maps. CycleGAN goes further, achieving image style transfer without paired data—for example, transforming ordinary horse photos into zebras, or applying Monet's painting style to any landscape photo.
	
	\item Super-Resolution: Models like SRGAN can restore high-resolution details from low-resolution images, with important applications in fields such as medical imaging and satellite imagery.
\end{itemize}
Despite the challenges of unstable training, GANs remain a milestone in the field of image generation due to their unique adversarial learning paradigm and stunning generative results.
\vspace{3mm}

\noindent\textcolor{structure3}{\textbf{Diffusion Models: From Chaos to Order}}

Diffusion Models are a technical approach that has risen to prominence in recent years and have become the ``ace'' in the current field of image generation—mainstream text-to-image tools like Stable Diffusion, DALLE 2, and Midjourney are all based on this technology. Its inspiration comes from the diffusion process in non-equilibrium thermodynamics.
\\
Its core idea is: Diffusion models consist of two opposite processes—the forward diffusion process and the reverse generative process.

Forward Diffusion Process: This process is fixed and non-trainable. We start with a real image \(x_0\) and gradually add small Gaussian noise to it. After \(T\) steps, the original image is completely corrupted, becoming a pure noise image \(x_T \sim \mathcal{N}(0, I)\). This process can be seen as the progressive destruction of information, with the noise addition at each step following a fixed pattern:

\[
q(x_t | x_{t-1}) = \mathcal{N}(x_t; \sqrt{1-\beta_t} x_{t-1}, \beta_t I)
\]
where \(\beta_t\) is a predefined noise schedule parameter that increases over time. A clever point is that, due to the additivity of Gaussian distributions, we can directly calculate the noisy result at any time step \(t\) from \(x_0\) in one step:

\[
x_t = \sqrt{\bar{\alpha}_t} x_0 + \sqrt{1-\bar{\alpha}_t} \epsilon, \quad \epsilon \sim \mathcal{N}(0, I)
\]
where \(\bar{\alpha}_t = \prod_{s=1}^t (1-\beta_s)\). This closed-form solution greatly simplifies the training process.

Reverse Generative Process: This is the part the model actually learns—how to start from pure noise \(x_T\), gradually denoise, and finally reconstruct a clear image \(x_0\). Each reverse step \(p_\theta(x_{t-1} | x_t)\) is modeled as a Gaussian distribution, with its mean and variance predicted by a neural network:

\[
p_\theta(x_{t-1} | x_t) = \mathcal{N}(x_{t-1}; \mu_\theta(x_t, t), \Sigma_\theta(x_t, t))
\]

In practical implementations (such as DDPM, Denoising Diffusion Probabilistic Models), the model is usually trained to predict the noise \(\epsilon\) added to \(x_0\), rather than directly predicting the image itself. The training objective simplifies to:

\[
\mathcal{L}_{\text{simple}} = \mathbb{E}_{x_0, \epsilon, t} \left[ \|\epsilon - \epsilon_\theta(x_t, t)\|^2 \right]
\]

This means the model learns: given a noisy image \(x_t\) and time step \(t\), predict what noise was added at that time. After mastering this ability, during generation, the predicted noise can be gradually subtracted to rebuild order from chaos.

To extend diffusion models to text-to-image tasks, the basic diffusion model described above can only generate images unconditionally. To achieve ``generating images based on text descriptions'', conditional control needs to be introduced. This is typically achieved through a cross-attention mechanism:
\begin{itemize}
	\item Use a text encoder (such as CLIP or T5) to convert the user's input text description into a text embedding vector.
	
	\item During the denoising process of the diffusion model, through cross-attention layers, let image features interact with text features, guiding the denoising process towards a direction that conforms to the text description.
\end{itemize}

Stable Diffusion further introduces the idea of Latent Diffusion Models: instead of performing diffusion directly in pixel space, first train an autoencoder to compress the image into a low-dimensional latent space, and then perform diffusion in the latent space. This significantly reduces computational costs while maintaining generation quality.

Example of the Generation Process: When a user inputs ``a corgi in an astronaut suit'' into Stable Diffusion, the model performs the following operations in sequence:
\begin{enumerate}
	\item The text encoder converts this sentence into a vector representation.
	\item The model starts from a random noise latent representation.
	\item After dozens of iterative denoising steps, each step is guided by referencing the text vector.
	\item The final latent representation is decoded by the decoder into a pixel image—a corgi wearing a miniature astronaut suit, with a serious expression, appears vividly on the page.
\end{enumerate}

Compared to GANs, diffusion models train more stably, are less prone to mode collapse, and can generate more diverse results. Their step-by-step denoising process also gives users more control—they can intervene mid-generation, perform image inpainting, editing, and other operations. Of course, the trade-off is slower generation speed (requiring dozens to hundreds of iterative steps), but accelerated sampling methods proposed in recent years (such as DDIM, DPM-Solver) have significantly improved efficiency.
\vspace{3mm}

\noindent\textcolor{structure3}{\textbf{Transformer Architecture and Large Language Models}}

This is the cornerstone of current generative AI (especially in the language field). Its core self-attention mechanism enables the model to process in parallel and deeply understand the complex relationships between all words in a text. By pre-training on terabytes of massive text, large language models acquire general knowledge and logic of language. Then, through alignment techniques like instruction fine-tuning, they are enabled to follow human instructions for creative work. The detailed principles of the Transformer architecture will be discussed in the next section, Introduction to Large Language Models, and will not be elaborated here.

Generative AI is evolving from a cool technology into a fundamental creative infrastructure. It is reshaping the ways of content creation, software development, and scientific research. Its core value lies in greatly unleashing human creativity, freeing us from repetitive labor, and allowing us to focus on higher-level strategic, aesthetic, and innovative decisions. With the deepening of multimodal integration (such as models that can simultaneously understand and generate text, images, and sound), an era of intelligent interaction driven by natural language has begun.

\section{Introduction to Large Language Models}

To understand how large language models (LLMs) work, we need to approach it from two levels: first, the overall architecture of the model, examining how information flows in and out; second, its core mechanisms, understanding how the model comprehends language.
\vspace{3mm}

\noindent\textcolor{structure3}{\textbf{I. The Overall Framework of Large Language Models: Input, Processing, Output}}

At its core, a large language model is a neural network system that predicts the next word based on input text. Taking the GPT series of models as an example, its workflow can be summarized in three stages:
\begin{enumerate}
	\item Input Processing: Converting text into vectors
	
	Computers cannot directly understand text, so the first step is to convert the input text into a numerical representation. This process includes:
	\begin{itemize}
		\item Tokenization: Splitting the text into smaller units (such as words or subwords). For example, ``I like AI'' might be tokenized into `[``I'', ``like'', ``AI'']`, resulting in a token sequence of length \(L\).
		
		\item Word Embedding: Mapping each token to a high-dimensional vector. Let the vocabulary size be \(V\) and the embedding dimension be \(d_{\text{model}}\). Then each token, based on its index, looks up its corresponding vector from an embedding matrix \(E \in \mathbb{R}^{V \times d_{\text{model}}}\):
		\[
		\mathbf{x}_i = E[\text{token}_i] \in \mathbb{R}^{d_{\text{model}}}
		\]
		These vectors carry semantic information about the words—words with similar meanings are close to each other in the vector space.
		
		\item Positional Encoding: Since the self-attention mechanism itself is order-insensitive, additional positional information needs to be injected. For position \(i\), its positional encoding \(\mathbf{p}_i\) is added to the word embedding:
		\[
		\mathbf{h}_i^{(0)} = \mathbf{x}_i + \mathbf{p}_i
		\]
		The positional encoding can be a fixed sine/cosine function (as in the original Transformer):
		\[
		\text{PE}(i, 2j) = \sin\left(\frac{i}{10000^{2j/d_{\text{model}}}}\right), \quad \text{PE}(i, 2j+1) = \cos\left(\frac{i}{10000^{2j/d_{\text{model}}}}\right)
		\]
		or it can be a learnable parameter. Finally, we obtain the input representation matrix \(H^{(0)} \in \mathbb{R}^{L \times d_{\text{model}}}\).
	\end{itemize}
	
	\item Core Processing: Feature extraction through multiple Transformer blocks
	
	This is the ``brain'' of the large language model. The input vector sequence passes sequentially through dozens or even hundreds of identical processing modules—Transformer blocks. Each block performs a ``global information fusion and local feature transformation'' on the sequence, allowing the representation at each position to continuously absorb contextual information.
	
	\item Output Generation: Predicting the next word
	
	After multi-layer processing, the model outputs a high-dimensional representation for each position. For the output \(\mathbf{h}_L^{(N)}\) at the last position, the model transforms it into a probability distribution over all words in the vocabulary via a linear transformation and a Softmax function:
	\[
	P(\text{next token} = v | \text{context}) = \frac{\exp(\mathbf{w}_v^T \mathbf{h}_L^{(N)})}{\sum_{v'=1}^{V} \exp(\mathbf{w}_{v'}^T \mathbf{h}_L^{(N)})}
	\]
	where \(\mathbf{w}_v \in \mathbb{R}^{d_{\text{model}}}\) is the output weight vector for word \(v\) (often shared with the input embedding matrix). The model selects the word with the highest probability (or samples according to the probability) as the prediction result. For example, given the input ``I like'', the model might predict the next word as ``AI'' (probability 0.3), ``eat'' (probability 0.2), ``study'' (probability 0.1), etc.
\end{enumerate}
This ``input-processing-output'' cycle constitutes the basic way large models generate text: predict a word, append the new word to the input, predict the next, and so on, until a complete sentence is generated.
\vspace{3mm}

\noindent\textcolor{structure3}{\textbf{II. Core Mechanism: Attention—Teaching the Model to ``Focus on What Matters''}}

In the above process, the most critical question is: when processing each word, how does the model know which parts of the context it should focus on? This is precisely the problem the attention mechanism solves.

\noindent \textcolor{structure3}{\textbf{Origin of the Idea: Allocating Attention Like Humans}}

When we read the sentence ``The cat in the detention center chases its own tail,'' to understand what ``its'' refers to, we need to look back to find ``cat''—this is the naive form of attention. The attention mechanism is designed to give the model this ability of ``selective focus'': when processing the current position, it dynamically decides which words in the context are more important and fuses their information.

Mathematically, the attention mechanism is a dynamic weighted average. It involves three basic roles:
\begin{itemize}
	\item Query (Q): Represents the current focus point, equivalent to asking ``Where should I look now?''
	
	\item Key (K): Represents the identifier for each context word, equivalent to ``Who am I''
	
	\item Value (V): Represents the actual content of each context word, equivalent to ``What information do I have''
\end{itemize}
Given a query vector \(\mathbf{q} \in \mathbb{R}^{d_k}\) and a set of key-value pairs \(\{(\mathbf{k}_i, \mathbf{v}_i)\}_{i=1}^{L}\), where \(\mathbf{k}_i \in \mathbb{R}^{d_k}, \mathbf{v}_i \in \mathbb{R}^{d_v}\), the attention output is:
\[
\text{Attention}(\mathbf{q}, K, V) = \sum_{i=1}^{L} \alpha_i \mathbf{v}_i
\]
where the weight \(\alpha_i\) is calculated from the similarity between the query and the key and normalized via softmax:
\[
\alpha_i = \frac{\exp(s(\mathbf{q}, \mathbf{k}_i))}{\sum_{j=1}^{L} \exp(s(\mathbf{q}, \mathbf{k}_j))}
\]
The similarity function \(s(\cdot, \cdot)\) is most commonly the scaled dot product:
\[
s(\mathbf{q}, \mathbf{k}) = \frac{\mathbf{q}^T \mathbf{k}}{\sqrt{d_k}}
\]
Here, \(\sqrt{d_k}\) is the scaling factor, used to prevent the dot product from becoming too large when the vector dimension is high, which could cause the softmax to enter a region with extremely small gradients.

In practical implementation, we compute attention for all positions in the entire sequence simultaneously, which can be written in matrix form:
\[
\text{Attention}(Q, K, V) = \text{softmax}\left(\frac{QK^T}{\sqrt{d_k}}\right) V
\]
where \(Q \in \mathbb{R}^{L \times d_k}, K \in \mathbb{R}^{L \times d_k}, V \in \mathbb{R}^{L \times d_v}\), and softmax is applied row-wise.

The weight \(\alpha_i\) intuitively reflects the degree of attention the current query pays to the \(i\)-th piece of information. Through learnable linear transformations, the model can dynamically generate \(Q, K, V\) vectors from the input sequence. This means that at every position when generating a sequence, the model can flexibly ``look back'' and focus on the most relevant content—this is the essence of the attention mechanism.

\begin{figure}[htbp]
	\centering
	\includegraphics[width=0.6\linewidth]{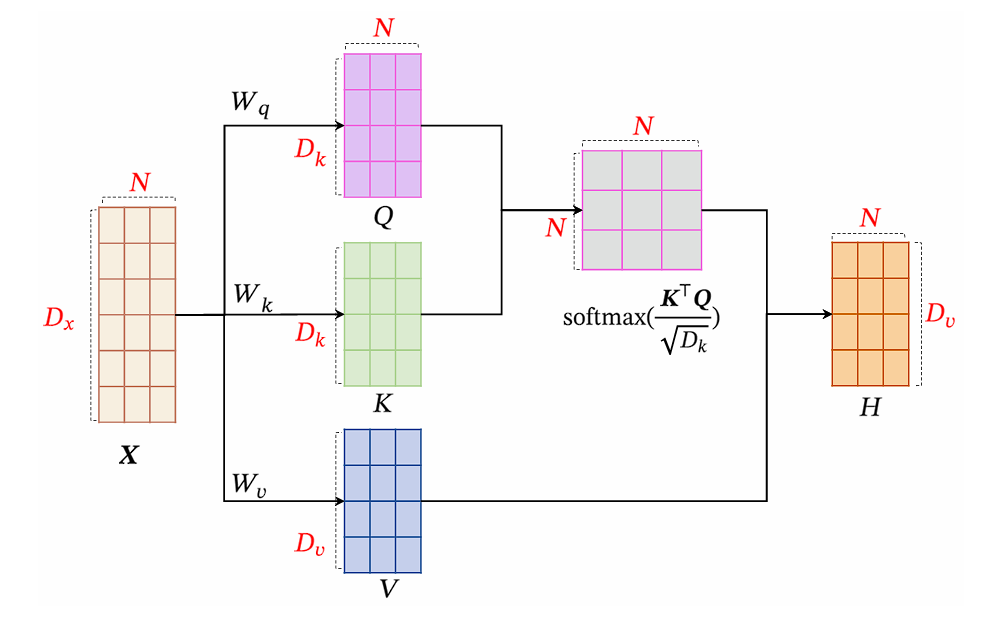}
	\caption{Attention computation process \label{fig:Attention}}
\end{figure}

The computational process of the attention mechanism is shown in Figure \ref{fig:Attention} (cited from Qiu Xipeng's ``Neural Networks and Deep Learning'' Figure 8.4).

\noindent\textcolor{structure3}{\textbf{Self-Attention}}

In large language models, we use a special form—Self-Attention. The ``self'' here means that the queries, keys, and values all come from different linear transformations of the same input sequence. Let the input matrix be \(X \in \mathbb{R}^{L \times d_{\text{model}}}\). We project it into query, key, and value spaces using three learnable weight matrices \(W^Q, W^K, W^V \in \mathbb{R}^{d_{\text{model}} \times d_k}\):
\[
Q = XW^Q, \quad K = XW^K, \quad V = XW^V
\]
Then we compute self-attention:
\[
\text{SelfAttention}(X) = \text{softmax}\left(\frac{QK^T}{\sqrt{d_k}}\right) V
\]
The ingenuity of this design lies in the fact that every word in the sequence is both a ``questioner'' (issuing queries) and a ``respondent'' (providing keys and values). Regardless of how far apart two words are in the sentence, self-attention allows them to establish a direct connection. In traditional models (like RNNs), the influence of ``cat'' on ``its'' needs to pass through multiple time steps, causing information to easily decay; whereas in self-attention, this influence is direct in one step, fundamentally solving the long-range dependency problem.
\vspace{3mm}

\noindent\textcolor{structure3}{\textbf{Multi-Head Attention: Understanding Language from Multiple Perspectives}}

Instead of computing only one set of attention, it's better to let the model observe from multiple angles simultaneously. Multi-Head Attention projects the queries, keys, and values into \(h\) different low-dimensional subspaces (each head has dimension \(d_k = d_{\text{model}} / h\)), computes attention independently on each head, and then concatenates the outputs of all heads and projects them again:

\[
\begin{aligned}
	&\text{head}_i = \text{Attention}(X W_i^Q, X W_i^K, X W_i^V) \\
	&\text{MultiHead}(X) = \text{Concat}(\text{head}_1, ..., \text{head}_h) W^O
\end{aligned}
\]
where \(W_i^Q, W_i^K, W_i^V \in \mathbb{R}^{d_{\text{model}} \times d_k}\), and \(W^O \in \mathbb{R}^{h d_v \times d_{\text{model}}}\) are all learnable projection matrices.

This is akin to having multiple ``experts'' read the same text simultaneously: one expert focuses on grammatical relationships (subject-verb-object), another on referential relationships (who refers to whom), and another on semantic associations (synonyms). The multi-head mechanism allows the model to capture different types of relationships in parallel, greatly enhancing its expressive power.
\vspace{3mm}

\noindent \textcolor{structure3}{\textbf{Structure of a Transformer Block}}

Combining the above components yields a standard Transformer encoder layer (or Transformer block). Its structure is as follows:
\begin{itemize}
	\item Multi-Head Self-Attention Layer: Computes the interrelationships among all positions in the sequence, allowing each position to aggregate information from other positions.
	
	\item First Residual Connection and Layer Normalization: Adds the input of the self-attention layer to its output (residual connection), then performs layer normalization. The residual connection provides a ``highway'' for gradients, greatly alleviating the vanishing gradient problem in deep networks; layer normalization stabilizes the training process.
	
	\item Feed-Forward Network: A simple two-layer fully connected network that independently transforms the representation at each position. This step introduces non-linearity and enhances the expressive power of each position.
	
	\item Second Residual Connection and Layer Normalization: Applies residual connection and layer normalization again, completing the processing of the entire block.
\end{itemize}
By stacking multiple such Transformer blocks, the model can build increasingly abstract representations layer by layer—lower layers may focus on lexicons and phrases, middle layers on syntactic structures, and higher layers may capture semantics and discourse relationships.

The Transformer network architecture is shown in Figure \ref{fig:Transformer} (cited from [Vaswanietal.,2017]):

\begin{figure}[htbp]
	\centering
	\includegraphics[width=0.5\linewidth]{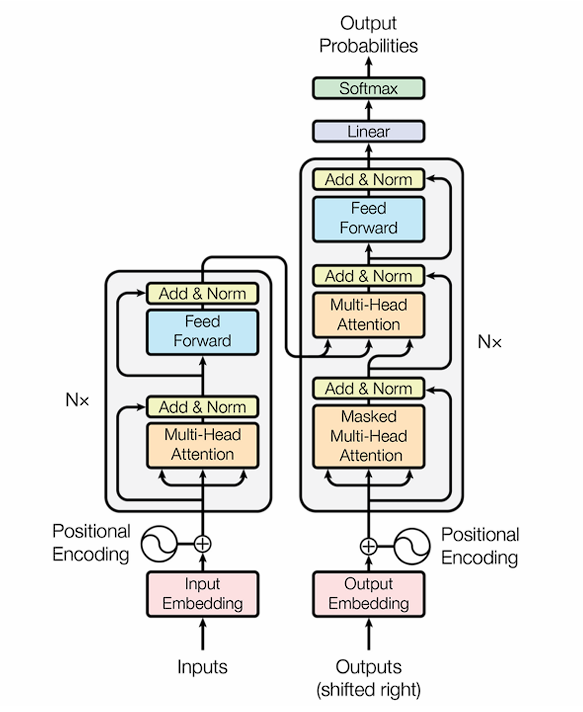}
	\caption{Transformer network architecture \label{fig:Transformer}}
\end{figure}

After understanding the above mechanisms, we can see why the Transformer architecture has become the ideal framework for processing large-scale sequential data (such as natural language, code, mathematical symbol sequences):
\begin{itemize}
	\item Parallel Computation: Self-attention can compute relationships among all positions at once, unlike RNNs which must process sequentially. This makes large-scale training possible.
	
	\item Long-Range Dependencies: Any two positions can interact directly, regardless of distance, without information decay.
	
	\item Hierarchical Understanding: Through multi-layer stacking, the model can build a deep understanding of language from local phrases to global semantics, layer by layer.
	
	\item Scalability: The architectural design of the Transformer is naturally suited to stacking more layers, using more heads, and processing longer sequences—this is the embodiment of the ``large'' in ``large language models''.
\end{itemize}
It is these advantages that enable Transformer-based large language models to learn the patterns of language from massive amounts of text and demonstrate astonishing capabilities in tasks such as understanding, generation, and reasoning.
\vspace{3mm}

\noindent\textcolor{structure3}{\textbf{III. Training Mechanisms and Process of Large Language Models}}

With the powerful Transformer architecture in place, the next core question emerges: How can we, through an engineered training process, endow the model with language understanding, logical reasoning, and even mathematical capabilities? The training of modern large language models is not a single stage but a meticulously designed multi-stage pipeline. Each stage has a different mission, collectively shaping the final, highly capable model.
\\
1. Pretraining: Building the Foundation of Language and Knowledge

Pretraining is the most resource-intensive and data-heavy stage of the entire training process. Its goal is to lay a solid foundation of language and broad world knowledge for the model. This stage belongs to self-supervised learning—the supervision signal does not come from manual annotation but is automatically generated from the structure of the data itself.

Data Format: Large-scale, diverse text corpora are used, including web pages, books, academic papers, code repositories, social media, etc. After tokenization, the data forms continuous token sequences as input to the model. This data requires no manual labeling.

Training Objective: Depending on the model architecture, there are two main paradigms for pretraining objectives:
\begin{itemize}
	\item Masked Language Modeling (MLM): Encoder models like BERT[Devlin J, Chang M W, Lee K, et al. BERT] adopt this strategy. The specific method is to randomly mask a portion of tokens in the input sequence (typically 15\%) and then train the model to predict the original masked tokens. This is similar to a ``cloze'' task, enabling the model to understand bidirectional context information. The labels here—the masked tokens—are extracted from the input sequence itself.
\end{itemize}
\begin{itemize}
	\item Next Token Prediction: Autoregressive decoder models like GPT adopt this strategy. Given a preceding token sequence, the model is trained to predict the next most likely token. This is similar to a ``word chain'' game, giving the model inherent text generation capability. The labels here—the actual next token in the sequence—also come from the data itself. Due to the universality of generation tasks, this paradigm has become the mainstream for pretraining current large language models.
\end{itemize}

Mathematically, for a sequence \(\mathbf{x} = (x_1, x_2, ..., x_T)\) of length \(T\), the training objective of an autoregressive language model is to maximize the following likelihood:

\[
\mathcal{L}_{\text{LM}} = \sum_{t=1}^{T} \log P(x_t | x_{1:t-1}; \theta)
\]
where \(\theta\) represents the model parameters, and \(P(x_t | x_{1:t-1}; \theta)\) is the conditional probability predicted by the model at the \(t\)-th position. Note that the supervision signal \(x_t\) here is precisely part of the input sequence itself—this is the core characteristic of self-supervised learning.

Pretraining such large-scale models requires a series of sophisticated engineering techniques:
\begin{itemize}
	\item Large-Scale Distributed Training: When model parameters reach hundreds of billions and training data reaches trillions of tokens, the memory and computational power of a single GPU are far from sufficient. Complex parallelization strategies must be employed to distribute computation across thousands of GPUs:
	\begin{itemize}
		\item Data Parallelism: Splits the training data across multiple devices, each holding a complete copy of the model, with periodic gradient synchronization.
		\item Tensor Parallelism: Splits the weight matrix of a single layer across multiple devices, collaboratively performing matrix operations.
		\item Pipeline Parallelism: Assigns different layers to different devices, forming a computational pipeline that processes multiple micro-batches simultaneously.
	\end{itemize}
	\item Mixed Precision Training: Using FP16 (half-precision) or BF16 floating-point formats for forward and backward propagation can significantly save GPU memory and accelerate computation. Meanwhile, parameters and optimizer states are kept in FP32 format in the master copy to ensure numerical stability. This is combined with dynamic loss scaling to avoid gradient underflow in half-precision.
	\item Activation Checkpointing: During backpropagation, all intermediate activations are typically saved for gradient computation, occupying a large amount of memory. Activation checkpointing selectively saves some activations and recomputes the rest when needed, trading time for space, which is particularly effective when memory is limited.
\end{itemize}

\noindent 2. Supervised Fine-Tuning (SFT): Aligning Instructions and Output Formats

Although pretrained models have acquired rich knowledge, they may not follow human instructions well or output specific formats. The Supervised Fine-Tuning (SFT) stage uses high-quality labeled data to fine-tune the model, aligning its behavior with human expectations. Unlike pretraining, this stage belongs to standard supervised learning—each training sample has a clear, manually annotated target.

Data Format: Uses carefully crafted human-written ``instruction-output'' pair datasets. For example, in the mathematics domain, the data format might be:
\begin{itemize}
	\item Instruction example: Solve the equation $x^2 - 5x + 6 = 0$
	\item Output example: Factoring gives $(x-2)(x-3)=0$, so the solutions are $x=2$ or $x=3$.
\end{itemize}

SFT requires extremely high data quality, covering diverse task types, and outputs should include correct reasoning steps. Datasets typically contain tens of thousands to hundreds of thousands of such examples.

Supervised fine-tuning is a standard supervised learning process. The instruction and desired output are concatenated as input, and the model is trained to generate the target sequence in the output part. The loss function is still the next token prediction loss, but typically only the loss on the output part is computed to avoid meaningless modeling of the instruction part.

Mathematically, for an instruction \(x\) and target output \(y\), the loss function is:

\[
\mathcal{L}_{\text{SFT}}(\theta) = -\sum_{t=1}^{|y|} \log P(y_t | x, y_{1:t-1}; \theta)
\]
Engineering techniques include:
\begin{itemize}
	\item Parameter-Efficient Fine-Tuning (PEFT): Large models have enormous parameters; fully fine-tuning all parameters is costly and prone to overfitting. LoRA (Low-Rank Adaptation) is currently the most widely used technique. Its core idea is to assume that the weight update \(\Delta W\) has a low-rank property. Therefore, the original weights \(W_0\) are frozen, and two much smaller trainable matrices \(A\) and \(B\) are introduced, such that \(\Delta W = BA\), and the updated weight is \(W = W_0 + BA\). During fine-tuning, only \(A\) and \(B\) are updated, reducing trainable parameters to one ten-thousandth or even less of the original, greatly lowering memory usage and overfitting risk.
	\item High-Quality Data Construction: The quality of SFT data directly determines the upper limit of fine-tuning effectiveness. Besides manual writing, the following strategies can be employed:
	\begin{itemize}
		\item Model-Assisted Construction: Use stronger base models to generate candidate answers, which are then screened and corrected by experts.
		\item Mining Community Resources: Collect and organize from high-quality resources like math competition solutions, technical Q\&A communities, etc.
		\item Data Augmentation: Expand the dataset through rewriting, translation, etc., to increase diversity.
	\end{itemize}
\end{itemize}

\noindent 3. Reinforcement Learning (RL): Optimization Based on Feedback

After SFT, the model can follow instructions, but output quality can still be improved—such as correctness, conciseness, safety, alignment with human preferences, etc. The Reinforcement Learning (RL) stage introduces a reward signal to further optimize the model, making its outputs better align with higher-order goals. This is a key stage for enhancing the model's mathematical reasoning ability and aligning with human values.
Before delving into specific techniques, let's review the basic framework of reinforcement learning. In the language model scenario:
\begin{itemize}
	\item Agent: The language model itself, whose policy \(\pi_\theta(a|s)\) is the probability distribution of choosing the next token \(a\) given the context \(s\) (the sequence of tokens generated so far).
	\item Environment: The interactive process of text generation, including the user's input prompt and the autoregressive generation environment of the model.
	\item State: The current text context, i.e., all tokens generated so far.
	\item Action: Choosing the next token.
	\item Reward: A quality score calculated based on the complete generated sequence, provided by an external evaluation mechanism.
\end{itemize}
The goal is to optimize the policy parameters \(\theta\) to maximize the expected cumulative reward:

\[
J(\theta) = \mathbb{E}_{\tau \sim \pi_\theta}[R(\tau)]
\]

where \(\tau = (s_0, a_0, s_1, a_1, ...)\) is a complete text generation trajectory, and \(R(\tau)\) is the total reward obtained for that trajectory.

Below are several typical reinforcement learning methods for large models:
\vspace{3mm}

\noindent\textcolor{second}{\textbf{Reinforcement Learning from Human Feedback (RLHF)}}

RLHF is the core alignment technology used by models like ChatGPT. It incorporates human preferences into the optimization objective, making model outputs more aligned with human values. RLHF typically consists of three steps:
\begin{enumerate}
	\item Collect Human Preference Data: Given a prompt, the initial model after SFT generates multiple different answers. Human annotators rank these answers (which is better, which is worse). This process produces a large number of preference pairs \((x, y_w, y_l)\), where \(x\) is the prompt, \(y_w\) is the preferred response, and \(y_l\) is the less preferred response.
	\item Train a Reward Model (RM): Use the collected preference pairs to train an independent reward model \(r_\varphi(y|x)\), teaching it to assign higher scores to text outputs that better align with human preferences. Preference modeling typically uses the Bradley-Terry model:
	
	\[
	P(y_w \succ y_l | x) = \sigma(r_\varphi(y_w|x) - r_\varphi(y_l|x))
	\]
	
	where \(\sigma\) is the sigmoid function. The training objective of the reward model is to maximize this likelihood probability:
	
	\[
	\mathcal{L}_{\text{RM}}(\varphi) = -\mathbb{E}_{(x, y_w, y_l)} [\log \sigma(r_\varphi(y_w|x) - r_\varphi(y_l|x))]
	\]
	
	After training, the reward model can provide a scalar score for any text output, reflecting its degree of alignment with human preferences.
	\item Use Reinforcement Learning to Optimize the Policy Model: Freeze the trained reward model \(r_\varphi\) and use the SFT-finetuned policy model \(\pi_{\text{SFT}}\) as the starting point for fine-tuning. Given a prompt \(x\), the policy model \(\pi_\theta\) generates a response \(y\) and receives a reward \(r_\varphi(y|x)\) from the reward model. Simultaneously, to prevent the policy model from deviating too much from the initial \(\pi_{\text{SFT}}\) (to avoid losing language capability or producing meaningless output), a KL divergence penalty term is added to the reward. Therefore, the total reward is:
	
	\[
	R(y|x) = r_\varphi(y|x) - \beta \cdot \text{KL}(\pi_\theta(\cdot|x) \| \pi_{\text{SFT}}(\cdot|x))
	\]
	
	where \(\beta\) is a penalty coefficient controlling the degree of deviation. The KL divergence term is calculated as:
	
	\[
	\text{KL}(\pi_\theta \| \pi_{\text{SFT}}) = \mathbb{E}_{y \sim \pi_\theta} \left[ \log \frac{\pi_\theta(y|x)}{\pi_{\text{SFT}}(y|x)} \right]
	\]
	
	Then, policy gradient algorithms are used to optimize the policy model \(\pi_\theta\) to maximize the expected reward \(J(\theta)\). Proximal Policy Optimization (PPO) is currently the most commonly used algorithm. It achieves stable and efficient policy updates by introducing importance sampling and a clipping mechanism.
\end{enumerate}
\vspace{3mm}

\noindent\textcolor{second}{\textbf{Direct Preference Optimization (DPO)}}

The RLHF process is complex, requiring training a separate reward model and using reinforcement learning algorithms, which can be unstable and involve many hyperparameters. Direct Preference Optimization (DPO) provides a more concise alternative, bypassing explicit reward model modeling and directly using human preference data to optimize the policy.

The core insight of DPO is: Under the Bradley-Terry preference model, there exists a closed-form analytical relationship between the optimal policy \(\pi^*\), the reward function \(r^*\), and the reference policy \(\pi_{\text{ref}}\):

\[
r^*(x, y) = \beta \log \frac{\pi^*(y|x)}{\pi_{\text{ref}}(y|x)} + \beta \log Z(x)
\]
where \(Z(x)\) is the partition function, independent of \(y\). Substituting this expression into the preference probability formula, a loss function directly based on the policy \(\pi_\theta\) can be derived:

\[
\mathcal{L}_{\text{DPO}}(\theta) = -\mathbb{E}_{(x, y_w, y_l)} \left[ \log \sigma \left( \beta \log \frac{\pi_\theta(y_w|x)}{\pi_{\text{ref}}(y_w|x)} - \beta \log \frac{\pi_\theta(y_l|x)}{\pi_{\text{ref}}(y_l|x)} \right) \right]
\]
Here, \(\pi_{\text{ref}}\) is typically the initial model after SFT. Optimizing this loss function directly drives the policy model to assign higher likelihood (relative to \(\pi_{\text{ref}}\)) to the preferred response \(y_w\) and lower likelihood to the dispreferred response \(y_l\).
DPO eliminates the need for reward model training and the complex reinforcement learning loop, making implementation simpler and training more stable. It has become a popular alternative to RLHF, achieving comparable or even better results on many tasks.
\vspace{3mm}

\noindent\textcolor{second}{\textbf{Group Relative Policy Optimization (GRPO) and Reinforcement Learning for Mathematical Reasoning}}

In tasks such as mathematical reasoning, answer quality often lies on a spectrum rather than a simple binary of correct or incorrect. This raises the question of how to effectively use diverse candidate answers to guide model learning. Group Relative Policy Optimization (GRPO) addresses this by modeling the relative quality of different candidate answers under the same input.

The core idea of GRPO is that it does not rely on binary preferences, but instead optimizes based on the relative quality within a group of answers. For the same mathematical problem, the model generates or collects multiple candidate answers, which can be grouped according to criteria such as correctness, completeness of reasoning steps, and conciseness.

For a given problem \(x\), the model generates a set of candidate outputs \(\{y_1, y_2, \dots, y_n\}\). Each candidate answer is assigned a quality score \(r(x, y_i)\), for example using a reward model or rule-based evaluation. GRPO does not directly optimize these absolute reward values; instead, it focuses on their relative relationships within the same group.

To this end, we define the \emph{relative advantage} of each candidate as the difference between its reward and the group mean:
\[
A(x, y_i) = r(x, y_i) - \frac{1}{n} \sum_{j=1}^{n} r(x, y_j)
\]
This quantity captures how a particular answer compares to others in the same group. If an answer has a score above the average, its advantage is positive; otherwise, it is negative.

Based on this formulation, the optimization objective of GRPO can be written as:
\[
\mathbb{E}_{y \sim \pi_\theta(\cdot|x)} \left[ A(x, y) \cdot \log \pi_\theta(y|x) \right]
\]

Intuitively, this objective increases the probability of answers that are better than the group average, and decreases the probability of those that are worse. In this way, the model gradually learns to prefer higher-quality outputs for the same problem.

From a higher-level perspective, the key idea of GRPO is to replace absolute evaluation with relative comparison. This design has several advantages. First, it reduces dependence on the exact scale of the reward function, making training more robust to reward magnitudes. Second, by comparing answers within the same input, it helps reduce variance and improves training stability. In addition, it naturally leverages ranking information rather than relying only on binary labels, which is particularly useful in tasks such as mathematical reasoning and code generation.

Therefore, GRPO can be understood as a policy optimization method based on within-group relative ranking. By reinforcing the principle that better answers should be more likely to be generated, it guides the model toward more reliable generation behavior.

\noindent GRPO is well-suited for mathematical reasoning because:
\begin{itemize}
	\item Learning \emph{why an answer is wrong} is as important as learning \emph{how to obtain the correct answer}
	
	\item Answer quality often exists at multiple levels, and binary preferences cannot fully capture this information
	
	\item Between-group comparison encourages the model to learn more robust and efficient reasoning patterns
\end{itemize}

The idea of GRPO can be flexibly combined with methods such as DPO or reward model-based approaches. For example, a calibrated reward model can be used to score each answer, after which answers are dynamically grouped based on their scores, followed by group-based comparative optimization.

In summary, the training of modern large language models is a multi-stage, progressive process, with three learning paradigms each playing its role. These three stages build upon each other, each with its own focus, collectively shaping the final large language model. Pretraining provides breadth and depth, supervised fine-tuning provides format alignment, and reinforcement learning provides value alignment and fine-grained optimization. It is this meticulously designed training pipeline that enables models to demonstrate near-human-level capabilities in complex tasks such as mathematical reasoning, code generation, and conversational interaction.

\begin{table}[htbp]
	\caption{Three Reinforcement Learning Paradigms\label{tab:强化学习范式}}
	\centering
	\begin{tabular}{p{4cm} p{4cm} p{4cm} p{4cm}}
		\toprule
		\textcolor{structure3}{\textbf{Stage}} & \textcolor{structure3}{\textbf{Learning Paradigm}} & \textcolor{structure3}{\textbf{Source of Supervision Signal}} &
		\textcolor{structure3}{\textbf{Core Objective}}\\
		\midrule
		Pretraining	& Self-Supervised Learning	& Structure of the data itself	& Learn language patterns and world knowledge, establish foundational capabilities \\
		\addlinespace
		
		Supervised Fine-Tuning	& Supervised Learning	& Annotations	& Teach the model to follow instructions and output formats \\
		\addlinespace
		
		Reinforcement Learning	& Reinforcement Learning & Reward signal & Optimize output quality, align with human preferences and values \\
		\bottomrule
	\end{tabular}
\end{table}

\section{Engineering Techniques}

The core idea of machine learning engineering practice can be summarized as ``control and guidance''—through a series of carefully designed technical means, control the inherent uncertainty of the optimization process and guide the model parameters to converge to regions with good generalization performance. This is far from a simple accumulation of technical details; it is a profound understanding and comprehensive application of optimization dynamics, generalization theory, and computational resource management.

The final performance of a model depends not only on the algorithm itself but also on the meticulous design and coordinated cooperation of every link in the engineering practice. This section will delve into the key engineering aspects of the machine learning pipeline—data preparation, model initialization, training optimization, and regularization—revealing the mathematical intuition and design principles behind them.
\vspace{3mm}

\noindent\textcolor{structure3}{\textbf{1. Data Preparation: The Foundation of Model Learning}}

Machine learning models ``learn'' patterns from data; therefore, the quality, scale, and diversity of the data directly determine the upper limit of model performance. As the computer science adage goes, ``Garbage In, Garbage Out.'' No matter how sophisticated the algorithm, if the input data is flawed, the model ultimately cannot learn correct knowledge.

\noindent 1. Data Collection: Building a High-Quality Data Source

In machine learning for mathematical tasks, data sources exhibit rich diversity, each with its unique value and challenges.

Symbolic data is the most direct carrier of mathematical knowledge, typically extracted from structured mathematical databases. For example, the On-Line Encyclopedia of Integer Sequences (OEIS) contains hundreds of thousands of integer sequences and their mathematical properties; arXiv's mathematical papers, textbooks, and theorem-proof documents contain vast amounts of formulas, theorems, and proof steps. The characteristics of such data are clear structure and precise semantics, but they often require complex parsing techniques to restore them from PDF or LaTeX source code into usable structured forms.

Generated data is automatically synthesized according to specific mathematical rules or constraints and plays an increasingly important role in mathematical reasoning tasks. For instance, we can design a random polynomial generator to produce quadratic equations like \(ax^2 + bx + c = 0\) and automatically compute their discriminants and roots; or generate adjacency matrices of specific graph structures and their corresponding graph-theoretic properties (such as diameter, chromatic number). The advantages of generated data are scalability, naturally existing labels, and precise control over the coverage of the data distribution. However, generation rules must be carefully designed to avoid producing ``pseudo-data'' that contradicts real mathematical patterns.

Simulation data is common in machine learning for solving scientific and engineering problems, especially tasks involving differential equations and physical simulations. For example, when solving partial differential equations, we can sample points on the computational domain grid and obtain approximate solutions of the equation through numerical solvers, forming ``input-output'' pairs. The quality of simulation data depends on the accuracy of the numerical method and the rationality of the sampling strategy.

\noindent 2. Data Cleaning: Stripping Noise, Preserving Essence

Collected raw data is often ``dirty,'' especially when sourced from the internet. The goal of data cleaning is to reduce noise in the data, allowing the model to focus on genuine mathematical patterns. For mathematical text, cleaning involves multiple levels:

First is format normalization. Mathematical expressions have multiple representation forms; for example, the multiplication symbol might be ``×'', ``·'', or ``*'', which need to be unified into a standard form; commands in LaTeX source code like ``\verb|\frac{a}{b}|'' need to be correctly parsed into structured fraction representations. Second is noise filtering. Mathematical content crawled from the internet is often mixed with advertisements, navigation bars, irrelevant comments, and other interference, requiring heuristic rules or trained classifiers to filter them out. For scanned literature, errors from Optical Character Recognition (OCR) also need correction—for instance, misidentifying ``$x^2$'' as ``$x$2'' is a common problem that requires correction based on context or mathematical grammar.

In the training of large language models, data cleaning is even more massive and crucial. Taking the training of a model with hundreds of billions of parameters as an example, the training data volume typically reaches trillions of tokens. The cleaning process involves deduplication (removing highly similar documents), toxic content filtering, privacy information desensitization, format standardization, and other steps. This process requires designing efficient distributed processing pipelines to ensure data quality while controlling computational costs.

\noindent 3. Data Representation: Bridging the Mathematical World and Neural Networks

Data representation is the key step of transforming raw mathematical objects into forms processable by models (usually numerical vectors or tensors), and it is also a core challenge in AI for Math. A good representation should preserve the essential structure of mathematical objects while being convenient for neural network processing.

Text and symbolic representation is the most common approach. For mathematical expressions, such as a first-order logic formula \(\forall x \exists y (x + y = 0)\), we first tokenize it into a discrete token sequence: `[“$\forall$”, “x”, “$\exists$”, “y”, “(”, “x”, “+”, “y”, “=”, “0”, “)”]`. This requires the tokenizer to understand the grammatical roles of mathematical symbols—for example, distinguishing ``x'' as a variable from ``×'' as a multiplication symbol. Current large language models widely adopt subword tokenization algorithms, such as Byte-Pair Encoding (BPE), which can discover frequently occurring subword units from a statistical perspective, effectively handling combinations of rare words and mathematical symbols. The tokenized sequence is then mapped to dense vectors through an embedding layer, where each token is represented as a high-dimensional vector, and semantically similar tokens are close to each other in the vector space.

Graph and combinatorial structure representation is a common way to handle discrete mathematical objects. A graph can be represented by an adjacency matrix, where element \(A_{ij} = 1\) indicates node \(i\) is connected to node \(j\), otherwise 0. For weighted graphs, the elements of the adjacency matrix can be real-valued weights. For polyhedra or more complex topological spaces, their combinatorial structure can also be described by similar 0-1 matrices—for example, the face-edge incidence matrix of a polyhedron. The emerging Graph Neural Networks (GNNs) in recent years are specifically designed to handle such structures, learning vector representations for nodes, edges, and even entire graphs through message passing and neighbor aggregation on graphs.

Representation of abstract algebraic structures is more difficult, requiring deep mathematical knowledge to design. For an abstract group, we cannot directly input it into a neural network, but we can compute its numerical invariants—for example, the distribution of element orders, the number of conjugacy classes, the character table, etc., and form these numerical values into a feature vector. For a manifold, we can compute its topological invariants—such as Betti numbers (the ranks of homology groups in various dimensions), homotopy groups, curvature integrals, etc. Another approach is to use coefficients of function spaces on the manifold as representations, such as expansion coefficients of eigenfunctions (harmonic functions) of the Laplace operator on the manifold. The core requirement of these representations is to maintain invariance under transformations—for example, regardless of how we relabel the elements of a group, the representation should remain unchanged.

\noindent 4. Data Augmentation: Creating Infinite Possibilities from Limited Data

When real data is scarce, data augmentation artificially expands the dataset by applying reasonable transformations to existing data to improve model robustness and generalization ability. Its mathematical essence is to apply a transformation that preserves invariance to the input space without changing the ``semantics'' of the data labels.

In scenarios where real data is scarce, data augmentation artificially expands the dataset by applying reasonable transformations to existing data to improve model robustness and generalization ability. Its mathematical essence is: without changing the ``semantics'' of the data labels, apply a transformation to the input space that preserves invariance.

In the field of computer vision, data augmentation is already mature—rotating, flipping, cropping, or adding noise to images, these transformations do not change the image category (a cat rotated is still a cat). In mathematical tasks, data augmentation can also play an important role, with its core being the utilization of symmetry, invariance, and equivalence relations of mathematical objects.

Variable substitution and scaling are the most intuitive augmentation methods. In algebraic expressions, systematically replacing variable names—for example, replacing all \(x, y\) in the equation \(x^2 + y = 1\) with \(a, b\)—keeps the solution set structure unchanged. Multiplying both sides of an equation by a non-zero constant simultaneously, or performing linear combinations on a system of equations, can also generate new equivalent forms. This utilizes the isomorphic relationship of mathematical objects—variable names are just labels; the real mathematical structure lies in the interactions between variables.

Rewriting into equivalent forms utilizes mathematical identities for transformation. For example, rewriting the expression \(\log(ab)\) as \(\log a + \log b\) (within the domain), or rewriting \(\sin^2 x + \cos^2 x\) as 1. This type of augmentation teaches the model to focus on the equivalence class of expressions rather than specific syntactic forms, helping the model learn algebraic rules of mathematical operations.

Logical transformations are particularly useful in theorem proving and logical reasoning tasks. Given a theorem ``if \(A\) then \(B\)'', we can generate its contrapositive ``if not \(B\) then not \(A\)'' as new data; for the equivalence relation ``\(A\) if and only if \(B\)'', we can generate four directional implication relationships. This type of augmentation helps the model understand the symmetry of logical structures.

Parameter perturbation is effective in numerical computation and geometric problems. For geometry problems, we can randomly change the dimensions (such as side lengths of a triangle) or angles within a reasonable range while keeping the geometric relationships of the problem unchanged; for algebraic calculation problems, we can perturb equation coefficients within the range allowed by numerical stability to generate new instances. This forces the model to learn the structural features of the problem rather than specific numerical coincidences.

Data augmentation is an effective, low-cost means of injecting mathematical knowledge (such as symmetry, invariance, equivalence relations), which can significantly improve the model's generalization ability and data efficiency.
\vspace{3mm}

\noindent\textcolor{structure3}{\textbf{2. Model Initialization: The Starting Point of the Training Journey}}

The initial parameter values determine the starting point of the optimization process. In the vast landscape of non-convex optimization, the choice of this starting point is crucial—a poor starting point may cause the model to converge to a poor local minimum or encounter numerical instability early in training. Just as a climber needs to choose a suitable base camp to start the ascent, model initialization is a key step that lays the foundation for the entire training journey.

\noindent 1. Avoiding Zero Initialization

Intuitively, initializing all weights and biases to zero seems like a ``neutral'' starting point. However, for multi-layer neural networks, this initialization leads to severe symmetry issues, preventing the model from learning effectively.

Consider a fully connected layer as an example. If all weights \(w_{ij} = 0\), then the linear input for each neuron in this layer is
\[
z_j = \sum_i w_{ij} x_i + b_j = b_j
\]
It can be seen that at this point, the neuron's output does not depend on the input \(x\) at all but is determined solely by the bias.

If we further assume all biases \(b_j\) are also the same (e.g., all initialized to 0), then the outputs of all neurons in this layer will be completely identical. In this case, each neuron produces the same activation value during forward propagation; during backpropagation, since the error signals and inputs they receive are also identical, their gradient updates will be exactly the same. Therefore, throughout the training process, these neurons always maintain the same parameter values and learn exactly the same features.

Even if the biases \(b_j\) are different, since the weights are zero, the neuron outputs still contain no information from the input features. The network in the initial stage is equivalent to a model relying only on constant offsets, making it difficult to effectively learn the relationship between input and output.

Therefore, the root of the problem is: when weights are initialized to the same value (especially zero), the model cannot break the symmetry between neurons, causing multiple neurons to degenerate into functionally equivalent units, thereby significantly reducing the model's expressive power. This phenomenon is called failure to break symmetry, which also explains why in practice, weights must be randomly initialized to introduce differences between parameters.

\noindent 2. Random Initialization: Scale is Key

The most direct method to break symmetry is random initialization—sampling each parameter independently from some probability distribution. However, random does not mean arbitrary; the choice of scale becomes the key to success or failure. Suppose we sample weights \(W\) from a distribution with zero mean and variance \(\sigma^2\). Consider a simple forward propagation: \(z = Wx + b\), where \(x\) is the input vector. If \(\sigma\) is too large, the weight values may cause the magnitude of \(z\) to rapidly inflate, and after multi-layer propagation, activation values tend to infinity, leading to gradient explosion; if \(\sigma\) is too small, activation values rapidly shrink to zero, leading to gradient vanishing. Both scenarios can paralyze training—the former causes numerical overflow, the latter makes parameter updates almost stall.

So, what scale is ``just right''? The answer lies in variance analysis.

\noindent 3. Variance Scaling-Based Initialization

The core idea of this class of initialization methods is: during initialization, try to keep the variance of activation values in forward propagation and the variance of gradients in backpropagation stable. This essentially designs a delicate statistical balance to allow signals to flow smoothly through the network.

Xavier initialization (also known as Glorot initialization) is an early exemplar of this idea. It assumes the activation function is linear and symmetric near the origin (like the tanh function) and derives that the variance of weights should be set to:

\[
\text{Var}(W) = \frac{2}{n_{\text{in}} + n_{\text{out}}}
\]
where \(n_{\text{in}}\) is the number of input neurons and \(n_{\text{out}}\) is the number of output neurons. This setting ensures that the variance of activation values in each layer remains equal during forward propagation, while the variance of gradients also remains equal during backpropagation. When sampling from a uniform distribution, the corresponding range is \(W \sim \mathcal{U}[-\sqrt{\frac{6}{n_{\text{in}}+n_{\text{out}}}}, \sqrt{\frac{6}{n_{\text{in}}+n_{\text{out}}}}]\).

However, Xavier initialization targets linear activation functions. When ReLU became mainstream, researchers found Xavier performed poorly on ReLU networks. The reason is: ReLU sets half of the negative input values to zero, halving the output variance. To compensate for this ``information loss,'' He initialization (also known as Kaiming initialization) emerged:

\[
\text{Var}(W) = \frac{2}{n_{\text{in}}}
\]

For a uniform distribution, the corresponding range is \(W \sim \mathcal{U}[-\sqrt{\frac{6}{n_{\text{in}}}}, \sqrt{\frac{6}{n_{\text{in}}}}]\); for a normal distribution, it is \(W \sim \mathcal{N}(0, \frac{2}{n_{\text{in}}})\). The intuition of He initialization is: since ReLU halves the variance, double the variance during initialization to maintain overall variance stability. This simple adjustment made training deep ReLU networks possible.

\noindent 4. Orthogonal Initialization

Variance scaling methods ensure stable signal propagation from a statistical perspective, while orthogonal initialization provides another elegant idea from a geometric viewpoint.

Consider a linear layer \(y = Wx\). We want to keep the vector norm stable throughout the network—i.e., \(\|y\|\) as close as possible to \(\|x\|\). If \(W\) is an orthogonal matrix (i.e., \(W^T W = I\)), then for any vector \(x\), we have \(\|Wx\| = \|x\|\). Orthogonal matrices perfectly preserve vector norms without introducing scaling effects.

Extending this idea to neural network initialization: we initialize the weight matrix as an (approximately) orthogonal matrix. In specific implementation, we can sample a random matrix from a standard normal distribution and then orthogonalize it via QR decomposition or singular value decomposition. Although activation functions and nonlinearities in real networks break this ideal norm-preserving property, orthogonal initialization still greatly alleviates gradient vanishing and explosion problems, especially suitable for Recurrent Neural Networks (RNNs) and scenarios requiring long-term dependency modeling.

The deep charm of orthogonal initialization lies in that it is not merely a numerical trick but embodies a geometric understanding of information flow—we want the input signal to preserve its ``energy'' as it propagates through the network, avoiding attenuation or explosion due to layer-by-layer scaling. This geometric intuition complements the statistical intuition of variance scaling methods, together forming the theoretical foundation of modern deep learning initialization.
\vspace{3mm}

\noindent\textcolor{structure3}{\textbf{3. Training Process Optimization: Navigating Non-Convex Terrain}}

Training deep neural networks is essentially searching for the lowest point on a high-dimensional non-convex function landscape—a complex terrain filled with peaks, valleys, saddle points, and flat regions. The core goal of training process optimization is to enable gradient descent and its variants to traverse this complex terrain efficiently and stably, descending quickly while avoiding getting stuck in poor local minima, ultimately reaching parameter regions with good generalization.

\noindent 1. The Family of Optimization Algorithms: From Naive Gradient Descent to Adaptive Methods

Stochastic Gradient Descent and Mini-batch Gradient Descent form the foundation of all modern optimization algorithms. Unlike Batch Gradient Descent, which uses all data in each iteration, Mini-batch Gradient Descent randomly samples a small batch of examples and uses the average of their gradients as an estimate of the true gradient. Let the parameters at step \(t\) be \(\theta_{t-1}\), and a small batch of samples randomly sampled from the training set be \(B\). Then the gradient estimate is:

\[
g_t = \frac{1}{|B|} \sum_{i \in B} \nabla_\theta \mathcal{L}_i(\theta_{t-1})
\]
where \(|B|\) is the mini-batch size, \(\mathcal{L}_i\) is the loss function value for the \(i\)-th sample, and \(\nabla_\theta \mathcal{L}_i\) is the gradient of the loss function with respect to parameters \(\theta\). This approach brings a dual effect: on one hand, computational cost is greatly reduced, making training on large-scale datasets possible; on the other hand, the random noise introduced by mini-batch sampling actually becomes a ``beneficial perturbation'' for the optimization process—it can help the model escape poor local minima and explore a broader parameter space. The mini-batch size embodies the classic bias-variance trade-off: large batches provide more accurate gradient estimates (low variance) but are computationally expensive and may get stuck in sharp minima; small batches introduce more noise (high variance) but may lead to better generalization performance.

Momentum is inspired by physics—imagine a ball rolling down a hill; it accumulates speed, traverses flat areas, and resists local bumps. Mathematically, momentum maintains a velocity vector \(v_t\), which is an exponentially weighted moving average of historical gradients:

\[
v_t = \beta v_{t-1} + \eta g_t, \quad \theta_t = \theta_{t-1} - v_t
\]
where \(\beta\) is the momentum decay coefficient, controlling the decay rate of historical information; \(\eta\) is the learning rate; \(g_t\) is the gradient at the current step. The effect of momentum is: accelerating progress in regions where gradient directions are consistent, smoothing updates in regions where gradient directions oscillate, thereby effectively alleviating ill-conditioned curvature problems and helping the model traverse saddle points and flat regions.

Adaptive learning rate algorithms push optimization to new heights—they assign independent adaptive learning rates to each parameter, dynamically adjusting the update step size based on the parameter's historical gradient information. The most outstanding representative is the Adam algorithm. Adam maintains two state variables: the first moment estimate of gradients \(m_t\) (i.e., gradient mean with momentum) and the second moment estimate \(v_t\) (i.e., mean of squared gradients, reflecting the magnitude of gradient variation):

\[
m_t = \beta_1 m_{t-1} + (1-\beta_1) g_t, \quad v_t = \beta_2 v_{t-1} + (1-\beta_2) g_t^2
\]
where \(\beta_1\) and \(\beta_2\) are decay coefficients; \(g_t\) is the gradient at the current step; \(g_t^2\) denotes element-wise square. Since \(m_t\) and \(v_t\) are biased towards zero in the initial stages, Adam also introduces bias correction steps:

\[
\hat{m}_t = \frac{m_t}{1-\beta_1^t}, \quad \hat{v}_t = \frac{v_t}{1-\beta_2^t}
\]

Finally, parameters are updated as:

\[
\theta_t = \theta_{t-1} - \frac{\eta}{\sqrt{\hat{v}_t} + \epsilon} \hat{m}_t
\]
where \(\eta\) is the global learning rate, and \(\epsilon\) is a tiny constant to prevent division by zero. The ingenuity of Adam lies in: for parameters with large and frequently changing gradients, \(\sqrt{\hat{v}_t}\) is large, effectively reducing the learning rate; for parameters with small and stable gradients, the effective learning rate increases. This adaptive mechanism makes Adam relatively robust to hyperparameter choices, converges quickly, and has become the most popular default optimizer today. Of course, Adam is not a panacea—in some tasks (such as certain areas of computer vision), carefully tuned Stochastic Gradient Descent with momentum can still achieve better generalization performance.

\noindent 2. Learning Rate Scheduling

The learning rate is one of the most important hyperparameters in training, and the core idea of learning rate scheduling is simple yet profound: use a larger learning rate in the early stages of training for rapid descent and exploration of the vast parameter space; reduce the learning rate in the later stages for fine-tuning in local regions.

Common learning rate scheduling strategies include:

\begin{itemize}
	\item Step decay: Multiply the learning rate by a decay factor (e.g., 0.1) every certain number of epochs. This strategy is simple and effective, especially suitable for long-term training.
	
	\item Cosine annealing: Adjust the learning rate periodically according to a cosine function, causing the learning rate to first decrease slowly, then accelerate, and finally approach zero. Its smooth decay curve sometimes leads to better convergence.
	
	\item Early stopping-style scheduling: Monitor validation set performance and proactively reduce the learning rate when performance stagnates. This adaptive strategy avoids mismatches between manually preset scheduling strategies and training dynamics.
\end{itemize}

\noindent 3. Gradient Clipping

When training deep networks, especially Recurrent Neural Networks or Transformers, gradient explosion is a common threat—the gradient norm may grow exponentially, causing parameter update steps to become too large, sending the loss function to infinity instantly and crashing the training.

Gradient clipping provides a simple and effective solution. Let the gradient vector be \(g\), its norm be \(\|g\|\), and the preset threshold be \(c\). Then the clipped gradient \(\tilde{g}\) is:

\[
\tilde{g} = \begin{cases}
	g, & \text{if } \|g\| \leq c \\
	g \cdot \frac{c}{\|g\|}, & \text{if } \|g\| > c
\end{cases}
\]
This operation does not change the direction of the gradient, only limiting its norm within the threshold. It can be understood as a fuse in a circuit—when the current (gradient) is too high, the fuse blows but protects the entire system; gradient clipping actively scales the gradient back to a safe range when it is about to go out of control, ensuring training stability.

The beauty of gradient clipping is that it acknowledges the uncertainty in the optimization process and sets up a ``safety net.'' In training large language models, gradient clipping is almost standard, preventing gradient explosion without overly interfering with the normal optimization process.

\noindent 4. Curriculum Learning

The inspiration for curriculum learning comes directly from human teaching processes—we don't let beginners face the most difficult problems directly but start with simple concepts and gradually increase difficulty. This idea also applies to machine learning: don't let the model face the most difficult samples from the beginning; start with simple samples and gradually introduce more complex cases.

Curriculum learning can be implemented in various ways:
\begin{itemize}
	\item Data ordering: Sort training data according to difficulty metrics (e.g., sentence length, number of problem steps), sample simple examples in the early stages, and gradually increase the proportion of difficult examples later.
	
	\item Dynamic sampling: Adjust sampling weights based on the model's current performance—sampling probability can be reduced for sample types the model performs well on and increased for those it performs poorly on.
	
	\item Progressive tasks: In reinforcement learning, train in a simplified environment first, then transfer to the full environment.
\end{itemize}
The mathematical intuition of curriculum learning is: simple samples provide stable gradient signals, helping the model quickly establish correct representations in the early stages; as the model's capability improves, the challenge of difficult samples matches its current level, avoiding premature entrapment in poor local minima. In mathematical reasoning tasks, curriculum learning is particularly effective—for example, first let the model learn linear equations in one variable, then gradually introduce systems of linear equations in two variables, quadratic equations, and transcendental equations.
\vspace{3mm}

\noindent\textcolor{structure3}{\textbf{4. Regularization: Combating Overfitting, Improving Generalization}}

The core goal of regularization is to reduce overfitting and improve the model's generalization ability on unseen data. Its philosophical essence is: impose constraints or disturbances on the optimization process, guiding the model towards simpler, smoother, more general solutions, rather than merely memorizing the training data.

\noindent 1. Explicit Constraints: L1 and L2 Regularization

The most intuitive regularization method is to add a norm penalty term of the parameters to the loss function, directly constraining model complexity.
L2 regularization (also known as weight decay) adds the sum of squared weights to the loss function. Let the original loss function be \(\mathcal{L}_{\text{original}}(\theta)\), then the regularized loss is:

\[
\mathcal{L}_{\text{reg}}(\theta) = \mathcal{L}_{\text{original}}(\theta) + \frac{\lambda}{2} \|\theta\|_2^2
\]
where \(\lambda\) is the regularization coefficient, controlling the strength of the penalty; \(\|\theta\|_2^2 = \sum_i \theta_i^2\) is the squared L2 norm of the parameter vector. Its gradient update becomes:

\[
\theta \leftarrow \theta - \eta \nabla \mathcal{L}_{\text{original}} - \eta \lambda \theta
\]

It can be seen that each update step subtracts a small portion of the weight value (\(\eta \lambda \theta\)), driving parameters to shrink towards zero. L2 regularization tends to make all parameter absolute values smaller, making the model output more smooth in response to input changes, thereby reducing overfitting risk. From a Bayesian perspective, L2 regularization is equivalent to introducing a Gaussian prior on the parameters.

L1 regularization adds the sum of absolute values of weights:

\[
\mathcal{L}_{\text{reg}}(\theta) = \mathcal{L}_{\text{original}}(\theta) + \lambda \|\theta\|_1
\]
where \(\|\theta\|_1 = \sum_i |\theta_i|\) is the L1 norm of the parameter vector. The unique aspect of L1 regularization is that it tends to drive some parameters exactly to zero, achieving feature selection and obtaining a sparse model. This is because the L1 norm is non-differentiable at zero, making the optimization process more likely to push parameters to zero. L1 regularization is equivalent to introducing a Laplace prior on the parameters.

\noindent 2. Implicit Regularization: Dropout

Dropout is one of the most imaginative regularization techniques in deep learning. It does not directly modify the loss function but randomly interferes with the network structure during training.
Operation: During each forward propagation, randomly ``drop'' a portion of neurons with probability \(p\) (usually 0.5)—i.e., temporarily set the outputs of these neurons to zero. This means each iteration trains a different sub-network, and these sub-networks share weights. During testing, all neurons are retained, but outputs need to be multiplied by \(p\) to maintain consistency of expected output.
Intuition and Effects: The effectiveness of Dropout comes from multiple levels:
\begin{itemize}
	\item Preventing co-adaptation: It forces each neuron not to overly rely on a few other specific neurons because those neurons might be dropped in the next iteration. Each neuron must learn to work with many random subsets, making the extracted features more robust and independent.
	
	\item Approximation of model averaging: Each dropout instance is equivalent to training a different sub-network. During testing, using the full network with weight scaling (multiplying by the retention probability) can be approximately viewed as averaging the outputs of these exponentially many sub-networks (corresponding to geometric mean under specific conditions). This is similar to ensemble learning, which is proven to effectively reduce variance and improve generalization performance.
	
	\item Sparse activation: Dropout creates a sparse activation pattern, sharing a similar spirit with L1 regularization—both encourage the network to use fewer features for prediction.
\end{itemize}
From an information theory perspective, Dropout can be understood as applying random perturbations to the network structure during training, an efficient data-agnostic form of ``data augmentation.''

\noindent 3. Early Stopping

Early stopping might be the simplest and most effective regularization method. Its idea is straightforward yet profound: stop training before the model starts overfitting the training data.

Specific method: Split data into training and validation sets. During training, periodically evaluate model performance (e.g., loss or accuracy) on the validation set. When it is observed that the error on the validation set no longer decreases or even starts to increase—even though the training error may still be decreasing—stop training and revert to the parameter state with the best validation set performance.

Mathematical explanation: In the early stages of optimization, parameters move in a direction that reduces both training error and test error. The model is learning the true patterns in the data, which are consistent across training and test sets. However, as training continues, optimization begins to overfit the noise and specific cases in the training data, which only exist in the training set. At this point, parameters move in a direction that only reduces training error but may increase test error; training error and validation error start to diverge. Early stopping essentially truncates the number of optimization iterations when validation error reaches its minimum, preventing further worsening of this divergence.

From an optimization perspective, early stopping limits the distance parameters move in the parameter space, similar to the constraint L2 regularization imposes on parameter norms. From a Bayesian perspective, it concentrates the parameter prior near the initial values. The elegance of early stopping is that it does not require modifying the loss function or network structure; simply monitoring validation set performance yields a regularization effect.
\vspace{3mm}

\noindent\textcolor{structure3}{\textbf{5. Engineering Practice for Large Language Models on Mathematical Problems}}

Applying the above engineering practices to the training of mathematical large language models allows us to more concretely understand how these techniques work together and what special challenges the mathematical domain brings.
\begin{enumerate}
	\item Data Preparation: Training mathematical large language models requires massive, high-quality mathematical text—from arXiv papers, mathematics textbooks, Stack Exchange discussions to mathematical implementations in code repositories. This data needs fine-grained cleaning, including normalization of LaTeX expressions, standardization of mathematical symbols, cross-document deduplication, etc. For supervised fine-tuning, mathematical question-answer pairs need to be carefully constructed, not only containing questions and answers but also emphasizing the presentation of reasoning processes—chain-of-thought data teaches the model to derive step-by-step rather than jumping directly to conclusions.
	
	\item Training Pipeline: As mentioned earlier, follow the three-stage pipeline of ``pre-training → supervised fine-tuning → reinforcement learning.'' In the reinforcement learning stage, the particularity of mathematical tasks is especially prominent—we can utilize program-assisted execution, letting the model generate executable Python code to solve mathematical problems, then verify results through a code interpreter. This provides a more objective and scalable reward signal than human preference: whether the code runs correctly is deterministic. We can also train specialized verifier models to judge the correctness of generated answers and the rigor of reasoning steps, serving as reward models in reinforcement learning.
	
	\item Regularization and Stability: In mathematical reasoning, overfitting may manifest as the model memorizing solutions to specific problems rather than learning general problem-solving logic. Dropout and early stopping are equally applicable here. Furthermore, the Transformer architecture's inherent residual connections and layer normalization already have built-in effects for stabilizing training, reducing reliance on certain explicit regularizations.
	
	\item Curriculum Learning: Mathematical knowledge has a natural hierarchical structure—from arithmetic to algebra, from geometry to calculus. This provides an ideal application scenario for curriculum learning. We can first let the model learn simple arithmetic operations, gradually introduce equations, functions, derivatives, integrals; or in the reinforcement learning stage, first let the model solve simple problems with fewer steps, then gradually increase problem difficulty. This progressive training aligns with the cognitive laws of mathematical learning, guiding the model to establish a solid knowledge foundation.
\end{enumerate}
It is worth pondering that these engineering practices are not isolated tricks but an interconnected system. Good initialization allows the use of larger initial learning rates; residual connections and layer normalization reduce gradient vanishing problems and also lower dependence on complex regularization; early stopping and learning rate scheduling work together—when validation performance stagnates, we can either stop training or lower the learning rate to continue exploration; and the multi-stage training pipeline itself is a macro-level ``curriculum learning''—first let the model master basic language abilities (pre-training), then learn to follow instructions (supervised fine-tuning), and finally optimize output quality (reinforcement learning).

In the exploration of AI for Math, combining prior knowledge of specific mathematical problems to design and adjust these practices is key to success. For example, designing mathematics-specific data augmentation strategies (such as variable substitution, equivalent form rewriting), designing special loss functions for physics-informed neural networks, and designing program-verification-based reinforcement learning rewards for mathematical reasoning models. Ultimately, the common goal of all these engineering practices is to achieve ``control and guidance'' over the optimization process and model behavior—taming complex non-convex optimization with limited resources, guiding the model towards parameter regions with powerful generalization ability, reliable reasoning capability, and mathematical problem-solving ability.

\nocite{*}

\printbibliography[heading=subbibliography,title=References]

\end{refsection}

\begin{refsection}[ref3.bib]
\chapter{AI for Discovering Mathematical Patterns}
\section{Overview}

The previous chapter's introduction to the fundamental techniques of machine learning has laid out a powerful toolkit for us. Neural networks, deep learning, supervised and unsupervised learning—the success of these methods in fields like image recognition and natural language processing has fully demonstrated their ability to extract patterns from complex data. However, when we turn our gaze to mathematical research, a deeper question emerges: Can these techniques transcend the role of ``efficient calculators'' and truly participate in the core process of mathematical discovery?

To answer this question, we need to re-examine the inherent dilemmas of mathematical research itself. The discovery of mathematical laws has always relied on the interplay of two core capabilities: intuitive insight and logical deduction. Mathematicians, through a profound understanding of known structures, generate conjectures about unknown relationships, which are then verified through rigorous proofs, forming a ``conjecture-proof'' cycle. However, as the objects of mathematical research become increasingly complex—such as topological invariants on high-dimensional manifolds, algebraic structures in high-degree polynomial rings, or the long-term behavior of nonlinear dynamical systems—the boundaries of human intuition begin to show. We excel at imagining symmetries in three-dimensional space but struggle to intuitively grasp the geometry of fifty-dimensional space; we can derive explicit solutions for low-degree equations but find it difficult to discern hidden connections between high-dimensional algebraic varieties.

It is precisely within this dilemma that artificial intelligence reveals its unique value. But the ``artificial intelligence'' here is not a generalized concept; rather, it points to an emerging interdisciplinary field taking shape: using machine learning as a cognitive tool, with mathematical structures themselves as the object of study, and mathematical understanding as the core goal. This field has its own distinct methodology, theoretical concerns, and research questions. It is neither a simple application of AI technology nor a passive extension of mathematical theory, but a genuine form of intersection—the birth of a new ``cognitive paradigm.''

The core of this new paradigm lies in this: when we face a mathematical problem that lies beyond the boundaries of human intuition, we can leverage AI's pattern recognition capabilities to ``probe'' for potential regularities or signals from mathematical data; then, through interpretability techniques, ``map'' these signals back to concepts understandable by mathematicians, ultimately refining them into testable mathematical conjectures. This is AI-assisted mathematical discovery—the machine is responsible for exploring structures in high-dimensional spaces that are difficult for humans to access directly, while the mathematician is responsible for interpretation, refinement, and proof. The two work in a division of labor, jointly expanding the boundaries of mathematical cognition.

This collaborative logic can be clearly articulated as: Mathematical Problem → Suitable AI Technology → Concrete Implementation Process. We do not start from the technology to find application scenarios—that would lead to a mere piling up of tools; nor do we start from theory to fit in examples—that would obscure the essence of the problem. Instead, we start from the dilemma of the mathematical problem itself and examine what kind of AI technology can provide substantive assistance. For example, faced with the deep problem of ``understanding the relationship between geometric and algebraic invariants of knots,'' the challenge lies in learning the mapping relationships between different mathematical structures from unstructured data—this is precisely what neural networks excel at. Therefore, we design experiments for AI to find patterns in this data, then through interpretability analysis, reveal the key features the model relies on, and finally translate these features into conjectures that mathematicians can test.

This problem-oriented perspective reveals the unique character of AI for Math as an interdisciplinary field. Here, the relationship between AI and mathematics is no longer a one-way ``application'' or ``assistance,'' but a deep ``collaboration'': AI is responsible for probing, the mathematician is responsible for understanding.

In the following sections, we will present this collaborative paradigm in its entirety through concrete case studies. Each case will follow the narrative thread of ``Mathematical Problem → AI Technology → Implementation Process,'' demonstrating how AI, as a new methodology, helps us explore deep relationships between knot invariants, discover hidden patterns in algebraic structures, and even predict the existence of certain mathematical conjectures. These cases are not merely demonstrations of technology; they are witnesses to a new research paradigm—an era belonging to AI for Math is quietly unfolding.

\section{AI Discovers Hidden Relationships Between Knot Invariants: A Supervised Learning Approach}

Imagine a rope tied into a complex knot in three-dimensional space, with its ends then connected—this is a ``knot'' in mathematics. How do we precisely describe and study this ``knot''? The key lies in finding quantities that remain unchanged no matter how we continuously stretch or twist the rope (as long as we don't cut it or let it pass through itself). Such quantities are called knot invariants. They are like the ``fingerprints'' of a knot, helping us distinguish different knots and understand their deep structure.

The main knot invariants can be divided into two broad categories:
The first category is geometric (hyperbolic) invariants. For the vast majority of knots, their complement space (the part of three-dimensional space with the knot itself removed) can be endowed with a very elegant geometric structure—hyperbolic geometry. Just as a saddle surface is a two-dimensional hyperbolic space, a knot's complement can be a three-dimensional hyperbolic space. This leads to a series of refined geometric quantities, such as:
\begin{itemize}
	\item Volume: Hyperbolic volume, measuring the size of this geometric space, is a fundamental geometric characteristic of a knot.
	\item Chern-Simons invariant: A topological invariant originating from theoretical physics, related to the ``twisting'' of the space.
	\item Cusp geometry: The knot itself can be thought of as an infinitely thin tube. The boundary of its tubular neighborhood is a torus. The geometric information on this torus (such as the meridional translation and longitudinal translation) encodes the structure of the knot ``at infinity''.
	\item Injectivity radius: Measures the scale of the ``narrowest point'' in the knot complement, reflecting a geometric ``bottleneck''.
\end{itemize}
The second category is algebraic invariants, defined through algebraic or combinatorial methods, such as:
\begin{itemize}
	\item Jones polynomial: A polynomial invariant discovered in 1984, whose appearance revolutionized knot theory.
	\item Signature: An integer invariant that encodes many important algebraic properties of a knot, for example, it is closely related to whether a knot can be the boundary of some surface in four-dimensional space (i.e., the minimal genus of such a surface).
\end{itemize}

For a long time, mathematicians have vaguely felt that there should be profound connections between invariants from these two different worlds. A famous example is the ``Volume Conjecture'', which boldly proposes that the hyperbolic volume of a knot might be hidden in the asymptotic behavior of its Jones polynomial. However, discovering new, concretely provable relationships faces enormous challenges. The reason is that the relationships between these invariants can be extremely subtle, involving nonlinear patterns in high-dimensional spaces. From vast, complex data, it is difficult to reliably extract mathematically valuable clues relying solely on human observation and traditional statistical methods. We are good at visualizing in three dimensions, but find it hard to intuitively grasp the structure in the high-dimensional space spanned by multiple invariants.

This is precisely the point where AI can intervene. When the core difficulty of a mathematical problem is ``high-dimensional complexity beyond human intuition'', the powerful pattern recognition capabilities of machine learning come into play.

In 2021, the DeepMind team published a breakthrough result in the journal \textit{Nature}. Their developed AI system could help mathematicians discover new connections between knot invariants. This research demonstrated that AI can guide mathematicians' intuition to discover mathematical connections that humans might overlook. The entire study follows a clear logical thread, which we can break down into four key steps.

\noindent Step 1: Transforming a Mathematical Conjecture into a Machine Learning Problem and Data Generation

Any exploration begins with an initial intuition or hypothesis. In this study, the mathematicians' initial hypothesis was: ``There exists some undiscovered predictable relationship between the geometric (hyperbolic) invariants of a knot and its algebraic invariants (especially the signature).'' To test this hypothesis, it needed to be transformed into a machine learning problem. Specifically:
\begin{itemize}
	\item Define Input (X): We need to select a set of geometric invariants to serve as the feature vector describing each knot. The researchers chose volume, the Chern-Simons invariant, four quantities describing Cusp geometry (real and imaginary parts of the meridional translation, real and imaginary parts of the longitudinal translation), injectivity radius, etc. In this way, each knot $K$ is represented as a numerical vector $X(K)$.
	\item Define Output (Y): We need to choose an algebraic invariant as the prediction target. Here, the signature $\sigma(K)$ was chosen. Choosing the signature was wise: it is an integer, making it easy to model as a classification problem; it is easy to compute; and it contains rich mathematical information, making it a quantity ``worth understanding''.
	\item Build the Dataset: Machine learning requires data. The researchers constructed three complementary datasets to ensure the robustness and generalization ability of subsequent discoveries:
	\begin{itemize}
		\item Systematic Census Data: From the ``Regina'' database, containing all knots with crossing number (a measure of knot complexity) up to 16, totaling about 1.7 million samples. This dataset provides ``systematic coverage'' of small, simple knots.
		\item Random Complex Data: Using the professional knot software SnapPy, randomly generated complex knots with about 80 crossings, yielding about 1 million samples after deduplication. This dataset represents a more complex, more ``generic'' distribution of knots, used to verify the generalization of discoveries.
		\item Specially Constructed Data: By controlling braid parameters, constructed tens of thousands of knots with specific algebraic structures (e.g., 4-braids, 5-braids, 6-braids). This dataset was not used for training but specifically for subsequent ``stress testing'' of conjectures—to see if the discovered patterns still hold on these special knots.
	\end{itemize}
\end{itemize}

Thus, an abstract mathematical intuition was transformed into a concrete machine learning problem: Given a knot's geometric feature vector $X(K)$, can we train a model to accurately predict the knot's signature $\sigma(K)$?

\noindent Step 2: Supervised Learning and Pattern Detection

Next, a neural network was used to train on the prepared dataset. In the knot case, the researchers used a standard fully connected feedforward neural network. This is a multi-class classification problem because the signature is an integer. Model performance was evaluated based on its classification accuracy on an independent test set.

The experimental results were encouraging: The trained model achieved a test accuracy of 78\% in predicting the signature, and prediction errors never exceeded ±2. In contrast, a baseline of random guessing had a much lower accuracy (e.g., uniform guessing within the signature's value range). This result strongly suggests that between a knot's geometric features and its signature, there exists a robust mathematical pattern that can be captured by the machine learning model. This gave mathematicians confidence to explore further—their intuitive direction was likely valuable.

\noindent Step 3: Interpretability Analysis and Clue Extraction

The model's successful prediction is just the beginning. More importantly: ``How'' does the model make predictions? Which geometric features does it rely on? How do these features interact? Answering these questions is key to transforming machine learning output into mathematical conjectures.

To this end, the researchers used an interpretability technique called ``gradient-based saliency analysis''. Its core idea is very intuitive: We want to know how much each input feature ``contributes'' to the model's final prediction. Specifically, for each sample in the dataset, we compute the absolute value of the partial derivative of the model's loss function with respect to each input feature $|\partial L / \partial x_i|$. The intuitive meaning of this value is: If we make a small change to a feature $x_i$, how much will the model's prediction loss change? A larger change indicates the model is more ``sensitive'' to that feature, meaning the feature is more important in the decision.

For each sample $x$ in dataset $X$, compute the absolute value of the gradient of the model's loss function $L$ with respect to the $i$-th input feature $x_i$, $|\partial L/\partial x_i|$. This value reflects how much the model's prediction loss would change if feature $x_i$ were slightly altered. A larger change indicates the model is more sensitive to that feature, suggesting it may be more important.

To obtain the overall importance of each feature across the entire dataset, we average the absolute gradient values over all samples:

$$
r_i = \frac{1}{|X|} \sum_{x \in \mathcal{X}} \left| \frac{\partial L(x)}{\partial x_i} \right|
$$
where $r_i$ is the average gradient saliency score for the $i$-th feature, and $|X|$ is the total number of samples in the dataset.

Applying this saliency analysis to the knot model yielded a clear pattern. The analysis indicated that among all geometric invariants, three features stood out, with their $r_i$ values significantly higher than others, being crucial for predicting the signature:
\begin{itemize}
	\item Real part of the meridional translation $Re(\mu)$
	\item Imaginary part of the meridional translation $Im(\mu)$
	\item Real part of the longitudinal translation $Re(\lambda)$
\end{itemize}
These three quantities together describe the geometry of the ``Cusp'' (the boundary torus of the infinitely thin tubular neighborhood) in the knot complement, i.e., the so-called Cusp geometry. In other words, the model told us: To predict the signature, the most important information is almost entirely concentrated in the Cusp geometry.

To verify this discovery, the researchers trained a second model, this time using only these three Cusp geometry features as input. The result was astonishing: The prediction accuracy of this simplified model was almost identical to that of the original model using all dozen geometric features! This was undoubtedly decisive evidence: Information about the signature is almost completely concentrated in these three Cusp geometric quantities. This step sharply focused the mathematicians' attention from a dozen possible geometric invariants down to three core features. The problem was greatly simplified.

\noindent Step 4: Mathematical Refinement and Conjecture/Theorem Formation

Now, the machine learning task was complete. It detected the pattern and located the key features. Next came the moment for mathematicians to exercise their irreplaceable creative wisdom. Armed with the clue provided by AI—``Cusp geometry is key''—they began to apply their deep mathematical intuition and expertise for in-depth analysis and re-creation.

First, they visualized the relationship between these key features and the signature. Observations revealed that the relationship between the signature and these quantities was not a simple linear one but exhibited a complex pattern. After repeatedly examining the data distribution and deeply mining the geometric meaning, the mathematicians creatively defined a new, composite geometric quantity, which they called the natural slope:

$$\text{slope} (K) = Re(\lambda / \mu)$$
This newly defined quantity has a clear geometric interpretation: On the Euclidean torus of the Cusp, there is a special curve (a geodesic starting from a meridian). When it travels around the torus and intersects the meridian again, the path it takes is equivalent to some multiple of the meridian plus a longitude. This ``multiple'' is the natural slope. It is a quantity directly derived from Cusp geometry, with clear geometric meaning.

An even more exciting discovery followed: When plotting the signature $\sigma (K)$ against this newly defined natural slope $\text{slope} (K)$, a strong linear trend emerged between the two! This prompted the mathematicians to propose the first concrete mathematical conjecture:
\begin{quote}
	\textcolor{structure3}{\textbf{Initial Conjecture:}} There exist constants $c_1$ and $c_2$ such that for all hyperbolic knots $K$,
	$$| 2\sigma(K) - \text{slope}(K)| < c_1 \cdot \text{vol}(K) + c_2 $$
\end{quote}

This conjecture is concise and elegant, linking an algebraic invariant (signature) with a new geometric invariant (natural slope) and another fundamental geometric quantity (volume). However, the journey of scientific discovery is never smooth sailing. The researchers used the previously constructed ``specially constructed data'' (those special knots constructed from braids) to stress-test this conjecture. They found that on some special knots, this conjecture did not hold. The appearance of counterexamples did not signal the failure of the exploration but instead guided the mathematicians to a more refined analysis. They noted that in the saliency analysis, another feature—the injectivity radius $\text{inj}(K)$—though not as prominent as the Cusp geometry, also showed some importance. Incorporating this clue, they ultimately proposed a revised conjecture:
\begin{quote}
	\textcolor{structure3}{\textbf{Final Conjecture:}} There exists a constant $c$ such that for all hyperbolic knots $K$,
	$$| 2\sigma(K) - \text{slope}(K)| \leq c \cdot \text{vol}(K) \cdot \text{inj}(K)^{-3} $$
\end{quote}

Subsequently, the researchers rigorously proved the above conjecture in a separate mathematical paper, establishing it as a reliable mathematical theorem. This theorem has direct applications; for example, it implies that the signature controls non-hyperbolic Dehn surgery on the knot, and the natural slope controls the genus of surfaces bounded by the knot in $\mathbb{R}^+_4$. A new mathematical truth, discovered with AI assistance, was thus born.

Reviewing the entire knot research case, we can clearly see a logical thread running through it, ensuring that AI assistance always serves the core goal of mathematical discovery:
\begin{itemize}
	\item Object Definition: The core object of study is the hyperbolic knot $K$. This is the starting point of the entire exploration.
	\item Representation Learning: Transforming the abstract knot structure, via its geometric invariants, into a numerical vector $X(K)$ that machine learning models can process. Simultaneously, the target property of interest is its algebraic invariant, the signature $Y(K)$. This step is the bridge connecting the mathematical world and the data world.
	\item Function Approximation: Transforming the vague goal of ``discovering a relationship'' into a clear supervised learning problem: finding a function $f$ such that $f(X(K))$ can well predict $Y(K)$. The neural network acts as a high-dimensional approximator for this unknown function $f$.
	\item Structure Inversion: Using interpretability techniques to ``probe'' the trained neural network, reverse-engineering the key mathematical structure (i.e., Cusp geometry) it relies on for decisions. This is equivalent to ``translating'' the complex high-dimensional patterns learned by the model back into a subset of features with clear geometric meaning that mathematicians can understand.
	\item Conjecture Proposal: Based on the identified key structure, mathematicians use domain knowledge to create new composite concepts (natural slope) and construct precise conjectures expressible in traditional mathematical language, ultimately establishing them as theorems through proof. This completes the decisive leap from ``patterns in data'' to ``statements in mathematics''.
	\item Verification and Refinement: Using specially constructed data and mathematical proof to rigorously verify and refine the conjecture, ultimately forming mathematically sound results.
\end{itemize}
This thread emphasizes that AI's role is not to replace mathematicians' reasoning but to serve as a powerful pattern detector and focusing lens. It handles the analysis of vast, high-dimensional data that humans are not good at, and points to ``veins'' worth digging deeper. The final ``mining'' and ``smelting'' work (i.e., concept creation, conjecture proposal, and theorem proof) is completed by mathematicians relying on their profound professional intuition and strict logical reasoning. More precisely, the clues provided by AI guided the mathematicians to perform the following creative work:
\begin{enumerate}
	\item Define a New Concept: Based on the key features $\mu$ and $\lambda$, mathematicians defined a new, geometrically meaningful composite quantity—the natural slope: $\text{slope} (K) = Re(\lambda/\mu)$.
	\item Propose and Refine a Conjecture: After observing data trends, they first proposed a conjecture that the signature $\sigma (K)$ has an approximately linear relationship with the natural slope and is constrained by volume. Subsequently, using AI's performance on another dataset, they found counterexamples to this conjecture.
	\item Conjecture the Final Conclusion: Incorporating another feature of relatively high importance (the injectivity radius $\text{inj} (K)$), the mathematicians successfully conjectured the following conclusion:
	$$| 2\sigma (K) - \text{slope} (K) | \leq c \cdot \text{vol}(K) \cdot \text{inj}(K)^{-3}$$
\end{enumerate}
In the above research work, the workflow of the AI model can be summarized as follows:
\begin{itemize}
	\item Data Sources: Building a Multi-Layered Data Foundation
	
	Any data-driven exploration begins with data. To ensure the comprehensiveness and robustness of subsequent discoveries, the researchers constructed three complementary datasets, each playing a different role:
	
	\begin{itemize}
		\item Systematic Census Data: From the ``Regina'' database, containing all knots with crossing number (a common measure of knot complexity) up to 16, totaling about 1.7 million samples. The value of this dataset lies in its ``systematicity''—it covers all structurally relatively simple knots, providing a complete benchmark for the study.
		\item Random Complex Data: Using the professional knot software SnapPy, randomly generated complex knots with about 80 crossings, yielding about 1 million samples after deduplication. This dataset represents a more complex, more ``generic'' distribution of knots, used to verify the generalization of discoveries—i.e., whether the discovered patterns still hold across a broader family of knots.
		\item Specially Constructed Data: By controlling braid parameters, constructed tens of thousands of knots with specific algebraic structures (e.g., 4-braids, 5-braids, 6-braids). This dataset was not used for the initial model training but specifically for subsequent ``stress testing'' of conjectures—to see if the discovered patterns still hold on these specially constructed knots, or even to find counterexamples.
	\end{itemize}
	These three datasets form an organic complement in terms of scale and distribution: census data provides systematic coverage, random data provides generalization verification, and constructed data is used for targeted boundary testing. This multi-layered data strategy is an important foundation for ensuring that subsequent machine learning discoveries have statistical robustness and mathematical significance.
	
	\item Input and Output: Translating the Mathematical Problem into a Machine Learning Task
	
	With data in hand, the next step is to ``translate'' the mathematical problem into a language machine learning can understand:
	
	\begin{itemize}
		\item Input Features (X): Each knot is represented as a numerical vector composed of multiple geometric invariants. These features include volume, the Chern-Simons invariant, four quantities describing Cusp geometry (real part $Re(\mu)$ and imaginary part $Im(\mu)$ of the meridional translation, real part $Re(\lambda)$ and imaginary part $Im(\lambda)$ of the longitudinal translation), injectivity radius, etc. This vector $X(K)$ is the perspective through which the model ``observes'' the knot.
		\item Output Label (Y): The model's prediction target is the knot's algebraic invariant—the signature $\sigma(K)$. The signature is an integer, so this problem is modeled as a multi-class classification task.
	\end{itemize}
	
	\item Network Architecture and Model: Choosing Suitable Tools
	
	Faced with this prediction task, the researchers chose the most classic and mature tool:
	
	\begin{itemize}
		\item Model Type: A standard fully connected feedforward neural network. Although simple in structure, this type of network can theoretically approximate arbitrarily complex functional relationships, sufficient to meet the needs of detecting unknown mathematical patterns.
		\item Training Task: Supervised learning with the geometric feature vector as input and the signature category as output. The model's goal is to learn a mapping function $f$ from geometric features to signature, such that $f(X(K))$ is as close as possible to the true $\sigma(K)$.
	\end{itemize}
	
	\item Interpretability Analysis: Opening the Black Box, Locating the Key
	
	However, the model's high prediction accuracy is just the beginning. More important is answering ``why''—which geometric features does the model actually rely on to make judgments? This step is key to transforming machine learning output into mathematical insight.
	
	\begin{itemize}
		\item Analysis Method: The researchers used gradient-based saliency analysis. They computed the absolute value of the gradient of the loss function with respect to each input feature and averaged it over the dataset to obtain an ``importance score'' for each feature.
		\item Key Discovery: The analysis results showed that among all geometric features, three features had saliency scores far higher than the others—they stood out like three prominent peaks: the real part of the meridional translation $Re(\mu)$, the imaginary part of the meridional translation $Im(\mu)$, and the real part of the longitudinal translation $Re(\lambda)$. Retraining a model using only these three features yielded accuracy comparable to using all features, confirming the concentration of information.
	\end{itemize}
	
	\item Method Boundaries: A Cautious View of AI's ``Discoveries''
	
	In the same paper, the researchers also successfully applied this methodology to the field of representation theory, further validating its generality. However, despite the remarkable success of this research, we must also soberly recognize the inherent limitations of this AI-assisted discovery approach:
	
	\begin{itemize}
		\item Data Dependence and Representation Bias: All knowledge learned by the model comes from the training data. If the data fails to cover certain important types of knots, or if the chosen feature representation fails to effectively capture the essential properties of knots, the model may draw partial or even misleading conclusions.
		\item The Gulf Between Correlation and Causality: Machine learning identifies statistical correlations, not mathematical necessity. A high-accuracy predictive model merely indicates a stable association between certain geometric features and the signature in the observed data, but this does not guarantee that this association is a universal mathematical theorem. Final confirmation must rely on rigorous mathematical proof.
		\item The Indirectness of Interpretability Techniques: The ``importance'' information provided by methods like gradient saliency is essentially an indirect inference based on model behavior, not a direct interpretation of mathematical structure. The clues it provides need to be judged cautiously and cross-validated with domain knowledge; they cannot be accepted blindly.
		\item The Creative Leap from Pattern to Conjecture: AI outputs data and feature importance. Transforming this into a mathematically elegant, appropriately conditioned conjecture is a process highly dependent on the mathematician's creativity and professional expertise. The refinement from the initial conjecture to the final theorem in this study fully demonstrates the complexity and irreplaceability of this creative leap.
		\item Computational Cost and Reproducibility: Large-scale data generation and model training require considerable computational resources. A complete study must meticulously document the data generation protocol, model parameters, and random seeds to ensure its discoveries can be reproduced and verified by others.
	\end{itemize}
	These limitations do not negate the value of AI but rather delineate a boundary for caution: AI is a powerful assistant, but not yet an omniscient oracle. Its discoveries need to be understood, tested, and proven.
	
\end{itemize}

\section{AI Discovers New Algorithms: Based on Reinforcement Learning and LLMs}

Matrix multiplication is one of the most fundamental and core computational tasks in computer science and mathematics. From scientific computing, linear algebra, and signal processing to neural networks in modern artificial intelligence, countless complex computations ultimately boil down to large-scale matrix multiplication operations. Therefore, improving the efficiency of matrix multiplication, even by a tiny theoretical or practical margin, can have a huge ``multiplier effect'' across a wide range of scientific and technological fields.

However, since German mathematician Volker Strassen proposed the famous Strassen algorithm in 1969, substantial improvements to algorithms for fixed-size matrix multiplication have stagnated for decades. The core insight of the Strassen algorithm is: computing the product of two $2 \times 2$ matrices does not necessarily require $8$ scalar multiplications (the standard ``dot product'' method), but can be accomplished with only $7$ multiplications through a clever combination. This algorithmic idea can be applied recursively to larger block matrices, reducing the time complexity of an $N \times N$ matrix multiplication from the classical $O(N^3)$ to approximately $O(N^{\log_2 7}) \approx O(N^{2.81})$.

This breakthrough inspired an esoteric research direction in computational complexity theory called the ``matrix multiplication exponent,'' which seeks the smallest exponent $\omega$ that can reduce the complexity of matrix multiplication to $O(N^{\omega})$. After more than half a century of research, the current best theoretical result is $\omega < 2.37286$, but these asymptotically optimal constructions are often highly complex ``existence proofs'' or non-explicit methods, not directly corresponding to explicit algorithms that we can run efficiently on actual hardware.

From a practical perspective, a more fundamental and perplexing question is: For a given fixed-size matrix (e.g., $3 \times 3$, $4 \times 4$, $5 \times 5$), what is the minimum number of multiplications required to complete the multiplication? This question seems basic, but the answer is extremely difficult to determine. For example, for $3 \times 3$ matrices, the currently known best explicit algorithm was discovered by J. Laderman in 1976 and requires $23$ multiplications. Despite nearly 50 years of effort, we still do not know if this is optimal. For $4 \times 4$ matrices, the best-known method is to recursively apply the Strassen algorithm twice (the so-called Strassen-square algorithm), requiring $7^2 = 49$ multiplications, but whether this is optimal is also unknown.

Why is such a seemingly simple, well-defined problem so difficult? The root cause is that finding an efficient matrix multiplication algorithm is essentially equivalent to finding a low-rank decomposition of a specific three-dimensional ``tensor'' (a multi-dimensional array). This tensor is called the matrix multiplication tensor, which completely encodes the bilinear operation of ``multiplication.'' Finding a matrix multiplication algorithm is finding a set of simple ``primitives'' (rank-one tensors) whose sum exactly equals this complex matrix multiplication tensor. Each primitive corresponds to one multiplication operation in the algorithm. Therefore, the number of multiplications required by the algorithm is the rank of that tensor decomposition.

Transforming this mathematical problem into a search problem immediately reveals its difficulty. Taking $4 \times 4$ matrices as an example, the size of its matrix multiplication tensor $\mathcal{T}_4$ is $16 \times 16 \times 16$. Searching for its rank-$R$ decomposition over a finite set (e.g., $F = \{-2, -1, 0, 1, 2\}$) results in a combinatorial explosion. Specifically, each decomposition consists of $3R$ vectors of length $16$. For $\mathcal{T}_4$, its action space (possible combinations of rank-one factors) is $10^{10}$ times larger than that of $\mathcal{T}_3$, far exceeding traditional games like chess or Go. The search space is so vast that it surpasses any human method or computer method based on exhaustive or traditional combinatorial search. Therefore, the discovery of matrix multiplication algorithms has long relied on mathematicians' ingenious constructions, heuristic searches guided by domain-specific knowledge, or successive numerical optimization followed by manual rounding adjustments. These methods are often highly personal, difficult to generalize, and hard to scale.

This leads to the core question of this section: Can we leverage modern artificial intelligence technology to automatically explore this vast algorithmic space and discover new, efficient, and correct algorithms that surpass human intuition? This question is not only about matrix multiplication itself but also represents a new paradigm—viewing ``algorithm discovery'' itself as a process that can be explored and optimized by machines.

AlphaTensor and AlphaEvolve are two landmark works under this paradigm. They start from different technical paths—the former based on deep reinforcement learning gamifying the problem, the latter based on large language model-guided program evolution—both demonstrating the enormous potential of artificial intelligence in automatically discovering mathematical algorithms and constructions, and achieving substantial breakthroughs on a classic problem that has puzzled the academic community for decades.

To deeply understand the work of AlphaTensor and AlphaEvolve, we must first grasp the core mathematical idea of ``algorithm as tensor decomposition.'' This is the bridge connecting specific computational problems with abstract search spaces.

Matrix multiplication $\mathbf{C} = \mathbf{A}\mathbf{B}$ is a bilinear operation: each element $c_{ij}$ in the output $\mathbf{C}$ is linear with respect to both inputs $\mathbf{A}$ and $\mathbf{B}$. Any bilinear operation can be uniquely determined by a three-dimensional tensor.

Specifically, consider the multiplication of $2 \times 2$ matrices:
$$ \mathbf{C} = \begin{pmatrix} a_{11} & a_{12} \\ a_{21} & a_{22} \end{pmatrix} \begin{pmatrix} b_{11} & b_{12} \\ b_{21} & b_{22} \end{pmatrix} = \begin{pmatrix} c_{11} & c_{12} \\ c_{21} & c_{22} \end{pmatrix} $$
Here, $c_{11} = a_{11}b_{11} + a_{12}b_{21}$, $c_{12} = a_{11}b_{12} + a_{12}b_{22}$, and so on.

We can flatten the elements of the input matrices $\mathbf{A}$ and $\mathbf{B}$ into vectors. For example, in row-major order: $\mathbf{a} = (a_{11}, a_{12}, a_{21}, a_{22})^T = (a_1, a_2, a_3, a_4)^T$, similarly $\mathbf{b} = (b_1, b_2, b_3, b_4)^T$ and $\mathbf{c} = (c_1, c_2, c_3, c_4)^T$.

Then, $c_1 = c_{11} = a_1 b_1 + a_2 b_3$. This relationship can be fully encoded by a $4 \times 4 \times 4$ tensor $\mathcal{T}_2$: define the elements $\mathcal{T}_{i,j,k}$ of $\mathcal{T}_2$ such that
$$ c_i = \sum_{j=1}^{4} \sum_{k=1}^{4} \mathcal{T}_{i,j,k} \cdot a_j b_k $$
For the computation of $c_1$, we need $\mathcal{T}_{1,1,1} = 1$ and $\mathcal{T}_{1,2,3} = 1$, with all other $\mathcal{T}_{1,j,k} = 0$. This tensor $\mathcal{T}_2$ is the matrix multiplication tensor for $2 \times 2$ matrix multiplication. It is a sparse tensor with elements only 0 or 1, and the positions of its non-zero entries precisely describe which products of input terms contribute to which output term.

More generally, $n \times n$ matrix multiplication corresponds to a tensor $\mathcal{T}_n$ of size $n^2 \times n^2 \times n^2$. Similarly, the multiplication of an $n \times m$ matrix with an $m \times p$ matrix corresponds to the tensor $\mathcal{T}_{n,m,p}$.

Understanding how tensors encode matrix multiplication, the next question is: How to decompose this tensor?

First, we need to define the most basic building block—the rank-one tensor. A three-dimensional tensor is called a rank-one tensor if it can be written as the outer product of three vectors. Specifically, given three vectors $\mathbf{u} \in \mathbb{R}^I$, $\mathbf{v} \in \mathbb{R}^J$, $\mathbf{w} \in \mathbb{R}^K$, their outer product $\mathbf{u} \otimes \mathbf{v} \otimes \mathbf{w}$ produces a three-dimensional $I \times J \times K$ tensor whose element at position $(i,j,k)$ is $u_i v_j w_k$.

Intuitively, this is the simplest, indecomposable tensor primitive.

If a tensor $\mathcal{T}$ can be expressed as the sum of $R$ rank-one tensors:
$$ \mathcal{T} = \sum_{r=1}^{R} \mathbf{u}^{(r)} \otimes \mathbf{v}^{(r)} \otimes \mathbf{w}^{(r)} $$
then we say the tensor rank of $\mathcal{T}$ is at most $R$, denoted $\operatorname{Rank}(\mathcal{T}) \le R$. This is a natural generalization of the concept of matrix rank to higher dimensions. The rank of a matrix is the minimum dimension of the space spanned by its column (or row) vectors, while the rank of a tensor is the minimum number of terms (outer products) required to decompose it into rank-one terms.

Here emerges a profound and elegant connection:

\begin{theorem}
	A rank $R$ decomposition of the matrix multiplication tensor $\mathcal{T}_n$ directly corresponds to an algorithm for computing the product of two $n \times n$ matrices, using exactly $R$ scalar multiplications.
\end{theorem}

How is this correspondence established? Let's derive it step by step.

Given the decomposition $\mathcal{T}_n = \sum_{r=1}^{R} \mathbf{u}^{(r)} \otimes \mathbf{v}^{(r)} \otimes \mathbf{w}^{(r)}$. By the definition of the tensor, for any inputs $\mathbf{a}, \mathbf{b}$, the output $\mathbf{c}$ satisfies:
$$ c_i = \sum_{j,k} \mathcal{T}_{i,j,k} a_j b_k = \sum_{j,k} \left( \sum_{r=1}^{R} u_i^{(r)} v_j^{(r)} w_k^{(r)} \right) a_j b_k = \sum_{r=1}^{R} u_i^{(r)} \left( \sum_{j} v_j^{(r)} a_j \right) \left( \sum_{k} w_k^{(r)} b_k \right) $$

\noindent This expression directly gives an algorithm:
\begin{enumerate}
	\item For $r = 1$ to $R$, compute:
	$$m_r = \left( \sum_{j} v_j^{(r)} a_j \right) \cdot \left( \sum_{k} w_k^{(r)} b_k \right)$$
	Here, the expressions in parentheses are linear combinations of input elements (only additions), and then the two are multiplied—this is one scalar multiplication. There are $R$ such multiplications in total.
	
	\item For $i = 1$ to $n^2$, compute:
	$$c_i = \sum_{r=1}^{R} u_i^{(r)} m_r$$
	This step involves only additions and scalar multiplications (multiplying by coefficients $u_i^{(r)}$), introducing no new multiplications.
\end{enumerate}
This algorithm uses exactly $R$ multiplications. The coefficients of the vectors $\mathbf{u}^{(r)}$, $\mathbf{v}^{(r)}$, $\mathbf{w}^{(r)}$ define the specific linear combinations in the algorithm. For example, Strassen's classic algorithm is precisely an $R=7$ decomposition of $\mathcal{T}_2$, and its corresponding 7 sets of $(\mathbf{u}^{(r)}, \mathbf{v}^{(r)}, \mathbf{w}^{(r)})$ define those famous intermediate products $M_1, \dots, M_7$.

More importantly, a fast algorithm for a small matrix size can be recursively applied to large matrices via the ``block matrix'' idea. For example, viewing a large $N \times N$ matrix as composed of $2 \times 2$ sub-blocks, applying Strassen's $2 \times 2$ algorithm to each sub-block (now viewed as operations on ``blocks,'' where each block operation requires 7 scalar multiplications) can recursively reduce the overall complexity to $O(N^{\log_2 7}) \approx O(N^{2.81})$.

More generally, if an $n \times n$ matrix multiplication algorithm uses $R$ multiplications (i.e., $\operatorname{Rank}(\mathcal{T}_n) \le R$), then through recursive blocking, the asymptotic complexity for multiplying two $N \times N$ matrices is $O(N^{\log_n R})$. Therefore, searching for a lower $R$ directly impacts the theoretical limit of matrix multiplication.

Finding low-rank decompositions of tensors is a notoriously difficult problem. For matrices (two-dimensional tensors), computing their rank can be solved in polynomial time (e.g., via singular value decomposition). But for tensors of three or more dimensions, computing their exact rank is an NP-hard problem and extremely difficult in practice.

Therefore, algorithm discovery has long relied on mathematicians' flashes of insight, heuristic searches for specific structures, or successive numerical optimization followed by manual rounding adjustments. These methods are inefficient, poorly scalable, and heavily dependent on human-designed heuristic rules that may not be optimal. This is precisely where AI methods can shine: automatically searching this vast space that is difficult for human intuition to reach.

DeepMind's AlphaTensor work's core innovation lies in cleverly formalizing the NP-hard problem of tensor decomposition as a single-player game (called TensorGame) and using deep reinforcement learning technology based on AlphaZero to play this game, thereby automating the search for efficient decompositions.

\noindent The design of TensorGame ingeniously transforms the mathematical goal of ``finding a low-rank decomposition'' into a sequential decision-making problem:
\begin{itemize}
	\item State: The game state $S_t$ is a three-dimensional tensor. The initial state $S_0$ is precisely the target tensor $\mathcal{T}$ we want to decompose (e.g., the tensor $\mathcal{T}_4$ corresponding to $4 \times 4$ matrix multiplication).
	
	\item Action: At each step $t$, the player (the AI agent) chooses an action—this action corresponds to a rank-one tensor, formed by the outer product of three vectors $(\mathbf{u}^{(t)}, \mathbf{v}^{(t)}, \mathbf{w}^{(t)})$. The elements of these vectors must come from a predefined discrete set $F$, e.g., $\{-2,-1,0,1,2\}$. The purpose of discretization is practical: to ensure the coefficients of the ultimately discovered algorithm are concise, interpretable, and avoid issues from floating-point precision.
	
	\item State transition: After executing an action, the state is updated to the current tensor minus this rank-one tensor:
	$$S_t = S_{t-1} - \mathbf{u}^{(t)} \otimes \mathbf{v}^{(t)} \otimes \mathbf{w}^{(t)}$$
	The intuitive meaning of this operation is: ``remove a primitive from the remaining part to be decomposed.'' Each step gradually ``dissolves'' the target tensor.
	
	\item Termination and reward: The goal of the game is to reduce the tensor to zero in as few steps as possible, i.e., reach $S_R = \mathbf{0}$. At this point, the sequence of actions constitutes a valid decomposition:
	$$\mathcal{T} = \sum_{t=1}^{R} \mathbf{u}^{(t)} \otimes \mathbf{v}^{(t)} \otimes \mathbf{w}^{(t)}$$
	The number of steps $R$ here is precisely the rank of the decomposition, corresponding to the number of multiplications in the algorithm.
\end{itemize}
The reward design directly serves this goal: a reward of $-1$ is given for each step. Therefore, the total return for the entire game is $-R$, and maximizing the return is equivalent to minimizing the number of steps $R$—that is, finding a decomposition with as low a rank as possible.

If the game fails to reach zero within a preset maximum number of steps $R_{\text{limit}}$, an additional penalty is given based on an upper bound on the rank of the remaining tensor $S_{R_{\text{limit}}}$. This guides the agent to reduce the complexity of the remaining part as much as possible even if it cannot fully decompose it.

A key advantage of this formalization is its flexibility. The reward function can be easily modified to optimize other objectives, not just the theoretical number of multiplications. For example, at the end of the game, the algorithm corresponding to the action sequence can be run on specific hardware, and its negative runtime can be used as an additional reward—thus directly discovering algorithms customized and practically efficient for specific hardware like GPUs and TPUs.

AlphaTensor is built upon the famous AlphaZero framework, whose core is a neural network Monte Carlo Tree Search (MCTS) planner. MCTS is a tree search strategy that combines random simulation with value estimation to efficiently select promising actions in large action spaces.

\begin{itemize}
	\item Neural network $f_{\theta}$: It takes the current state (tensor $S_t$) as input and outputs two key pieces of information:
	\begin{itemize}
		\item Policy $\pi$: A probability distribution over all possible actions $(\mathbf{u}, \mathbf{v}, \mathbf{w})$. It predicts which actions are more likely to lead to a good decomposition. Due to the huge action space, AlphaTensor adopts the strategy of Sampled AlphaZero, letting the network output a relatively compact set of candidate actions rather than enumerating all of them.
		\item Value $z$: An estimate of the distribution of future returns (cumulative reward). This is essentially the network's belief about ``starting from the current state, how small a rank can be achieved to finish the game.'' Using a distribution rather than a single value can better capture uncertainty.
	\end{itemize}
	
	\item Monte Carlo Tree Search: At each step, AlphaTensor does not directly use the policy output by the neural network. Instead, it uses that policy and value as guidance to perform multiple rounds of simulation to build a search tree. In the tree, it explores more deeply those action paths that appear promising and synthesizes the simulation results to obtain a search policy superior to the original network policy. This process balances ``exploitation'' (choosing what currently seems best) and ``exploration'' (trying new possibilities).
	
	\item Training loop: The agent trains through self-play. In each game, it uses MCTS to select actions. After the game ends, the resulting trajectory (states, actions, final reward) is used as training data to update the neural network parameters $\theta$, making its policy and value predictions increasingly accurate. This is a continuously self-improving closed loop.
\end{itemize}

Directly applying standard AlphaZero to TensorGame faces huge challenges because its action space and state representation complexity far exceed those of board games. AlphaTensor introduces several key innovations for this:

\begin{itemize}
	\item Specialized neural network architecture: Adopts a Transformer-based architecture but deeply customized for three-dimensional tensor input.
	\begin{itemize}
		\item Input projection: Projects the $S \times S \times S$ input tensor onto three $S \times S$ two-dimensional grids. Each grid corresponds to a pair of tensor ``modes'' (e.g., rows and columns correspond to indices of input matrices $\mathbf{A}$ and $\mathbf{B}$, respectively).
		\item Generalized axial attention: The core of the model is a series of attention blocks, each operating between pairs of grids. This is more efficient than standard full self-attention and explicitly models the relationships between different slices of the tensor. The architecture design respects the mathematical property of tensor rank invariance under slice permutations—this customized inductive bias enables the network to learn and reason about tensor structure more effectively.
	\end{itemize}
	
	\item Synthetic demonstrations and mixed training: Tensor decomposition is hard, but its inverse process—constructing a tensor given a decomposition—is easy. AlphaTensor leverages this to generate a massive amount of ``synthetic demonstration'' data: randomly generate sets of rank-one tensors $\{(\mathbf{u}^{(r)}, \mathbf{v}^{(r)}, \mathbf{w}^{(r)})\}$, then construct the corresponding tensor $\mathcal{D} = \sum_r \mathbf{u}^{(r)} \otimes \mathbf{v}^{(r)} \otimes \mathbf{w}^{(r)}$. This yields (tensor, decomposition) pairs.
	
	The neural network is trained on a mixture of two types of data: one is reinforcement learning on the target tensor $\mathcal{T}_n$ (via self-play); the other is supervised learning on synthetic demonstration data (imitating known decompositions). This mixed strategy greatly improves performance and sample efficiency, even though randomly generated tensors have different properties from the target matrix multiplication tensor.
	
	\item Basis transformation to inject diversity: A tensor has different representations under different bases (coordinate systems), but its rank is invariant. At the start of each game, AlphaTensor applies a random invertible linear transformation (basis change) to the target tensor $\mathcal{T}_n$, then plays the game in the transformed space. After finding a decomposition, it transforms back to the standard basis to obtain the final algorithm.
	
	This strategy offers a dual advantage: First, it provides a powerful exploration mechanism because the same target appears in different ``guises'' under different bases; Second, the decomposition in the transformed basis may have simpler coefficients (restricted to $F$), but when transformed back to the original basis, it may produce coefficients outside $F$, thus unexpectedly expanding the coverage of the algorithm space.
	
	\item Action normalization: To reduce the network learning redundant representations, equivalent actions (differing only in sign) are normalized. At the same time, additional training data is extracted from completed games by methods like swapping action orders to improve data utilization efficiency.
\end{itemize}

AlphaTensor trained a single model to decompose matrix multiplication tensors of various sizes ($n, m, p \leq 5$) and explored different number systems (standard arithmetic $\mathbb{R}$ and modulo-2 arithmetic $\mathbb{Z}_2$). Its achievements are fruitful and landmark:

\begin{itemize}
	\item \textcolor{structure3}{\textbf{Rediscovery of classic algorithms}}: AlphaTensor autonomously rediscovered the Strassen algorithm ($2 \times 2$, rank $7$) and the Laderman algorithm ($3 \times 3$, rank $23$) from scratch, validating the effectiveness of its method on known problems.
	
	\item \textcolor{structure3}{\textbf{First algorithm surpassing Strassen for $4 \times 4$}}: This is the most theoretically groundbreaking result.
	\begin{itemize}
		\item In modulo $2$ arithmetic ($\mathbb{Z}_2$), AlphaTensor discovered a rank-47 decomposition for $\mathcal{T}_4$, the first improvement over the rank-49 corresponding to the two-level recursive Strassen method.
		\item In standard arithmetic ($\mathbb{R}$), it improved the best-known upper bounds for multiple matrix sizes. For example, for $\mathcal{T}_{4,5,5}$ ($4 \times 5$ multiplied by $5 \times 5$ matrix), it reduced the known best multiplication count from 80 to 76; for $\mathcal{T}_{5,5,5}$, from 98 to 96.
	\end{itemize}
	
	\item \textcolor{structure3}{\textbf{Enriched algorithm database and recursive improvements}}: AlphaTensor discovered thousands of inequivalent decompositions (not convertible to each other via simple symmetry operations) for each size. For example, for the rank-49 decomposition of $4 \times 4$ alone, it discovered over 14,000 non-equivalent forms. This reveals the richness of the algorithm space. By intelligently recursively combining these discovered small-size algorithms, AlphaTensor further improved known upper bounds for over 70 larger sizes ($n, m, p \leq 12$) of matrix multiplication.
	
	\item \textcolor{structure3}{\textbf{Beyond standard matrix multiplication}}: The framework's generality was validated.
	\begin{itemize}
		\item Skew-symmetric matrix-vector multiplication: AlphaTensor discovered fast decompositions for small sizes and induced and proved a general algorithm from them: for $n \times n$ skew-symmetric matrix-vector multiplication, only $\sim \frac{1}{2}n^2$ multiplications are needed, achieving asymptotic optimality and improving the previous $n^2$ algorithm.
		\item Discovery of discrete Fourier transform bases: Over finite fields, AlphaTensor was used to find decompositions of cyclic convolution tensors. As a result, it autonomously discovered factors corresponding to the discrete Fourier transform and its inverse, demonstrating its ability to recognize deep mathematical structures.
	\end{itemize}
	
	\item \textcolor{structure3}{\textbf{Hardware-customized algorithm discovery}}: By modifying the reward function to include actual runtime on specific hardware (Nvidia V100 GPU and Google TPU v2) as feedback, AlphaTensor successfully discovered matrix multiplication algorithms optimized for hardware. Although the theoretical multiplication count of these algorithms might be the same as Strassen-square (e.g., $4 \times 4$ block matrix multiplication), because the operation sequences generated by their decompositions better match the hardware's memory hierarchy, computational units, and compiler optimization strategies, they achieved significant speedups in actual tests. This proves that AI can not only optimize theoretical complexity but also directly optimize key performance metrics in engineering practice.
\end{itemize}
The success of AlphaTensor shows that deep reinforcement learning can tackle core NP-hard computational mathematical problems like tensor decomposition and systematically produce new knowledge that is provably correct, theoretically valuable, and practically significant. It transforms algorithm discovery from an ``art'' reliant on inspiration into a scalable, automated ``scientific'' process.
\vspace{3mm}

\noindent \textcolor{structure3}{\textbf{AlphaEvolve: Program Evolution and Algorithm Discovery Based on Large Language Models}}

If AlphaTensor is a ``specialist,'' customizing a sophisticated reinforcement learning solution for the tensor decomposition problem, then Google DeepMind's AlphaEvolve is more like a ``generalist.'' It is a code evolution agent based on large language models, with a broader design goal: by combining evolutionary computation with the code generation and understanding capabilities of LLMs, to search the program space for better algorithms or constructions that can solve various scientific and engineering problems (especially those that can be automatically evaluated).

Before delving into the technical details, we need to first understand a fundamental question: Why not directly ask a large language model to generate a perfect algorithm based on the problem description? There are three levels of consideration behind this:
\begin{itemize}
	\item The reliability dilemma: The success rate of ``one-shot generation'' for complex algorithms is extremely low. The ``hallucination'' phenomenon of large language models leads to a large number of invalid outputs—they may look plausible but are riddled with errors when run. Pure generative methods lack verification mechanisms, making it difficult to guarantee algorithm correctness.
	
	\item The necessity of cumulative innovation: Algorithm discovery is rarely achieved in one step. The working style of human mathematicians is often incremental—making small improvements on an existing algorithm, accumulating to a certain extent before forming a breakthrough. The evolutionary framework perfectly simulates this process of ``cumulative innovation'': a small effective modification is retained and becomes the basis for the next, larger improvement. Iterating on code that has already been validated as effective is far more reliable for an LLM than creating from scratch.
	
	\item The value of search guidance: Pure code generation is aimless. The database and evaluation feedback in the evolutionary framework provide clear direction for the LLM's creation—it is not randomly wandering in a sea of possibilities but continuously optimizing towards the goal of performance improvement. This ``goal-directed generation'' is the core of intelligent search.
\end{itemize}
Based on these considerations, AlphaEvolve established its core philosophy: represent solutions as programs and leverage LLMs as powerful, domain-knowledgeable ``mutation operators'' in an evolutionary loop guided by an automatic evaluation function, continuously improving these programs.

The workflow of AlphaEvolve can be clearly decomposed into four key steps:

\noindent Step 1: Task definition and automatic evaluation loop

The user needs to provide the core of the problem: an automatic evaluation function \texttt{evaluate(program)}. This function receives a candidate program (algorithm), runs it, and returns one or more scalar scores to measure its performance. Correctness can be ensured by verification during execution—for example, using the discovered matrix multiplication algorithm to compute random instances and compare with standard results; any errors are automatically caught and penalized.

Here, the real results of program execution replace subjective evaluation by humans or LLMs, enabling long-term, stable iterative optimization without fear of LLM ``hallucinations'' or biases. It is this closed loop that distinguishes AlphaEvolve from purely conversational AI.

\noindent Step 2: Initialization

The user provides an initial program (which can be very simple, inefficient, or even a naive implementation) and marks the parts of the code allowed to evolve with special comments (e.g., \texttt{\# EVOLVE-BLOCK-START/END}). The remaining parts serve as a fixed framework, unchanged. This design ensures both freedom for evolution and stability of the program's basic structure.
\vspace{3mm}

\noindent Step 3: Evolutionary loop

\begin{itemize}
	\item Prompt sampling and construction: Sample a batch of high-performing programs from the ``program database'' (an archive storing historically explored high-scoring, diverse programs and their scores). Construct an information-rich prompt for the LLM, including: task description and instructions, one or more high-scoring programs and their detailed evaluation results (scores, outputs, etc.), possible additional context (relevant mathematical formulas, paper excerpts, optimization objective descriptions, etc.). Finally, ask the LLM to analyze these programs and propose improvements, presented in the form of code diffs.
	
	\item LLM generation and creative deduction: The LLM (such as the Gemini series models) digests the prompt and generates code modification suggestions. The key here is that the LLM can not only make small syntax-level modifications but also deeply understand program logic and propose profound algorithmic changes—such as introducing new optimization steps, changing data structures, adding heuristic rules, or even changing the entire algorithmic paradigm. It draws not only on information in the current context but also on the vast algorithm knowledge base accumulated during its pre-training on massive amounts of code.
	
	\item Program creation and evaluation: Apply the LLM's generated results to the parent program to produce new candidate programs. Then, a cascaded evaluation mechanism starts: first run on a small, fast set of test cases to quickly filter out programs with errors or extremely poor performance; after passing the preliminary screening, conduct formal evaluation on a more comprehensive, more time-consuming test set. This layered strategy balances efficiency and accuracy.
	
	\item Database update and evolutionary strategy: New programs and their scores are fed into the program database. The database management strategy draws inspiration from advanced evolutionary algorithms like MAP-Elites, aiming to simultaneously maintain an ``elite archive'' (the best in each performance region) and overall diversity, balancing ``exploitation'' (deepening known advantageous regions) and ``exploration'' (trying entirely new possibilities).
\end{itemize}

\noindent Step 4: Iteration and convergence

As the loop continues, the performance of algorithms in the program database continuously improves. The evolutionary process can stop after reaching a preset performance threshold, or human researchers can intervene to examine the evolved algorithms and extract new mathematical insights.

AlphaEvolve was applied to the same tensor decomposition problem as AlphaTensor, but it adopts a higher-order ``meta-search'' strategy. The uniqueness of this strategy lies in changing the fundamental question of ``what to search,'' including:

\begin{itemize}
	\item Transition of the search object: In AlphaTensor, the user directly searches for the decomposition factors $(\mathbf{u}, \mathbf{v}, \mathbf{w})$ themselves. But in AlphaEvolve, the user does not ask the LLM to directly output decomposition factors; instead, they ask it to evolve a ``program that can find tensor decompositions.''
	
	The initial program can be very simple—for example, an off-the-shelf gradient descent optimizer using the Adam optimizer to minimize the reconstruction loss $\| \mathcal{T} - \sum \mathbf{u}\otimes\mathbf{v}\otimes\mathbf{w} \|^2$, with a preset rank $R$. This program itself may not be efficient, but it is the starting point for evolution.
	
	\item Transition of the evaluation object: Correspondingly, the evaluation function no longer directly measures the quality of the decomposition but measures the performance of this ``decomposition-finding algorithm.'' Specifically, the evaluation function runs the evolved algorithm, attempting it on multiple target tensors (e.g., $\mathcal{T}_{4,4,4}$) and multiple random seeds, returning the best rank it can find and the success rate of achieving that rank. This evaluation method incentivizes the evolution of more powerful, robust optimization algorithms—those that can stably and repeatably discover high-quality decompositions.
	
	\item The role of the LLM: In this framework, the LLM reads the current best-performing ``decomposition-finding algorithms,'' understands their structure (e.g., what optimizer is used, what regularization terms are in the loss function, how initialization is done, whether random restarts are included, etc.), and then proposes modifications. These modifications can be very deep, far beyond the scope of daily adjustments by human experts, for example:
	\begin{itemize}
		\item Changing gradient descent to a quasi-Newton method for faster convergence;
		\item Adding regularization terms promoting coefficient discretization to the loss function, guiding the algorithm to find decompositions with concise coefficients;
		\item Designing an alternating projection algorithm iterating between factor space and tensor space;
		\item Even writing a small search logic mimicking the AlphaTensor style, combining local search with global planning.
	\end{itemize}
	
	The workflow of AlphaEvolve is shown in Figure \ref{fig:AlphaEvolve} (cited from [alpha evolve paper]).
	
	\begin{figure}[htbp]
		\centering
		\includegraphics[width=0.7\linewidth]{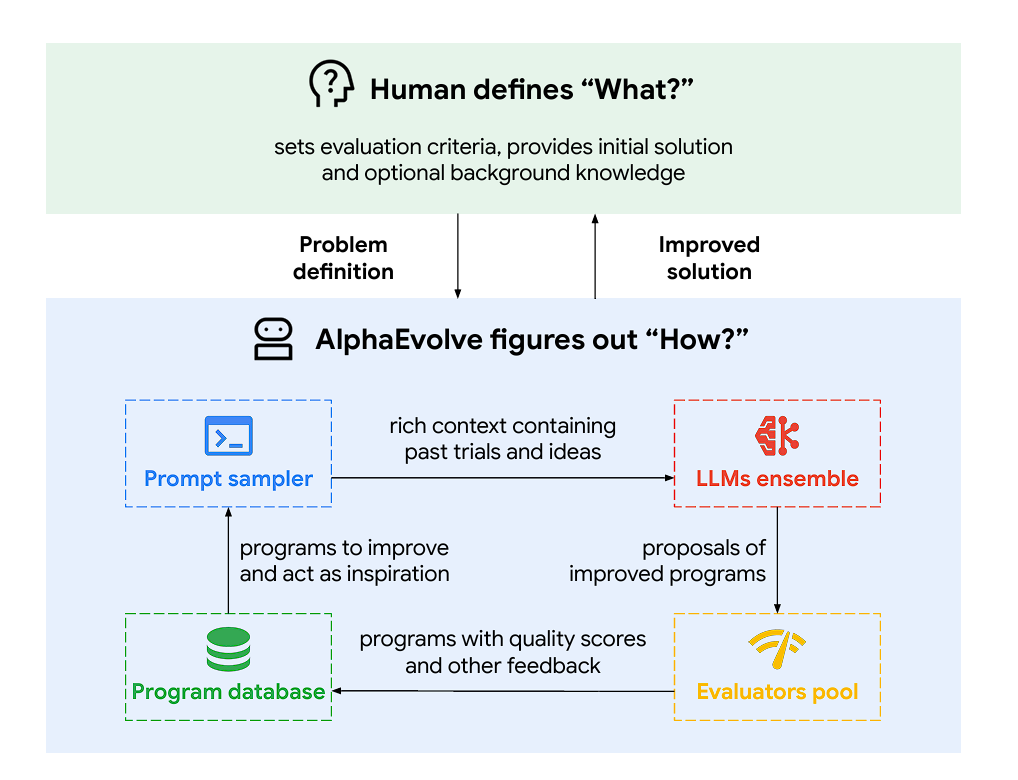}
		\caption{AlphaEvolve Workflow \label{fig:AlphaEvolve}}
	\end{figure}
	
\end{itemize}

The reason the LLM can make these creative modifications is that it has ``read'' a vast amount of algorithm code, optimization literature, and mathematical knowledge during its pre-training. It is not imagining out of thin air but performing well-founded combination and innovation within a huge prior knowledge base.

On the matrix multiplication problem, AlphaEvolve achieved results complementary to AlphaTensor and, in some aspects, more historically significant:

\begin{enumerate}
	\item \textcolor{structure3}{\textbf{Broad matching and surpassing}}: Among the 54 tested matrix multiplication sizes, the programs discovered by AlphaEvolve matched known best results in 38 cases and achieved substantial improvements in 14 sizes. For example, it reduced the multiplication count for $\langle 2,4,5 \rangle$ from 33 to 32, and for $\langle 3,4,7 \rangle$ from 66 to 63. Although these improvements are numerically small, in the field of algorithm optimization, each reduction of ``1'' may signify a breakthrough in theoretical bounds.
	
	\item \textcolor{structure3}{\textbf{Historical breakthrough: rank-48 algorithm for $4 \times 4$ complex matrices}}: This is AlphaEvolve's most striking achievement. It discovered the first rank-48 algorithm for computing $4 \times 4$ matrix multiplication over the complex field (and also the real field). Since the Strassen algorithm was proposed in 1969, the best record for $4 \times 4$ matrices over general number systems has been rank-49 (the result of two-level Strassen recursion). This record stood for over half a century, attempted by countless mathematicians without success. AlphaEvolve changed this record to 48, leading recursively to an algorithm with complexity approximately $O(N^{\log_4 48}) \approx O(N^{2.792})$—a new matrix multiplication method faster than the Strassen algorithm.
\end{enumerate}
The true power of AlphaEvolve lies in its generality. Matrix multiplication is just the first showcase of its capabilities. In broader mathematical and engineering fields, it has also demonstrated astonishing potential.

\noindent 1. Mathematical construction problems: The research team tested AlphaEvolve on over 50 open problems from analysis, combinatorics, number theory, and geometry. The results are stunning:
\begin{itemize}
	\item On about 75\% of the problems, it rediscovered known best constructions—this validates the method's reliability.
	\item On about 20\% of the problems, it discovered new, provably better constructions—this demonstrates the method's creativity.
\end{itemize}

\noindent Specific cases include:
\begin{itemize}
	\item Improved the upper bound for the Erdős minimal overlap problem, slightly improving the previous record.
	\item Increased the known lower bound for the kissing number problem in 11-dimensional space from 592 to 593. The kissing number problem studies how many disjoint spheres of equal size can simultaneously touch a central sphere, and is extremely difficult in high-dimensional spaces.
	\item Improved the best-known densities or sizes for various geometric packing problems (e.g., hexagonal packing, packing circles in a circle, Heilbronn triangle problem).
\end{itemize}

The key to its methodology is: AlphaEvolve is designed to evolve ``heuristic search algorithms,'' not directly evolve ``mathematical constructions'' themselves. Each evolved program is required to perform local search and optimization based on the previous best construction within a fixed time. This multi-stage, adaptive strategy enables it to automatically attack complex optimization problems that are difficult for traditional methods to handle.

\noindent 2. Critical engineering optimization: This is one of the most astonishing application scenarios of AlphaEvolve: it can optimize complex engineering systems without any prior hardware knowledge, solely through automatic evaluation feedback.

\begin{itemize}
	\item Data center scheduling: Evolved a simple heuristic function for Google's Borg cluster management system to evaluate machine fitness during scheduling. This AI-evolved function reclaimed an average of 0.7\% of global computing resources—in a hyperscale data center, this means huge cost savings and value creation.
	
	\item AI training kernel optimization: Evolved a ``tiling strategy'' heuristic for the key matrix multiplication kernel training the Gemini large model. Tiling is a core technique for optimizing the runtime speed of matrix multiplication on hardware, but the optimal tiling strategy is highly dependent on hardware architecture. Without being told any hardware details, AlphaEvolve, through automatic feedback from runtime, found tiling strategies superior to those designed by human experts, improving kernel speed by an average of 23\% and reducing overall training time.
	
	\item Hardware circuit design: In highly optimized TPU arithmetic circuit Verilog code, AlphaEvolve discovered simplification schemes that could remove redundant logic.
	
	\item Compiler intermediate code optimization: Directly optimized the intermediate representation code generated by the XLA compiler for the Transformer attention mechanism, improving execution efficiency. This means it not only optimizes at the algorithmic level but can also delve into the compiler level to optimize how code is actually executed.
\end{itemize}

These cases collectively prove one point: AlphaEvolve is a powerful, general-purpose ``scientific discovery and engineering optimization engine'' that combines the creativity of cutting-edge large language models with the reliability of automatic evaluation. It is not confined to specific mathematical problems but can be applied to any domain that can be expressed as a program and automatically evaluated.

AlphaTensor and AlphaEvolve represent two different but complementary paradigms for AI-driven algorithm discovery. They start from disparate origins, ultimately converge on solving the same class of core problems, and exhibit their own characteristics and potential.

\begin{table}[htbp]
	\caption{Methodological Comparison of AlphaTensor and AlphaEvolve\label{AlphaTensor与AlphaEvolve的方法论对比}}
	\centering
	\begin{tabular}{l p{6cm} p{6cm}}
		\toprule
		\multicolumn{1}{c}{\textcolor{structure3}{\textbf{Dimension}}} & \multicolumn{1}{c}{\textcolor{structure3}{\textbf{AlphaTensor}}} & \multicolumn{1}{c}{\textcolor{structure3}{\textbf{AlphaEvolve}}} \\
		\midrule
		Core Method & Deep Reinforcement Learning (based on AlphaZero/MCTS), customized for a specific problem formalization (game) & Large Language Model-guided program evolution, employing a general problem-solving framework \\
		\addlinespace
		
		Problem Formalization & Formalizes ``finding decomposition'' as a single-player game (TensorGame), state is a tensor, action is a rank-one factor & Formalizes ``solving the problem'' as a code optimization task, state/knowledge is implicit in the program database, action is code modification \\
		\addlinespace
		
		Search Space & Directly searches the solution space (decomposition factors $\mathbf{u}, \mathbf{v}, \mathbf{w}$) & Searches the meta-space (programs that can generate or find solutions) \\
		\addlinespace
		
		Core Technology & Specialized neural network (Transformer variant designed for tensors), trained via reinforcement learning & General large language model (e.g., Gemini), its knowledge comes from pre-training, invoked via prompt engineering and in-context learning \\
		\addlinespace
		
		Advantages & Highly targeted, high search efficiency; good theoretical guarantees; extreme performance within the target domain & Extremely flexible, requires almost no modification of the core framework for different problems; can handle complex, multi-component codebases; can seamlessly incorporate rich natural language domain knowledge; outputs are often more interpretable and deployable \\
		\addlinespace
		
		Limitations & Requires designing specialized game rules, state representations, and reward functions for each new problem; neural network architecture is tightly coupled with the problem, high migration cost & Heavily reliant on the existence and quality of the automatic evaluation function; evolutionary process can be slow and depends on LLM generation quality; may have high computational resource requirements \\
		\addlinespace
		
		Output Nature & Direct mathematical objects (decomposition factors), immediately convertible to an algorithm & Programs that generate those mathematical objects, or directly optimized, runnable code \\
		\bottomrule
	\end{tabular}
\end{table}

These two paradigms are not in competition but constitute two poles of a methodological spectrum. They each capture different aspects of intelligent search—AlphaTensor represents ``focused depth,'' AlphaEvolve represents ``breadth and flexibility.'' In the future, these two paradigms are likely to move towards deep integration:

\begin{itemize}
	\item Complementarity of search levels: AlphaTensor searches at the ``object level,'' AlphaEvolve searches at the ``meta level.'' The latter can be seen as a generalization and abstraction of the former. An interesting idea is: using AlphaEvolve to evolve a better AlphaTensor—for example, evolving more effective neural network architectures, cleverer reward function designs, or even new MCTS variants. In this way, meta-level evolution provides better tools for object-level search, and object-level discoveries in turn enrich meta-level knowledge.
	
	\item Building hybrid workflows: A powerful human-machine collaborative research paradigm might look like this:
	\begin{itemize}
		\item Exploration phase: Use a system like AlphaEvolve for broad, heuristic exploration, quickly generating a large number of candidate ideas or rough algorithms. The focus of this phase is breadth—covering as many different possibilities as possible, identifying promising directions.
		
		\item Deepening and verification phase: For promising candidate directions, use more specialized, powerful methods (like a customized AlphaTensor-style reinforcement learning) for focused, in-depth search and optimization. The focus of this phase is depth—pushing as close to the optimal solution as possible in selected directions. Simultaneously, use formal verification tools to ensure the correctness of final results.
		
		\item Interpretation and conjecture phase: From high-performance algorithms or objects discovered by AI, humans and AI collaborate to extract patterns, symmetries, and potential new mathematical conjectures. AI may have discovered a new structure, and human mathematicians are responsible for understanding the mathematical principles behind this structure and formalizing them into provable theorems.
		
		\item Feedback and iteration phase: New insights from human mathematicians are formalized into new constraints, heuristics, or evaluation criteria, fed back to the AI system, initiating a new round of more efficient search. Thus, human intuition and AI search capabilities form a positive feedback loop, mutually enhancing each other.
	\end{itemize}
\end{itemize}

AlphaTensor and AlphaEvolve together depict such a future: algorithm discovery is no longer an occasional flash of inspiration but a systematically advanceable, deeply human-machine intelligent collaborative exploration process. In this future, AI does not replace mathematicians' thinking but expands the boundaries of mathematicians' capabilities—they are responsible for probing possibilities in vast spaces difficult for humans to reach, while humans are responsible for understanding, refining, and proving. It is precisely this synergy that constitutes the core appeal of the emerging field of AI for Math.

The work of AI discovering mathematical laws is burgeoning, with achievements emerging one after another. Limited by space, we cannot detail them all. Interested readers can refer to the references at the end of this chapter, which include important recent advances in this field.

\nocite{*}

\printbibliography[heading=subbibliography,title=References]
\end{refsection}

\begin{refsection}[ref4.bib]
\chapter{AI for Proving Mathematical Theorems}
\section{Overview}

\noindent \textcolor{structure3}{\textbf{1. The Development History of Automated Theorem Proving}}

Proving mathematical theorems is the most central activity of the discipline of mathematics. From the axiomatic deduction of Euclid's \textit{Elements}, to the logical construction of Whitehead and Russell's \textit{Principia Mathematica}, to contemporary mathematicians' exploration of the Millennium Prize Problems, ``proof'' has always been the cornerstone of mathematical knowledge—it not only confirms the truth or falsity of a proposition but, more importantly, reveals the profound logical connections between propositions, making mathematics a solid and elegant edifice.

Automated theorem proving is precisely a pursuit in the field of artificial intelligence that has lasted nearly seventy years: enabling machines to derive new conclusions from known axioms and theorems through logical reasoning, much like human mathematicians. The history of this pursuit not only witnesses the evolution of AI technology but also reflects the changing understanding of the fundamental question: ``How do machines think?''

The development of automated theorem proving can be roughly divided into three stages, each corresponding to different technical approaches and philosophical reflections on ``how machines understand mathematics.''

\noindent Stage One: Early Exploration and Rule-Based Automation

The fundamental idea of this period originated from the theoretical foundations of computability laid by Turing, Church, and others in the 1930s. Since mathematical reasoning can be formalized as symbolic manipulation, in theory, machines should be able to perform such operations.

In 1956, the ``Logic Theorist'' program developed by Allen Newell, Herbert Simon, and others became the first realization of this ideal. It successfully proved 38 theorems from Chapter 2 of Russell and Whitehead's \textit{Principia Mathematica}—this was the first time a machine demonstrated logical reasoning ability, also marking the birth of artificial intelligence as an independent discipline.

The core methodology of this stage was symbolism and heuristic search. Computers were seen as searchers navigating a ``state space'' composed of axioms, theorems, and inference rules: the initial state was known facts, the goal state was the theorem to be proved, and each action was the application of an inference rule. However, this method soon encountered a fundamental challenge: combinatorial explosion. Even for moderately complex theorems, the number of possible reasoning paths is astronomical, with the vast majority being dead ends. Early systems heavily relied on carefully designed, domain-specific heuristic rules crafted by mathematicians to guide the search; their ``intelligence'' was essentially the internalization of human expert knowledge. This dependence meant that the system's capability boundaries were limited to what humans could foresee and encode in advance, making it difficult to truly break through the human cognitive framework.

\noindent Stage Two: The Rise of Interactive Theorem Provers and Formal Mathematics

Faced with the dilemma of combinatorial explosion, the research paradigm underwent a profound shift in the 1970s and 1980s. Researchers gradually realized that, in the foreseeable future, pursuing fully automated proofs might be unrealistic. Instead, they turned to a more pragmatic question: How can computers become guardians of mathematical rigor, rather than replacements for creativity?

This line of thinking gave birth to Interactive Theorem Provers (ITPs), such as Isabelle, HOL Light, Coq, and later Lean. The core design philosophy of these tools is exceptionally elegant: a small, rigorously verified logical ``kernel'' responsible solely for checking whether each step of a proof conforms to the most basic logical rules, without concern for how those steps were conceived. The mathematician's work becomes interacting with the prover: using high-level ``tactics'' to decompose and prove goals, while the prover acts like a tireless proofreader, ensuring every step of reasoning is absolutely rigorous.

This collaborative model had profound implications. On one hand, it allowed mathematicians to focus on high-level insights and strategies, delegating tedious detail-checking to the machine. On the other hand, with the help of ITPs, mathematicians achieved complete formal verification of milestone proofs like the Four Color Theorem and the Kepler conjecture—proofs whose complexity had surpassed the capacity of any human to verify alone.

More importantly, formal mathematics libraries began to be systematically constructed, such as Isabelle's Archive of Formal Proofs and Lean's Mathlib. These libraries are high-quality, structured, and absolutely reliable data sources of mathematical knowledge that have been rigorously verified by machines. They provide foundational infrastructure for subsequent AI models—a profoundly significant foundational work that organizes mathematical knowledge in a machine-readable, understandable, and operable form for the first time.

\noindent Stage Three: The Integration of Large Language Models and Symbolism

The rise of deep learning, particularly the development of large language models and reinforcement learning, has brought new possibilities to automated theorem proving. Researchers realized that the reason human mathematicians can quickly find correct paths in the vast space of reasoning relies not only on logical deduction but also on an ineffable ``mathematical intuition''—a vague sense of which proof directions are promising. This intuition might be learnable from data.

Large language models, pre-trained on billions of lines of code and mathematical text, have indeed learned the structure of mathematical language and common reasoning patterns. They can be fine-tuned to take the current proof state as input and predict the next likely effective proof strategy—in other words, they act as heuristic policy networks, providing a sense of direction for the search.

Systems like DeepMind's AlphaProof have pushed this concept to its extreme. They formalize theorem proving entirely as a reinforcement learning problem:
\begin{itemize}
	\item Environment: Interactive theorem provers like Lean, which provide deterministic state transitions and perfect correctness feedback—an ideal reinforcement learning environment with clear goals (proof completion) and objective rewards (success or failure of the proof).
	\item Agent: A large neural network that simultaneously learns a policy function (choosing the next tactic) and a value function (evaluating how far the current state is from completing the proof).
	\item Training: Through self-play on a massive number of automatically generated formal problems, the agent continuously optimizes its proof search strategy, learning to navigate the vast reasoning space efficiently.
\end{itemize}
This neuro-symbolic integration—combining the reasoning capabilities of neural networks with the absolute rigor of formal systems—represents the current frontier of automated theorem proving.
\vspace{3mm}

\noindent \textcolor{structure3}{\textbf{2. Core Achievements of Current AI-Assisted Theorem Proving}}

The AlphaProof system has achieved remarkable success along this path. At the 2024 International Mathematical Olympiad, it solved three non-geometry problems, including the most difficult Problem 6 of that competition. Combined with its dedicated geometry reasoning system, AlphaGeometry 2, AlphaProof's overall performance reached a silver medal level. This is the first time artificial intelligence has achieved medal-level performance in a top-tier mathematics competition—and, crucially, every one of its proofs has been fully verified by the Lean kernel, possessing absolute logical reliability.

Behind this breakthrough lies a series of methodological innovations:

\begin{itemize}
	\item Automated Formalization: Utilizing large language models to transform vast numbers of natural language mathematical problems into formal propositions, constructing a training set containing tens of millions of problems, thereby breaking through the bottleneck of manual formalization. This means the system can ``read'' and ``understand'' mathematically described problems in natural language and convert them into a symbolic form it can operate on.
	
	\item Neural-Guided Proof Search: Combining the pattern recognition capabilities of neural networks with the lookahead planning capabilities of Monte Carlo Tree Search, significantly improving search efficiency in the vast proof space. The neural network tells the search ``where might be worth exploring,'' while the tree search is responsible for systematically evaluating and comparing different paths.
	
	\item Test-Time Reinforcement Learning: For a single extremely difficult problem, the system can perform targeted reinforcement learning by quickly generating variants of it, enabling ``specialized training.'' This capability allows the system to dynamically adapt to the characteristics of the problem, demonstrating powerful adaptive ability.
\end{itemize}
\vspace{3mm}

\noindent \textcolor{structure3}{\textbf{3. Core Limitations and Future Challenges}}

Despite significant achievements, current AI theorem proving systems still face fundamental challenges:

Computational cost and accessibility are practical issues. Systems at the level of AlphaProof require enormous computational resources for training and operation (especially test-time reinforcement learning). This makes them more akin to ``national laboratory-level'' research apparatuses rather than tools that ordinary researchers or students can use daily. How to reduce computational costs and democratize this technology is an urgent problem to solve.

The dependence on formalization infrastructure is equally significant. The system's capability boundaries are limited by the coverage of formal mathematics libraries. Currently, successes are mainly concentrated in competition mathematics and classical mathematics areas with well-established formalizations. Formalizing new concepts and theories from cutting-edge mathematical research is itself an extremely time-consuming and expertise-intensive endeavor—creating a paradox: AI needs formalized data to learn, but formalized data requires substantial human effort to create.

Perhaps the most profound challenge lies in the gap from ``problem-solving'' to ``theory-building.'' Current AI, no matter how powerful, is essentially a super strategy searcher and combinatorial optimizer. It operates within a relatively closed formalized knowledge system to solve given problems, rather than engaging in pioneering exploration. The work of human mathematicians extends far beyond solving given problems—they also pose new questions, identify meaningful patterns, evaluate the value of different mathematical directions (so-called ``mathematical taste''), and even create entirely new concepts and theoretical frameworks. These activities require a level of metacognitive ability that current technology is far from achieving.

The history of automated theorem proving is an evolution from symbolic search to neural learning, from independent verification to human-machine collaboration. Systems represented by AlphaProof, through the deep integration of the ``intuition'' of large language models with the ``rigor'' of formal systems, have for the first time enabled machines to reach a level of high-quality mathematical reasoning close to that of human experts. Although the road ahead remains long, especially in terms of high-level creativity and theory-building, it has already opened a new door for mathematical research.

\section{Mathematical Capabilities of Large Language Models: Technical Implementation}

The mathematical reasoning capabilities exhibited by large language models do not stem from some mysterious ``mathematical intuition,'' but are the result of a rigorous, engineerable training pipeline. This pipeline begins with the infusion of vast knowledge, guides the model to understand task formats, and ultimately achieves precise alignment with specific objectives through reinforcement learning. Understanding this pipeline is to understand the source and boundaries of current AI mathematical capabilities. The mathematical capabilities of modern large language models are typically acquired through a carefully designed three-stage training pipeline. Each stage has its unique data composition, algorithmic objectives, and focus on capability shaping.
\vspace{3mm}

\noindent Stage One: Large-scale Pre-training -- Building the ``Knowledge Foundation''

The starting point for any capability is knowledge. The goal of the pre-training stage is to make the model a ``knowledgeable'' general learner.

\begin{itemize}
	\item Data: Massive unlabeled corpora. The model is exposed to text data on the scale of trillions of tokens, covering high-quality web pages, books, academic papers, open-source code, etc. Scale is key -- only with sufficient data can the breadth and depth of human knowledge be covered.
	
	\item Algorithm: Next-token prediction and the Transformer architecture. The core task of pre-training appears simple: given the preceding context, predict the next most likely token. However, this ``simple'' task, with sufficiently large models and data, gives rise to astonishing emergent capabilities. The engine powering this process is the Transformer architecture, whose self-attention mechanism allows the model to dynamically evaluate relationships between all tokens in a sequence, efficiently capturing long-range dependencies and complex patterns. Through distributed training on large-scale clusters, the model's hundreds of millions or even hundreds of billions of parameters gradually learn the ability to map text into a high-dimensional semantic space.
\end{itemize}
This stage produces a ``knowledgeable but undisciplined'' base model. It possesses a rich reservoir of knowledge, knowing a vast number of facts, formulas, and reasoning patterns about mathematics, but it does not yet understand specific task formats -- you can ask it any question, and its response might be factual, fictional, or completely irrelevant. It needs further ``disciplining.''
\vspace{3mm}

\noindent Stage Two: Supervised Fine-tuning -- Teaching ``Task Format'' and ``Chain-of-Thought''

If pre-training is about instilling knowledge, then supervised fine-tuning is about teaching the model how to ``use'' this knowledge to respond to human instructions.

\begin{itemize}
	\item Data: High-quality instruction-response pairs. Researchers construct tens of thousands to hundreds of thousands of carefully crafted (instruction, input, expected output) data pairs. For example, the instruction is ``Solve the equation: $2x + 5 = 13$'', and the expected output is ``$x = 4$''. These data demonstrate how the model should respond to different types of tasks.
	
	\item Algorithm: Supervised instruction fine-tuning. The base model is fine-tuned in a supervised manner using this instruction data. This process essentially teaches the model a new conditional probability distribution: $P(\text{output} \mid \text{instruction}, \text{input})$. The model gradually learns that when it sees the instruction ``solve the equation,'' it should output a numerical answer; when it sees the instruction ``prove,'' it should output a sequence of logical derivations.
\end{itemize}
Chain-of-thought fine-tuning is a key innovation for eliciting the model's explicit reasoning capabilities. Standard instruction fine-tuning only requires the model to output the final answer, whereas chain-of-thought fine-tuning requires the model to output a complete solution including step-by-step reasoning.
\\
For example, for the same equation problem:
\begin{itemize}
	\item Standard fine-tuning output: $x = 4$
	\item Chain-of-thought fine-tuning output: First, subtract $5$ from both sides of the equation to get $2x = 8$. Then, divide both sides by $2$ to get $x = 4$. Therefore, the answer is 4.
\end{itemize}
By learning from a large amount of such chain-of-thought data, the model internalizes the paradigm that ``for complex problems, logical derivation steps should be shown before the final answer.'' This does not grant the model a new reasoning ability but guides the logical abilities already implicitly learned during pre-training to be expressed explicitly through demonstration data. This explicit reasoning process not only makes the model's decisions more interpretable but also provides intermediate steps that can be evaluated for subsequent reinforcement learning.

After this stage, the model transforms into a ``student who follows instructions and can show its thought process.'' However, the optimization objectives for its output -- such as conciseness, rigor, or innovativeness -- are not yet precisely defined. It knows how to answer, but not yet what constitutes the best answer.
\vspace{3mm}

\noindent Stage Three: Reinforcement Learning -- Achieving ``Precise Task Alignment''

The goal of the third stage is to strongly align the model's output with specific, objectively evaluable task objectives. Its core is designing a rule-based reward function and then optimizing the model via reinforcement learning so that its behavior becomes highly consistent with these rules.

\begin{itemize}
	\item \textbf{Design of the Reward Function}: The reward function $R$ needs to automatically score the model's output based on a set of clear rules. The design of the reward function directly determines the behavior the model ultimately learns.
	
	\begin{itemize}
		\item \textbf{For computational problems}: The reward function is usually very simple and deterministic. The rule is: $R(\text{output}) = 1$ if the final numerical answer is correct, otherwise $R(\text{output}) = 0$. The system can automatically extract the final numerical answer from the model's output and compare it with the standard answer.
		
		\item \textbf{For logical proof problems}: The design of the reward function is much more complex, representing the current technological frontier.
		\begin{itemize}
			\item \textbf{Formal verification}: In formal mathematics settings, the strictest rule is to submit the entire proof process output by the model (e.g., Lean or Coq code) to a theorem prover kernel for verification. The rule is: $R(\text{output}) = 1$ if the prover fully accepts the proof, otherwise $R(\text{output}) = 0$. This is the core method used by systems like AlphaProof, providing an absolutely reliable correctness signal.
			
			\item \textbf{Challenges with natural language proofs}: For informal natural language proofs, automatically judging the validity of each reasoning step is extremely difficult. A compromise method is to check whether key reasoning steps appear in the proof (e.g., ``used induction,'' ``applied the Cauchy-Schwarz inequality''), or to use another verification model to check the logical consistency between steps. However, the completeness and reliability of such rules are limited.
		\end{itemize}
		
		\item \textbf{The dilemma of judging innovativeness}: It is currently almost impossible to effectively evaluate innovativeness through automated rules. Innovativeness involves the degree of ``difference'' from known proofs, the ``cleverness'' of conception, etc. These are high-level, fuzzy semantic concepts difficult to quantify as deterministic reward signals.
	\end{itemize}
	
	\item \textbf{Reinforcement Learning Algorithm}: Policy gradient algorithms like Proximal Policy Optimization (PPO) are typically used. The process is: treat the supervised fine-tuned model as the policy to be optimized; for a task prompt, the policy model generates an output; the rule-based reward function automatically scores this output; the PPO algorithm uses this reward signal to update the model parameters, making it more likely to generate outputs that receive high scores in the future.
\end{itemize}
This method is direct and efficient, particularly suitable for tasks where objectives can be clearly quantified. It strongly binds the model's behavior to the objective metric of ``correctness,'' achieving a leap from ``knowing how to answer'' to ``knowing how to answer correctly.''

Although rule-based reinforcement learning provides a clear alignment direction for the model, its limitations are fully exposed in complex reasoning tasks. Understanding these limitations helps us accurately grasp the boundaries of current AI mathematical capabilities.
\begin{enumerate}
	\item Ambiguity and Complexity of Proof Process Judgment
	Natural language proofs are full of ambiguity for machines. When a model outputs ``it obviously follows'' or ``by a standard lemma,'' these expressions may be clear enough for human readers but difficult for machines to verify. The ``jumps'' common in human proofs -- omitting intermediate steps considered ``obvious'' -- need to be fully expanded for machine verification. Automatically judging whether a jump is acceptable requires nearly complete background knowledge, which is itself an AI-complete problem.
	
	\item The Non-computability of Innovativeness Evaluation
	What is the ``innovation'' of a proof? Is it using a novel combination of lemmas? Simplifying a known proof? These definitions themselves cannot be fully formalized. To judge whether a proof is novel requires semantic-level comparison with the entire database of existing proofs, which is infeasible in practice. At best, the reward function can reward outputs that are ``different from common proof methods in the training data,'' but this may be far from genuine mathematical innovation and could even reward incorrect or more verbose proofs.
	
	\item Sparsity of Reward Function and Exploration Challenges
	In proof tasks, the reward function returns a positive signal (+1) only when the final proof is completely correct; all intermediate steps before that receive a reward of 0. This sparse reward is like searching for a specific planet in the vast universe -- the guiding signal is extremely weak, leading to low training efficiency.
\end{enumerate}
A more subtle problem is the trap of local optima. The model may learn to generate safe, verbose proofs that mimic known ones to ensure correctness rewards, but this strategy strongly discourages its motivation to explore more concise, clever new proofs. This runs counter to the ``creativity'' we desire -- the model is guided by the reward mechanism towards conservatism, not innovation.

Surveying the entire training pipeline, we can clearly see the source and boundaries of the mathematical capabilities of current large language models. It is an extremely diligent polymath. Through pre-training, it absorbs the vast mathematical knowledge accumulated by humanity; through supervised fine-tuning, it learns how to respond to mathematical problems in standard formats; through reinforcement learning, it optimizes its behavior to maximize correctness.

However, there remains an essential gap from becoming a pioneering creator with intuition and inspiration. The core of this gap lies in the fact that what we can define and optimize are objectives that can be quantified by clear rules -- answer correctness, proof acceptance. But the qualities required for high-level mathematical creation -- deep insight, ingenious conception, theoretical taste -- are precisely those that cannot be reduced to clear rules.

When we define the reward function as ``answer correct,'' we optimize for reliability and rigor; but how to define, quantify, and set ``creativity'' and ``deep insight'' as optimization objectives remains an unsolved mystery. This reveals the boundaries of current AI capabilities in mathematical reasoning and general scientific discovery: it excels in closed, rule-defined problem spaces, but to play a role in open, creative work requiring value judgments, fundamental methodological breakthroughs are still needed.

Future progress may depend on how more abstract, higher-order meta-cognitive objectives -- such as ``elegance of explanation,'' ``unification of theories,'' ``depth of concepts'' -- can be modeled as learnable signals. This is not only a technical challenge but also a rethinking of the fundamental question: ``What is mathematical creativity?''

\section{Current Success Case 1: AlphaProof}

After understanding the development history of automated theorem proving and the technical implementation of large language models, we can finally delve into a specific, milestone system: AlphaProof developed by DeepMind. It is not only a top achievement in the current field of AI theorem proving but also a deep integration of many concepts we previously discussed—formalized mathematics, neural-guided search, reinforcement learning—within a unified framework.

AlphaProof solved three high-difficulty problems from the 2024 International Mathematical Olympiad, including the hardest Problem 6 of that competition. Combined with its dedicated geometry system AlphaGeometry2, its overall performance reached a silver medal level. This is the first time artificial intelligence has achieved medal-level performance in a top-tier mathematical competition, and each of its proofs has been rigorously verified by the Lean theorem prover, ensuring absolute logical reliability.

Behind this achievement lies a meticulously designed technical architecture. We will use AlphaProof as the main case study to dissect in detail how it transforms abstract mathematical reasoning into a computable, optimizable problem.

\noindent \textcolor{structure3}{\textbf{Mathematical Model: Viewing Theorem Proving as a Markov Decision Process}}

The starting point for any reinforcement learning system is to formalize a real-world problem into an environment with which an agent can interact. AlphaProof defines theorem proving as the following Markov Decision Process:

\begin{itemize}
	\item State ($S$): The state $s_t$ is the ``tactic state'' of the Lean proof assistant at time $t$. It is a text string that clearly lists all current goals to be proven and the available hypotheses for each goal. For example, when proving ``for all natural numbers $n$, $n$ is either even or odd'', the initial state might be the goal ``$\vdash \forall n \in \mathbb{N},\ \mathrm{Even}(n) \lor \mathrm{Odd}(n)$''. After applying a tactic like ``intro $n$'', the state becomes ``$n : \mathbb{N} \vdash \mathrm{Even}(n) \lor \mathrm{Odd}(n)$''. This textual representation contains all information about the problem structure—it tells the agent: what you have (hypotheses) and what you need to prove (goals).
	
	\item Action ($A$): An action $a_t$ is a Lean tactic that can be executed in the current state $s_t$, also represented as a text string. Tactics can be basic (such as \texttt{intro}, \texttt{apply}, \texttt{have}, \texttt{exact}) or complex combined automated strategies (such as \texttt{linarith}, \texttt{ring}), or even user-defined ones. The action space is open and vast—this is precisely the richness of mathematical proof.
	
	\item State Transition ($P$): The transition function $P(s_{t+1} | s_t, a_t)$ is defined by Lean's proof assistant execution engine and is deterministic. Given the current state and a tactic string, Lean attempts to execute it. If the tactic is legal and successful, the environment transitions to a new proof state $s_{t+1}$; if the tactic is inapplicable or results in an error, the current attempt fails—meaning the agent needs to learn to choose ``feasible'' actions.
	
	\item Reward ($R$): The design of the reward function needs to guide the agent to find proofs that are as short as possible. The system assigns a small negative reward for each non-terminal step (e.g., $r_t = -1$). When the agent successfully completes the proof of the entire theorem, the reduction in cumulative negative reward itself reflects the optimization goal—the fewer steps required to find a proof, the higher the total return. A more sophisticated design handles the case of ``subgoals'': if a tactic decomposes a goal into multiple independent subgoals (e.g., proving $P \land Q$ decomposes into proving $P$ and proving $Q$), its return is defined as the sum of the negative step counts of the longest branch among all subgoal proof paths. This encourages the agent to balance the difficulty of each subgoal rather than just optimizing the total number of steps.
	
	\item Discount Factor ($\gamma$): Typically set to 1, because proving is a finite-horizon episodic task, and there is no need for exponential discounting of the future.
\end{itemize}

The agent's goal is to learn a policy \( \pi(a|s) \) that maximizes the expected cumulative reward (return) over the entire proof episode, which is equivalent to minimizing the average number of steps required to find a proof.
\vspace{3mm}

\noindent Core Component One: The Proof Network—Understanding State and Generating Intuition

To learn the policy \( \pi \), the agent needs a model capable of understanding and representing proof states. AlphaProof uses a massive, 30-billion-parameter encoder-decoder Transformer model as its ``proof network''. This network plays two key roles, analogous to the two capabilities of a human mathematician:
\begin{enumerate}
	\item Policy Network: Corresponds to the intuition of ``what to do next''. The decoder part of the network, receiving the state representation produced by the encoder, outputs a probability distribution $\pi(a|s)$ over possible tactics. This distribution is not uniform; rather, based on patterns learned from millions of training runs, the network ``ranks'' the tactics most likely to be effective in the current state. It tells the agent: in this proof state, which tactics are more worth trying.
	\item Value Network: Corresponds to the evaluation of ``how good/difficult is this situation''. Another output head of the network estimates the expected return $V(s) \approx E[G_t | s_t = s]$ from the current state \( s \) to the final completion of the proof. Intuitively, $V(s)$ reflects the ``difficulty estimate'' of completing the remaining proof. A $V$ value close to 0 means nearing completion; a large negative value means there is still a long way to go, or perhaps a dead end has been reached.
\end{enumerate}
The training of this network is divided into three stages. First is large-scale pre-training on hundreds of billions of tokens of code and mathematical text corpora using standard language modeling, enabling it to grasp the basic syntax of mathematical language and logical structure. Next is supervised fine-tuning on paired data of ``state-next tactic'' from human-written formal proof libraries like Mathlib, allowing the network to initially learn to imitate common steps of human provers. But the most crucial stage is the subsequent reinforcement learning, which is key for the agent to form its own ``intuition''—through millions of trials and errors in self-play, the network gradually learns to distinguish promising proof paths from dead ends.
\vspace{3mm}

\noindent Core Component Two: Interaction with the Environment and Search—Planning and Trial-and-Error

Having just the proof network is not enough; the policy and value estimates it provides are immediate and local. To find a complete proof, this local intuition needs to be combined with forward-looking search. AlphaProof adopts a Monte Carlo Tree Search algorithm inspired by AlphaZero but with key adaptations for the proving task.

The search process constructs a tree where nodes are proof states and edges are tactic actions. Starting from the initial theorem state (root node), the following cycle is repeated:

\begin{itemize}
	\item Selection: Starting from the root node, recursively select child nodes until reaching a leaf node that is not fully expanded. The selection criterion balances ``exploitation'' (choosing branches with high value estimates) and ``exploration'' (choosing branches with few visits but potential), using a formula similar to PUCT, where the prior probability $\pi(a|s)$ from the policy network guides exploration. This mechanism ensures the agent neither rigidly sticks to known good paths nor blindly tries all possibilities.
	
	\item Expansion and Evaluation: When a leaf node $s_L$ is reached, the proof network is called to generate $K$ candidate tactics for it (sampled according to the policy distribution), and their execution is attempted in the Lean environment. Each successful tactic produces a new child node (a new proof state). Then the value network evaluates this new state $s_L$, producing an estimate $V(s_L)$.
	
	\item Backpropagation: The leaf node's value estimate $V(s_L)$ is propagated back along the selected path, updating the visit counts and average value estimates for each node-action pair on the path.
\end{itemize}

A key adaptation is for the common case of ``subgoals'' in proofs. For example, a tactic might decompose the goal ``prove $A \land B$'' into two independent subgoals ``prove $A$'' and ``prove $B$''. In the search tree, this creates a special ``AND node''. From this node, the agent needs to choose which subgoal to tackle first. The search algorithm prioritizes the subgoal with the worst value estimate (i.e., the one that appears most difficult), because the completion time of the entire proof depends on the longest branch—this embodies the proof strategy of ``tackling the hardest part first'', consistent with the intuition of human mathematicians.

At the end of the search, based on the visit counts of actions under the root node, an improved policy refined by forward-looking search can be obtained. The agent can then choose its next action accordingly or directly use the proof path found by the entire search tree.

AlphaProof's exceptional capability stems not only from its sophisticated network and search algorithms but also from its revolutionary training paradigm, particularly two core ideas: building a large-scale training curriculum through automated formalization, and test-time reinforcement learning.

\noindent \textcolor{structure3}{\textbf{Core Innovation One: Automated Formalization and Large-Scale Curriculum Learning}}

To make a reinforcement learning agent powerful, it needs to be ``well-informed''—to practice on an extremely diverse and massive set of problems. However, manually formalizing mathematical problems into Lean is an extremely time-consuming task. AlphaProof's method to break this bottleneck is automated formalization. It uses a fine-tuned large language model (based on Gemini) to automatically translate approximately 1 million natural language mathematical problems (from a wide range of sources, from middle school math to Olympiad problems) into Lean statements. This process does not require 100\% accuracy, because even if the translation is biased, the generated Lean statement itself is a syntactically correct, verifiable new mathematical proposition. This system ultimately generated about 80 million formalized problems, constituting an unprecedented, massive-scale training dataset.

A central ``matcher'' system dynamically selects problems from this database and assigns them to distributed ``executor'' agents to attempt to prove or disprove. Problem selection priority is based on ``interestingness''—problems that are unsolved, or sometimes solvable and sometimes not, are prioritized. For problems that fail after multiple attempts, the system allocates more computational resources (search simulations) to tackle them. The successful proof/disproof data (state-tactic sequences) generated by agents during attempts are collected and used to continuously update the parameters of the proof network. This process forms a positive feedback loop: the stronger the agent's problem-solving ability, the more diverse and difficult problems it can solve, thereby generating higher-quality training data, which further improves the network. This learning from large-scale, autonomously generated experience is key to AlphaProof surpassing previous supervised learning methods based on fixed datasets.

\noindent \textcolor{structure3}{\textbf{Core Innovation Two: Test-Time Reinforcement Learning}}

Even after large-scale training, when facing a brand-new, extremely difficult Olympiad problem, a general-purpose agent may still fail to find a proof within a limited search budget. Another breakthrough of AlphaProof is the introduction of Test-Time Reinforcement Learning (TTRL). The idea is simple yet powerful: since general training gives the agent broad mathematical capabilities, can we conduct a brief, intensive ``special training'' for a single ``hard nut'' problem?

The specific process is as follows: When presented with a target difficult problem $T$, the system first automatically generates a large, related set of ``variant problems'' $\{V_T\}$ around $T$. These variants are generated by a large language model and could be simplified versions of $T$ (e.g., removing a condition), analogical versions (e.g., replacing an algebraic structure with a topological one), generalized versions, or those obtained by programmatically making minor perturbations to the assumptions of $T$. These variants, together with $T$ itself, constitute a tailored training curriculum for that specific problem.

Then, AlphaProof initiates a focused reinforcement learning loop, but this time the training data is no longer the 80 million general problems, but only the set of $T$ and its variants. The agent (initialized from a trained general model) rapidly performs self-play and learning on this small-scale, highly relevant curriculum. This process may last for several days, consuming significant computational resources.

The result is that the agent can deeply adapt to the specific mathematical structures and difficulties involved in $T$, often discovering proof strategies that the general model could not find even after searching for days. This mimics the behavior of human mathematicians working on a difficult problem: trying various special cases, searching for lemmas, generalizing from simple cases. Test-time reinforcement learning successfully automates this mechanism of ``problem-specific'' deep exploration.

AlphaProof demonstrates its powerful capabilities in multiple dimensions:

\begin{itemize}
	\item Solving High-Difficulty Problems: In the 2024 International Mathematical Olympiad, it solved three non-geometry problems, including the hardest Problem 6 of the competition. Combined with its dedicated geometry system AlphaGeometry 2, its total score reached a silver medal level. This is the first time artificial intelligence has achieved medal-level performance in a top-tier mathematical competition.
	
	\item Verified Reliability: All outputs are formal proofs verified by the Lean kernel; their reliability is equivalent to the mathematical axiomatic system itself, completely eliminating ``hallucinations'' or specious reasoning.
	
	\item Discovering Novel Proofs: Through reinforcement learning and self-play, the agent sometimes finds proof paths that are different from known human proofs but equally correct and potentially more concise. This capability goes beyond mere imitation, exhibiting a certain sense of ``creativity''.
	
	\item Efficiency Improves with Experience: As training progresses, the agent not only becomes more capable of solving problems but also requires fewer search steps to solve the same problems, indicating that its value and policy networks indeed learn more effective heuristics. It is not only ``doing it right'' but also ``doing it better''.
\end{itemize}
However, while appreciating these achievements, we must also be soberly aware of AlphaProof's current limitations:
\begin{itemize}
	\item Massive Computational Cost: Both the training phase (tens of thousands of TPU-days) and the test-time reinforcement learning for a single problem (several TPU-days) require enormous computational resources, far exceeding the typical budgets of academic institutions. This makes AlphaProof more akin to a ``national laboratory-level'' research apparatus rather than a tool usable by ordinary researchers, raising concerns about accessibility and reproducibility.
	
	\item Domain Still Limited: Current successes are mainly concentrated within the scope of Olympiad mathematics and lower-division undergraduate competition mathematics. Although these problems are difficult, they belong to a ``closed world''—all necessary concepts and theorems are already well-formalized in libraries like Mathlib. Extending the system to the frontiers of mathematical research faces significant challenges, as that involves defining new concepts and proposing new conjectures, not merely solving already formalized problems.
	
	\item Shortcomings in Geometry Handling: Due to insufficient support for Olympiad-style plane geometry in current formal geometry libraries (Mathlib), AlphaProof had to delegate geometry problems to the specialized AlphaGeometry 2 system. While this ``division of labor'' solves the problem, it also exposes that the development of formal mathematical libraries itself is a major challenge: if knowledge in a certain domain has not yet been formalized, AI cannot access it.
	
	\item The Nature of ``Creativity'' is Debated: AlphaProof's ``creativity'' is mainly manifested in searching and combining existing knowledge to find new paths. This is still fundamentally different from the ``high-level creativity'' of human mathematicians who propose entirely new concepts and construct new theoretical frameworks. As revealed by benchmarks like DeepMath-Creative, current large language models still frequently exhibit directional errors, logical flaws, or ineffective lengthy reasoning when faced with tasks requiring truly constructive thinking (such as constructing specific counterexamples) or open-ended problems. Their performance heavily relies on memorizing and recombining patterns in the training data, rather than deep conceptual understanding or genuine inspiration. Is this ``creativity'' a new form of intelligence or merely clever recombination of existing knowledge? This question will likely accompany the entire development journey of AI for Math.
\end{itemize}

The success of AlphaProof marks a key leap for automated theorem proving from ``toy problems'' to ``competition-level difficult problems''. It proves that the path of neuro-symbolic integration is feasible—the intuition guidance of neural networks, combined with the rigorous verification of formal systems, can yield superior performance. However, the road ahead from Olympiad problems to research frontiers, from problem-solving to theory-building, remains long. Nonetheless, a new door has undoubtedly been opened.

\section{Current Success Case 2: LLM-based Agents}

Mathematical research has long relied on the intuition, reasoning, and creativity of individual mathematicians. However, with the rapid advancement in the capabilities of large language models (LLMs), a new research paradigm is emerging, where AI agents are deeply involved or even lead mathematical discovery. This section briefly introduces Aletheia: a mathematical research agent built upon Gemini Deep Think. It can perform iterative generation, verification, and correction for research-level mathematical problems and has successfully produced mathematical papers entirely generated by AI.

An agent based on a large language model refers to a system that uses a large language model as its core, integrated with mechanisms such as planning, memory, and tool usage, enabling it to perceive its environment, make autonomous decisions, and execute complex tasks. Aletheia is precisely an agent designed following this concept. Its architecture revolves around three core pillars, allowing it to handle high-difficulty reasoning while leveraging external tools to ensure the accuracy and verifiability of results.

First, Aletheia is equipped with Gemini Deep Think as its deep reasoning engine. This is an advanced language model optimized for complex mathematical and logical reasoning, possessing powerful multi-step deduction capabilities and long-context understanding. It can handle Olympiad-level challenging problems and doctoral-level specialized questions while remembering and relating multiple intermediate results. Second, it implements reasoning-time extended computation, continuously deepening through multiple rounds of ``think-verify-correct'' loops until reliable conclusions are reached. Furthermore, Aletheia deeply integrates Google Search and web browsing functionalities, enabling it to verify theorems in real-time, search literature, avoid fabrication errors, and check complex expressions using external computational tools. This tool-calling mechanism liberates the model from closed knowledge recall, allowing it to explore based on the latest information, thereby effectively ending ``hallucinations'' and literature errors.

\noindent Aletheia's workflow is a typical iterative generate-verify-correct closed loop, applicable to various tasks from specific calculations to theoretical construction:

\begin{enumerate}
	\item Problem Input: The researcher poses a problem in natural language. For example, ``Calculate the eigenweights in the arithmetic Hirzebruch proportionality principle'' or ``Analyze the current status of Conjecture No. 47 in the Erdős problem database.''
	
	\item Generate Multi-round Reasoning: Based on the input problem, the model first generates a preliminary solution idea or proof sketch. Then, using reasoning-time extended computation, the model deeply deduces each step. If encountering an obstacle, it backtracks and tries a new path. During generation, the model can call search tools at any time to check for known results or verify whether a certain intermediate lemma has already been proven.
	
	\item Verification and Correction: After completing the preliminary result, the model performs self-verification: checking logical consistency and computational correctness. If errors or inconsistencies are found, it returns to the reasoning stage for correction, repeating the iteration until satisfied.
	
	\item Output Solution: Finally, it outputs the complete problem solution, accompanied by the reasoning process and literature citations.
\end{enumerate}
This process may seem simple, but the ``autonomous iteration'' capability it embodies is a key step in transforming large language models from ``question-answering tools'' to ``research partners.'' Aletheia has demonstrated powerful capabilities on several cutting-edge mathematical problems, with the following two cases being the most representative.
\vspace{3mm}

\noindent Case One: A Fully Autonomously Generated Mathematical Paper

Aletheia's most striking achievement is that it completely autonomously generated the mathematical content of a mathematical paper, from problem understanding and theory construction to final calculations, without any human intervention.

\begin{itemize}
	\item Problem Background: Feng, Yun, and Zhang established the (higher) arithmetic Hirzebruch proportionality principle in previous work. This is a profound result linking the arithmetic volume of shtukas moduli stacks to differential operators of L-functions. The formula involves certain structural constants called ``eigenweights,'' but the original authors only calculated these weights in simple cases; the general case remained unsolved.
	
	\item Aletheia's Contribution: The research team tasked Aletheia with calculating the eigenweights for all classical groups. During exploration, the agent autonomously discovered a profound connection between these weights and the representation theory of symmetric groups. It then used algebraic combinatorics tools (such as Young diagrams, Schur functors) to derive a closed formula for the general case. The entire research process, from problem understanding and theory construction to final calculations, was completed entirely independently by Aletheia.
\end{itemize}
The significance of this case lies not only in solving a specific problem but also in demonstrating that AI can make conceptual discoveries—it recognized the connection between eigenweights and representation theory and proactively invoked relevant theoretical tools to construct a solution. This goes beyond simple pattern matching or computational execution, entering the core realm of mathematical research.
\vspace{3mm}

\noindent Case Two: Semi-autonomously Tackling an Open Erdős Problem

If the first case demonstrates the ``depth'' of AI, the second case demonstrates its ``breadth'' and ``systematicity.''

\begin{itemize}
	\item Problem Background: The Erdős problem database, compiled by mathematician Bloom, contains about 700 conjectures marked as ``open.'' These problems vary in difficulty, and many have remained unsolved for a long time due to scattered literature or ambiguous formulation. For human mathematicians, examining these 700 problems one by one is a time-consuming and tedious task.
	
	\item Research Strategy: The research team adopted a hybrid approach: AI first performed natural language verification and literature retrieval to narrow the search space, followed by human expert evaluation for correctness and novelty. This collaborative model of ``AI preliminary screening + human confirmation'' leverages the strengths of both sides: AI is tireless and reads extensively, while humans possess deep judgment and creative understanding.
	
	\item Aletheia's Contribution: Aletheia systematically evaluated all 700 open problems, with remarkable results:
	\begin{itemize}
		\item Autonomously solved 4 problems: The solutions generated by AI appeared novel and were confirmed correct and previously unrecorded by human experts. These problems themselves may not be extremely difficult, but their ``open'' status meant no one had successfully solved them before; AI filled this gap.
		\item Assisted in identifying existing solutions for 9 problems: Through literature retrieval, AI assisted in discovering that these problems had actually been solved in some obscure papers, which might have been published in niche journals or written in obscure language, thus missed by mainstream databases. Their status was updated from ``open'' to ``solved.''
	\end{itemize}
	
	The research team noted that the ``open'' status of many of these problems stemmed more from the obscurity of the literature than from their inherent difficulty. In other words, some problems remained unsolved for a long time not because they were too hard, but because their solutions were buried in the vast ocean of literature. AI, with its powerful search and reasoning capabilities, effectively overcomes this. It can not only read and understand but also bridge language and journal barriers to establish connections in knowledge.
\end{itemize}
The success of the Aletheia agent marks AI's leap from a ``problem-solving tool'' to a ``research partner.'' It combines a deep reasoning engine, reasoning-time extended computation, and tool-calling capabilities to jointly construct an autonomous system capable of handling cutting-edge mathematical problems.

From fully autonomously generating papers to systematically tackling Erdős problems, Aletheia demonstrates that large language models can play a substantive role in mathematical discovery. In the future, with further improvements in model reasoning capabilities and the refinement of tool ecosystems, we can reasonably expect more agents to deeply participate in mathematical research.

\section{Creativity in Large Language Models}

Currently, research on the application of large language models in the field of mathematics is experiencing explosive growth. Mainstream evaluation benchmarks (such as GSM8K, MATH, etc.) and model optimization directions are almost entirely focused on the model's reasoning ability—that is, whether the model can derive the correct answer step-by-step and solve standardized mathematical problems. The achievements in this direction are exciting, with AlphaProof's performance on the IMO being the best proof.

However, mathematics is not merely a mechanical combination of logical deduction. The soul of mathematics lies in creativity—proposing unprecedented concepts, inventing ingenious methods, constructing counterexamples that overturn cognition. From the leap from Euclidean geometry to non-Euclidean geometry, from the conceptual extension of integers to complex numbers, from Fermat's Last Theorem to the final proof of the Poincaré conjecture, every major breakthrough in the history of mathematics has essentially been a victory of creativity. If reasoning ability makes AI an excellent ``problem solver,'' then creativity is the key to making it a true ``mathematician.''

But what is creativity? Can it be evaluated? Do current AI models possess this ability? These questions are becoming the next frontier in the exploration of the AI for Math field.

Currently, academic research on the mathematical creativity of LLMs receives very little attention, lacking systematic evaluation standards and specialized datasets. To scientifically evaluate creativity, we need an operational analytical framework. Drawing on research in the history and philosophy of mathematics, the paper DeepMath-Creative proposes an evaluation model based on three key dimensions:

\noindent \textcolor{structure3}{\textbf{Dimension One: Generation of New Concepts}}

This is the highest level of creativity, manifested in the introduction of unprecedented mathematical concepts or ideas, thereby opening up entirely new research fields. Concepts are the language of mathematics; the birth of a new concept often signifies a restructuring of the entire discipline.

The emergence of the Riemannian metric laid the foundation for modern differential geometry. It not only defined curvature at an abstract level but, more importantly, provided the precise mathematical language for Einstein's theory of general relativity—matter tells spacetime how to curve, spacetime tells matter how to move. Without Riemannian geometry, general relativity might have remained at the level of philosophical speculation.

The concept of a differentiable manifold perfectly combines local feasibility of calculus with global complex topological structures. It allows us to perform calculus on curved spaces like spheres and tori, becoming a cornerstone of theoretical physics and geometric modeling. Today, from the manifold hypothesis in machine learning to spacetime models in cosmology, manifolds are ubiquitous.

Group theory was born from Galois's exploration of the solvability of algebraic equations. This mathematician, only in his twenties at the time, pioneered a completely new structural way of thinking in the manuscripts written on the eve of his duel. Today, group theory unifies the study of symmetry in geometry, algebra, and number theory, with applications everywhere from crystal structures to quantum mechanics.

Topological spaces transcend the traditional concept of distance (metric spaces), defining continuity and proximity solely through ``open sets.'' This extreme abstraction elevates mathematical analysis to a more fundamental level—it turns out that the essence of continuity lies not in distance, but in the concept of ``neighborhood'' itself.

\noindent \textcolor{structure3}{\textbf{Dimension Two: Invention of New Methods}}

Another form of creativity is inventing entirely new techniques or tools to solve complex problems that were previously difficult to overcome. New methods often do not directly provide the answer but rather open a path towards it.

For example, the invention of the theory of generalized functions originated from the confusion caused by Dirac's $\delta$ function in quantum mechanics. This function is infinite at one point and zero elsewhere, yet its integral is 1—within the framework of classical function theory, it simply could not be defined. Mathematicians had to break through the inherent notion that ``a function is a point-to-point mapping,'' viewing functions instead as functional actions on test functions. This breakthrough not only provided a rigorous mathematical foundation for quantum mechanics but also became a core tool in the modern theory of partial differential equations.

The Bochner technique is a milestone in geometric analysis. It cleverly links geometry (curvature) with analysis (the Laplace operator), using curvature conditions to constrain the topological properties of manifolds. This method of ``attacking geometry with analysis'' provides an elegant pathway for proving many profound geometric theorems.

\noindent \textcolor{structure3}{\textbf{Dimension Three: Creation of New Examples}}

Testing the boundaries of propositions by constructing concrete examples (especially counterexamples) is an important driver for theoretical development. A clever counterexample often deepens our understanding of a concept more than ten positive examples.

In 1956, John Milnor constructed a seven-dimensional manifold that is topologically equivalent to the standard seven-dimensional sphere (homeomorphic) but different in its differential structure (not diffeomorphic). This ``exotic sphere'' overturned people's intuitive understanding of high-dimensional spaces—it turns out that different smooth structures can exist on the same topological space. This discovery directly initiated the theory of exotic manifolds and the entirely new field of differential topology.

Earlier, in 1872, Weierstrass constructed a function that is continuous everywhere but differentiable nowhere. Before this, mathematicians generally believed that continuous functions were always differentiable, at least at most points. This counterexample directly challenged the foundations of analysis at the time, forcing mathematicians to re-examine basic concepts like function, continuity, and differentiability, and promoting the rigorization of real analysis theory. It reminds us: intuition may deceive us; only strict logic can guide us to truth.

These three dimensions—new concepts, new methods, new examples—together constitute the complete spectrum of mathematical creativity. They exist at different levels but are equally important. A new concept may open up a field, a new method may solve a batch of problems, and a new example may correct the direction of an era.

The process by which human mathematicians create new concepts and methods (such as drafts, failed attempts, moments of inspiration) is often chaotic, unstructured, and full of chance. It is difficult for us to reconstruct these complex cognitive trajectories into large-scale, standardized datasets suitable for training large language models. More importantly, when a model generates a ``novel'' concept or method, it is difficult for us to judge: is this genuine originality, or merely a clever recombination of training data? Existing automatic evaluation mechanisms can hardly distinguish between these two situations.

Unlike abstract concepts and methods, concrete, constructible mathematical examples are ``tangible.'' Whether an example is correct or constitutes a counterexample can be judged through rigorous mathematical verification, which is a unique advantage of the discipline of mathematics. This verifiability provides a relatively objective and feasible standard for evaluating machine creativity.

The value of mathematical creativity extends far beyond solving known problems. Looking back at every leap in the development of mathematics—from the birth of non-Euclidean geometry to the rise of topology, from the proposal of Galois theory to Wiles's proof of Fermat's Last Theorem—each is a crystallization of creativity. It is precisely the introduction of new concepts, the invention of new methods, and the construction of new examples that continuously reshape the landscape of mathematics and expand the boundaries of human reason. In the era of artificial intelligence, endowing large language models with mathematical creativity will have even more profound significance: it is not only a touchstone for testing whether machines possess true intelligence, but it may also give rise to an ``AI mathematician'' or ``AI collaborator'' capable of exploring mathematical conjectures, generating proof ideas at unprecedented speeds, and even discovering hidden patterns and connections within complex structures beyond human reach, thereby accelerating the process of scientific discovery and driving transformative breakthroughs in mathematics and related disciplines (such as theoretical physics, cryptography, computational biology).

However, on the path to this vision lie major core technical challenges. Among them, the most central problem is the difficulty of defining and designing reward functions that can effectively evaluate mathematical creativity within the reinforcement learning framework. The effectiveness of reinforcement learning highly depends on a clear, quantifiable reward signal that can guide the model to gradually approach the goal. But in the context of mathematical creativity, this requirement encounters fundamental dilemmas:
\begin{enumerate}
	\item The subjectivity and context-dependence of novelty: What is ``novel''? A construction may be unknown to the model itself, but perhaps it is common knowledge to the entire mathematical community. True creativity requires stepping outside the distribution of training data to produce entirely new knowledge with ``historical'' significance, and the judgment of this novelty often requires deep mathematical insight and cannot be measured by simple statistical indicators (such as dissimilarity from training data). Whether a construction is ``clever'' or ``profound'' is a higher-order aesthetic judgment, difficult to formalize into a scalar reward.
	
	\item The unpredictability of long-term impact: Just as the value of the Riemannian metric or exotic spheres was fully recognized only decades later, the true impact of mathematical creation often only becomes apparent in the distant future. At each step of reinforcement learning, the model cannot predict what chain reaction its current construction fragment might produce in the future. This delayed and sparse reward makes traditional optimization methods based on short-term returns (such as policy gradient) almost ineffective.
	
	\item The exponential sparsity of the exploration space: The construction space of mathematical objects (such as manifolds, groups, functions) is infinite and highly unstructured. Truly creative constructions (like the Weierstrass function) are like a drop in the ocean. For a model to randomly wander in such a vast space, it is almost impossible to stumble upon a valuable example by mere ``trial and error.'' Without effective intrinsic motivation or heuristic guidance, reinforcement learning will fall into endless ineffective exploration.
	
	\item The tension between correctness and creativity: Mathematical creativity must be built on a foundation of logical rigor. A ``novel'' but self-contradictory counterexample is worthless. Therefore, the reward function must achieve a delicate balance between encouraging novelty and enforcing correctness. Overemphasizing correctness will make the model tend to conservatively replicate known knowledge; overemphasizing novelty may lead the model to generate a large number of meaningless, logically broken constructions. How to design a reward mechanism that can tolerate errors during exploration while ultimately guiding the model towards logical closure is a blind spot in current technology.
	
	\item The evaluation dilemma of process vs. result: The value of mathematical creation is often embedded in the thinking process itself—from vague intuition to rigorous construction, from failed attempts to moments of insight. However, existing reinforcement learning primarily focuses on the final output result (e.g., whether the counterexample is correct). A reward signal capable of capturing the ``insight,'' ``analogical transfer ability,'' or ``handling of contradictions'' demonstrated by the model during the construction process almost does not exist.
\end{enumerate}
These challenges collectively point to a deeper issue: we still lack a computational theory capable of understanding and evaluating mathematical creation. Creativity is not a simple quantifiable metric but a complex, multi-dimensional phenomenon deeply bound to context. To make it a computable, optimizable goal, we need a deeper understanding of creativity itself.

Future research may need to go beyond a single, external reward function, exploring instead composite paradigms such as intrinsic motivation mechanisms (e.g., rewards based on curiosity or information gain), metacognitive mechanisms (allowing the model to learn to evaluate its own thinking paths), and human-machine collaborative evaluation (involving human mathematicians in reward shaping).

Regardless, facing and overcoming this core technical challenge will be the necessary path towards intelligent systems with genuine mathematical creativity. Because ultimately, what we anticipate is not just a machine that can solve problems, but a partner that can think, create, and explore the unknown realms of mathematics with us. This partner may not have moments of inspiration like human mathematicians, but it may discover mathematical truths that are difficult for humans to see alone—through massive exploration, precise pattern recognition, and tireless combination—in a completely different way.

\nocite{*}

\printbibliography[heading=subbibliography,title=References]

\end{refsection}

\begin{refsection}[ref5.bib]
\chapter{AI for Constructing Counterexamples}
\section{Overview}

The core driving force of mathematics is not only to prove theorems, but also to delineate their boundaries. Constructing counterexamples is precisely the art of revealing ``where a conjecture fails.'' A beautiful proof tells us ``truth ends here,'' while a brilliant counterexample tells us ``falsehood begins there.''

Constructing counterexamples is the most subversive art in mathematical logic—it not only declares the end of a path but also delineates the insurmountable boundaries of truth. In the history of mathematics, an ingenious counterexample often reveals the inherent fissures within a concept more powerfully than ten conventional proofs, forcing the revision of axiomatic systems, the reconstruction of definitions, and even giving rise to entirely new research paradigms. Artificial intelligence, through brute-force traversal and intelligent pruning via deep search, reinforcement learning, or generative models, can quietly infiltrate the sparse corners beyond human imagination. This approach does not attempt to replace the mathematician's stroke of genius that ignites a counterexample with inspiration. Instead, it transforms the discovery of counterexamples from a personal art reliant on serendipitous insight into an engineering process that is scalable and systematically reproducible.

\section{AI Constructs Counterexamples in Graph Theory: Based on Reinforcement Learning}

Combinatorics and graph theory are fascinating and challenging branches of mathematics. The objects they study are often discrete, finite, and structured. In these fields, mathematicians have proposed numerous conjectures concerning the relationships between extremal combinatorial and graph parameters. These conjectures typically arise from observations in numerical experiments, inductions from specific instances, or generalizations of known results. They serve as beacons, guiding the direction of exploration.

However, the history of mathematical conjectures tells us: not all seemingly plausible propositions are correct. Determining whether a conjecture holds, especially finding a counterexample to refute it, is often an extremely arduous task. The essence of the problem lies in the vastness of the search space, where the target (a structure violating the conjecture's conditions) might be just a ``needle in a haystack''.

Take graph theory as an example. For a simple undirected graph with \( n \) vertices, the number of all possible graphs is \( 2^{\binom{n}{2}} \), a number that grows exponentially with \( n \). When \( n=20 \), the number of possible graphs already exceeds \( 10^{57} \). Clearly, it is completely infeasible to examine all possibilities through manual enumeration or traditional exhaustive algorithms.

For a long time, mathematicians have relied on intuition, structural insights, and theoretical deduction to guess the possible forms of counterexamples. They also use algorithmic tools such as heuristic search and linear programming for assistance. For instance, for certain problems that can be expressed as linear programs, commercial solvers can find optimal solutions or counterexamples very efficiently. However, many combinatorial problems cannot be concisely expressed as linear programs, or the structure of their search space is very complex, making it difficult for traditional optimization methods to explore effectively.

Thus, we face a core question: when confronted with a vast, discrete, and structurally complex search space where we need to find a ``rare'' structure satisfying specific (often ``pathological'' or non-intuitive) conditions, does there exist a systematic method that can effectively explore and discover in a ``black-box'' manner, without relying on deep problem-specific insights?

Such a method should possess the following characteristics: generality (able to handle a wide range of problem formulations), minimal reliance on domain knowledge (reducing dependence on problem-specific heuristics), effective exploration of high-dimensional discrete spaces, and some ``learning'' ability to accumulate experience from failed attempts. This is precisely the stage where reinforcement learning, especially policy search methods, can play a role.

The core idea of reinforcement learning is to let an agent learn how to perform a sequence of actions through interaction with an environment to maximize cumulative reward. This paradigm can be mapped very intuitively to our mathematical construction problem:

\begin{itemize}
	\item Agent: An algorithm attempting to construct a mathematical object (e.g., a graph, matrix, set family).
	\item Environment: Defines the rules for constructing the object (e.g., constructing a simple graph of order $n$) and the metric for evaluating the quality of the final object.
	\item State: At a certain point in the construction process, the partial decisions made so far (e.g., some edges already determined).
	\item Action: A specific decision the agent can make in the current state (e.g., adding a specific edge next, or not adding it).
	\item Reward: A score calculated after the entire object is constructed, based on whether it violates the conjecture (or the degree of violation). Our goal is to minimize this score (for lower-bound conjectures) or maximize it (for upper-bound conjectures) to find a counterexample.
\end{itemize}

Specifically, the process of constructing a combinatorial object can be viewed as generating a string in a certain order. For example, to construct a graph with $n$ vertices, we can consider all possible \( \binom{n}{2} \) edges in a fixed order, deciding for each edge to ``keep'' (encoded as 1) or ``delete'' (encoded as 0). Thus, any graph uniquely corresponds to a binary string of length $\binom{n}{2}$. The agent's task is to generate this string step by step.

The key idea is that the agent knows nothing about the ``mathematical problem'' it is solving. It merely learns a policy—a mapping from the ``current generated string fragment'' to a probability distribution over the ``next character''. This policy is adjusted through trial and error: it generates many complete strings (i.e., constructs many graphs), each receiving a score based on the reward function. Then, the agent analyzes those constructions with high (or low, depending on the goal) scores and adjusts its policy to make decisions more similar to these successful constructions in the future.

To better understand the above method, we introduce Wagner's work. Wagner chose the deep cross-entropy method as the foundational tool.

The cross-entropy method is essentially an evolutionary algorithm based on policy search. Its core is to maintain a parameterized policy network (typically a neural network) and improve the policy through iterative ``generate-evaluate-evolve'' steps. We break down its workflow in the context of combinatorial construction.

Suppose we want to construct a discrete object \( x \), generated by a series of decisions \( a_1, a_2, \ldots, a_T \), where each decision \( a_t \) belongs to a finite action set \( \mathcal{A} \). For example, in constructing a graph, \( T = \binom{n}{2} \), \( \mathcal{A} = \{0, 1\} \). There exists a reward function \( R(x) \) evaluating the quality of the final object \( x \). Our goal is to find the object \( x^* \) that maximizes (or minimizes) \( R(x) \).

A policy \( \pi_\theta \) is a function with parameters \( \theta \), which gives the probability distribution over actions given the current partial sequence (state \( s_t \)): \( \pi_\theta(a | s_t) \). In the sequential generation setting, the state \( s_t \) is typically the sequence of decisions made so far \( (a_1, \ldots, a_{t-1}) \).

We wish to learn parameters \( \theta \) such that objects \( x \) generated according to policy \( \pi_\theta \) have high expected reward \( \mathbb{E}_{x \sim \pi_\theta}[R(x)] \).

The deep cross-entropy method approximates the optimal policy through the following iterative steps:

\begin{enumerate}
	\item Sampling Phase: Generate $N$ complete objects $x_1, x_2, \ldots, x_N$ independently according to the current policy $\pi_\theta$, and compute the reward $r_i = R(x_i)$ for each object.
	
	\item Selection Phase: Sort all samples by reward, retaining only the top $K = \lceil \rho N \rceil$ elite samples, where $\rho$ is the elite retention ratio (typically 5\% to 20\%).
	
	\item Update Phase: Analyze the generation trajectories of these elite samples, i.e., the decision sequences at each step. For each elite sample's generation sequence $(a_1, \ldots, a_T)$, construct training data: for each step $t$, state $s_t = (a_1, \ldots, a_{t-1})$, action $a_t$. Then, update the policy by minimizing the cross-entropy loss:
	$$L(\theta) = -\sum_{(s,a) \in \text{elite trajectories}} \log \pi_\theta(a | s)$$
\end{enumerate}
The cross-entropy form of the loss function \( L(\theta) \) is the core. For a given (state, action) pair \( (s, a) \), we want the policy network's output probability \( \pi_\theta(a|s) \) for action \( a \) in that state to be as large as possible. \( -\log \pi_\theta(a|s) \) measures the difference between the predicted probability and the perfect prediction (probability $1$). Minimizing the sum of these differences over all elite trajectories is ``stretching'' the policy network, aligning its decision distribution closer to the empirical distribution of the elite samples.

This process repeats until a satisfactory construction is found or a preset number of iterations is reached. Algorithm\ref{alg:deep_cross_entropy} is the pseudocode for this deep cross-entropy method.

To put the above algorithm into practice, we need to encode the partial construction sequence \( s_t \) into a fixed-dimensional vector that the neural network can process. Wagner adopted a simple yet general encoding scheme, particularly suitable for tasks generating binary strings (e.g., the upper triangular part of a graph's adjacency matrix).

The input is the concatenation of two vectors:
\begin{itemize}
	\item \textbf{Decision Vector}: Length $T$, where the $i$-th component indicates whether a decision has been made at the $i$-th position. If decided, the component is the actual value ($0$ or $1$); if not yet decided, the component is $0$. This vector records the ``history'' so far.
	\item \textbf{Position Indicator Vector}: This is a one-hot vector, with only the $t$-th component being $1$ and the rest $0$. It explicitly tells the network which position it is currently making a decision for.
\end{itemize}

Why are two vectors needed? The decision vector provides context, and the position vector provides focus. The network needs to combine them to understand: ``Based on the decisions I have already made (decision vector), what is the higher probability I should choose $0$ or $1$ for this specific position (position vector)?''

Regarding the neural network architecture, Wagner used a simple Multi-Layer Perceptron (MLP):

\begin{itemize}
	\item Input Layer: Size $2T$ (concatenation of two vectors)
	\item Hidden Layers: Typically 2-3 fully connected layers with ReLU activation function. For example, the paper mentions a three-layer structure with 128, 64, and 4 neurons.
	\item Output Layer: Size $|\mathcal{A}|$, with two neurons for binary decisions, converted into a probability distribution \( \pi_\theta(0|s_t) \) and \( \pi_\theta(1|s_t) \) via the softmax function.
\end{itemize}

\[
\begin{array}{c}
	\text{[State } s_t\text{]} \\
	\downarrow \\
	\text{[Decision Vector } \oplus \text{ Position Vector]} \\
	\downarrow \\
	\text{[Fully Connected Layer (128) + ReLU]} \\
	\downarrow \\
	\text{[Fully Connected Layer (64) + ReLU]} \\
	\downarrow \\
	\text{[Fully Connected Layer (4) + ReLU]} \\
	\downarrow \\
	\text{[Output Layer (2) + Softmax]} \\
	\downarrow \\
	\text{[Probability Distribution } (P(0), P(1))\text{]}
\end{array}
\]

\begin{algorithm}[htbp]
	\caption{Deep Cross-Entropy Method}
	\label{alg:deep_cross_entropy}
	
	\KwIn{
		Reward function $R(\cdot)$,\;
		Action space $\mathcal{A}$, sequence length $T$,\;
		Policy network $\pi_\theta$, initial parameters $\theta_0$,\;
		Samples per round $N$, elite retention ratio $\rho \in (0,1)$ (elite count $K = \lceil \rho N \rceil$),\;
		Learning rate $\alpha$, stopping condition
	}
	
	\KwOut{Best construction found during training $x^* = \arg\max_{x} R(x)$}
	
	Initialize policy network parameters $\theta \leftarrow \theta_0$\;
	
	\While{stopping condition not met (e.g., counterexample found or max iterations reached)}{
		\textbf{Sampling Phase}:\;
		\begin{enumerate}
			\item Initialize sample set $\mathcal{S} \leftarrow \emptyset$\;
			\For{$i = 1$ \KwTo $N$}{
				Initialize empty sequence $s \leftarrow \epsilon$\;
				\For{$t = 1$ \KwTo $T$}{
					Sample action $a_t$ according to $\pi_\theta(\cdot|s)$\;
					$s \leftarrow s \oplus a_t$\;
				}
				Decode complete object $x_i$ from $s$, compute reward $r_i = R(x_i)$\;
				Add $(x_i, r_i, s)$ to $\mathcal{S}$\;
			}
		\end{enumerate}
		
		\textbf{Selection Phase}:\;
		Sort samples in $\mathcal{S}$ by reward $r_i$\;
		Let $\mathcal{E}$ be the set of top $K$ elite samples\;
		
		\textbf{Update Phase}:\;
		Initialize training set $\mathcal{D} \leftarrow \emptyset$\;
		\ForEach{generation sequence $(a_1, \ldots, a_T) \in \mathcal{E}$ of an elite sample}{
			\For{$t = 1$ \KwTo $T$}{
				$s_t \leftarrow (a_1, \ldots, a_{t-1})$\;
				$\mathcal{D} \leftarrow \mathcal{D} \cup \{(s_t, a_t)\}$\;
			}
		}
		
		Compute cross-entropy loss:
		\[
		L(\theta) = -\sum_{(s,a) \in \mathcal{D}} \log \pi_\theta(a|s)
		\]
		
		Gradient descent update:
		\[
		\theta \leftarrow \theta - \alpha \cdot \nabla_\theta L(\theta)
		\]
	}
	
	\Return Best construction found during training $x^* = \arg\max_{x} R(x)$
\end{algorithm}

This architecture, though simple, is sufficient to capture dependencies between decisions in many combinatorial problems. For objects like graphs with strong structural relationships, Graph Neural Networks (GNNs) might be a more natural choice, as they can directly take the graph structure (vertices, edges) as input and utilize message-passing mechanisms to aggregate neighborhood information. However, the generality of a simple MLP allows it to be quickly applied to various problems without designing specialized network structures for each new problem.

Wagner's paper demonstrates the successful application of this method to several graph theory and combinatorial conjectures. Below, we select two typical examples to elaborate on how AI works and what mathematical insights it reveals.
\vspace{3mm}

\noindent Example One: Conjecture on the Sum of Eigenvalue and Matching Number
\begin{quote}
	\textcolor{second}{\textbf{Conjecture:} For any connected graph $G$ with $n \geq 3$, the sum of its largest eigenvalue $\lambda_1$ and matching number $\mu$ satisfies $\lambda_1 + \mu \geq \sqrt{n-1} + 1$.}
\end{quote}

\noindent AI Setup Details:
\begin{itemize}
	\item Object Representation: Graph with $n$ vertices, encoded as a binary string of length $T = \binom{n}{2}$.
	\item Reward Function: $R(G) = -(\lambda_1(G) + \mu(G))$, aiming to minimize this sum to find a potential counterexample.
	\item Task Scale: Search for $n = 19$.
	\item Computational Details: Each reward evaluation requires computing the largest eigenvalue of the graph's adjacency matrix and the maximum matching. Total training iterations approximately 5000 rounds.
\end{itemize}

\noindent AI's Discovery Process:
\begin{enumerate}
	\item Initial Phase: Random policy, generating various dense and sparse graphs.
	\item Learning Trend: The network quickly (within a few hundred iterations) ``learns'' from reward feedback that sparse graphs tend to have smaller \( \lambda_1 \). Simultaneously, to maintain connectivity (which the reward function may not explicitly require, but sparsity easily leads to disconnection), it discovers trees are a good candidate category: they are minimally connected graphs.
	\item Structural Evolution: From random graphs, to sparse disconnected graphs, to trees, and finally focusing on a specific tree structure—``balanced double star'': a central edge connecting the centers of two stars, each star having a similar number of leaves. This structure effectively keeps the largest eigenvalue low while also having a small matching number, all while maintaining connectivity.
	\item Successful Counterexample: When $n=19$, the finally found graph satisfies $\lambda_1 = \sqrt{10} \approx 3.162$, $\mu = 2$, sum approximately 5.162, which is less than $\sqrt{18}+1 \approx 5.243$, successfully refuting the conjecture.
\end{enumerate}

This case demonstrates how AI, starting from scratch, through pure trial-and-error learning, discovered a graph structure that human mathematicians might not have considered and successfully used it to overturn the conjecture.
\vspace{3mm}

\noindent Example Two: Conjecture on Distance Spectrum and Proximity

\begin{quote}
	\textcolor{second}{\textbf{Conjecture:} Concerning the distance matrix eigenvalues $\partial_i$ and the proximity parameter $\pi$ of a graph, they satisfy $\pi + \partial_{\lfloor 2D/3 \rfloor} > 0$, where $D$ is the diameter.}
\end{quote}

\noindent AI Setup Details:
\begin{itemize}
	\item Reward Function: $R(G) = -(\pi(G) + \partial_{\lfloor 2D/3 \rfloor}(G))$, aiming to find graphs making this sum as small as possible (even negative).
	\item Task Scale: Initial search for $n = 30$.
	\item Computational Challenge: Computing distance matrix eigenvalues and proximity is more time-consuming than for adjacency matrices, especially for non-tree graphs requiring all-pairs shortest path algorithms. This limits the number of samples $ N $ that can be evaluated per round.
\end{itemize}

\noindent AI's Discovery Process:
\begin{enumerate}
	\item Failed to directly find a counterexample: After several days of training, the best graph found by AI for $n=30$ had a sum value of about $0.4$, still positive, not a counterexample.
	\item Key Insight—Structural Convergence: Although not a counterexample, a strong pattern emerged: among all elite samples, the graph structures were highly consistent. They all consisted of a long path, connected near its midpoint to a large star, whose neighbors themselves formed small cliques. The only variation was the sizes of these small cliques, as shown in Figure\ref{fig:Counterexample} (cited from the original paper).
	\begin{figure}[htbp]
		\centering
		\includegraphics[width=0.5\linewidth]{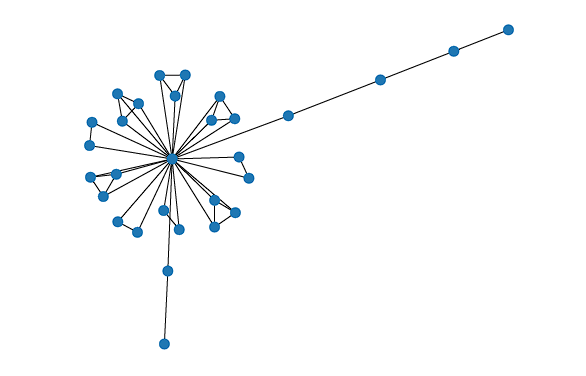}
		\caption{Reinforcement Learning Constructing Counterexamples \label{fig:Counterexample}}
	\end{figure}
	\item From Pattern to Counterexample: This pattern provided human researchers with a clear ``design blueprint''. Based on this, Wagner et al. manually constructed a parameterized family of graphs: a path of length \( d \), with \( m \) leaves attached (i.e., a star) to a vertex near the midpoint. Through analysis, they found that when \( d=12 \) (diameter \( D=12 \), \( \lfloor 2D/3 \rfloor = 8 \)) and \( n \) is sufficiently large (\( \geq 190 \)), this ``double-tailed comet'' graph indeed satisfies \( \pi + \partial_{8} < 0 \), thus becoming a counterexample.
\end{enumerate}

This case highlights another role of AI in mathematical discovery—a super pattern detector. When directly searching for a counterexample is computationally difficult, AI can reveal the topological features that potential extremal structures must satisfy by outputting a series of ``near-optimal'', structurally similar candidate solutions. This greatly reduces the human search space, transforming the problem from ``searching among all graphs'' to ``optimizing parameters within a graph family of a specific pattern''.

Based on the above examples, the general workflow of using reinforcement learning for mathematical construction can be seen as follows:

\noindent Step 1: Problem Formulation

Transform the specific mathematical construction problem into a reinforcement learning framework, defining state representation, action space, reward function, and termination conditions. This is the most critical step, requiring a deep understanding of the nature of the mathematical problem. A good formulation should allow the reward function to accurately reflect the problem's objective while ensuring the state representation contains sufficient information for decision-making.

\noindent Step 2: Algorithm Selection and Implementation

Choose an appropriate reinforcement learning algorithm. For mathematical construction problems with sparse rewards and discrete action spaces, policy-based methods like policy gradient methods and the cross-entropy method often perform well. These methods directly optimize the policy without needing to learn complex value functions, making them more suitable for combinatorial problems with huge exploration spaces.

\noindent Step 3: Training Process

The agent learns through interaction with the environment (i.e., the construction process). In each training cycle, the agent uses the current policy to generate multiple complete constructions, computes the reward for each, and then uses this information to update the policy. This process repeats until the policy converges or a satisfactory construction is found.

\noindent Step 4: Result Extraction and Analysis

After training, use the learned policy to generate constructions and perform mathematical verification. If a counterexample or extremal construction is found, further analyze its structure to gain mathematical insights. Sometimes, even if no counterexample is found, the ``near-optimal'' constructions discovered by AI can provide valuable clues for human researchers.

\noindent The essential points of the method can be summarized as:
\begin{enumerate}
	\item General Framework: The same reinforcement learning code can be used to attack different mathematical conjectures, requiring only a change of the reward function. This generality is difficult to match with traditional methods.
	
	\item Learning from Experience: The reinforcement learning agent accumulates experience through extensive trial and error, gradually learning which decision patterns are more likely to produce high-reward constructions. This is similar to how human mathematicians guide research by accumulating intuition and experience, but reinforcement learning can explore possibilities more systematically and extensively.
	
	\item Handling High-Dimensional Combinatorial Spaces: Reinforcement learning methods, especially when combined with neural networks, can handle extremely high-dimensional state and action spaces, which is unattainable for traditional exhaustive methods.
	
	\item Exploring Unconventional Structures: Since reinforcement learning algorithms do not rely on human prior intuition, they have the potential to discover counterintuitive, unconventional constructions that human researchers might easily overlook.
	
	\item Sparse Reward Challenge: Mathematical construction problems typically have only final rewards, making learning very difficult. Specialized techniques, such as curriculum learning (starting from simple problems and gradually increasing difficulty) or intrinsic rewards (encouraging exploration of new states), are needed to assist the learning process.
	
	\item Scalability: Once a policy is trained, it can quickly generate a large number of high-quality constructions, which is valuable for studying the distribution of constructions or finding multiple different counterexamples.
\end{enumerate}
It is worth noting that while reinforcement learning provides powerful search capabilities, it does not guarantee finding the optimal solution or a counterexample. The algorithm's success depends on various factors, including the quality of problem formulation, the design of the reward function, the choice of algorithm, and the tuning of hyperparameters.

From a broader perspective, the application of reinforcement learning in discovering graph theory counterexamples represents an important paradigm in AI for Math: when human intuition and traditional algorithms struggle to reach the problem space, AI can discover hidden, counterintuitive structures through systematic exploration and learning. These discoveries not only verify or refute conjectures but, more importantly, enrich our understanding of mathematical structures and inspire new theoretical directions.

\section{The PatternBoost Method}

In the previous section, we explored how to leverage reinforcement learning and heuristic search for efficient construction within discrete combinatorial spaces. We saw the great potential of pure reinforcement learning methods (such as the cross-entropy method), while also recognizing their limitations: when the size of the constructed object (e.g., sequence length) becomes large (for instance, several hundred symbols), a naive neural network struggles to effectively learn and predict the entire structure. This difficulty is fundamentally rooted in the inherent conflict between ``exploration'' and ``exploitation,'' which is particularly pronounced in the vast, discrete, and structurally complex space of mathematical constructions.
\\
Specifically, we can view the mathematical construction task from two perspectives:
\begin{enumerate}
	\item Local Perspective: Starting from a given construction (such as a graph or a matrix), attempt to improve the degree to which it satisfies the target constraints through a series of ``small-step'' modifications (e.g., adding or deleting an edge, an element). This is a ``fine-tuning'' style of optimization. Its advantage lies in the ability to conduct detailed exploration near known ``good solutions,'' but its drawback is the tendency to get stuck in local optima, lacking the ability to break out of the current structural pattern. For example, for a triangle-free graph, local search can try adding new edges to existing non-triangle edges, but it struggles to conceive an entirely new, structurally distinct bipartite graph layout.
	
	\item Global Perspective: Grasping the overall patterns and structural characteristics of ``good constructions'' from a macro level. For instance, by analyzing hundreds of excellent triangle-free graphs, a human mathematician might discern that most exhibit a ``bipartite graph'' structure. The global perspective can provide novel ideas to escape local patterns, but its disadvantage is the difficulty in precisely generating concrete instances that satisfy all micro-level constraints. A pure generative model might learn the pattern of ``bipartiteness,'' but the graphs it randomly generates may still contain some triangles.
\end{enumerate}
In mathematical research, this ``local adjustment'' (e.g., trying to modify a known counterexample) and ``global conception'' (e.g., conjecturing a new structural framework based on symmetry or extremal principles) are typically alternated. A method purely based on reinforcement learning or local search often lacks effective global conception capability. Conversely, a pure generative model (e.g., directly training a Transformer to output constructions) may generate a large number of invalid outputs due to its inability to precisely satisfy complex, combinatorial constraints.

Therefore, a natural question arises: Can we design a hybrid method that allows machines, like mathematicians, to effectively iterate between local fine-tuning and global pattern learning? This method needs to leverage both the precision of local search and the creativity of global learning. This is precisely the core motivation behind the PatternBoost method. It aims not to replace humans, but to build an automated system that can simulate the cycle of ``hands-on experimentation'' and ``reflective abstraction'' in human mathematical research. Its target mathematical problems are clear: in those structurally complex combinatorial spaces, find constructions that satisfy specific constraints, possess extremal properties (e.g., maximum number of edges, maximum size), or serve as counterexamples to conjectures.

The essence of PatternBoost lies in its simple yet powerful iterative framework. It decomposes the entire construction search process into two alternating phases: local search and global learning. This cycle can be understood through a clear analogy: imagine a community of bicycle designers (local phase) each fine-tuning and optimizing their own bicycle designs to make them more comfortable and efficient. These excellent bicycles are put into use, filling the streets. A new generation of designers (global phase) observes the various optimized bicycles on the streets, abstracts the common features and patterns of successful designs, and conceives new bicycle designs based on these patterns, yet in a similar style. These new designs are then taken back by the designers for fine-tuning optimization, and a new round of the cycle begins.
\\
In the mathematical context, this cycle is formalized into a repeatable algorithmic process:

\begin{enumerate}
	\item Local Phase (Producing Good Solutions): Use a (typically simple) local search algorithm, starting from one or more ``seed'' constructions, to explore the nearby space through a series of allowed local operations (e.g., adding/deleting elements, swapping positions). The goal is to obtain one or more ``high-quality'' constructions. Here, ``high-quality'' is quantified by a scoring function that measures how well the construction satisfies our objective (e.g., for the problem of maximizing edges in a triangle-free graph, the score is the number of edges; to guide the search, a scoring function that penalizes the existence of triangles can also be designed).
	
	\item Global Phase (Learning Patterns): Collect a batch of optimal constructions produced in the local phase, using them as training data. Use a machine learning model capable of capturing sequential or structural patterns (in PatternBoost, primarily Transformer-based generative models) to learn the implicit ``construction patterns'' within these good solutions. After training, let this model generate a batch of new constructions ``similar'' to the training data, serving as new ``seeds.''
	
	\item Iteration: Feed the new seeds generated in the global phase back into the local search algorithm, starting a new round of the cycle. As the cycle progresses, the starting points for local search become increasingly better (because they come from a model that has already learned ``good patterns''), and the quality of the dataset used to train the model also improves (because they are the results of continuously optimized local search). Ideally, this process forms a positive feedback loop, continuously approaching or even discovering optimal or groundbreaking constructions.
\end{enumerate}
The beauty of this framework lies in its modularity and generality. The local search algorithm and the global learning model can be customized or replaced according to the specific problem. The authors of the PatternBoost paper specifically emphasize their desire to provide a mathematician-friendly tool that does not require deep machine learning expertise. Therefore, in their experiments, they extensively used very simple local search and a lightweight Transformer implementation to demonstrate that the method remains effective even in this ``naive'' setting.

Next, we will delve into each key component of the PatternBoost workflow, using the paper's core example—``constructing the graph with the maximum number of edges among graphs on n vertices that contains no triangles (i.e., verifying/discovering the extremal graph of Mantel's theorem)''—as a running case study.
\vspace{3mm}

\noindent \textcolor{structure3}{\textbf{Core Component One: Local Search – From Seeds to Good Solutions}}

Local search is the foundational engine of PatternBoost. It needs to accomplish two tasks: produce a feasible, improved construction from a starting point; and define what constitutes a ``local'' move.

Let's understand this using the paper's core example: constructing the graph with the maximum number of edges among graphs on $n$ vertices that contains no triangles—this is precisely the problem described by the famous Mantel's theorem in graph theory, whose extremal structure is the complete bipartite graph $K_{\lfloor n/2\rfloor, \lceil n/2\rceil}$.

For this problem, the paper adopts an extremely simple two-stage greedy strategy as the local search:

\begin{enumerate}
	\item Eliminate Violations: If the input graph contains triangles, repeatedly delete a random edge that participates in the most triangles until the graph contains no triangles. This operation is greedy, aiming to quickly reduce the ``violation'' degree.
	
	\item Expand and Optimize: After obtaining a triangle-free graph, repeatedly attempt to add a new random edge. If adding this edge does not create a new triangle, keep it; otherwise, reject the addition. Repeat this process until no more edges can be added without creating a triangle.
\end{enumerate}

This algorithm is very simple; it does not utilize any deep knowledge about graph structure (such as bipartiteness), relying purely on random attempts. Its scoring function is also straightforward: in the final triangle-free graph, the score is the number of edges. The authors use this simple setup to highlight the improvement brought by the subsequent Transformer's global learning, rather than the intelligence of the local search itself.

Why is local search needed? Because the constructions directly output by generative models (like Transformers) may not be perfect in terms of micro-level constraints. For example, a graph generated by a Transformer that has learned the ``rough structure of a bipartite graph'' might still contain a few sporadic triangles. Local search acts like a ``proofreader'' or ``polisher,'' responsible for refining these rough, potentially flawed conceptions into mathematical objects that strictly satisfy all constraints and can be precisely evaluated. Without this step, a large number of generated results would be invalid and unable to enter subsequent evaluation and iteration.
\vspace{3mm}

\noindent \textcolor{structure3}{\textbf{Core Component Two: Global Learning – Capturing Construction ``Patterns'' with Transformers}}

After collecting a batch of high-quality constructions (e.g., those graphs with the highest edge counts selected from numerous local search runs), PatternBoost enters the global learning phase. The core idea is: these good solutions must contain some generalizable ``success pattern,'' and the task of the machine learning model is to discover and learn this pattern.

Why choose Transformers? The Transformer architecture excels at processing sequential data, particularly in capturing long-range dependencies. A mathematical construction (such as a graph's adjacency matrix, a matrix's entry sequence, a point set's coordinate list) can be ``serialized'' into a string of symbols. Through its self-attention mechanism, the Transformer can learn the relationships between elements at any two positions in the sequence, which is crucial for understanding the overall layout of combinatorial structures (e.g., the connection pattern between two parts of a graph). Compared to simple feedforward neural networks used in prior work, Transformers can handle longer sequences, thus enabling modeling of larger-scale constructions.

To input a mathematical object into a Transformer, it first needs to be represented as a series of discrete symbols. Taking a triangle-free graph on 20 vertices as an example:

\begin{enumerate}
	\item Matrix Representation: The graph can be represented by a $20\times 20$ symmetric adjacency matrix, with zeros on the diagonal. Due to symmetry, we only need the 190 $0/1$ elements in the upper triangular part.
	
	\item Flattening: Concatenate these 190 elements row-wise into a one-dimensional binary sequence, e.g., 010010110...
	
	\item Tokenization: Directly using a length-190 sequence of 0s and 1s is inefficient because the vocabulary has only two symbols, and overly long sequences increase the model's learning difficulty. PatternBoost employs Byte Pair Encoding (BPE)—a common compression technique in natural language processing. BPE automatically analyzes all sequences in the training data, identifies frequently occurring substrings, and assigns a new, unique Token ID to each such substring. After BPE processing, the original long binary sequence is compressed into a shorter sequence composed of multiple tokens, where each token may represent 2, 3, or more original bits. This not only shortens the sequence length, improving training and generation efficiency, but these tokens themselves may correspond to meaningful local patterns.
\end{enumerate}
In practice, the size of the BPE vocabulary (i.e., the number of tokens) is a hyperparameter that needs tuning. Typically, a balance must be struck between model capacity (a larger vocabulary means shorter sequences but a larger token embedding layer) and the informativeness of the sequence representation. A common rule of thumb is that vocabulary size can be related to the size of the training dataset and the model's embedding dimension. For mathematical construction datasets, due to potentially stronger regularity in patterns, a relatively small vocabulary (e.g., 100-500) may sometimes be sufficiently efficient. Another practical issue is dynamism: when the dataset of good solutions is updated during iteration, should the BPE dictionary be retrained? In principle, for consistency, it's best to use a fixed tokenization scheme throughout the PatternBoost cycle. If new data introduces entirely new high-frequency patterns, retraining BPE might be more optimal, but this makes it harder to compare model outputs across different iterations. A compromise is to train BPE on a larger initial training set and keep it fixed in subsequent iterations.

The training process is standard language model training: given a token sequence (representing a good solution graph), train the Transformer model to predict the next token in the sequence. By minimizing the prediction error (cross-entropy loss) over all good solution sequences, the model gradually learns the probability distribution of ``what a good construction sequence should look like.''

After training, generating a new construction is like letting the model ``continue the story'': starting from an initial token, let the model predict the probability distribution for the next token based on the already generated tokens, then sample the next token from this distribution, and repeat until a complete sequence (ending with an end-of-sequence token) is generated. Finally, decode this token sequence back to the original binary representation (adjacency matrix), yielding a new graph ``conceived'' by the model.

The key point is: the new graph generated by the model has sequence patterns ``similar'' to the good solution graphs in the training set, but it is not a simple copy. It is a new sample from the learned probability distribution, thus potentially producing new constructions that are structurally similar but differ in specific details. This embodies the creativity of ``global learning''—it can generate new starting points that follow successful patterns yet are entirely novel.

Connecting the two phases above forms the main loop of PatternBoost. The experiments in the paper clearly demonstrate the effect of iteration:

\begin{itemize}
	\item Generation Zero (Initialization): Starting from empty graphs, run 40,000 simple local searches. The resulting distribution peaks around 66 edges, with the best result being 99 edges—appearing only twice. The known theoretical optimum is 100 edges, i.e., the complete bipartite graph $K_{10,10}$.
	
	\item Generation One: Take the top 25\% of good solutions to train a small Transformer. Let the Transformer generate 100,000 new sequences, of which about 37,000 can be correctly decoded into valid 20-vertex adjacency matrices. Use these 37,000 graphs as seeds, each undergoing local search. The result is surprising: 46 graphs reach the theoretical optimum, and 47 graphs have 99 edges. In contrast, the local search baseline found only two 99-edge graphs in 40,000 attempts. Moreover, these 46 optimal graphs are structurally isomorphic—all are the complete bipartite graph $K_{10,10}$ or its isomorphic variants.
	
	\item Subsequent Iterations: Add the newly found good solutions to the training set, continue fine-tuning the Transformer, and repeat generation and search. After several rounds of iteration, the model almost exclusively generates complete bipartite graphs and quickly learns to generate the optimal structure with two equal-sized parts.
\end{itemize}
This example perfectly illustrates the power of PatternBoost: a very simple local search, paired with a lightweight Transformer, through a few iterations, automatically ``discovers'' the extremal structure of the problem (the complete bipartite graph) starting from complete ignorance (random search from empty graphs). The Transformer successfully abstracts the global pattern of ``bipartiteness'' from a collection of successful cases and uses it to dramatically improve the starting point quality for local search.

The PatternBoost paper does not only showcase successful cases; it carefully selects a series of problems, forming a spectrum from ``method effective but not surpassing human'' to ``method makes breakthrough discoveries.'' This helps us objectively assess its capabilities and scope of applicability.

\noindent 1. Difficult Problem: Maximum Number of Edges in Graphs Without 4-Cycles

This problem requires constructing a graph on $n$ vertices that contains no cycles of length 4 and has as many edges as possible. This problem is mathematically much more difficult than the triangle-free graph problem. Using the same simple local search, after running 50 million local searches, the lowest score obtained was 68, the highest was 89, the distribution peak was at 81, while for \(n = 33\), the maximum possible score is 96. After applying PatternBoost, performance improved significantly; the best result was a graph with 91 edges. By using a larger Transformer model and an improved tokenization strategy (adding separators after each matrix row), the method eventually found the theoretically optimal 96-edge graph after 116.5 million local searches. However, the search volume required to reach the optimum was enormous, and for larger $n$, even with PatternBoost, the found solutions still lag behind known lower bounds. This indicates that for some problems with extremely complex, hard-to-capture structures, even the enhanced PatternBoost faces challenges, though its performance is still far superior to pure local search.

\noindent 2. Well-Performing Problem: Maximizing the Permanent of Matrices Avoiding the ``312'' Pattern

In this problem, the goal is to find 0-1 matrices that avoid a specific ``312'' pattern and have the maximum permanent. PatternBoost used a setup similar to before and was compared against a carefully designed, problem-specific search algorithm crafted by human experts. The result was that the specialized human algorithm won only by a narrow margin (found matrix permanent ~5.2e6, PatternBoost ~5.1e6). This is a highly instructive result: a general, almost domain-knowledge-free ML method can perform close to a specialized algorithm painstakingly designed by humans. More importantly, the optimal solution sets found by the two methods were almost disjoint, meaning they explored different regions of the search space. When the good solutions found by the human algorithm were added to PatternBoost's training set, PatternBoost was able to discover a series of new high-quality constructions. This demonstrates the potential of human-machine collaboration: human intuition can provide high-value seeds, and machines can conduct large-scale, tireless exploration and generalization based on them.

\noindent 3. Breakthrough Problem: Minimum Number of Edges in Spanning Subgraphs of Hypercubes

This is a 30-year-old problem: concerning how many edges must be retained in a $d$-dimensional hypercube so that its spanning subgraph still has diameter $d$. A conjecture estimates that such a subgraph must have at least \(2^{d} + \binom{d}{d / 2} - 2\) edges. For $d=5$, the conjectured construction appears optimal. But when the authors applied PatternBoost to $d=6$, the method successfully found a construction with only 81 edges, better than the conjectured construction with \(2^{6} + \binom{6}{3} - 2 = 82\) edges, thereby disproving the conjecture. This is one of PatternBoost's most striking achievements: it not only optimized known bounds but genuinely solved an open mathematical problem, making a new discovery. The process of discovering this counterexample fully demonstrates the closed loop from automated construction to mathematical discovery: the machine proposes candidate counterexamples, and humans verify and theorize.

Furthermore, the paper demonstrates PatternBoost's application on multiple other problems, including ``point sets in grids with no isosceles triangles,'' ``point sets in space with no 5 points on a sphere,'' and ``saturated Sperner families,'' some of which improved known best constructions. These cases collectively indicate that PatternBoost is a general framework applicable to various combinatorial extremal problems, and its effectiveness largely depends on whether the problem itself possesses ``implicit patterns'' that can be captured by machine learning models.

Although PatternBoost has achieved impressive results, it is not a universal key. Its successful application relies on a set of conditions and faces numerous challenges:
\begin{enumerate}
	\item Problem Suitability: The method is best suited for problems where ``optimal constructions have clear but complex patterns.'' If the optimal construction is highly disordered or random (e.g., constructions for lower bounds of certain Ramsey numbers), or if the scoring function is extremely rugged and lacks gradients, PatternBoost may struggle to learn effective patterns. The ``no 4-cycle'' problem in the paper is much harder than the ``no triangle'' problem, partly because the structure of the extremal graph is more complex and harder to encode and learn.
	
	\item Quality of Local Search: The overall performance of PatternBoost heavily depends on the effectiveness of the local search component. The simple greedy search used in the paper proves the concept, but in practical applications, designing more intelligent local search for specific problems (e.g., leveraging problem symmetry, designing finer neighborhood operations) can greatly enhance loop efficiency. A poor local search may fail to fully optimize the good starting points generated by the Transformer, thereby hindering the entire process.
	
	\item The Art of Representation and Tokenization: How to effectively represent a mathematical object as a sequence is crucial for the Transformer's learning. Different flattening orders (row-wise, column-wise, diagonal), whether to add separators, the size of the BPE vocabulary, etc., all significantly affect the proportion of valid sequences generated by the model and the learning speed. Currently, choices in this area rely more on experience and experimentation, lacking systematic theoretical guidance. This constitutes an important direction for engineering and theoretical research.
	
	\item Interpretability and Mathematical Insight: PatternBoost is a powerful ``discovery engine,'' but it is essentially an optimization tool. Does the counterexample or extremal construction it finds reveal deeper mathematical principles? Can the ``patterns'' learned by the model be translated into human-understandable mathematical concepts or conjectures? For example, in the case of disproving the hypercube conjecture, can the specific graph structure with 81 edges inspire mathematicians to propose new theorems about extremal graph structures? At present, the transition from machine-generated constructions to their elevation into human mathematical knowledge still demands substantial involvement from mathematicians. Enabling models not only to generate constructions, but also to offer explanations or formulate hypotheses about them, remains an open and frontier challenge.
	
	\item Computational Cost and Stopping Criteria: Although the models used in PatternBoost are relatively lightweight, large-scale iteration still requires considerable GPU computational resources, especially when dealing with large-scale problems (large $n$). The local search phase may involve millions or even hundreds of millions of evaluations of candidate solutions, which itself can be computationally intensive. In practice, clear stopping criteria need to be set to avoid infinite computation. Common stopping conditions include: model performance on a validation set (e.g., the proportion of valid samples generated) no longer improves; learning curves (e.g., training loss) plateau; the quality of the best solution found does not improve significantly over several consecutive iterations; or a predetermined computation time or resource budget is reached. When these conditions are met, iteration should stop, and the current best construction should be analyzed.
\end{enumerate}

\section{AI Constructing Counterexamples: Based on LLM Agents}

In the previous sections, we explored two methods for constructing counterexamples: one based on reinforcement learning, which systematically searches the vast combinatorial space through a policy network; and another based on the iterative fusion of local search and global learning, enabling the machine to learn the implicit patterns of ``good constructions.'' Both methods have their merits, but they share a common prerequisite: a well-defined formal environment—whether it's an explicit action space, a reward function, or an executable scoring criterion.

This subsection systematically explains how to build an LLM agent system capable of autonomously constructing counterexamples to mathematical conjectures, incorporating verification mechanisms to ensure the correctness of the constructions.
\\
A complete counterexample construction agent system can be abstracted into the following three-layer architecture:

\begin{itemize}
	\item Core Engine Layer: Includes the generative model used to produce counterexamples. This layer provides the agent's fundamental capabilities: understanding mathematical language, performing reasoning, and generating constructions.
	
	\item Agent Layer: Implements the core logic for intelligent decision-making, including how to decompose complex problems, how to learn from mistakes, how to retrieve relevant knowledge, etc. This layer determines whether the agent can, like a human researcher, strategically adjust its approach when facing difficulties, rather than blindly trying.
	
	\item Application Layer: The goal-oriented layer for specific tasks, responsible for translating the high-level goal of ``refuting this conjecture'' into an executable action plan and ultimately outputting a verified counterexample.
\end{itemize}
In the above architecture, a key design choice is to separate the generation and verification of counterexamples: a large language model is responsible for generating candidate counterexamples, which are then submitted to an independent verification module for correctness judgment. The core advantages of this design are:

\begin{itemize}
	\item No need for a complete formal environment: It is not necessary to fully formalize the entire conjecture and its background theory in advance. The agent can directly handle problems described in natural language, relying on formal tools or human checks only in the final verification stage.
	
	\item Relatively simple verification target: It only needs to judge whether the construction satisfies the conditions and conclusion of the proposition, without verifying complex intermediate reasoning steps. This ``final result check'' is simpler than verifying a complete proof chain.
	
	\item Fast iteration speed: After generation, it is immediately judged by the verification module, forming a ``generate-verify-correct'' closed loop, allowing a large number of attempts in a short time.
\end{itemize}
In this architecture, the design of the verification module is crucial. It can be a formal verification tool (such as Lean, Coq), a symbolic computation system (such as Mathematica, SageMath), or involve human expert intervention. Regardless of the form, the verification module needs to output a clear judgment result and provide specific feedback information when an error is determined—for example, ``this construction does not satisfy condition A'' or ``it actually does not negate conclusion B.'' This interpretable output not only allows the generation module to make targeted corrections but also provides a basis for subsequent human review.

\noindent \textcolor{structure3}{\textbf{Core Mechanism One: Iterative Verification and Repair}}

Iterative verification and repair is the core mechanism for interaction between the agent and the LLM verifier. Its basic workflow is as follows:

\begin{enumerate}
	\item Initial Generation: The agent generates a candidate counterexample construction and a brief explanation based on the given mathematical conjecture.
	
	\item Verification: Submit the candidate counterexample to the verifier, requesting it to judge its correctness. The verifier outputs the judgment result along with detailed reasoning.
	
	\item Error Feedback: If the verifier judges it as incorrect, it outputs the specific reason for the error—for example, ``this construction does not satisfy condition A'' or ``it actually does not negate conclusion B.''
	
	\item Reflection and Correction: The generation module receives the error feedback, combines it with the problem context and previous attempt history, and generates a corrected candidate counterexample. This process is not a simple ``try again,'' but a targeted improvement based on feedback.
	
	\item Loop Iteration: Repeat steps 2-4 until verification passes or the maximum number of attempts is reached.
\end{enumerate}
The technical key point here is: the quality of error feedback is crucial. If the verifier only outputs ``incorrect,'' the generation module does not know how to correct it. Therefore, prompts need to be designed to make the verifier output detailed, actionable feedback. Meanwhile, the correction process needs to retain historical context to avoid repeating the same mistakes—the agent should ``remember'' which approaches have been tried and failed.

\noindent \textcolor{structure3}{\textbf{Core Mechanism Two: Reflection and Decomposition}}

For complex counterexample construction tasks, directly generating a complete counterexample often exceeds the LLM's capability. The reflective decomposition mechanism gradually breaks down complex problems into manageable subproblems:

\begin{enumerate}
	\item Planning: The agent first understands the original problem and generates a general construction idea. This step is similar to a human mathematician's ``thinking about the general direction'' when facing a difficult problem.
	
	\item Subgoal Decomposition: Decompose the construction process into multiple subtasks. For example, ``first construct a function satisfying condition A, then adjust it to satisfy condition B, and finally verify that it breaks conclusion C.'' Each subgoal is relatively simple and easier to achieve.
	
	\item Recursive Solving: For each subgoal, the iterative verification and repair mechanism can be applied again—or the subgoal can be further decomposed. This recursive decomposition enables the agent to handle problems whose complexity far exceeds its single-generation capability.
	
	\item Result Synthesis: After all subgoals are verified, the system combines them into a complete counterexample description and performs a final overall verification.
\end{enumerate}
The quality of subgoal decomposition directly affects the final success rate. Decomposition that is too coarse leaves subgoals still difficult to solve; decomposition that is too fine may introduce excessive combinatorial complexity. The dependencies between subgoals also need to be properly managed—some subgoals may need to be completed sequentially, while others can be explored in parallel.

\vspace{3mm}

\noindent \textcolor{structure3}{\textbf{Case Study – Successful Counterexample Construction Practice}}

To understand how the above mechanisms operate in practice, let's examine a successful case: the handling of the Anderson conjecture in commutative ring theory by the AI4Math team at Peking University.

The Anderson conjecture was proposed by American mathematician David F. Anderson in 2014, concerning a profound property of ``quasi-complete local rings'' in commutative algebra, a rather deep problem in commutative ring theory. For over a decade after its proposal, no substantial breakthrough was made. The AI4Math team, formed by Professor Bin Dong's group at the Beijing International Center for Mathematical Research, Peking University, and collaborators, decided to use this difficult problem as a touchstone to test their agent's capabilities. They independently built a dual-agent collaboration framework—consisting of the natural language reasoning agent Rethlas and the formal verification agent Archon. The two agents have clear division of labor: Rethlas is responsible for literature retrieval and mathematical reasoning, while Archon is responsible for translating the reasoning results into rigorous formal proofs.

The mathematical reasoning agent Rethlas did not attempt to directly prove the conjecture. Instead, through cross-domain retrieval, it connected the theory of ``integral domain completion'' with the Anderson conjecture and constructed a counterexample, thereby refuting the conjecture. This means the Anderson conjecture was negated rather than proven—in mathematical research, negating a conjecture is as valuable as proving one, as it delineates the boundaries of theory.

Subsequently, the formal verification agent Archon transformed this counterexample construction into approximately 19,000 lines of Lean formalization code. During the formalization process, it autonomously discovered and corrected logical flaws in the initial plan. When a required mathematical concept was not yet included in Lean's current formalized mathematics library, Archon, through retrieval and comparison, autonomously found an equivalent alternative path. Finally, Archon completed the generation of all the code.

This case clearly demonstrates the core advantages of the ``generate-verify'' paradigm in counterexample construction tasks. In the generation phase, the agent dares to propose a negative construction rather than solely seeking a proof—Rethlas, through cross-domain association, linked the seemingly unrelated theory of ``integral domain completion'' with the Anderson conjecture. This kind of analogy and transfer belongs to the ``conceptual leap'' traditionally considered the core of human intelligence. In the verification phase, the formal verification step provided rigorous final assurance for this seemingly bold construction—Archon not only completed the generation of tens of thousands of lines of code but also autonomously discovered and repaired logical flaws during the process. The two agents collaborated, jointly solving a problem that human mathematicians had failed to resolve for over a decade.

Counterexample construction agents based on large language models represent a new paradigm for AI for Math. Compared to traditional methods requiring a complete formal environment, it has significant advantages:

\begin{itemize}
	\item Low Barrier to Entry: It does not require pre-formalizing the entire conjecture and its background theory; it can directly handle conjectures described in natural language. This enables AI exploration of a large number of mathematical problems that have not yet been formalized.
	
	\item High Flexibility: It can handle various types of mathematical objects—from sets of integers in number theory, to functions in analysis, to constructions in geometry. As long as it can be described in language, it can be attempted to be constructed in language.
	
	\item Interactivity: The agent's output is human-readable natural language, which mathematicians can understand, evaluate, and correct. This human-machine readability makes collaboration possible.
	
	\item Fast Iteration: The closed loop of generate-verify-correct allows a large number of attempts in a short time, exploring avenues that human researchers might overlook.
\end{itemize}

However, this method also has inherent limitations. The generation module (the large language model) may still make mistakes or get stuck in loops when dealing with problems requiring multi-step complex reasoning. The reflective decomposition mechanism can alleviate this issue but cannot eliminate it completely. The model's capabilities are limited by its training data; for extremely cutting-edge or niche mathematical fields, the model may lack the necessary background knowledge.

Constructing counterexamples to mathematical conjectures is a touchstone for testing the mathematical creativity of LLMs. By combining the generative capabilities of LLMs with external verification mechanisms, we can build an efficient paradigm for counterexample discovery. This paradigm does not require pre-formalizing the entire problem, has a lower barrier to entry, iterates faster, and is expected to benefit more mathematical researchers. The successful handling of the Anderson conjecture by the AI4Math team at Peking University is strong proof of the effectiveness of this paradigm: the agents not only constructed a counterexample that human mathematicians had failed to find for over a decade but also ensured its correctness through formal verification, with both steps completed autonomously by AI agents. Although the reliability of large language models themselves still needs improvement, by positioning them in the role of ``generating candidates'' rather than ``verifying conclusions,'' we can fully leverage their exploratory advantages while entrusting the responsibility of rigor to specialized verification modules. This division of labor and collaboration model makes counterexample construction tasks an ideal testing ground for the mathematical creativity of LLMs.

\nocite{*}

\printbibliography[heading=subbibliography,title=References]

\end{refsection}

\begin{refsection}[ref6.bib]
\chapter{AI for PDEs (Numerical Computation)}
\section{Overview of PDE Problems and Traditional Numerical Methods}

Partial differential equations (PDEs) are the core mathematical tools for describing the evolution of continuous media systems in fields such as physics, chemistry, biology, engineering, and even socio-economics. From heat conduction and fluid motion to quantum mechanics and financial option pricing, PDEs model the laws governing how variables change with space and time in the real world. Traditionally, solving these equations relies on well-developed numerical methods, such as finite difference methods, finite element methods, and spectral methods, which form the cornerstone of scientific and engineering computation. However, as the complexity of problems continues to increase—whether due to high-dimensional spaces, complex geometric boundaries, strong nonlinearities, or data-scarce inverse problems—the limitations of traditional methods are becoming increasingly apparent. This chapter will delve into the basic classification of PDE problems, the mathematical ideas, implementation details, theoretical limits, and computational challenges of traditional numerical methods. Using this as a starting point, it will reveal why artificial intelligence-based solution methods in recent years, particularly Physics-Informed Neural Networks (PINNs), are seen as a promising new paradigm for addressing these challenges. Through detailed mathematical examples, algorithmic pseudocode, complexity analysis, and visual comparisons, we will construct a comprehensive and in-depth framework for understanding.

To understand how AI can intervene, we must first understand the nature of the problem. A partial differential equation establishes a relationship between an unknown function (usually a function of space and time) and its partial derivatives with respect to the independent variables. It is typically formulated as:

\[
F\left( x_1, \dots, x_n, u, \frac{\partial u}{\partial x_1}, \dots, \frac{\partial u}{\partial x_n}, \frac{\partial^2 u}{\partial x_1^2}, \dots \right) = 0, \quad \mathbf{x} \in \Omega
\]

where \( u = u(\mathbf{x}) \) is the unknown function, \( \mathbf{x} = (x_1, \dots, x_n) \) are the independent variables (typically spatial and temporal coordinates) defined in the domain \( \Omega \subset \mathbb{R}^n \), and \( F \) is a given operator. This equation acts like a ``syntactic rule,'' specifying the local constraints that the function \( u \) must satisfy at every point in its domain. However, this rule alone is usually insufficient to uniquely determine a solution; we need additional ``contextual'' information, namely boundary conditions and (for time-dependent problems) initial conditions. A complete PDE problem, consisting of the governing equation, initial conditions, and boundary conditions together, is called a well-posed problem. Mathematically, one needs to study the existence, uniqueness, and stability (well-posedness) of its solution. In computational practice, our goal is to construct a numerical solution that approximates the true solution as closely as possible when an analytical solution is unavailable.
\\
From a mathematical structure perspective, PDEs can be classified based on the characteristics of their highest-order derivatives. This classification profoundly influences the behavior of the equation's solutions and the properties of the numerical methods used to solve them. For second-order linear PDEs, they can generally be categorized into the following three classic types:
\begin{enumerate}
	\item Elliptic Equations (Elliptic PDEs)
	Elliptic equations typically describe steady-state (time-independent) equilibrium or distribution problems, and their solutions possess ``smoothing'' and ``global dependence'' properties. A classic example is Poisson's equation, which describes the steady state of a potential field (such as electrostatic or gravitational potential) under a given source distribution:
	
	\[
	-\nabla^2 u = f(\mathbf{x}), \quad \mathbf{x} \in \Omega
	\]
	
	where \( \nabla^2 \) is the Laplacian operator and \( f \) is a known source term. In particular, when \( f=0 \), it is called Laplace's equation. Elliptic equations have no time dimension; their solution is determined simultaneously across the entire domain, and any disturbance on the boundary instantly affects all interior points (though the influence strength decays with distance). Typical boundary conditions are Dirichlet conditions (specifying the function value \( u \) on the boundary), Neumann conditions (specifying the normal derivative \( \partial u / \partial n \) on the boundary), or a mixture of both (Robin conditions). Elliptic problems mathematically correspond to finding the minimizer of an energy functional.
	
	\item Parabolic Equations (Parabolic PDEs)
	Parabolic equations introduce a time variable and describe dissipative processes such as diffusion and heat conduction. The most famous representative is the heat equation:
	
	\[
	\frac{\partial u}{\partial t} = \alpha \nabla^2 u + f(\mathbf{x}, t), \quad \mathbf{x} \in \Omega, \, t > 0
	\]
	
	where \( \alpha > 0 \) is the diffusion coefficient. This type of equation evolves unidirectionally in time (irreversible) and has the property of ``smoothing'' initial disturbances. Solving them requires initial conditions \( u(\mathbf{x}, 0) = u_0(\mathbf{x}) \) and boundary conditions. Information propagates inward from the initial time and the boundaries, making time-stepping a natural solution strategy.
	
	\item Hyperbolic Equations (Hyperbolic PDEs)
	Hyperbolic equations also describe time evolution processes, but their characteristics are wave-like and non-dissipative, such as the propagation of sound waves, light waves, and elastic waves. The standard form is the wave equation:
	
	\[
	\frac{\partial^2 u}{\partial t^2} = c^2 \nabla^2 u + f(\mathbf{x}, t), \quad \mathbf{x} \in \Omega, \, t > 0
	\]
	
	where \( c \) is the wave speed. These equations possess characteristic lines; disturbances propagate along these lines at finite speeds and may maintain discontinuities (such as shocks). They require initial conditions (including initial displacement and initial velocity) and boundary conditions.
\end{enumerate}
In addition to the three classic types mentioned above, the real world contains a large number of nonlinear PDEs (such as the Navier-Stokes equations describing fluids, the Allen-Cahn equation describing phase transitions) and systems of equations (coupling multiple unknown functions). These complexities pose significant challenges for both analytical and numerical solutions.

The core idea of traditional numerical methods is to discretize the continuous PDE problem, transforming it into a discrete, finite-dimensional algebraic problem (usually a system of linear or nonlinear equations), which is then solved using a computer. This process can be viewed as finding an approximation to the original PDE solution within a finite-dimensional space spanned by a grid or a set of basis functions. Different discretization strategies give rise to different families of numerical methods, each with its own mathematical principles, advantages, and applicable scenarios. Table \ref{tab:四大主流数值方法对比} summarizes the core concepts and typical application scenarios of these four mainstream methods:
\newpage
\begin{table}[htbp]
	\centering
	\caption{Comparison of Four Mainstream Numerical Methods}
	\label{tab:四大主流数值方法对比}
	\begin{tabular}{p{2.8cm} p{3.2cm} p{3.2cm} p{3.2cm} p{3.2cm}}
		\toprule
		\textcolor{structure3}{\textbf{Method Name}} & \textcolor{structure3}{\textbf{Core Discretization Idea}} & \textcolor{structure3}{\textbf{Typical Application Fields}} & \textcolor{structure3}{\textbf{Key Advantages}} & \textcolor{structure3}{\textbf{Main Challenges}} \\
		\midrule
		Finite Difference Method (FDM) & Approximate derivatives using difference quotients of function values at grid points & Fluids, heat conduction on regular domains & Simple concept, easy implementation, mature theory & Poor adaptation to complex geometry, curse of dimensionality \\
		\addlinespace
		
		Finite Element Method (FEM) & Partition domain into elements, approximate solution using local polynomial basis functions, based on variational principles & Structural mechanics, fluids with complex geometry, electromagnetics & Strong geometric adaptability, rigorous theory, widest application & Complex mesh generation, high computational cost in high dimensions \\
		\addlinespace
		
		Finite Volume Method (FVM) & Divide control volumes, integrate conservation laws, maintain physical flux balance & Computational fluid dynamics, combustion, multiphase flow & Strictly maintains physical conservation, strong robustness & Complex high-order schemes, difficult theoretical analysis \\
		\addlinespace
		
		Spectral Method & Expand solution using global smooth basis functions (e.g., Fourier series, Chebyshev polynomials) & Turbulence simulation, quantum chemistry, weather forecasting & Exponential convergence accuracy for smooth solutions & Requires smooth solution, poor geometric adaptability, dense matrices \\
		\bottomrule
	\end{tabular}
\end{table}

Next, we will delve into the mathematical details, implementation examples, and analyze the computational complexity and error characteristics of each method.
\vspace{3mm}
\\
\textcolor{structure3}{\textbf{1. Finite Difference Method (FDM)}}

Imagine laying a regular grid with nodes over the solution domain, like a sheet of graph paper. The PDE tells us the relationship between the rate of change (derivative) of the function at each point and the function value at that point. The core of FDM is to use the difference (difference quotient) of function values at adjacent nodes to approximate the derivative at that point. This is the most direct numerical implementation of the fundamental definition of calculus: ``the derivative is the limit of the difference quotient.''

Mathematical formulation and example: Consider a simple but classic two-dimensional Poisson equation with Dirichlet boundary conditions on a unit square domain:

\[
-\left( \frac{\partial^2 u}{\partial x^2} + \frac{\partial^2 u}{\partial y^2} \right) = f(x, y), \quad (x, y) \in (0,1)\times(0,1)
\]

\[
u(x, y) = g(x, y), \quad (x, y) \in \partial\Omega
\]

We discretize uniformly in the \( x \) and \( y \) directions with step size \( \Delta x = \Delta y = h = 1/(N+1) \), obtaining grid points \( (x_i, y_j) = (ih, jh) \), where \( i, j = 0, 1, ..., N+1 \). The function values at boundary points \( (i=0, N+1 \text{ or } j=0, N+1) \) are given by the boundary condition \( g \). For interior points \( (i, j) \), we approximate the second derivatives using central differences:

\[
\frac{\partial^2 u}{\partial x^2} \bigg|_{(x_i, y_j)} \approx \frac{u_{i+1,j} - 2u_{i,j} + u_{i-1,j}}{h^2}, \quad
\frac{\partial^2 u}{\partial y^2} \bigg|_{(x_i, y_j)} \approx \frac{u_{i,j+1} - 2u_{i,j} + u_{i,j-1}}{h^2}
\]

where \( u_{i,j} \approx u(x_i, y_j) \). Substituting these approximations into the Poisson equation yields an equation for each interior point:

\[
-\frac{u_{i+1,j} + u_{i-1,j} + u_{i,j+1} + u_{i,j-1} - 4u_{i,j}}{h^2} = f_{i,j}
\]

Rearranging:

\[
4u_{i,j} - u_{i+1,j} - u_{i-1,j} - u_{i,j+1} - u_{i,j-1} = h^2 f_{i,j}
\]

This forms a large system of linear equations for all interior unknowns \( u_{i,j} \). This system has a sparse, banded coefficient matrix (each equation involves only 5 unknowns) and can be solved using efficient iterative methods (such as the conjugate gradient method, multigrid method).

\begin{algorithm}[htbp]
	\caption{Finite Difference Method for Solving the 2D Poisson Equation}
	\label{alg:poisson_fdm}
	
	\KwIn{
		Number of grid divisions $N$ (number of interior points in each direction), \;
		Source term function $f(x,y)$,\;
		Boundary condition function $g(x,y)$,\;
		Solver parameters (iteration count, convergence tolerance, etc., optional)
	}
	
	\KwOut{
		Numerical solution matrix $U \in \mathbb{R}^{(N+2)\times(N+2)}$ (including boundary points)
	}
	
	\BlankLine
	Initialize step size $h = 1/(N+1)$, create solution matrix $U$ of dimension $(N+2) \times (N+2)$\;
	
	\BlankLine
	\textbf{Apply boundary conditions}: \;
	\For{all boundary points $(i,j)$ ($i=0$ or $i=N+1$ or $j=0$ or $j=N+1$)}{
		$U[i][j] \leftarrow g(x_i, y_j)$\;
	}
	
	\BlankLine
	\textbf{Assemble linear system} $A \mathbf{u} = \mathbf{b}$: \;
	\For{each interior point $(i,j)$ ($i=1,\ldots,N$, $j=1,\ldots,N$)}{
		Construct equation: $4u_{i,j} - u_{i+1,j} - u_{i-1,j} - u_{i,j+1} - u_{i,j-1} = -h^2 f_{i,j}$\;
		\If{neighboring point is on boundary}{
			Move its value (known quantity) to the right-hand side $\mathbf{b}$
		}
	}
	\textit{Note: The coefficient matrix $A$ is a sparse, symmetric positive definite block tridiagonal matrix (under natural ordering), with at most about $5$ non-zero entries per row, dimension $N^2 \times N^2$}\;
	
	\BlankLine
	\textbf{Solve sparse linear system}: \;
	Choose a solver (e.g., conjugate gradient method, multigrid method)\;
	Solve $A \mathbf{u}_{\text{vec}} = \mathbf{b}$, obtaining interior solution vector $\mathbf{u}_{\text{vec}}$\;
	
	\BlankLine
	\textbf{Reconstruct solution matrix}: \;
	Map solution vector $\mathbf{u}_{\text{vec}}$ back to interior point positions in $U$\;
	
	\BlankLine
	\Return numerical solution matrix $U$\;
\end{algorithm}

For a problem in \( d \)-dimensional space, if discretized into \( N \) interior points in each dimension, the total number of degrees of freedom (number of unknowns) is:

$$
N_{\text{dof}} = N^d
$$

This relationship reveals the fundamental challenge of the finite difference method: the number of unknowns grows exponentially as the dimension \( d \) increases. For example, when \( N=100 \), a two-dimensional problem has \( 10^4 \) unknowns, a three-dimensional problem increases to \( 10^6 \), and a four-dimensional problem reaches \( 10^8 \). This ``Curse of Dimensionality'' imposes enormous computational and storage pressure on traditional numerical methods when dealing with high-dimensional PDE problems.

The computational cost of the finite difference method mainly comes from two stages: system assembly and system solution.
\\
The system assembly stage requires generating the corresponding discrete equation for each interior node. Since the total number of interior nodes is \(N^d\), and the discrete stencil for each node (e.g., five-point difference) involves only a constant number of neighboring nodes, the time complexity of the assembly stage is

\[
T_{\text{assemble}} = O(N^d)
\]
The system solution stage is the main bottleneck of the entire computational process. For the resulting sparse linear system \(A\mathbf{u} = \mathbf{b}\), different solution strategies exhibit markedly different complexity characteristics.

Direct methods like Gaussian elimination, while theoretically general-purpose, can lead to significant fill-in during the elimination process for sparse matrices, dramatically increasing the density of the coefficient matrix. For general two-dimensional problems, the computational complexity can be as high as \(O((N^d)^3) = O(N^{3d})\), which is usually unacceptable in practice.
\\
Classical iterative methods like Jacobi require only \(O(N^d)\) operations per iteration. The convergence rate of such methods is typically linear, meaning the error decays by a fixed ratio each iteration. For elliptic problems, the convergence rate is determined by the spectral radius of the coefficient matrix, and the required number of iterations is inversely proportional to the square of the mesh size \(h\), i.e., \(O(h^{-2}) = O(N^2)\). Therefore, the total complexity of classical iterative methods is

\[
T_{\text{classical}} = O(N^2 \cdot N^d) = O(N^{d+2})
\]

Modern iterative methods like the conjugate gradient method can significantly improve convergence efficiency. For the symmetric positive definite system resulting from discretizing the Poisson equation, its condition number satisfies \(\kappa(A) = O(h^{-2}) = O(N^2)\). The number of iterations for the conjugate gradient method is approximately \(O(\sqrt{\kappa(A)}) = O(N)\), resulting in a total complexity of

\[
T_{\text{CG}} = O(N \cdot N^d) = O(N^{d+1})
\]

Optimal methods like the multigrid method, by alternating error correction on grids of different scales, can achieve solution efficiency proportional to the number of unknowns. For sufficiently smooth problems, the computational complexity of the multigrid method can reach

\[
T_{\text{solve}} = O(N^d)
\]
This is the theoretically optimal complexity achievable by the finite difference method, keeping the overall computational complexity at the \(O(N^d)\) level.

Although the linear system \(A\) is sparse—each equation involves only a constant number of unknowns—storage requirements still need careful handling.
\\
If using sparse storage formats like compressed row storage or compressed column storage, only the non-zero elements and their corresponding row and column indices need to be recorded. For the five-point stencil, each interior point corresponds to 5 non-zero elements, so the total storage is \(O(N^d)\).
\\
If using a dense storage strategy, i.e., retaining all elements of the matrix (including many zeros), the storage would be as high as \(O(N^{2d})\). This approach is completely infeasible for most practical problems, so practical applications must adopt sparse storage strategies to control space complexity at \(O(N^d)\).

The error of the finite difference method originates from approximating derivatives with difference quotients. Taking central differences as an example, Taylor expansion yields

\[
\frac{u(x+h) - u(x-h)}{2h} = u'(x) + \frac{h^2}{6}u'''(x) + O(h^4)
\]
For the second derivative, its central difference approximation is

\[
\frac{u(x+h) - 2u(x) + u(x-h)}{h^2} = u''(x) + \frac{h^2}{12}u^{(4)}(x) + O(h^4)
\]
Thus, the local approximation error (i.e., truncation error) of the central difference scheme is of order \(O(h^2)\). This means that as the grid is refined, the local error at each discrete point tends to zero at a rate of \(h^2\).

The discretization error refers to the overall difference between the numerical solution and the true solution. When the true solution is sufficiently smooth, the discretization error has the same order as the truncation error. Specifically, there exists a positive constant \(C\) independent of \(h\) such that

\[
\|u - u_h\| \leq C h^2
\]
where \(\|\cdot\|\) can be the \(L^2\) norm or the \(L^\infty\) norm. This relationship indicates that when the grid step size is halved, the error is reduced to about one-quarter of its original value. Therefore, the above central difference scheme is said to have second-order convergence accuracy.

More generally, if a difference scheme of order \(p\) is used, its discretization error satisfies

\[
\|u - u_h\| = O(h^p)
\]
where \(p\) is determined by the construction of the difference scheme. It should be noted that the actual convergence order is limited by the smoothness of the true solution: if the true solution lacks sufficient high-order derivatives, the actual convergence rate may be lower than the theoretical value.

The core advantage of the finite difference method lies in its intuitive concept and simple implementation. The method directly constructs discrete schemes based on the definition of derivatives, making the physical meaning clear and easy for beginners to understand and master. For problems on regular domains (such as rectangles, cubes), the finite difference method can efficiently generate sparse linear systems, and its theoretical framework—including stability analysis, convergence proofs, and error estimation—is well-developed, providing solid theoretical guarantees for the reliability of numerical solutions.

However, the finite difference method also faces significant limitations. First, the method has poor adaptability to the geometric shape of the computational domain. When dealing with complex or irregular boundaries, constructing high-accuracy difference schemes that satisfy boundary conditions is often very difficult, usually requiring complex techniques like coordinate transformations or immersed boundary methods. Second, as the spatial dimension increases, the degrees of freedom grow exponentially, leading to a sharp increase in computational cost and storage requirements. This ``curse of dimensionality'' makes the traditional finite difference method difficult to apply directly to high-dimensional problems.
\vspace{3mm}
\\
\textcolor{structure3}{\textbf{2. Finite Element Method (FEM)}}

If the finite difference method (FDM) imposes rules ``point by point,'' then the finite element method (FEM) constructs the solution ``piece by piece.'' Its core idea is: partition the complex solution domain into many non-overlapping small pieces, such as triangles or quadrilaterals; these small pieces are called ``elements.'' On each element, we use a simple function (e.g., a low-degree polynomial) to approximate the true solution, much like using many small polygonal tiles to mosaic a floor of complex shape, drawing simple patterns on each tile. By cleverly combining these ``local approximations,'' we can obtain a global approximate solution over the entire domain.

Unlike the finite difference method, which directly discretizes the partial differential equation, the finite element method deals with the ``weak form'' (integral form) of the original equation. The weak form, through ``averaging'' over the integration domain, reduces the smoothness requirements on the solution—allowing the solution to have discontinuous derivatives in some places, which provides convenience for handling complex problems. Meanwhile, certain boundary conditions (like natural boundary conditions) are automatically incorporated into the weak form without needing extra treatment. The core of the finite element method is the Galerkin method: we seek an approximate solution within a finite-dimensional function space spanned by piecewise polynomial basis functions, and require the residual of the partial differential equation to be orthogonal to all ``test functions'' in that space, thereby ensuring the approximate solution is as close as possible to the true solution in an overall sense.

\begin{figure}[htbp]
	\caption{Schematic of the 2D finite element method: Left figure shows triangular element mesh, node $i$ connected to several elements; Right figure illustrates the corresponding linear basis function $\phi_i(x,y)$ as a ``tent function,'' taking value $1$ at node $i$, $0$ on the support boundary, and varying linearly on each adjacent triangular element.}
	\label{fig:fem_2d_linear_basis}
	\centering
	\begin{tikzpicture}[
		>=Stealth,
		thick,
		every node/.style={font=\small}
		]
		
		\begin{scope}[xshift=0cm]
			
			\coordinate (A) at (0,0);
			\coordinate (B) at (2,0);
			\coordinate (C) at (4,0);
			\coordinate (D) at (1,1.7);
			\coordinate (E) at (3,1.7);
			\coordinate (I) at (2,0.85);   
			
			\fill[gray!8] (A)--(B)--(I)--cycle;
			\fill[gray!8] (B)--(C)--(I)--cycle;
			\fill[gray!8] (A)--(D)--(I)--cycle;
			\fill[gray!8] (D)--(E)--(I)--cycle;
			\fill[gray!8] (E)--(C)--(I)--cycle;
			
			\draw[black] (A)--(B)--(C);
			\draw[black] (A)--(D)--(E)--(C);
			\draw[black] (A)--(I)--(C);
			\draw[black] (D)--(I)--(E);
			\draw[black] (B)--(I);
			
			\node[circle, fill=black, inner sep=1.8pt, label={[red!80]above right:$i$}] at (I) {};
			\node[circle, fill=black, inner sep=1.5pt] at (A) {};
			\node[circle, fill=black, inner sep=1.5pt] at (B) {};
			\node[circle, fill=black, inner sep=1.5pt] at (C) {};
			\node[circle, fill=black, inner sep=1.5pt] at (D) {};
			\node[circle, fill=black, inner sep=1.5pt] at (E) {};
			
			\node at (0.9,0.45) {$K_1$};
			\node at (3.1,0.45) {$K_2$};
			
			\node[align=center] at (2,-0.8) {2D triangular mesh};
			
		\end{scope}
		
		\begin{scope}[xshift=7.2cm]
			
			\coordinate (P1) at (0,0);
			\coordinate (P2) at (2,0.8);
			\coordinate (P3) at (4,0);
			\coordinate (P4) at (2,-0.8);
			
			\coordinate (Top) at (2,2.6);
			
			\fill[structure3!8] (P1)--(P2)--(P3)--(P4)--cycle;
			
			\fill[structure3!15] (Top)--(P1)--(P2)--cycle;
			\fill[structure3!20] (Top)--(P2)--(P3)--cycle;
			\fill[structure3!12] (Top)--(P3)--(P4)--cycle;
			\fill[structure3!18] (Top)--(P4)--(P1)--cycle;
			
			\draw[blue!70!black, dashed] (P1)--(P2)--(P3);
			\draw[blue!70!black] (P3)--(P4)--(P1);
			
			\draw[blue!70!black] (Top)--(P1);
			\draw[blue!70!black, dashed] (Top)--(P2);
			\draw[blue!70!black] (Top)--(P3);
			\draw[blue!70!black] (Top)--(P4);
			
			\node[circle, fill=red!80, inner sep=1.8pt, label={[red!80]below:$i$}] at (2,0) {};
			
			\node[above right] at (Top) {$\phi_i=1$};
			
			\node[below] at (2,-1.05) {$\phi_i=0$ (adjacent boundary)};
			
			\draw[->, gray!80] (4.6,-0.6) -- (4.6,2.8) node[right] {$\phi_i(x,y)$};
			
			\node[align=center] at (2,-1.8) {Linear basis function (tent function)};
			
		\end{scope}
		
	\end{tikzpicture}
\end{figure}

Next, using the two-dimensional Poisson equation as an example, we introduce the four main steps of the finite element method:

\[
-\left( \frac{\partial^2 u}{\partial x^2} + \frac{\partial^2 u}{\partial y^2} \right) = f(x, y), \quad (x, y) \in \Omega \subset \mathbb{R}^2
\]

\[
u(x, y) = g(x, y), \quad (x, y) \in \partial\Omega
\]

The finite element solution process can be divided into the following four main steps.
\\
Step 1: Domain Discretization

Partition the solution domain \(\Omega\) into many small triangular (or quadrilateral) elements, forming a triangular mesh. Unlike the finite difference method, finite element elements can flexibly conform to complex geometric boundaries. Denote the mesh as \(\mathcal{T}_h = \{K\}\), where \(h\) represents the characteristic size of the elements (usually taken as the diameter of the circumscribed circle of the largest element). The set of all element vertices constitutes the node set.
\\
Step 2: Construct Finite Element Space

On each element \(K\), we need a ``local blueprint'' to describe the shape of the solution on that element. This blueprint is the shape function. Taking linear elements on triangular elements as an example, define three shape functions \(N^K_1(x,y)\), \(N^K_2(x,y)\), \(N^K_3(x,y)\) on element \(K\), which satisfy:

\begin{itemize}
	\item On each element, the shape functions are linear functions of \(x\) and \(y\), simple in form, easy to integrate and differentiate
	\item Take value 1 at the corresponding vertex (node) and 0 at the other two vertices
\end{itemize}
This means that if we know the function values \(c_{i_1}, c_{i_2}, c_{i_3}\) at the three vertices of element \(K\), we can uniquely construct the approximate solution on the element via the shape functions:
\[
u_h(x,y)|_K = c_{i_1} N^K_1(x,y) + c_{i_2} N^K_2(x,y) + c_{i_3} N^K_3(x,y)
\]
Here, the shape functions act as ``interpolators,'' smoothly extending the discrete nodal values to the entire interior of the element.

Next, we need to piece these ``local blueprints'' together into a global approximate function. To do this, we define a global basis function \(\phi_i(x,y)\) for each global node \(i\). Its construction is very intuitive: ``stitch together'' the shape function corresponding to node \(i\) from all elements containing node \(i\), and set it to 0 on all other elements. The global basis function \(\phi_i\) has the following key properties:

\begin{itemize}
	\item Local support: \(\phi_i\) is non-zero only on the ring of elements surrounding node \(i\), and zero elsewhere. This ensures sparsity in subsequent computations.
	\item Nodal interpolation property: \(\phi_i\) takes value 1 at node \(i\) and 0 at all other nodes.
\end{itemize}
The approximate solution \(u_h(x,y)\) over the entire domain can be expressed as a linear combination of all nodal basis functions:

\[
u_h(x,y) = \sum_{i=1}^{M} c_i \phi_i(x,y)
\]
where \(M\) is the total number of nodes, and \(c_i\) are the unknown coefficients to be determined, whose physical meaning is the approximate solution value at the node. In this way, shape functions provide the ``local description'' on each element, while global basis functions are responsible for ``seamlessly stitching'' these local descriptions into a continuous function over the entire domain (for linear elements, the solution is automatically continuous across element boundaries).
\\
Step 3: Form Weak Form and Discrete System

To obtain the finite element equations, we first derive the weak form of the Poisson equation. Multiply the original equation by a ``test function'' \(v\) and integrate over the domain \(\Omega\). Using the divergence theorem (integration by parts in higher dimensions) to transform second derivatives into first derivatives:

\[
\int_\Omega \nabla u \cdot \nabla v \, d\Omega = \int_\Omega f v \, d\Omega + \int_{\partial \Omega} (\nabla u \cdot \mathbf{n}) v \, dS
\]
For Dirichlet boundary conditions (given function values on the boundary), we enforce the approximate solution to equal the known value \(g\) on the boundary, so the test function \(v\) is taken as \(0\) on the boundary, and the boundary integral term automatically vanishes. Thus, the problem transforms into: find \(u\) satisfying the boundary conditions such that the above integral equality holds for all admissible test functions.

Now, substitute the approximate solution \(u_h = \sum_{i=1}^{M} c_i \phi_i\) and take the test function \(v\) to be each basis function \(\phi_j\) (i.e., the Galerkin method). This yields \(M\) equations:

\[
\sum_{i=1}^{M} c_i \int_\Omega \nabla \phi_i \cdot \nabla \phi_j \, d\Omega = \int_\Omega f \phi_j \, d\Omega, \quad j = 1, 2, \dots, M
\]
Denote
\[
A_{ji} = \int_\Omega \nabla \phi_i \cdot \nabla \phi_j \, d\Omega, \quad b_j = \int_\Omega f \phi_j \, d\Omega
\]
Then the above system can be written in concise matrix form:

\[
A \mathbf{c} = \mathbf{b}
\]
where \(A\) is called the stiffness matrix, \(\mathbf{b}\) is the load vector, and \(\mathbf{c}\) is the vector of unknown coefficients.
\\
Step 4: Assembly and Solution

The computation of the stiffness matrix and load vector does not need to be performed directly over the global domain; instead, we utilize the local support property of the basis functions: each \(\phi_i\) is non-zero only on a few elements. Therefore, we can compute local contributions on each element separately, then ``assemble'' the contributions from all elements into the global matrix and vector. Specifically:

\begin{itemize}
	\item On each element \(K\), compute the element stiffness matrix and element load vector;
	\item According to the node numbering, accumulate the element contributions to the corresponding positions in the global matrix \(A\) and global vector \(\mathbf{b}\).
\end{itemize}
Since each basis function only overlaps with basis functions of adjacent nodes, the stiffness matrix \(A\) is sparse—the proportion of non-zero entries is low, and it is usually symmetric positive definite (for the Poisson equation). This sparsity allows us to solve the linear system using efficient iterative solvers (like the conjugate gradient method) or direct solvers (like sparse LU decomposition), completing the solution in reasonable time even when the number of nodes reaches millions.

After solving for the coefficient vector \(\mathbf{c}\), substitute it into \(u_h(x, y) = \sum c_i \phi_i(x, y)\) to obtain the finite element approximate solution of the original problem over the entire domain.

For a two-dimensional triangular mesh, the number of nodes \(M\) is roughly proportional to the number of elements. More generally, in \(d\)-dimensional space, to achieve discretization accuracy with characteristic size \(h\), the required degrees of freedom (number of nodes) is approximately \(O(h^{-d})\). This relationship is identical to that of the finite difference method, again revealing the essence of the curse of dimensionality—as the spatial dimension increases, degrees of freedom grow exponentially, directly leading to a sharp increase in computational cost and storage requirements.

The stiffness matrix \(A\) resulting from finite element discretization has three important properties: sparsity, symmetry, and positive definiteness (for elliptic problems). Sparsity arises because each basis function is non-zero only on a finite number of elements in its support, so each node couples only with adjacent nodes, and the vast majority of matrix entries are zero. When using sparse storage formats, the storage requirement is only \(O(M)\), i.e., linear in the number of degrees of freedom. Symmetry and positive definiteness allow efficient iterative algorithms like the conjugate gradient method to be applied.

The computational complexity of the finite element method consists of two stages: assembly and solution. The assembly stage requires integral calculations for each element; since the computation per element is constant, the total complexity of assembly is \(O(M)\). The complexity of the solution stage depends on the linear solver used. If using sparse direct methods (e.g., solvers based on multifrontal elimination), for two-dimensional problems, the complexity typically ranges between \(O(M^{1.5})\) and \(O(M^2)\); if using optimal iterative solvers like the multigrid method, the solution complexity can be reduced to \(O(M)\), achieving optimal efficiency linear in the number of degrees of freedom.

\begin{algorithm}[htbp]
	\caption{Finite Element Method for Solving 2D Poisson Equation (Triangular Linear Elements)}
	\label{alg:fem_poisson}
	
	\KwIn{
		Node coordinate array $\mathbf{nodes}$, dimension $M \times 2$,\;
		Element connectivity array $\mathbf{elements}$, each element contains $3$ node indices,\;
		Right-hand side function $f(x,y)$,\;
		Dirichlet boundary node list $\mathbf{dirichlet\_nodes}$ and corresponding boundary values
	}
	
	\KwOut{
		Nodal solution vector $\mathbf{u}$, where $u_i \approx u(\mathbf{x}_i)$
	}
	
	\BlankLine
	Initialize global stiffness matrix $\mathbf{A}$ (using sparse storage format) and load vector $\mathbf{b}$ to zero\;
	
	\BlankLine
	\For{each triangular element $K \in \mathbf{elements}$}{
		Get coordinates of the element's three nodes $\mathbf{x}_1, \mathbf{x}_2, \mathbf{x}_3$\;
		
		Compute element stiffness matrix $\mathbf{A}_K \in \mathbb{R}^{3 \times 3}$:
		\[
		(A_K)_{ij} = \int_K \nabla \phi_i \cdot \nabla \phi_j \, d\Omega
		\]
		
		Compute element load vector $\mathbf{b}_K \in \mathbb{R}^{3}$:
		\[
		(b_K)_i = \int_K f \phi_i \, d\Omega
		\]
		
		Assemble $\mathbf{A}_K$ into corresponding positions of global matrix $\mathbf{A}$ (according to element's local-to-global node mapping)\;
		
		Assemble $\mathbf{b}_K$ into corresponding positions of global vector $\mathbf{b}$\;
	}
	
	\BlankLine
	\textbf{Handle Dirichlet boundary conditions}: \;
	\For{each boundary node $i \in \mathbf{dirichlet\_nodes}$}{
		Enforce $u_i = g(\mathbf{x}_i)$\;
		
		Modify row $i$ and column $i$ of matrix $\mathbf{A}$ (typically set row $i$, column $i$ to $1$, adjust right-hand side accordingly)\;
	}
	
	\BlankLine
	Solve sparse linear system $\mathbf{A} \mathbf{u} = \mathbf{b}$ (using sparse direct method like UMFPACK, or iterative method like conjugate gradient)\;
	
	\BlankLine
	\Return nodal solution vector $\mathbf{u}$\;
\end{algorithm}

Regarding the convergence accuracy of the finite element method, for linear elements (i.e., using first-degree polynomial approximation), when the true solution is sufficiently smooth, the error satisfies the following estimates:

\[
\|u - u_h\|_{L^2} = O(h^2), \qquad \|u - u_h\|_{H^1} = O(h)
\]
where the \(L^2\) norm measures the error in the function values themselves, and the \(H^1\) energy norm measures the overall error in function values and their first derivatives. This means that when the mesh is refined by a factor of two, the function value error is reduced to about one-quarter, and the derivative error is reduced to about one-half.

More generally, if using elements of polynomial degree \(p\), the convergence order can be improved to

\[
\|u - u_h\|_{L^2} = O(h^{p+1})
\]

This property reveals the intrinsic relationship between accuracy and element order in the finite element method: increasing the polynomial degree can significantly accelerate convergence speed. However, this improvement in accuracy is not without cost—higher-order elements require more integration points and degrees of freedom per element, and the complexity of program implementation also increases accordingly. Therefore, in practical applications, a trade-off must be made between accuracy, efficiency, and implementation difficulty: linear elements (\(p=1\)) are widely popular due to their simplicity, quadratic elements (\(p=2\)) offer a good balance between accuracy and efficiency, while higher-order elements are typically used for special problems requiring extremely high accuracy.

The core advantage of the finite element method lies in its strong adaptability to complex geometric shapes. Through unstructured meshes (e.g., triangular, tetrahedral meshes), the finite element method can precisely conform to arbitrarily complex boundaries, giving it unparalleled flexibility when dealing with practical engineering problems. Furthermore, the finite element method has a rigorous mathematical theoretical foundation—based on Sobolev spaces and variational principles—and its stability analysis and error estimation framework are well-developed. For these reasons, the finite element method has become the most mainstream numerical method in engineering fields such as structural mechanics, fluid mechanics, and heat conduction.

However, the finite element method also faces non-negligible challenges. First, the curse of dimensionality in high-dimensional problems still exists—as the spatial dimension increases, the number of elements and nodes grows exponentially, causing computational costs to rise sharply. Second, generating high-quality computational meshes (especially three-dimensional unstructured meshes) is itself a highly challenging task, often requiring specialized pre-processing software and significant manual intervention, and mesh quality directly affects the accuracy and stability of the numerical solution. For dynamic problems involving moving boundaries, free surfaces, or large deformations, the mesh may need frequent regeneration, further increasing computational complexity and implementation difficulty.
\vspace{3mm}
\\
\textcolor{structure3}{\textbf{3. Finite Volume Method (FVM)}}

The idea of the finite volume method is rooted in the most fundamental conservation laws of the physical world. It divides the solution domain into many non-overlapping control volumes (usually the grid cells themselves), then directly applies the conservation law to each control volume: the rate of change of a physical quantity (such as mass, momentum, energy) equals the net flux entering through its boundaries plus internal source terms. This naive idea of ``from local conservation to global conservation'' makes FVM inherently conservative at the discrete level, which is its most distinctive feature compared to finite difference and finite element methods.

Unlike the finite difference method, which directly approximates derivatives, FVM discretizes the conservation law equation in integral form. Using the divergence theorem, it transforms volume integrals over a control volume into flux integrals over its boundary, thereby converting the continuous equation into a set of discrete balance equations. This approach makes FVM particularly suitable for handling flow problems with strong nonlinear phenomena like shocks and discontinuities—in these problems, derivatives of the solution may not exist, but the integral form of the conservation law still holds. Figure \ref{fig:fvm_1d} illustrates the division of control volumes and the physical picture of flux exchange for a one-dimensional case:
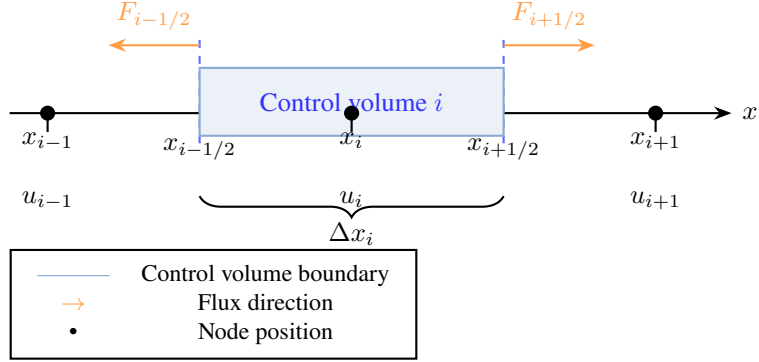
\begin{figure}[htbp]
	\caption{Schematic of 1D finite volume method: Each control volume is centered at node $x_i$, $u_i$ is the solution value at node $x_i$, $F_{i\pm1/2}$ are the fluxes at interfaces $x_{i\pm1/2}$, flux exchange occurs at interfaces $x_{i-1/2}$ and $x_{i+1/2}$}
	\label{fig:fvm_1d}
	\centering
	\begin{tikzpicture}[
		>=Stealth,
		thick,
		every node/.style={font=\small}
		]
		
		\def\xm{0}      
		\def\xmhalf{2}  
		\def\xi{4}      
		\def\xphalf{6}  
		\def\xp{8}      
		
		\draw[->, thick] (-0.5,0) -- (9,0) node[right] {$x$};
		
		\draw[dashed, blue!60, thick] (\xmhalf, -0.4) -- (\xmhalf, 1.0);
		\draw[dashed, blue!60, thick] (\xphalf, -0.4) -- (\xphalf, 1.0);
		
		\fill[structure3!10, draw=structure3!60, thick] (\xmhalf, -0.3) rectangle (\xphalf, 0.6);
		\node[blue!80, font=\small] at (4, 0.15) {Control volume $i$};
		
		\draw[thick] (\xm, 0.05) -- (\xm, -0.25);
		\node[circle, fill=black, inner sep=2pt, label={below:$x_{i-1}$}] at (\xm, 0) {};
		\draw[thick] (\xi, 0.05) -- (\xi, -0.25);
		\node[circle, fill=black, inner sep=2pt, label={below:$x_i$}] at (\xi, 0) {};
		\draw[thick] (\xp, 0.05) -- (\xp, -0.25);
		\node[circle, fill=black, inner sep=2pt, label={below:$x_{i+1}$}] at (\xp, 0) {};
		
		\node[below=0.9cm] at (\xm, 0) {$u_{i-1}$};
		\node[below=0.9cm] at (\xi, 0) {$u_i$};
		\node[below=0.9cm] at (\xp, 0) {$u_{i+1}$};
		
		\node[below=0.2cm] at (\xmhalf, 0) {$x_{i-1/2}$};
		\node[below=0.2cm] at (\xphalf, 0) {$x_{i+1/2}$};
		
		\draw[->, thick, second!80] (\xmhalf, 0.9) -- (\xmhalf-1.2, 0.9) 
		node[midway, above=2pt] {$F_{i-1/2}$};
		\draw[->, thick, second!80] (\xphalf, 0.9) -- (\xphalf+1.2, 0.9) 
		node[midway, above=2pt] {$F_{i+1/2}$};
		
		\draw[decorate, decoration={brace, amplitude=8pt, mirror}] 
		(\xmhalf, -1.1) -- (\xphalf, -1.1)
		node[midway, below=6pt] {$\Delta x_i$};
		
		\node[draw, fill=white, inner sep=4pt, anchor=north west, font=\footnotesize] at (-0.5, -1.8) {
			\begin{tabular}{cc}
				\textcolor{structure3!60}{———}  & Control volume boundary \\
				\textcolor{second!80}{$\rightarrow$} & Flux direction \\
				\textbullet & Node position
			\end{tabular}
		};
		
	\end{tikzpicture}
\end{figure}
\\
In one dimension, each control volume is centered at node \(x_i\), with left and right boundaries at \(x_{i-1/2}\) and \(x_{i+1/2}\) respectively. The conservation law requires:

\[
\text{Rate of change within control volume} = \underbrace{F_{i-1/2}}_{\text{entering left}} \;-\; \underbrace{F_{i+1/2}}_{\text{exiting right}} \;+\; \underbrace{\int_{\text{volume}} S \, dx}_{\text{internal source}}
\]
where \(F\) denotes the flux through the boundary (physical quantity transferred per unit time per unit area). This equation forms the physical basis for finite volume discretization.
\\
Consider the one-dimensional steady-state convection-diffusion equation:

\[
\frac{d}{dx}\left(a u - \nu \frac{du}{dx}\right) = f, \quad x \in (0, L)
\]
where \(a\) is the convection velocity (constant), \(\nu > 0\) is the diffusion coefficient, and \(f\) is the source term. This equation is already written in conservation form \(\frac{dF}{dx} = f\), where the total flux \(F\) consists of convection and diffusion terms:
\[
F = a u - \nu \frac{du}{dx}
\]
Integrate over control volume \(V_i = [x_{i-1/2}, x_{i+1/2}]\):

\[
\int_{x_{i-1/2}}^{x_{i+1/2}} \frac{dF}{dx} dx = F\big|_{x_{i+1/2}} - F\big|_{x_{i-1/2}} = \int_{x_{i-1/2}}^{x_{i+1/2}} f \, dx
\]
This integral form does not involve derivatives, allowing for discontinuous solutions, reflecting FVM's adaptability to weak solutions.

Next, discretize. Denote \(F_{i+1/2}\) as the flux approximation at interface \(x_{i+1/2}\), i.e., \( F_{i+1/2} \approx F(u_{i}, u_{i+1}) \), and approximate the right-hand side source term as \(f_i \Delta x_i\) (\(\Delta x_i\) is the control volume width). Then the discrete equation is:

\[
F_{i+1/2} - F_{i-1/2} = f_i \Delta x_i
\]
The core of the problem lies in how to construct the interface flux \(F_{i+1/2}\), i.e., approximate the physical flux at that interface based on neighboring node values \(u_i, u_{i+1}\). Different reconstruction methods are used for different physical mechanisms:

\begin{itemize}
	\item Diffusion term \(\displaystyle -\nu \frac{du}{dx}\): Usually central differencing is used, because diffusion is an isotropic physical process:
	\[
	-\nu \frac{du}{dx}\bigg|_{i+1/2} \approx -\nu \frac{u_{i+1} - u_i}{\Delta x}
	\]
	where \(\Delta x = x_{i+1} - x_i\) (assuming uniform grid).
	
	\item Convection term \(\displaystyle a u\): Convection has directionality; information propagates along the flow direction, so it is necessary to choose the ``upwind'' value based on the flow velocity direction to ensure numerical stability:
	\[
	a u|_{i+1/2} \approx 
	\begin{cases}
		a u_i, & a \ge 0 \quad (\text{flow from left to right}) \\[4pt]
		a u_{i+1}, & a < 0 \quad (\text{flow from right to left})
	\end{cases}
	\]
	This upwind scheme is physically reasonable: the convective flux at the interface is determined by the physical quantity upstream. The upwind scheme is unconditionally stable but only first-order accurate.
\end{itemize}

Combining the two parts above gives the discrete form of the total flux:

\[
F_{i+1/2} = a u_{\text{upwind}} - \nu \frac{u_{i+1} - u_i}{\Delta x}, \quad \text{where } u_{\text{upwind}} = \begin{cases} u_i, & \text{if } a \ge 0 \\ u_{i+1}, & \text{if } a < 0 \end{cases}
\]
Substituting into the discrete equation and rearranging yields a linear system for the node values \(u_i\):

\[
-\left( \frac{\nu}{\Delta x} + \max(0, -a) \right) u_{i-1}
+ \left( \frac{2\nu}{\Delta x} + |a| \right) u_i
- \left( \frac{\nu}{\Delta x} + \max(0, a) \right) u_{i+1}
= f_i \Delta x_i
\]
This is a tridiagonal system that can be solved directly using the efficient Thomas algorithm.

The greatest advantage of the finite volume method is its inherent conservation property. Since the method starts directly from the integral form of the conservation law, the discrete equation on each control volume precisely satisfies the balance of physical quantities; fluxes on internal interfaces cancel each other during global assembly, thus strictly maintaining overall conservation at the discrete level. This characteristic makes the finite volume method particularly suitable for handling multi-physics problems involving conservation of mass, momentum, and energy, such as fluid dynamics and heat transfer, ensuring physical reasonability of the solution even on coarse grids. Furthermore, the finite volume method has good adaptability to complex geometry, can handle complex computational domains in engineering using unstructured meshes, and coupled with its strong numerical stability (especially with upwind schemes), it has become the mainstream method in commercial software for computational fluid dynamics (CFD).

However, the finite volume method also has certain limitations. Constructing high-order accurate schemes is more complex than in the finite element method, requiring flux reconstruction at element interfaces and often introducing limiters to suppress unphysical oscillations, which imposes higher demands on scheme design and program implementation. At the same time, rigorous mathematical error analysis (such as a priori and a posteriori error estimates) is not as systematic and complete within the finite volume framework as it is for the finite element method, leaving its theoretical foundation somewhat less comprehensive. For high-dimensional problems on unstructured meshes, balancing accuracy, stability, and computational efficiency remains an ongoing research topic and challenge in finite volume method applications.

\newpage
\begin{algorithm}[H]
	\caption{Finite Volume Method for Solving 1D Steady Convection-Diffusion Equation (Upwind Scheme)}
	\label{alg:fvm_convection_diffusion}
	
	\KwIn{
		Grid node coordinate array $x[0..N]$, where $N$ is number of elements (number of nodes is $N+1$),\;
		Convection velocity $a$ (constant), diffusion coefficient $\nu > 0$,\;
		Source term function $f(x)$,\;
		Boundary values $u_0 = u(x_0)$, $u_N = u(x_N)$ (Dirichlet boundary conditions)
	}
	
	\KwOut{
		Nodal solution vector $\mathbf{u} = [u_0, u_1, \dots, u_N]^{\mathsf{T}}$, where $u_i \approx u(x_i)$
	}
	
	\BlankLine
	Compute lengths of each control volume $\Delta x_i = x_{i+1} - x_i$, $i = 0, 1, \dots, N-1$\;
	
	Initialize tridiagonal coefficient matrix $\mathbf{A} \in \mathbb{R}^{(N+1) \times (N+1)}$ and right-hand side vector $\mathbf{b} \in \mathbb{R}^{N+1}$ to zero\;
	
	\BlankLine
	\For{each interior control volume $i = 1, 2, \dots, N-1$}{
		Compute left interface flux coefficients ($i-1/2$ interface):
		\[
		a_{\text{conv}}^{\text{left}} = \max(a, 0) \quad \text{(upwind scheme)}
		\]
		\[
		d^{\text{left}} = \frac{\nu}{\Delta x_{i-1}}
		\]
		
		Compute right interface flux coefficients ($i+1/2$ interface):
		\[
		a_{\text{conv}}^{\text{right}} = \min(a, 0) \quad \text{(upwind scheme)}
		\]
		\[
		d^{\text{right}} = \frac{\nu}{\Delta x_i}
		\]
		
		Assemble into tridiagonal matrix:
		\[
		A_{i,i-1} = -\left( d^{\text{left}} + \max(0, -a) \right)
		\]
		\[
		A_{i,i} = d^{\text{left}} + d^{\text{right}} + |a|
		\]
		\[
		A_{i,i+1} = -\left( d^{\text{right}} + \max(0, a) \right)
		\]
		
		Compute source term contribution (midpoint integration approximation):
		\[
		b_i = f(x_i) \cdot \frac{\Delta x_{i-1} + \Delta x_i}{2}
		\]
	}
	
	\BlankLine
	
	\parbox[t]{\linewidth}{
		\textbf{Handle Dirichlet boundary conditions}: \;
		$A_{0,0} = 1$, $b_0 = u_0$ \quad (left boundary)\\
		$A_{N,N} = 1$, $b_N = u_N$ \quad (right boundary)\\
		Set other elements in boundary rows to zero (maintain matrix structure)
	}

	\BlankLine
	Solve tridiagonal linear system $\mathbf{A} \mathbf{u} = \mathbf{b}$ (using Thomas algorithm, time complexity $O(N)$)\;
	
	\BlankLine
	\Return nodal solution vector $\mathbf{u}$\;
\end{algorithm}
\vspace{3mm}

\noindent \textcolor{structure3}{\textbf{4. Spectral Methods}}

Spectral methods adopt a ``global'' strategy fundamentally different from the finite element method. While the finite element method uses locally supported ``small tent'' basis functions, each affecting only a small surrounding region, spectral methods use smooth, oscillatory global functions defined over the entire domain as basis functions, such as sine functions, cosine functions, Chebyshev polynomials, Legendre polynomials, etc. The idea is to expand the unknown solution as a series of these global basis functions, then determine the expansion coefficients through some criterion (e.g., requiring the residual of the partial differential equation to be zero at collocation points, or orthogonal to all basis functions).

Behind this global approximation strategy lies a profound trade-off: if the solution to the problem is itself smooth (i.e., infinitely differentiable), then approximating it with equally smooth global basis functions can be extremely efficient. Unlike the algebraic convergence of finite difference or finite element methods (error decreases as a power law with increasing degrees of freedom), spectral methods can achieve exponential convergence—meaning adding a few basis functions can cause the error to drop dramatically, achieving very high accuracy with very few degrees of freedom.

Taking Fourier basis functions \(\sin(k\pi x)\), \(\cos(k\pi x)\) defined on interval \([0,1]\) and Chebyshev basis functions \(T_k(x)=\cos(k\arccos x)\) defined on \([-1,1]\) as examples, low-order basis functions (small \(k\)) have smooth waveforms and long wavelengths, while high-order basis functions (large \(k\)) exhibit rapid oscillations and shorter wavelengths. In numerical implementation, Chebyshev-Gauss collocation points are often used; such nodes are densely distributed near the boundaries and become sparser towards the middle of the interval. This node distribution helps suppress the Runge phenomenon (i.e., violent oscillatory distortion near the ends of the interval when using equidistant nodes for high-degree polynomial interpolation), thereby improving the stability of high-order interpolation.
\\
Spectral methods use oscillatory basis functions over the entire interval; low-order basis functions describe large-scale features of the solution, while high-order basis functions capture fine structures.

Consider the one-dimensional Poisson equation defined on interval \([-1, 1]\):
\[
u''(x) = f(x), \quad x \in (-1, 1)
\]
with Dirichlet boundary conditions \(u(-1) = a\), \(u(1) = b\). We choose Chebyshev polynomials as basis functions:
\[
T_k(x) = \cos(k \arccos x), \quad k = 0, 1, 2, \dots
\]
Chebyshev polynomials are orthogonal on interval \([-1, 1]\) with respect to the weight function \(1/\sqrt{1-x^2}\), and naturally form denser node distributions near the boundaries, making them particularly suitable for handling boundary layer problems. The approximate solution is expressed as a truncated expansion:
\[
u_N(x) = \sum_{k=0}^{N} \hat{u}_k T_k(x)
\]
where \(\hat{u}_k\) are the unknown expansion coefficients.
\\
Spectral methods mainly have two implementation approaches:
\begin{itemize}
	\item Galerkin spectral method: Substitute the approximate solution into the differential equation and require the residual to be orthogonal to all test functions \(T_j(x)\) (\(j = 0, 1, \dots, N\)) in the weighted inner product sense:
	\[
	\int_{-1}^1 \left( u_N''(x) - f(x) \right) T_j(x) \frac{dx}{\sqrt{1-x^2}} = 0
	\]
	This leads to an algebraic system for the coefficients \(\hat{u}_k\). Due to the orthogonality of Chebyshev polynomials and the special structure of derivative relationships, for some linear equations with constant coefficients, the stiffness matrix can be diagonal or nearly diagonal, making computation relatively efficient.
	\item Collocation method (pseudospectral method): This is the more common form in engineering applications. We choose a special set of collocation points—Chebyshev-Gauss-Lobatto nodes:
	\[
	x_j = \cos\left(\frac{\pi j}{N}\right), \quad j = 0, 1, \dots, N
	\]
	These nodes are exactly \(-1\) and \(1\) at the interval endpoints, and become denser closer to the boundaries. The collocation method requires the approximate solution \(u_N(x)\) to satisfy the differential equation exactly at each interior collocation point:
	\[
	u_N''(x_j) = f(x_j), \quad j = 1, 2, \dots, N-1
	\]
	while boundary conditions are directly enforced at the endpoint nodes:
	\[
	u_N(x_0) = a, \quad u_N(x_N) = b
	\]
	
	To compute \(u_N''(x_j)\), a differentiation matrix is introduced. Denote \(u_j = u_N(x_j)\) as the function values at the nodes; then derivatives can be computed via matrix multiplication:
	\[
	u_N'(x_j) = \sum_{k=0}^{N} D_{jk} u_k, \quad u_N''(x_j) = \sum_{k=0}^{N} D^{(2)}_{jk} u_k
	\]
	where \(D\) is the first-order differentiation matrix, and \(D^{(2)} = D \cdot D\) is the second-order differentiation matrix. Both matrices are dense, and their elements can be given explicitly via analytical formulas for Chebyshev polynomials. Thus, the discretized equation at interior collocation points is written as:
	\[
	\sum_{k=0}^{N} D^{(2)}_{jk} u_k = f(x_j), \quad j = 1, 2, \dots, N-1
	\]
	Combined with boundary conditions, this yields a system of linear algebraic equations for the node values \(\{u_k\}\).
\end{itemize}

Algorithm \ref{alg:collocation_poisson} shows the complete process of the collocation spectral method for solving the one-dimensional Poisson equation.
\begin{algorithm}[htbp]
	\caption{Collocation Spectral Method for Solving 1D Poisson Equation (Chebyshev Basis Functions)}
	\label{alg:collocation_poisson}
	
	\KwIn{
		Truncation order $N$ (highest polynomial degree),\;
		Right-hand side function $f(x)$,\;
		Boundary values $a = u(-1)$, $b = u(1)$
	}
	
	\KwOut{
		Solution vector $\mathbf{u}$ at collocation points, where $u_j \approx u(x_j)$
	}
	
	\BlankLine
	Generate Chebyshev-Gauss-Lobatto nodes:
	\[
	x_j = \cos\left(\frac{\pi j}{N}\right), \quad j = 0, 1, \dots, N
	\]
	
	Construct first-order differentiation matrix $\mathbf{D} \in \mathbb{R}^{(N+1) \times (N+1)}$ (standard formula, dense matrix)\;
	
	Compute second-order differentiation matrix $\mathbf{D}^{(2)} = \mathbf{D} \cdot \mathbf{D}$\;
	
	\BlankLine
	\parbox[t]{\linewidth}{
		\textbf{Extract submatrix and right-hand side corresponding to interior nodes}: \;
		Interior node indices: $j = 1, 2, \dots, N-1$\;
		$\mathbf{A}_{\text{int}} = \mathbf{D}^{(2)}[1:N, 1:N]$ (size $(N-1) \times (N-1)$)\;
		$\mathbf{b}_{\text{int}} = f(x_{1:N}) - \mathbf{D}^{(2)}[1:N, 0] \cdot a - \mathbf{D}^{(2)}[1:N, N] \cdot b$
	}
	
	\BlankLine
	Solve dense linear system $\mathbf{A}_{\text{int}} \mathbf{u}_{\text{int}} = \mathbf{b}_{\text{int}}$ (using LU decomposition, time complexity $O(N^3)$)\;
	
	Combine full solution vector: $\mathbf{u} = [a, \mathbf{u}_{\text{int}}, b]^{\mathsf{T}}$\;
	
	\BlankLine
	\Return solution vector $\mathbf{u}$ at collocation points\;
\end{algorithm}

Spectral methods are extremely efficient in their use of degrees of freedom. In one-dimensional problems, the degrees of freedom of the solution are determined by the truncation order \(N\), merely \(O(N)\), independent of grid step size. Extending to \(d\)-dimensional space, if using tensor product basis functions, the degrees of freedom become \(O(N^d)\). This means that to achieve the same resolution, spectral methods require far fewer degrees of freedom than finite difference or finite element methods.

However, the cost of this efficiency is reflected in the structure of the algebraic system. The differentiation matrix produced by the collocation method is dense—the derivative calculation at each collocation point depends on function values at all nodes, so the discrete algebraic system matrix is also dense. In one-dimensional problems, the storage cost is \(O(N^2)\), and direct solution (e.g., LU decomposition) computational cost is \(O(N^3)\). For two-dimensional problems, if using \(N^2\) collocation points, the matrix size becomes \(N^2 \times N^2\), with storage and solution costs rising to \(O(N^4)\) and \(O(N^6)\) respectively, a growth trend that becomes unbearable in high-dimensional cases. Therefore, practical high-dimensional spectral methods often need to rely on fast transforms (like the Fast Fourier Transform) or iterative solution techniques to avoid directly handling dense matrices.

The most striking feature of spectral methods is their convergence rate. If the true solution is infinitely smooth (i.e., all derivatives exist and are continuous), then as the truncation order \(N\) increases, the approximation error decays exponentially:
\[
\| u - u_N \| \sim O(e^{-cN})
\]
where \(c > 0\) is a constant. This means that adding a few basis functions reduces the error by an order of magnitude. To achieve \(10^{-6}\) accuracy, the finite element method may require tens of thousands of degrees of freedom, while spectral methods often need only a few dozen. This exponential convergence rate is much faster than the algebraic convergence \(O(N^{-p})\) of finite difference or finite element methods, making spectral methods an ideal tool for high-precision computation.

However, this remarkable convergence speed has a key prerequisite: the solution must be smooth. If the true solution has discontinuities or sharp corners, the situation is completely different. Approximating a non-smooth function with smooth global basis functions leads to the famous Gibbs phenomenon—violent unphysical oscillations near the discontinuity, which do not disappear as \(N\) increases, only making the oscillation frequency higher. From a convergence perspective, when the solution is not smooth, the convergence rate of spectral methods deteriorates sharply to algebraic convergence \(O(N^{-1})\) or worse. This sensitivity constitutes the main application limitation of spectral methods.

The greatest advantage of spectral methods lies in their unparalleled accuracy and efficiency for smooth problems. Exponential convergence means that extremely high-precision numerical solutions can be obtained with very few degrees of freedom, making spectral methods the method of choice in fields requiring high-precision computation, such as direct numerical simulation (DNS) of turbulence, quantum chemistry calculations, and spectral models in weather forecasting. At the same time, for periodic problems and problems on regular domains, spectral methods can fully utilize techniques like the Fast Fourier Transform to achieve efficient computation.

However, the limitations of spectral methods are equally apparent. They have extremely stringent requirements on the smoothness of the solution; once the solution has discontinuities or local sharp variations, convergence deteriorates drastically, and even unphysical oscillations may appear. Furthermore, spectral methods have poor adaptability to complex geometric shapes—traditional spectral methods require the computational domain to be regular shapes like rectangles, annuli, etc., making it difficult to handle irregular domains common in engineering. Although variants like the spectral element method combine the geometric flexibility of finite elements with the high accuracy of spectral methods by partitioning the computational domain into multiple subdomains and using spectral approximation on each, the implementation complexity and computational cost increase significantly. Finally, the dense matrix nature of spectral methods causes computational costs to rise sharply for high-dimensional problems, limiting their widespread application in problems beyond three dimensions.

\section{Basic Framework and Mathematical Principles of PINNs}

Although traditional numerical methods have achieved great success and form the backbone of modern scientific computing, they reveal inherent limitations when facing the following emerging challenges. These limitations precisely constitute the core motivation for AI methods represented by PINNs to intervene in the field of PDE solving.
\\
\textcolor{structure3}{\textbf{Challenge 1: The ``Curse of Dimensionality'' in High-Dimensional Problems}}

This is the most fundamental challenge. The degrees of freedom (number of grid points, elements, basis functions) of traditional grid-based methods grow exponentially with the spatial dimension \( d \). For example, in a \( d \)-dimensional unit hypercube, if each dimension is discretized into \( N \) points, then the degrees of freedom for FDM and FEM (on quasi-uniform grids) are approximately \( N^d \). For a moderate dimension \( d=10 \), even with only \( N=10 \) points per dimension, the total degrees of freedom reach as high as \( 10^{10} \), which already exceeds conventional computing capabilities. While spectral methods also have degrees of freedom of \( O(N^d) \), the storage and computational costs of their dense matrices \( O(N^{2d}) \) are even more catastrophic. As a function approximator, the number of parameters (weights and biases) of a neural network mainly depends on the width and depth of the network, and not directly on the dimension of the input space. PINNs take spatial and temporal coordinates as input and directly output the value of the solution at that point. This ``point-to-point'' solving approach bypasses the need for global discretization of the entire high-dimensional space. Although the training process also requires computing the loss at many points, these points are randomly or adaptively sampled, and their number can be controlled independently of the dimension, offering a potential pathway to alleviate the curse of dimensionality.
\\
\textcolor{structure3}{\textbf{Challenge 2: Complicated Geometry, Inverse Problems, and Automation}}

For complex and irregular solution domains, although FEM and FVM can adapt through mesh generation, generating high-quality meshes itself is computationally expensive. Moreover, for scenarios involving topological changes (e.g., crack propagation), moving boundaries (e.g., free surfaces), or inferring geometry (inverse problems), the mesh may need dynamic adjustment, which is cumbersome and requires expertise. PINNs are inherently meshless methods. The network input is merely coordinate points \((x, y, z, t)\). As long as points can be sampled within the domain (including boundaries), the loss function can be constructed for training, making it highly insensitive to geometric complexity. This advantage is particularly evident when solving inverse problems: for example, in problems where only partial internal observation data (e.g., readings from several sensors) are known, and physical parameters, unknown source terms, or even boundary shapes need to be inferred simultaneously. PINNs can naturally formulate this as a unified optimization problem by incorporating the PDE residual, boundary conditions (if known), and data fitting terms into the loss function. The neural network parameters simultaneously encode the solution and the parameters to be inverted, and both can be obtained through a single training process, without the need to regenerate meshes and solve the forward problem for each possible parameter or geometric hypothesis.
\\
\textcolor{structure3}{\textbf{Challenge 3: Integrating Multi-Source Heterogeneous Data and Uncertainty Quantification}}

Traditional numerical solvers are typically ``purely physics-driven'': given complete governing equations and well-posed conditions, they output a deterministic solution. However, in practical applications, we often have sparse, noisy data from experiments or observations, and there may also be epistemic uncertainty about certain physical parameters or boundary conditions. Traditional methods struggle to seamlessly and naturally integrate these data and uncertainties into the solving process, often requiring complex data assimilation (e.g., 4D-Var) or stochastic PDE frameworks. The PINN loss function has a natural capacity for fusion and potential for probabilistic interpretation extension. It can sum the PDE residual loss, boundary/initial condition loss, and data fitting loss with weights. By minimizing this total loss, the solution learned by the neural network simultaneously satisfies physical laws and actual observation data, becoming a ``physics-driven + data-driven'' hybrid model. Furthermore, by employing a Bayesian framework or introducing random variables, PINNs can quantify prediction uncertainty, which is crucial for simulation-based decision-making.
\\
\textcolor{structure3}{\textbf{Challenge 4: Automation and Generality of the Solving Process}}

Implementing traditional numerical methods typically requires significant intervention from domain experts: selecting appropriate methods based on problem type (FEM vs. FVM), generating high-quality meshes, designing stable discretization schemes, handling complex boundary conditions, selecting and tuning solver parameters (e.g., iteration tolerance, preconditioners), etc. This process is highly specialized, difficult to automate, and has poor portability. PINNs provide a relatively unified framework: for a large class of PDE problems, the workflow can be abstracted as ``define network architecture, construct physics-informed loss (using automatic differentiation), sample coordinate points within the domain and on boundaries, train the network to optimize the loss.'' The automatic differentiation (Autograd) functionality provided by modern deep learning frameworks (e.g., PyTorch, TensorFlow) makes computing PDE residuals (often involving high-order partial derivatives) exceptionally simple—one only needs to code the original equation without manually deriving complex discretization schemes. This greatly lowers the barrier to entry for PDE solving and facilitates the construction of general-purpose solving tools and code libraries.

Next, we detail the basic framework and mathematical principles of PINNs.

From a mathematical perspective, solving a PDE problem essentially involves finding a function that satisfies specific equations and boundary/initial conditions. This is a classic function approximation problem: we wish to find a specific function satisfying complex constraints within an infinite-dimensional function space. Traditional numerical methods solve the continuous problem by discretizing it, transforming the infinite-dimensional problem into a finite-dimensional linear or nonlinear system. Although this discretization process is effective, it also, to some extent, ``solidifies'' our representation of the solution (e.g., a linear combination of basis functions) and tightly couples computational complexity with mesh resolution.

It is in this context that Physics-Informed Neural Networks emerged. The core idea is: Can we leverage the powerful function approximation capability of neural networks to directly approximate the solution function that satisfies the PDE constraints? This idea did not arise out of thin air. As early as the 1990s, researchers explored using neural networks to solve differential equations. However, its true revival and widespread application benefited from the revolutionary progress in deep learning in the 2010s: the emergence of efficient automatic differentiation frameworks, the proliferation of large-scale parallel computing hardware, and the deepening theoretical understanding of neural networks. The ``Physics-Informed Neural Networks'' framework proposed by Raissi et al. in 2019 clearly outlined the mathematical contours of this paradigm, making it a unified and flexible solver.

The fundamental motivation of PINNs lies in transforming PDE solving from a purely numerical discretization problem into a machine learning problem constrained by physical laws. The neural network is no longer merely a black-box fitting tool but becomes a ``white-box'' or ``gray-box'' model embedded with physical prior knowledge. This shift in perspective offers new possibilities for solving complex problems that are difficult for traditional methods to handle (e.g., high-dimensional, inverse problems, data assimilation).
\vspace{3mm}
\\
\textcolor{structure3}{\textbf{1. Core Idea: The Loss Function as the ``Guardian'' of Physical Laws}}

The basic idea of PINNs is intuitive and powerful. We can summarize it into the following key steps:

Step 1: Parameterize the unknown solution. Use a neural network to approximate the PDE solution. The neural network input is the spatial coordinates $\mathbf{x}$ and time $t$, and the output is the approximate solution value $\tilde{u}(\mathbf{x}, t; \boldsymbol{\theta})$, where $\boldsymbol{\theta}$ represents all adjustable parameters (weights and biases) of the network. Unlike traditional numerical methods, there is no concept of a grid here—the neural network itself is a global function approximator.

Step 2: Encode physical laws into the loss function. The PDE itself, initial conditions, and boundary conditions are no longer treated as ``hard constraints'' that must be strictly satisfied but are transformed into ``soft constraints'' that can be optimized, i.e., parts of a loss function. The goal of network training is to minimize this loss function. This approach reflects a profound shift in perspective: instead of satisfying the equation exactly at discrete points, we seek a function that ``as much as possible'' satisfies all constraints over the entire domain.

Step 3: Solve via optimization. Use gradient-based optimization algorithms (e.g., Adam, L-BFGS) to adjust the network parameters $\boldsymbol{\theta}$ so that the loss function continuously decreases. When the loss is sufficiently small, we consider the neural network output $\tilde{u}$ to be an acceptable approximate solution to the original PDE.

The cleverness of this framework lies in its full utilization of automatic differentiation technology in modern deep learning frameworks. To compute the PDE residual (i.e., the difference between the left and right sides of the equation), we need to compute the partial derivatives of the approximate solution $\tilde{u}$ with respect to input variables (e.g., $x, t$). In traditional programming, this requires manual derivation or symbolic computation, which is cumbersome and error-prone. However, in modern frameworks like TensorFlow and PyTorch, automatic differentiation can accurately compute derivatives of arbitrary order with computational efficiency comparable to network forward propagation. This makes embedding complex differential operators into the loss function exceptionally simple.

Therefore, the core mathematical principle of PINNs can be summarized as: transforming a PDE solving problem into a nonlinear optimization problem with neural network parameters as optimization variables and the weighted sum of physical law (PDE, initial/boundary conditions) and data fitting errors as the objective.
\vspace{3mm}
\\
\textcolor{structure3}{\textbf{2. Mathematical Formulation: A Unified Loss Function Framework}}

Now, let us formally describe a typical PDE problem and how a PINN models it.
\\
Consider a PDE problem defined on a spatial domain $\Omega \subset \mathbb{R}^d$ and time interval $[0, T]$. We seek a function $u(\mathbf{x}, t)$ that satisfies:

\begin{itemize}
	\item Governing equation (PDE):
	\[
	\mathcal{N}[u](\mathbf{x}, t) = f(\mathbf{x}, t), \quad (\mathbf{x}, t) \in \Omega \times [0, T]
	\]
	where $\mathcal{N}$ is a differential operator (e.g., for the heat equation, $\mathcal{N}[u] = \frac{\partial u}{\partial t} - \kappa \nabla^2 u$).
	
	\item Initial condition (IC):
	\[
	u(\mathbf{x}, 0) = h(\mathbf{x}), \quad \mathbf{x} \in \Omega
	\]
	
	\item Boundary condition (BC):
	\[
	\mathcal{B}[u](\mathbf{x}, t) = g(\mathbf{x}, t), \quad (\mathbf{x}, t) \in \partial\Omega \times [0, T]
	\]
	where $\mathcal{B}$ is a boundary operator (e.g., Dirichlet condition $\mathcal{B}[u]=u$, or Neumann condition $\mathcal{B}[u]=\nabla u \cdot \mathbf{n}$).
\end{itemize}
Additionally, sometimes we have some observation data points $\{ (\mathbf{x}_d^i, t_d^i), u_d^i \}_{i=1}^{N_d}$, which may come from experimental measurements or incomplete numerical solutions.

The PINN method approximates the true solution $u(\mathbf{x}, t)$ using a parameterized neural network $\tilde{u}(\mathbf{x}, t; \boldsymbol{\theta})$. Next, we construct a composite loss function to simultaneously encode all these constraints:

\[
\mathcal{L}(\boldsymbol{\theta}) = \lambda_f \mathcal{L}_f(\boldsymbol{\theta}) + \lambda_b \mathcal{L}_b(\boldsymbol{\theta}) + \lambda_i \mathcal{L}_i(\boldsymbol{\theta}) + \lambda_d \mathcal{L}_d(\boldsymbol{\theta})
\]
The meanings of each term are as follows:
\begin{enumerate}
	\item  PDE residual loss $\mathcal{L}_f$: Measures how well the approximate solution satisfies the governing equation inside the domain. By sampling a set of ``residual points'' or ``collocation points'' $\mathcal{T}_f = \{ (\mathbf{x}_f^i, t_f^i) \}_{i=1}^{N_f}$ within the domain, compute:
	\[
	\mathcal{L}_f(\boldsymbol{\theta}) = \frac{1}{N_f} \sum_{i=1}^{N_f} \left| \mathcal{N}[\tilde{u}](\mathbf{x}_f^i, t_f^i; \boldsymbol{\theta}) - f(\mathbf{x}_f^i, t_f^i) \right|^2
	\]
	This is the core of the entire loss function, forcing the neural network to learn the physical laws hidden in the PDE.
	
	\item  Boundary condition loss $\mathcal{L}_b$: Forces the approximate solution to satisfy given conditions on the boundary. Sample a set of points $\mathcal{T}_b = \{ (\mathbf{x}_b^i, t_b^i) \}_{i=1}^{N_b}$ on the boundary, compute:
	\[
	\mathcal{L}_b(\boldsymbol{\theta}) = \frac{1}{N_b} \sum_{i=1}^{N_b} \left| \mathcal{B}[\tilde{u}](\mathbf{x}_b^i, t_b^i; \boldsymbol{\theta}) - g(\mathbf{x}_b^i, t_b^i) \right|^2
	\]
	
	\item  Initial condition loss $\mathcal{L}_i$: Forces the approximate solution to satisfy given conditions at the initial time. Sample points $\mathcal{T}_i = \{ (\mathbf{x}_i^i, 0) \}_{i=1}^{N_i}$ on the initial time surface, compute:
	\[
	\mathcal{L}_i(\boldsymbol{\theta}) = \frac{1}{N_i} \sum_{i=1}^{N_i} \left| \tilde{u}(\mathbf{x}_i^i, 0; \boldsymbol{\theta}) - h(\mathbf{x}_i^i) \right|^2
	\]
	
	\item  Data loss $\mathcal{L}_d$ (optional): If observation data exist, used to force the approximate solution to match the data:
	\[
	\mathcal{L}_d(\boldsymbol{\theta}) = \frac{1}{N_d} \sum_{i=1}^{N_d} \left| \tilde{u}(\mathbf{x}_d^i, t_d^i; \boldsymbol{\theta}) - u_d^i \right|^2
	\]
\end{enumerate}
In the formula, $\lambda_f, \lambda_b, \lambda_i, \lambda_d$ are positive weighting coefficients used to balance the magnitude and importance of different loss terms. Setting these weights is crucial for successful training—they determine the network's priority between fitting the PDE residual, boundary conditions, initial conditions, and observation data.

Ultimately, solving the PDE is transformed into an optimization problem:
\[
\boldsymbol{\theta}^* = \arg\min_{\boldsymbol{\theta}} \mathcal{L}(\boldsymbol{\theta})
\]
By iteratively updating $\boldsymbol{\theta}$ using gradient descent or its variants (e.g., Adam), the final network $\tilde{u}(\mathbf{x}, t; \boldsymbol{\theta}^*)$ is the desired approximate solution.

In practice, setting the weighting coefficients $\lambda$ in the loss function is often an empirical challenge. The PDE residual loss $\mathcal{L}_f$ involves high-order derivatives, and its numerical value is usually much smaller than the boundary/initial condition losses $\mathcal{L}_b, \mathcal{L}_i$. If all weights are simply set to 1, the optimization process may be dominated by $\mathcal{L}_b$ and $\mathcal{L}_i$, causing the network to quickly satisfy boundary/initial conditions while the PDE residual remains large, resulting in a trivial solution that does not satisfy physical laws.

In practice, setting the weighting coefficients $\lambda$ in the loss function is often an empirical challenge. The PDE residual loss $\mathcal{L}_f$ involves high-order derivatives, and its numerical value is usually much smaller than the boundary/initial condition losses $\mathcal{L}_b, \mathcal{L}_i$. If all weights are simply set to 1, the optimization process may be dominated by $\mathcal{L}_b$ and $\mathcal{L}_i$, causing the network to quickly satisfy boundary/initial conditions while the PDE residual remains large, resulting in a trivial solution that does not satisfy physical laws.
\\
Common tuning strategies include:

\begin{itemize}
	\item Empirical trial and error: Manually set weights based on the initial magnitude ratio of loss terms, e.g., making $\lambda_f \mathcal{L}_f(0) \approx \lambda_b \mathcal{L}_b(0) \approx \lambda_i \mathcal{L}_i(0)$. This method is simple and intuitive but requires multiple attempts.
	
	\item Learning rate annealing: Dynamically adjust weights during training, e.g., adaptively adjusting based on the descent speed or current value of each loss term. A common practice is to make the weights inversely proportional to the average of the corresponding loss term, keeping the contribution of each loss to the total loss balanced.
	
	\item Gradient-based balancing: Adjust weights by monitoring the relative contribution of different loss terms to parameter updates (i.e., gradient norms), making the impact of each term on optimization roughly balanced. The basic idea is: if a loss term produces too large a gradient norm, reduce its weight; otherwise, increase it.
	
	\item Adaptive probabilistic models: Treat each loss term as the negative log-likelihood of a Gaussian distribution, with its variance (i.e., the reciprocal of the weight) as a trainable parameter, optimized simultaneously via maximum likelihood estimation. This method transforms the weight selection problem into an automatic learning process but is relatively complex to implement.
\end{itemize}

The computation of the loss function relies on discrete point sets sampled within the domain, on boundaries, and at the initial time. The distribution strategy of these points directly affects training efficiency and final solution accuracy. Commonly used sampling strategies include:

\begin{itemize}
	\item Uniform random sampling: The simplest method, uniformly random sampling within the domain. Suitable for problems with smooth solution variations.
	
	\item Latin hypercube sampling: A stratified random sampling method that divides each dimension into equiprobable intervals and then takes one sample from each interval. This method ensures uniform distribution of points projected onto each coordinate axis and has better coverage than pure random sampling in moderate dimensions.
	
	\item Quasi-Monte Carlo sequences: Low-discrepancy sequences such as Sobol sequences, Halton sequences, etc., can generate point sets with more uniform distribution than random sampling, often accelerating convergence and providing more stable training.
	
	\item Adaptive importance sampling: Dynamically adjust the distribution of sampling points during training based on the magnitude of the current PDE residual, collecting more points in regions with large residuals. This method can allocate computational resources more effectively, focusing on difficult regions.
\end{itemize}

The number and distribution of initial sampling points are hyperparameters that need tuning. Typically, boundary and initial condition points need to be dense enough to ensure constraints are accurately imposed, while the number of internal collocation points determines the sufficiency of the network's learning of the PDE. Empirically, the number of collocation points is usually much larger than that of boundary and initial points.

\begin{figure}[htbp]
	\centering
	\includegraphics[width=0.8\linewidth]{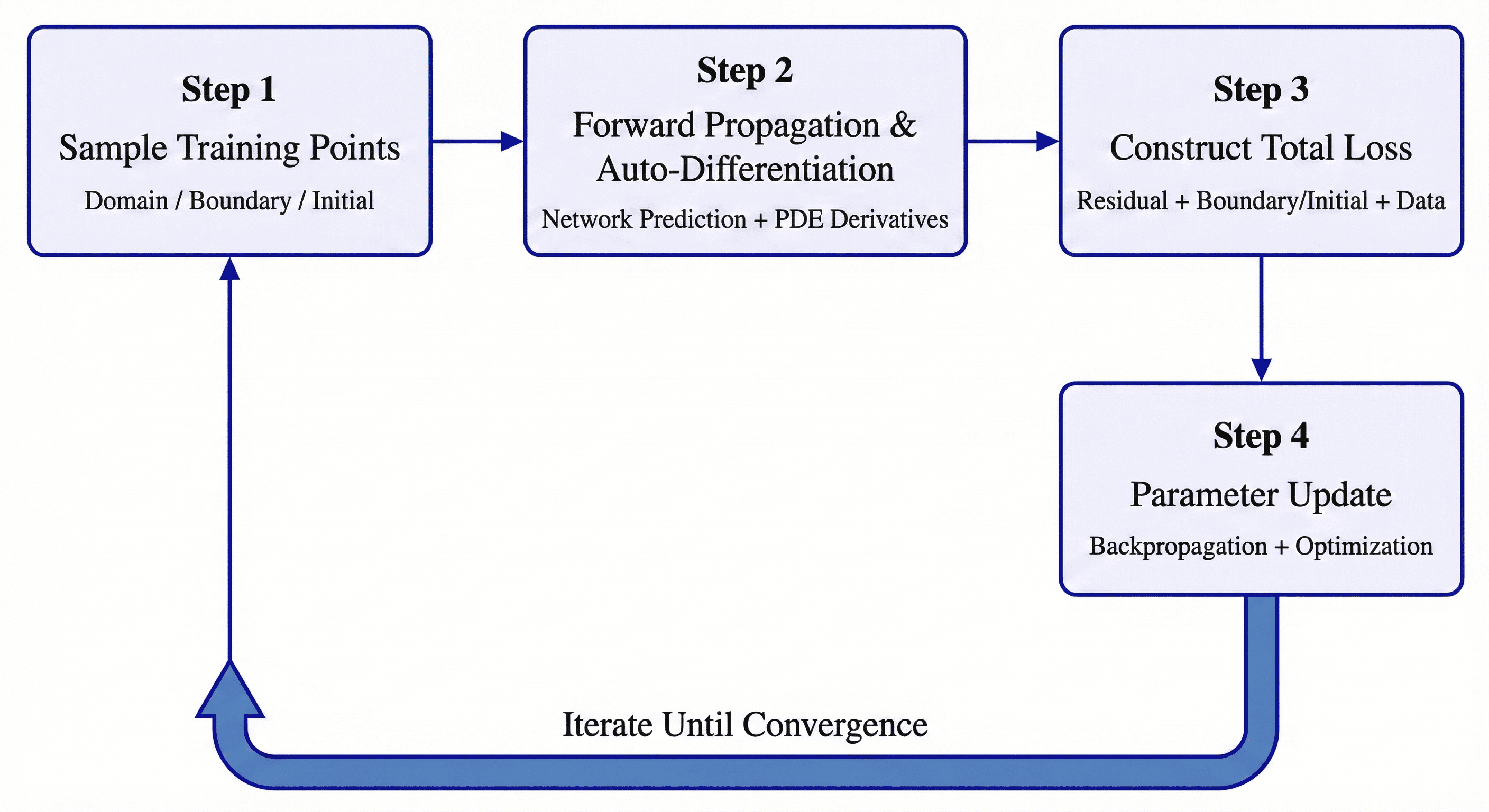}
	\caption{PINN Training Process \label{fig:PINN}}
\end{figure}

The PINN training process is shown in Figure\ref{fig:PINN}. PINN training includes training point sampling, forward propagation and automatic differentiation, loss construction, and parameter update, iteratively optimized until convergence.
\newpage

\vspace{3mm}
\noindent \textcolor{structure3}{\textbf{3. Network Architecture: Why Choose Fully Connected Networks?}}

In early and most PINN works, the foundational architecture chosen is the Multilayer Perceptron (MLP), also known as a fully connected feedforward neural network. This choice is not accidental but is jointly determined by the mathematical properties of MLPs and the characteristics of PDE problems.

Mathematically, MLPs are supported by the Universal Approximation Theorem. This theorem states that as long as there are enough neurons in a single hidden layer, an MLP can approximate any continuous function defined on a compact set to arbitrary accuracy. For deeper networks, their expressive power is even stronger. The solutions of PDEs are typically smooth functions we expect (at least inside the domain), so MLPs theoretically have the ability to approximate them.
\begin{theorem}{Universal Approximation Theorem}
	Let $\varphi(\cdot)$ be a non-constant, bounded, monotonically increasing continuous function, $\mathcal{I}_d$ be a $d$-dimensional unit hypercube $[0,1]^d$, and $C(\mathcal{I}_d)$ be the set of continuous functions defined on $\mathcal{I}_d$. For any function $f \in C(\mathcal{I}_d)$, there exists an integer $m$, and sets of real numbers $v_i, b_i \in \mathbb{R}$ and real vectors $\mathbf{w}_i \in \mathbb{R}^d$ ($i=1,\dots,m$), such that the function $F$ defined by
	
	\[
	F(\mathbf{x}) = \sum_{i=1}^{m} v_i \varphi(\mathbf{w}_i^\mathrm{T} \mathbf{x} + b_i)
	\]
	satisfies
	\[
	|F(\mathbf{x}) - f(\mathbf{x})| < \epsilon, \quad \forall \mathbf{x} \in \mathcal{I}_d
	\]
	where $\epsilon > 0$ is an arbitrarily small positive number.
\end{theorem}

The Universal Approximation Theorem guarantees the ability of MLPs as universal function approximators: as long as the network is wide enough, it can approximate any continuous function to arbitrary accuracy. This theoretical cornerstone provides rigorous mathematical support for the feasibility of PINNs.

From the problem structure perspective, the input of a PDE is usually spatial coordinates and time, with low dimensionality (1D, 2D, 3D plus time), and the output is the solution value, usually a scalar or low-dimensional vector. MLPs are naturally suited for handling this ``coordinates-to-value'' mapping relationship. Unlike convolutional neural networks (CNNs), which assume input has local spatial structure, or recurrent neural networks (RNNs), which assume input has temporal structure, MLPs make no prior assumptions about data structure, making them truly universal function approximators that can handle various geometric shapes and equation types equally.

A typical MLP used for PINNs can be expressed as:
\[
\begin{aligned}
	&\mathbf{z}^{(0)} = \mathbf{x} \quad (\text{input layer, containing spatial coordinates and time}) \\
	&\mathbf{z}^{(l)} = \sigma(\mathbf{W}^{(l)} \mathbf{z}^{(l-1)} + \mathbf{b}^{(l)}), \quad l = 1, \ldots, L-1 \quad (\text{hidden layers}) \\
	&\tilde{u} = \mathbf{W}^{(L)} \mathbf{z}^{(L-1)} + \mathbf{b}^{(L)} \quad (\text{output layer})
\end{aligned}
\]
where $\sigma$ is a nonlinear activation function, such as $\tanh$, $\sin$, or ReLU. In PINNs, $\tanh$ is particularly commonly used due to its smoothness (infinitely differentiable) and boundedness (output range $[-1, 1]$), because smoothness ensures stability in computing high-order derivatives via automatic differentiation, and boundedness helps stabilize training convergence.

However, MLPs are not the only choice. Depending on specific problem characteristics, researchers have also explored other architectures. Table\ref{tab:PINN 中常用神经网络架构对比} compares several common architectures in terms of their applicable scenarios and advantages/disadvantages in PINNs.

\begin{table}[htbp]
	\centering
	\caption{Comparison of Commonly Used Neural Network Architectures in PINNs\label{tab:PINN 中常用神经网络架构对比}}
	\begin{tabular}{p{2cm} p{3cm} p{3.6cm} p{3.2cm} p{3.2cm}}
		\toprule
		\textbf{Architecture Type} & \textbf{Core Idea} & \textbf{Applicable Scenarios} & \textbf{Advantages} & \textbf{Disadvantages and Challenges} \\
		\midrule
		MLP & Fully connected, universal function approximator & General, low-dimensional problems, irregular geometry & Solid theoretical guarantee (Universal Approximation Theorem), simple implementation, naturally compatible with automatic differentiation & Suffers from ``spectral bias'', low parameter efficiency for high-dimensional problems \\
		\addlinespace
		
		CNN & Local connectivity, weight sharing, translation invariance & Regular grid data (e.g., image-like physical fields), problems with local correlations & High parameter efficiency, effective at extracting local features, suitable for high-dimensional spatial processing & Complex to handle irregular geometry, requires projecting input onto regular grids \\
		\addlinespace
		
		RNN/LSTM & Recurrent connections, memory of historical states & Strongly time-dependent problems, sequence prediction & Naturally suited for temporal modeling, can capture long-term dependencies & Training prone to gradient vanishing/explosion, low parallelization \\
		\addlinespace
		
		Transformer & Self-attention mechanism, global dependency modeling & Complex geometry, long-range interactions, point cloud data & Can model dependencies at arbitrary distances, high flexibility & High computational and memory cost, training requires large amounts of data \\
		\addlinespace
		
		KAN & Learnable activation functions, based on Kolmogorov-Arnold theorem & Function approximation, scenarios requiring high interpretability & Fewer parameters, potentially better interpretability & Still in exploratory stage, theory and methods are still developing \\
		\bottomrule
	\end{tabular}
\end{table}
The choice of architecture should depend on the specific problem. For most standard, low-to-medium-dimensional forward PDE problems, a sufficiently deep and wide MLP is usually a reliable and convenient starting point. When the problem has obvious structural characteristics (e.g., spatiotemporal sequences, image data on regular domains), choosing corresponding specialized architectures (RNN, CNN) may improve efficiency and solution quality. Emerging architectures like Transformer and KAN represent frontier exploration directions, suitable for complex scenarios where traditional architectures perform poorly.
\vspace{3mm}
\\
\textcolor{structure3}{\textbf{4. The Triad of Error Sources: Approximation, Optimization, and Generalization}}

Understanding the sources of error in PINNs is crucial for evaluating their performance and reliability. Unlike traditional numerical methods, PINN error consists of three interrelated components, which we can call the ``error triad'':

\begin{enumerate}
	\item Approximation error: Even if there exists an optimal set of network parameters $\boldsymbol{\theta}^*$, the neural network function space $\{\tilde{u}(\cdot; \boldsymbol{\theta})\}$ may not perfectly represent the true solution $u$. This is determined by the limited expressive capacity of the network architecture (e.g., depth, width, choice of activation function). The Universal Approximation Theorem guarantees that error can be arbitrarily small with infinite width or depth, but in practice networks are finite in size, so their expressive capacity is necessarily limited. For solutions with singularities, boundary layers, or high-frequency oscillations, approximation error may be particularly significant—because these complex features require sufficient network capacity to accurately capture. Furthermore, the smoothness of the activation function also affects approximation ability: for example, ReLU networks can only represent piecewise linear functions, while $\tanh$ networks can approximate smooth functions.
	
	\item Optimization error: The training process may fail to find the global optimum $\boldsymbol{\theta}^*$, instead converging to a local optimum $\tilde{\boldsymbol{\theta}}$ such that $\mathcal{L}(\tilde{\boldsymbol{\theta}}) > \mathcal{L}(\boldsymbol{\theta}^*)$. Neural network loss functions are typically non-convex, with numerous saddle points and local minima in high-dimensional parameter spaces. The choice of optimization algorithm (Adam or L-BFGS?), learning rate scheduling strategy, and network parameter initialization all significantly affect the final converged parameter point. Moreover, differences in numerical magnitude between different terms in the loss function (as mentioned earlier) can also cause optimization difficulties, causing certain constraints to be ignored. Therefore, even if the network can theoretically represent the exact solution, inappropriate optimization may fail to find it.
	
	\item Generalization/Discretization error: Even if we find a network with low loss on the training sample points, we cannot guarantee it satisfies the PDE over the entire continuous domain. This is because the loss function is computed on a finite set of points $\{\mathcal{T}_f, \mathcal{T}_b, \mathcal{T}_i\}$, not over the entire continuous domain. If these sampling points are distributed unreasonably—for example, insufficient sampling in regions where the solution changes rapidly, or sparse sampling near boundaries—the network may produce large errors in unsampled regions. This is similar to discretization error in traditional numerical methods, but in PINNs, the selection and distribution of sampling points are more flexible, allowing both uniform random sampling and adaptive strategies to refine in difficult regions. However, this flexibility also brings new challenges: how to design sampling strategies to ensure accuracy over the entire domain remains an open research question.
\end{enumerate}
These three types of error are coupled, making the theoretical analysis of PINNs far more complex than that of traditional numerical methods. For example, a network with sufficient capacity (small approximation error) may be more difficult to optimize (large optimization error), and a network that fits training points perfectly may perform poorly at unsampled points (large generalization error). In practice, we rely more on posterior error assessment to judge solution quality: after training, evaluate the loss on an independent, dense set of test points, check the physical plausibility of the solution (e.g., symmetry, conservation laws), and compare it with known analytical solutions or high-precision numerical solutions. Furthermore, many physical systems satisfy conservation laws (e.g., mass, energy, momentum conservation). We can compute these conserved quantities corresponding to the PINN solution and check if they are physically reasonable (e.g., whether they remain constant over time in a closed system), which is an effective means of physical consistency verification.
\vspace{3mm}
\\
\textcolor{structure3}{\textbf{5. Theoretical Credibility: When Can We Trust PINNs?}}

Given the above challenges, a natural question arises: Under what circumstances can we have high confidence in PINN solutions? Although a complete theory is still under development, based on existing research and experience, we can outline some guiding principles:

\begin{itemize}
	\item Smoothness of the solution: When the PDE solution is sufficiently smooth (no singularities, discontinuities, or strong boundary layers), MLPs can approximate it more easily, the impact of spectral bias is smaller, and PINNs typically perform well. Elliptic equations (e.g., Poisson equation) under smooth domains and smooth boundary conditions are typical representatives of such problems. Conversely, for fluid problems containing shocks or singular perturbation problems with boundary layers, standard PINNs often struggle to capture these local sharp variations, requiring special treatments (e.g., adaptive sampling, domain decomposition, or shock-capturing techniques).
	
	\item Problem dimensionality: For problems with spatial dimensions not exceeding 3 (plus time), sampling and training for PINNs are relatively manageable, and satisfactory accuracy can be achieved with reasonable computational cost. For higher-dimensional problems (e.g., high-dimensional PDEs, stochastic PDEs in uncertainty quantification), standard PINNs face the ``curse of dimensionality''—exponentially increasing sampling points are needed to cover the entire space as dimension increases. Special architectures (e.g., tensor decomposition networks) or sampling strategies (e.g., sparse grids, low-discrepancy sequences) are then required to address this challenge.
	
	\item Well-posedness of the problem: PINNs are most suitable for solving well-posed forward problems—i.e., where the complete equation form, all parameters, and complete boundary and initial conditions are known. For inverse problems (requiring simultaneous inference of equation parameters or boundary conditions) or data-scarce situations, although PINNs can also handle them, the risk of failure increases significantly due to increased uncertainty in the solution space. In such cases, introducing additional regularization or leveraging physical prior knowledge becomes particularly important.
	
	\item Sufficient training and validation: Decreasing training loss does not equate to high solution accuracy. It is essential to validate solution accuracy using an independent, dense set of test points, not just relying on training loss. Checking the physical plausibility of the solution (e.g., symmetry, monotonicity, conservation laws) is also an important validation means. For time-dependent problems, one can check if the solution satisfies physical properties like energy dissipation or mass conservation.
	
	\item Consistency with traditional methods: Whenever possible, compare the PINN solution with high-precision traditional numerical solutions (e.g., spectral methods or fine-grid finite element solutions). Consistency is strong evidence for building confidence. Even if a global exact solution is unavailable, comparative validation in local regions or simplified cases is a beneficial practice.
\end{itemize}
When facing problems that do not meet the above conditions (e.g., fluid problems with shocks, high-dimensional stochastic PDEs, extremely complex geometries), one must resort to more advanced PINN variants or hybrid methods, such as domain decomposition, adaptive sampling, multi-scale network structures, etc.

However, we must clearly recognize that PINNs are not intended to completely replace traditional numerical methods. In domains where traditional methods excel (e.g., linear PDEs on regular domains, large-scale industrial steady-state simulations), they still hold overwhelming advantages in accuracy, efficiency, reliability, and maturity. The emergence of PINNs is more like adding a brand new, complementary set of tools to our PDE solving toolbox. They are particularly suitable for scenarios that are difficult or cumbersome for traditional methods to handle: high-dimensional problems, complex geometry and inverse problems, and scenarios requiring flexible integration of physical models with multi-source data.

\section{Applications}

In the previous sections, we systematically introduced the basic framework, training techniques, and robustification methods of Physics-Informed Neural Networks. Understanding these theories and techniques is foundational, while learning how they are applied to solve real scientific computing problems is the final step in mastering the PINN approach. In this section, we will demonstrate, through a series of representative application cases, how PINNs leverage their unique advantages when facing complex partial differential equation problems across multiple scientific and engineering fields.

This section will delve into the mathematical essence behind each case, explain the design motivation and core ideas of the PINN method, and outline the application boundaries and potential of PINNs through discussions on results and limitations. We will structure each case as follows: Problem Background and Mathematical Formulation, Limitations of Traditional Methods and Motivation for PINNs, Specific Design of the PINN Model, Result Analysis and Comparison, Summary and Insights. This structure aims to clearly reveal how PINNs intervene for a specific mathematical problem (PDE) and what new possibilities they offer at the methodological level.
\vspace{3mm}
\\
\textcolor{structure3}{\textbf{Example One: Solving Poisson's Equation with Complex Source Terms and Geometry (Forward Problem)}}
\begin{itemize}
	\item Problem Background and Mathematical Formulation\\
	Poisson's equation is one of the most fundamental elliptic partial differential equations in mathematical physics, describing the distribution of potential fields (such as electrostatic fields, gravitational fields, steady-state temperature fields). Its general form is:
	
	\[
	-\nabla^2 u(\mathbf{x}) = f(\mathbf{x}), \quad \mathbf{x} \in \Omega
	\]
	
	with appropriate boundary conditions, for example, Dirichlet boundary conditions \( u(\mathbf{x}) = g(\mathbf{x}), \mathbf{x} \in \partial\Omega \).
	
	\item Limitations of Traditional Methods and Motivation for PINNs\\
	Solving Poisson's equation is a classic problem in computational science, and traditional numerical methods like the Finite Element Method are very mature. However, when the source term \( f(\mathbf{x}) \) exhibits sharp variations, singularities, or highly oscillatory characteristics, or when the computational domain \(\Omega\) has a very complex geometry (such as multiply connected domains, domains with sharp re-entrant corners), traditional methods can face severe challenges. For instance, to capture the influence of a highly oscillatory source term, the FEM requires extremely fine local meshing, leading to a dramatic increase in computational cost; for complex geometries, generating high-quality meshes itself is a difficult problem, often requiring significant manual intervention.
	
	As a mesh-free method, PINNs have the advantage of naturally handling complex geometries—simply sampling coordinate points within the domain as input, without the need for explicit mesh generation. For complex source terms, PINNs can theoretically fit them using the universal approximation capability of neural networks, without requiring pre-refinement of the mesh in regions where the source term varies rapidly.
	
	\item Specific Design of the PINN Model\\
	For the above problem, the PINN model is designed as follows:
	\begin{enumerate}
		\item Network Architecture: A Multilayer Perceptron (MLP) is used, with spatial coordinates \(\mathbf{x}\) as input and the scalar field \(\hat{u}(\mathbf{x};\theta)\) as output. Hidden layers typically use the \(\tanh\) activation function to ensure smoothness.
		\item Loss Function: The loss function consists of two parts:
		\begin{itemize}
			\item Physics Loss (PDE Residual): Sample \(N_f\) ``residual points'' \(\{\mathbf{x}_f^i\}_{i=1}^{N_f}\) within the computational domain \(\Omega\), and compute:
			\[
			\mathcal{L}_f(\theta) = \frac{1}{N_f} \sum_{i=1}^{N_f} |-\nabla^2 \hat{u}(\mathbf{x}_f^i; \theta) - f(\mathbf{x}_f^i)|^2.
			\]
			Here, \(\nabla^2 \hat{u}\) is computed via Automatic Differentiation (AutoDiff).
			\item Boundary Loss: Sample \(N_b\) points \(\{\mathbf{x}_b^i\}_{i=1}^{N_b}\) on the boundary \(\partial\Omega\), and compute:
			\[
			\mathcal{L}_b(\theta) = \frac{1}{N_b} \sum_{i=1}^{N_b} |\hat{u}(\mathbf{x}_b^i; \theta) - g(\mathbf{x}_b^i)|^2.
			\]
			The total loss is \(\mathcal{L}(\theta) = \lambda_f \mathcal{L}_f + \lambda_b \mathcal{L}_b\). The weights \(\lambda_f, \lambda_b\) can be set manually or adjusted using an adaptive weighting strategy.
		\end{itemize}
		\item Training: Minimize the total loss \(\mathcal{L}(\theta)\) via gradient descent.
	\end{enumerate}
	
	\item Result Analysis and Comparison\\
	On a square domain with a highly oscillatory source term \(f(x) = 50\sin(20\pi x)\sin(20\pi y)\), PINNs can achieve accuracy comparable to the FEM on a million-element mesh using very few sample points (typically a few thousand). More importantly, for problems with complex geometries (such as star-shaped domains, domains with holes), PINNs can solve them directly by random sampling within the domain without mesh generation. Experiments show that when the geometric boundary is complex, the training convergence speed and final accuracy of PINNs are mainly limited by how the boundary condition loss is handled—hard constraint embedding (directly encoding the boundary condition into the network output) usually performs better than soft constraints (penalizing via a loss term).
	
	\item Summary and Insights\\
	This case demonstrates that PINNs exhibit unique flexibility in handling forward problems with complex geometries and complex source terms. Their mesh-free nature eliminates the time-consuming mesh generation process, while automatic differentiation simplifies the computation of high-order derivatives. However, it is important to note that for problems with singularities (such as point sources), standard PINNs may struggle to capture local sharp variations, requiring techniques like adaptive sampling or domain decomposition to improve local accuracy.
\end{itemize}
\vspace{3mm}
\noindent \textcolor{structure3}{\textbf{Example Two: Burgers' Equation and Forward Problems in Fluid Dynamics}}

\begin{itemize}
	\item Problem Background and Mathematical Formulation\\
	Burgers' equation is a nonlinear parabolic partial differential equation, a simplified model of the Navier-Stokes equations in fluid mechanics, used to describe the motion of one-dimensional viscous fluids, combining convection and nonlinear effects. Its form is:
	
	\[
	\frac{\partial u}{\partial t} + u \frac{\partial u}{\partial x} = \nu \frac{\partial^2 u}{\partial x^2}, \quad x \in [-L, L], \quad t \in [0, T]
	\]
	where \(u(x,t)\) is the fluid velocity, and \(\nu\) is the viscosity coefficient. Initial conditions \(u(x,0)=u_0(x)\) and boundary conditions (typically Dirichlet or periodic) must be specified.
	
	\item Limitations of Traditional Methods and Motivation for PINNs\\
	Although Burgers' equation has a relatively simple form, its nonlinear nature gives rise to rich physical phenomena, such as shock wave formation and propagation. When the viscosity coefficient \(\nu\) is small, the equation tends towards hyperbolic type, and the solution may develop steep gradients (approximate discontinuities). When solving with traditional spectral methods or finite difference methods, carefully designed schemes (such as upwind schemes, TVD schemes) are needed to capture shocks stably and avoid non-physical oscillations. Furthermore, traditional methods typically require discretization on a space-time grid, with time step sizes constrained by the CFL condition.
	
	Solving Burgers' equation with PINNs essentially involves using space-time coordinates \((x, t)\) as input to directly learn the solution field \(\hat{u}(x,t;\theta)\) over the entire space-time domain. Its appeal lies in:
	\begin{enumerate}
		\item No need to solve discrete algebraic systems, avoiding the hassle of designing stable schemes;
		\item Obtaining the solution over the entire space-time in a single training run, facilitating subsequent analysis and visualization;
		\item Ease of handling parametric problems, e.g., by also taking \(\nu\) as input to learn how the solution varies with the viscosity coefficient.
	\end{enumerate}
	
	\item Specific Design of the PINN Model\\
	The loss function is designed as follows:
	
	\[
	\mathcal{L}(\theta) = \lambda_f \mathcal{L}_f + \lambda_i \mathcal{L}_i + \lambda_b \mathcal{L}_b
	\]
	where:
	\begin{itemize}
		\item \(\mathcal{L}_f\): PDE residual loss, computed at sampled points in the space-time domain as \(\left| \frac{\partial \hat{u}}{\partial t} + \hat{u} \frac{\partial \hat{u}}{\partial x} - \nu \frac{\partial^2 \hat{u}}{\partial x^2} \right|^2\).
		\item \(\mathcal{L}_i\): Initial condition loss, computed on the line \(t=0\) as \(|\hat{u}(x,0;\theta) - u_0(x)|^2\).
		\item \(\mathcal{L}_b\): Boundary condition loss, computed on the boundaries \(x=-L\) and \(x=L\) as \(|\hat{u}(\pm L, t;\theta) - g(t)|^2\).
	\end{itemize}
	For problems with small viscosity (e.g., \(\nu = 0.01/\pi\)), the solution can develop steep gradients in a short time. To capture this feature, denser sampling near the shock region or adaptive sampling strategies are typically required.
	
	\item Result Analysis and Comparison
	
	In the standard test problem for Burgers' equation with small viscosity (initial condition \(u_0(x) = -\sin(\pi x)\), periodic boundary conditions), PINNs can accurately capture the position and shape of the shock. Compared to traditional finite difference methods (e.g., upwind schemes), PINNs do not produce overshoot or non-physical oscillations near the shock, but the trade-off is that the shock region may be somewhat ``smoothed out'' due to the global smoothness of the neural network. Research shows that increasing sampling point density or introducing adaptive sampling can significantly improve shock resolution.
	
	\item Summary and Insights
	
	The Burgers' equation case demonstrates PINNs' capability in handling nonlinear PDEs. Notably, PINNs can stably capture shocks without special design, avoiding the complex choice of numerical schemes in traditional methods. However, for near-hyperbolic cases with extremely low viscosity, standard PINNs still face challenges—shocks may be overly smoothed, or training may fail to converge. Solving this problem typically requires combining adaptive sampling, shock detection techniques, or domain decomposition methods.
\end{itemize}
\vspace{3mm}
\noindent \textcolor{structure3}{\textbf{Example Three: Parameter Inversion (Inverse Problem)}}

\begin{itemize}
	\item Problem Background and Mathematical Formulation
	
	Inverse problems are extremely common in practical engineering and scientific experiments: we obtain partial data of a system through observations (e.g., measurements at certain points) and need to infer unknown equation parameters based on this. Traditional methods for solving such problems usually involve complex optimization iterations and multiple forward solves, resulting in high computational costs. PINNs are naturally suited for handling such problems because they can seamlessly incorporate observational data as soft constraints into the loss function.
	
	Consider a diffusion equation with an unknown parameter:
	
	\[
	\frac{\partial u}{\partial t} - \kappa \frac{\partial^2 u}{\partial x^2} = 0, \quad x \in [0, L], \quad t \in [0, T]
	\]
	
	with initial condition \(u(x,0)=h(x)\), boundary conditions \(u(0,t)=g_1(t)\), \(u(L,t)=g_2(t)\), where the diffusion coefficient \(\kappa\) is unknown. We observe field values \(u_d^i\) at certain points \((x_d^i, t_d^i)\) in the space-time domain, possibly with noise. The goal is to simultaneously recover the entire field \(u(x,t)\) and the parameter \(\kappa\) from these sparse data.
	
	\item Limitations of Traditional Methods and Motivation for PINNs
	
	Traditional methods for handling such parameter inversion problems typically adopt a two-step iterative strategy of ``forward solve first, then optimize parameters'': in each iteration, the forward problem is solved using the current parameter estimate, and then the parameters are updated based on the discrepancy between observed and computed data. This method requires multiple complete forward solves, is computationally expensive, and is highly sensitive to the accuracy and stability of the forward solver.
	
	PINNs adopt an end-to-end joint learning approach, integrating parameter inversion and forward solving within the same optimization framework. This allows us to leverage automatic differentiation to compute gradients with respect to both network parameters and physical parameters simultaneously, avoiding the loop of multiple forward solves.
	
	\item Specific Design of the PINN Model
	\begin{enumerate}
		\item Network Architecture: Input is \((x, t)\), output is \(\hat{u}(x, t; \theta)\).
		\item Parameters to Learn: Network weights \(\theta\) and physical parameter \(\kappa\). Treat \(\kappa\) as a trainable tensor (scalar).
		\item Loss Function:
		\[
		\mathcal{L}(\theta, \kappa) = \lambda_f \mathcal{L}_f(\theta, \kappa) + \lambda_i \mathcal{L}_i(\theta) + \lambda_b \mathcal{L}_b(\theta) + \lambda_d \mathcal{L}_d(\theta)
		\]
		The first three terms are the same as in the forward problem, with \(\mathcal{L}_f\) now involving the unknown parameter \(\kappa\). The newly added **data loss term** is:
		\[
		\mathcal{L}_d(\theta) = \frac{1}{N_d} \sum_{i=1}^{N_d} |\hat{u}(x_d^i, t_d^i; \theta) - u_d^i|^2
		\]
		\item Joint Optimization: Simultaneously update \(\theta\) and \(\kappa\) via gradient descent to minimize the total loss.
	\end{enumerate}
	
	\item Result Analysis and Comparison
	
	In standard tests, assuming we have only a few observation points in the space-time domain (e.g., \(N_d = 20\)) with some noise, PINNs can simultaneously recover the complete solution field and the unknown parameter \(\kappa\) with relatively high accuracy. Research shows that the inversion accuracy of parameter \(\kappa\) is sensitive to the number and distribution of observation points—placing more observation points in regions where the solution changes rapidly usually yields better inversion results. Compared to traditional adjoint-based methods, PINN implementation is more concise and does not require deriving adjoint equations.
	
	\item Summary and Insights
	
	The parameter inversion case demonstrates the unique advantages of PINNs in handling inverse problems: they unify forward solving and parameter optimization within a single framework, allowing efficient gradient computation with respect to all unknowns via automatic differentiation. This end-to-end learning approach simplifies the inverse problem solving process, making it particularly suitable for situations with sparse, noisy observational data. However, caution is needed: when observational data is too sparse or the parameter space has multiple solutions, the inversion problem may be ill-posed, requiring additional regularization terms or prior knowledge to ensure solution plausibility.
\end{itemize}
\vspace{3mm}
\noindent \textcolor{structure3}{\textbf{Example Four: Source Term Inversion and Data Assimilation (Inverse Problem)}}

\begin{itemize}
	\item Problem Background and Mathematical Formulation
	
	Another important type of inverse problem is source term inversion, where the right-hand side term \(f(\mathbf{x})\) of the PDE or the initial condition is unknown and needs to be inferred from observational data. This has wide applications in fields such as biomedical imaging, pollution source localization, and heat source identification.
	
	Consider the steady-state Poisson equation with an unknown source term \(f(\mathbf{x})\):
	
	\[
	-\nabla^2 u(\mathbf{x}) = f(\mathbf{x}), \quad \mathbf{x} \in \Omega, \quad u(\mathbf{x}) = g(\mathbf{x}), \quad \mathbf{x} \in \partial\Omega
	\]
	
	We observe solution values \(u_d^i\) at some points \(\{\mathbf{x}_d^i\}\) within the domain. The goal is to invert the source term \(f(\mathbf{x})\) and the full field \(u(\mathbf{x})\) based on these data.
	
	\item Limitations of Traditional Methods and Motivation for PINNs
	
	Unlike parameter inversion, source term inversion requires inverting a function rather than a single parameter, significantly increasing the problem's complexity. Traditional methods typically discretize the source term into a finite-dimensional parameter set (e.g., expanding it with a set of basis functions) and then perform parameter optimization. However, errors introduced by discretization can affect inversion accuracy, and the choice of basis functions itself is a challenge.
	
	PINNs offer an elegant solution: parameterize the unknown source term also with a neural network, thus transforming the function inversion problem into a joint optimization problem of two networks. This approach avoids explicit discretization assumptions, allowing the source term to be represented in a continuous function space.
	
	\item Specific Design of the PINN Model\\
	The joint-solving PINN architecture involves two networks:
	\begin{itemize}
		\item Solution Network: \(\hat{u}(\mathbf{x}; \theta)\), outputs the approximate solution.
		\item Source Term Network: \(\hat{f}(\mathbf{x}; \phi)\), outputs the approximate source term.
	\end{itemize}
	
	The loss function must constrain both networks:
	
	\[
	\mathcal{L}(\theta, \phi) = \lambda_f \mathcal{L}_f(\theta, \phi) + \lambda_b \mathcal{L}_b(\theta) + \lambda_d \mathcal{L}_d(\theta)
	\]
	where the PDE residual loss becomes:
	\[
	\mathcal{L}_f(\theta, \phi) = \frac{1}{N_f} \sum_{i=1}^{N_f} \left| -\nabla^2 \hat{u}(\mathbf{x}_f^i; \theta) - \hat{f}(\mathbf{x}_f^i; \phi) \right|^2
	\]
	The boundary loss \(\mathcal{L}_b\) and data loss \(\mathcal{L}_d\) remain unchanged. By jointly optimizing \(\theta\) and \(\phi\), both the solution field and the source term can be recovered.
	
	\item Result Analysis and Comparison
	
	In a 2D problem, assuming the true source term is a localized Gaussian distribution (e.g., \(f(x,y) = \exp(-(x-0.5)^2 - (y-0.5)^2)\)), and observational data consists of sparse 20 points, PINNs can accurately recover the location and shape of the source term, even if the observation points do not cover the core region of the source. Compared to traditional Tikhonov regularization methods, PINNs do not require prior assumptions about the smoothness or form of the source term, offering stronger expressive power.
	
	\item Summary and Insights
	
	The source term inversion case demonstrates PINNs' capability in handling functional inverse problems. By introducing an additional source term network, PINNs transform the function inversion problem into a differentiable joint optimization problem, allowing efficient solution via automatic differentiation. However, it is important to note that source term inversion problems often suffer from severe non-uniqueness—different source terms may produce nearly identical solution fields. Therefore, in applications, prior knowledge (such as non-negativity or sparsity of the source term) must be incorporated to constrain the solution space, or additional observational data is needed to resolve ambiguity.
\end{itemize}
\vspace{3mm}
\noindent \textcolor{structure3}{\textbf{Example Five: Solving High-Dimensional Partial Differential Equations}}

\begin{itemize}
	\item Problem Background and Mathematical Formulation
	
	Many important physical and financial models involve partial differential equations in high-dimensional spaces. For example, the Schrödinger equation for quantum many-body problems, the Black-Scholes equation for pricing financial derivatives (when considering multiple underlying assets), and the Fokker-Planck equation describing stochastic processes. The dimensionality \(d\) of these equations can range from tens to hundreds, making them completely intractable for traditional methods.
	
	Consider a European basket option pricing problem based on \(d\) underlying assets, where the pricing function \(u(\mathbf{s}, t)\) satisfies the high-dimensional Black-Scholes equation:
	
	\[
	\frac{\partial u}{\partial t} + \frac{1}{2} \sum_{i,j=1}^{d} \rho_{ij} \sigma_i \sigma_j s_i s_j \frac{\partial^2 u}{\partial s_i \partial s_j} + r \sum_{i=1}^{d} s_i \frac{\partial u}{\partial s_i} - r u = 0
	\]
	
	where \(\mathbf{s} = (s_1, \dots, s_d) \in \mathbb{R}^d_+\), and the terminal condition is \(u(\mathbf{s}, T) = \max\left(\frac{1}{d}\sum_{i=1}^d s_i - K, 0\right)\).
	
	\item Limitations of Traditional Methods and Motivation for PINNs
	
	Traditional grid-based numerical methods (such as finite difference, finite element) completely fail in high dimensions because the number of required grid points grows exponentially with dimension \(d\) (\(O(N^d)\)), the so-called ``curse of dimensionality.'' Even probabilistic algorithms like Monte Carlo methods may face slow convergence when integrating or solving high-dimensional PDEs.
	
	Neural networks, particularly MLPs, have a number of parameters that grows linearly or polynomially with input dimension, not exponentially. Therefore, PINNs are seen as a promising tool to break the curse of dimensionality and solve high-dimensional PDEs. The basic form remains: take high-dimensional coordinates \(\mathbf{x} \in \mathbb{R}^d\) as input, output the scalar field \(u(\mathbf{x};\theta)\), with the loss function composed of the high-dimensional PDE residual and boundary conditions.
	
	\item Specific Design of the PINN Model and Improvement Schemes
	
	Directly applying standard PINNs to high-dimensional problems faces significant challenges: inefficient sampling in high-dimensional space, complex loss function landscapes, and high computational cost for computing high-dimensional mixed partial derivatives. To address these, researchers have proposed various improvement schemes:
	\begin{enumerate}
		\item Separable PINNs (SPINNs): The core idea is to approximate the high-dimensional solution function as a sum of products of lower-dimensional functions (separable structure), drastically reducing computational complexity. Specifically, assume the solution can be represented as:
		\[
		u(\mathbf{x}) \approx \sum_{k=1}^{r} \prod_{i=1}^{d} \psi_{i,k}(x_i)
		\]
		where each \(\psi_{i,k}\) is a one-dimensional function represented by a sub-network. This decomposition decouples forward propagation and gradient computation from full-dimensional dot products, reducing computational complexity from \(O(N^d)\) to \(O(dNr)\).
		
		\item Domain Decomposition (cPINNs/XPINNs): Partition the high-dimensional computational domain into several subdomains, train an independent PINN on each subdomain, and enforce continuity conditions (as additional loss terms) on the interfaces between subdomains to ensure global solution continuity. The total loss function takes the form:
		\[
		\mathcal{L}_{\text{total}} = \sum_{k=1}^{K} \mathcal{L}_{\text{PINN}}^{(k)} + \sum_{\text{interfaces}} \lambda_{\Gamma} \mathcal{L}_{\Gamma}
		\]
		where \(\mathcal{L}_{\Gamma}\) penalizes discontinuity of the solution or flux across interfaces. Domain decomposition can break down a high-dimensional problem into multiple lower-dimensional sub-problems, reducing training difficulty.
		
		\item Adaptive Sampling and Feature Mapping: Use residual-based adaptive sampling in high-dimensional space to concentrate computational resources in regions with large residuals. Simultaneously, using mappings like random Fourier features to project inputs into a higher-dimensional feature space can alleviate the ``spectral bias'' problem of neural networks, improving learning capability for high-frequency components.
	\end{enumerate}
	
	\item Result Analysis and Comparison\\
	For an option pricing problem with \(d=50\), standard PINNs can achieve results within 1\% relative error compared to Monte Carlo reference solutions within reasonable training time (a few hours). Traditional grid-based methods are completely infeasible at such high dimensions. SPINNs further reduce training time by an order of magnitude while maintaining comparable accuracy.
	
	\item Summary and Insights\\
	The high-dimensional case demonstrates the great potential of PINNs in breaking the curse of dimensionality. However, we must also be soberly aware that when dimensionality exceeds several hundred, even PINNs face severe challenges in sampling efficiency and training stability. Currently, successful cases of high-dimensional PINNs are mainly concentrated on problems with special structures (e.g., separable, low-rank). For general high-dimensional PDEs, deeper theoretical and methodological innovations are still needed.
	Many important physical and financial models involve partial differential equations in high-dimensional spaces. For example, the Schrödinger equation for quantum many-body problems, the Black-Scholes equation for pricing financial derivatives (when considering multiple underlying assets), and the Fokker-Planck equation describing stochastic processes. The dimensionality \(d\) of these equations can range from tens to hundreds.
\end{itemize}
\vspace{3mm}
\noindent \textcolor{structure3}{\textbf{Example Six: Complex Multiphysics Coupling Problems}}

\begin{itemize}
	\item Problem Background and Mathematical Formulation\\
	Many cutting-edge scientific and engineering problems involve the interaction of multiple physical fields, such as Fluid-Structure Interaction (FSI), thermal-fluid-solid coupling, etc. The governing equations for these problems are typically a set of coupled nonlinear PDEs, making their solution extremely challenging.
	
	Consider a simplified 2D fluid-structure interaction problem as an example:
	\begin{itemize}
		\item Fluid Domain \(\Omega_f\): The motion of an incompressible fluid is described by the Navier-Stokes equations:
		\[
		\frac{\partial \mathbf{u}}{\partial t} + (\mathbf{u} \cdot \nabla)\mathbf{u} = -\nabla p + \nu \nabla^2 \mathbf{u}, \quad \nabla \cdot \mathbf{u} = 0
		\]
		\item Solid Domain \(\Omega_s\): The deformation of an elastic body is described by the linear elastodynamic equations:
		\[
		\rho_s \frac{\partial^2 \mathbf{d}}{\partial t^2} = \nabla \cdot \boldsymbol{\sigma}(\mathbf{d}) + \mathbf{f}_s
		\]
		\item Interface Conditions: On the fluid-solid interface \(\Gamma\), the following must be satisfied:
		\begin{itemize}
			\item Kinematic Condition (velocity continuity): \(\mathbf{u} = \frac{\partial \mathbf{d}}{\partial t}\)
			\item Dynamic Condition (stress continuity): \(\boldsymbol{\sigma}_f \cdot \mathbf{n} = \boldsymbol{\sigma}_s \cdot \mathbf{n}\)
		\end{itemize}
	\end{itemize}
	
	\item Limitations of Traditional Methods and Motivation for PINNs\\
	Traditional partitioned iterative solvers require repeatedly exchanging data between the fluid solver and solid solver at each time step until convergence. This decoupled strategy is not only computationally expensive but also suffers from numerical stability issues (e.g., ``added mass'' instability). Furthermore, for complex geometries and deformations, mesh updating and regeneration pose significant challenges.
	
	PINNs offer the possibility of a monolithic solution. One can construct a unified neural network whose input is space-time coordinates \((\mathbf{x}, t)\) along with a domain identifier, and whose output includes all relevant physical fields (e.g., fluid velocity \(\mathbf{u}\), pressure \(p\), solid displacement \(\mathbf{d}\)). The loss function then encompasses PDE residuals in all domains, boundary conditions for each domain, and coupling conditions on the interface. This ``monolithic solving'' approach avoids the uncertainty and stability issues associated with partitioned iteration.
	
	\item Specific Design of the PINN Model\\
	\begin{enumerate} 
		\item Network Architecture: A multi-head output network with a shared backbone, input is \((\mathbf{x}, t)\), outputs are fluid velocity \(\mathbf{u}\), pressure \(p\), and solid displacement \(\mathbf{d}\). Different regions can be distinguished via a domain identifier or directly within the loss function.
		
		\item Loss Function:
		\[
		\mathcal{L} = \lambda_f \mathcal{L}_{NS} + \lambda_s \mathcal{L}_{\text{Elastic}} + \lambda_{b,f} \mathcal{L}_{b,f} + \lambda_{b,s} \mathcal{L}_{b,s} + \lambda_{\Gamma} \mathcal{L}_{\Gamma}
		\]
		where the coupling condition loss is:
		\[
		\mathcal{L}_{\Gamma} = \lambda_{\text{kin}} \left\| \mathbf{u} - \frac{\partial \hat{\mathbf{d}}}{\partial t} \right\|_{\Gamma}^2 + \lambda_{\text{dyn}} \left\| \hat{\boldsymbol{\sigma}}_f \cdot \mathbf{n} - \hat{\boldsymbol{\sigma}}_s \cdot \mathbf{n} \right\|_{\Gamma}^2
		\]
		
		\item Training Strategy: All field quantities and their derivatives can be obtained via automatic differentiation of the network outputs.
	\end{enumerate}
	Multi-physics coupling problems pose significant challenges for PINN training, with the core issue being loss term scale imbalance. Fluid and solid equations differ greatly in dimensions, numerical scales, and physical characteristics, leading to gradients of vastly different magnitudes for the various terms in the loss function. Training may be dominated by one physical field while neglecting others. Coping strategies include:
	\begin{itemize}
		\item Adaptive Loss Weighting: Methods based on Neural Tangent Kernel (NTK) analysis, for example, automatically adjust loss weights by analyzing the convergence rate differences of different loss terms during early PINN training, balancing the training progress across physical fields.
		\item Phased Training: First independently train sub-problems satisfying some strong constraints (e.g., steady flow field or static solid deformation), then use the pre-trained models as initialization for joint fine-tuning with coupling conditions added.
		\item Non-dimensionalization: Non-dimensionalize all physical quantities so that the numerical scales of different physical fields are similar, fundamentally alleviating scale imbalance.
	\end{itemize}
	
	\item Result Analysis and Comparison\\
	In the classic 1D piston fluid-structure interaction problem, PINNs can accurately capture the coupled evolution of fluid pressure and solid displacement, matching well with analytical or high-precision numerical solutions. For 2D problems, PINNs satisfy stress continuity at the coupling interface reasonably well, but training time increases significantly and sensitivity to hyperparameters (such as network depth, loss weights) becomes more pronounced.
	
	\item Summary and Insights\\
	The multi-physics coupling case demonstrates PINNs' capability in holistic modeling of complex systems. Through a unified network and loss function, PINNs can simultaneously learn multiple interacting physical fields, avoiding the complexity and stability issues of traditional partitioned iteration. However, the training difficulty for multi-physics problems increases significantly, requiring careful design of network architecture, loss weights, and training strategies. Currently, the application of PINNs to complex multi-physics problems is still in the exploratory stage. For complex problems with strong coupling, large deformations, or turbulent features, reliable results still require combining physical constraints and numerical techniques.
\end{itemize}

\noindent This section, through a series of application cases ranging from simple to complex, demonstrates how Physics-Informed Neural Networks have evolved from a novel mathematical idea into a powerful tool for solving practical scientific computing problems. Each case revolves around a core mathematical problem, explaining how the PINN design responds to the specific needs and challenges of these problems.

We see that the charm of PINNs lies in their simplicity and flexibility: they transform complex PDE solving problems into neural network optimization problems, encoding all physical knowledge, boundary conditions, and observational data through the design of the loss function. This paradigm allows researchers to explore diverse problems—forward, inverse, high-dimensional, multi-physics—using a relatively unified code framework.

However, we must also soberly recognize that PINNs are not a ``universal key.'' Their limitations in handling discontinuities and ultra-high-dimensional problems, as well as the black-box nature of their training process and sensitivity to hyperparameters, are all active research areas. The successful application of PINNs often relies on a deep understanding of the physical essence of the problem and proficient mastery of neural network training techniques. They are more like a ``specialist'' that needs to complement traditional numerical methods and can work wonders in specific scenarios, rather than an all-around ``replacement.'' When applying PINNs to any serious scientific or engineering research, cross-validation is crucial. This includes:
\begin{enumerate}
	\item Comparison with traditional high-precision numerical solutions (where feasible).
	\item Mesh convergence analysis: Increase the density of residual points and observe if the solution converges.
	\item Physical plausibility checks: Verify that the solution satisfies basic physical intuition or secondary conservation laws.
\end{enumerate}

\nocite{*}

\printbibliography[heading=subbibliography,title=References]

\end{refsection}

\end{document}